Normandie Université

# THÈSE

## Pour obtenir le diplôme de doctorat

### Spécialité INFORMATIQUE

### Préparée au sein de l'Université de Rouen Normandie

# Détection d'objets dans les documents numérisés par réseaux de neurones profonds

## Présentée et soutenue par
## MELODIE BOILLET

| Thèse soutenue le 10/01/2023<br>devant le jury composé de | | |
|---|---|---|
| M. ANDREAS FISCHER | PROFESSEUR DES UNIVERSITES, Haute école d'ingéniérie et d'archi. | Rapporteur du jury |
| M. HAROLD MOUCHERE | PROFESSEUR DES UNIVERSITES, UNIVERSITE NANTES | Rapporteur du jury |
| M. CHRISTOPHER KERMORVANT | , | Membre du jury |
| MME LAURENCE LIKFORMAN-SULEM | PROFESSEUR ASSOCIE, TELECOM PARISTECH | Membre du jury |
| MME CAROLINE PETITJEAN | PROFESSEUR DES UNIVERSITES, Université de Rouen Normandie | Membre du jury |
| M. THIERRY PAQUET | PROFESSEUR DES UNIVERSITES, Université de Rouen Normandie | Directeur de thèse |

**Thèse dirigée par THIERRY PAQUET (Laboratoire d'Informatique, du Traitement de l'Information et des Systèmes)**

UNIVERSITÉ DE ROUEN NORMANDIE

MIΣ

litis

# REMERCIEMENTS

Je tiens, en premier lieu, à remercier Thierry Paquet, qui a accepté d'encadrer et de superviser ma thèse, pour son expérience et ses précieux conseils. Je tiens également à remercier mon co-encadrant de thèse, Christopher Kermorvant, pour l'opportunité qu'il m'a offert en me proposant ce sujet, et pour m'avoir encadrée et soutenue durant ces trois années. Leur patience mais aussi leurs expériences ont permis des discussions enrichissantes et très intéressantes.

Je remercie les membres de mon jury d'avoir accepté d'évaluer mes travaux de thèse. Tout d'abord, mes deux rapporteurs Harold Mouchère et Andreas Fischer, pour le temps consacré à la lecture de ce manuscrit et leurs commentaires avisés. Merci également à Laurence Likforman et Caroline Petitjean d'avoir accepté de faire partie du jury de thèse.

Merci à Teklia de m'avoir permis de réaliser ma thèse dans les meilleures conditions, en m'encourageant à publier et en valorisant mon travail en l'intégrant dans de nombreux projets. J'aimerais également remercier toutes les personnes de l'entreprise avec qui j'ai eu la chance de collaborer. Je tiens tout particulièrement à remercier l'équipe de recherche : Martin, Marie-Laurence, Blanche, Chaza et Solène pour nos rencontres et discussions très inspirantes. Merci à toute l'équipe de Grenoble pour leur gentillesse et leur soutien, et plus spécialement à Bastien, avec qui j'ai pu échanger sur de nombreux points techniques, pour son aide et sa patience.

Je me dois aussi d'être très reconnaissante envers mon laboratoire, le LITIS, pour m'avoir offert tout le confort et le soutien matériel nécessaire au bon déroulement de ma thèse et plus particulièrement les membres de l'équipe Apprentissage pour leur accueil. Un chaleureux merci à Denis avec qui j'ai eu la chance de partager mon bureau pendant ces trois dernières années mais également de longues et captivantes discussions toujours remplies de joie.

Sur un plan plus personnel, je tiens à remercier toutes les personnes présentes durant ces trois années. Tout d'abord, mes parents et mes frères pour les nombreuses distractions qu'ils m'ont apportées mais aussi leur soutien. Merci également à ma belle famille pour leur présence et leurs relectures. Enfin, un grand merci à Quentin pour m'avoir soutenue et écoutée jour après jour.

# RÉSUMÉ

Qu'ils soient historiques ou modernes, imprimés ou manuscrits, les documents constituent un ensemble précieux d'informations souvent difficilement accessible dans leur forme originale. La transformation de ces documents en documents digitaux est désormais possible grâce à leur numérisation et à l'extraction automatique de leurs contenus. Cette extraction nécessite la détection de différents éléments tels que les lignes de texte, éléments cruciaux afin d'obtenir la transcription du texte présent dans les images. Bien que de nombreuses méthodes aient été proposées pour détecter ces éléments, l'analyse de la structure des documents reste un problème difficile : les modèles proposés souffrent de difficultés à généraliser à de nouvelles données et à des structures plus complexes, et ils nécessitent de nombreux exemples d'apprentissage.

Dans cette thèse, nous étudions différentes tâches liées à l'analyse de la mise en page de documents telles que la détection de lignes de texte, la séparation en actes ou encore la détection du support d'écriture (page). Ainsi, nous proposons deux modèles fondés sur des réseaux de neurones profonds suivant deux approches différentes. Les réseaux neuronaux ont démontré de bonnes capacités d'apprentissage dans de nombreux domaines d'application et notamment dans la détection d'objets. Récemment, de nouveaux types de réseaux neuronaux ont vu le jour, les réseaux à base de Transformers. Ceux-ci permettent de traiter plus efficacement les tâches de prédiction séquence-à-séquence telles que la traduction de texte. Leur adaptation aux tâches de vision a rapidement suscité l'engouement grâce à leurs performances élevées et leur capacité à produire des résultats séquentiels et structurés.

Notre objectif est de proposer un modèle permettant de détecter les objets en tenant compte des difficultés liées au traitement de documents, notamment le nombre restreint de données d'entraînement disponibles. De plus, les systèmes existants peuvent présenter des temps de traitement longs qui peuvent entraîner des coûts financiers importants et des impacts écologiques négatifs. Dans un cadre industriel, l'utilisation de tels systèmes ne semble pas appropriée, il est donc nécessaire de proposer des modèles plus parcimonieux en termes de nombre de paramètres afin d'obtenir des temps d'entraînement et d'inférence plus réduits.

Dans cette optique, nous proposons un modèle de détection niveau pixel et un second modèle de détection niveau objet. Nous commençons par proposer un modèle de détection comportant peu de paramètres, rapide en prédiction, et qui permet d'obtenir des masques de prédiction précis à partir d'un nombre réduit de données d'apprentissage. Le pré-entraînement de ce modèle sur différents jeux de données annotés a permis d'obtenir des gains significatifs de performances. Ces résultats nous ont donc conduits à mettre en place une stratégie de collecte et d'uniformisation de nombreux jeux de données, utilisés afin d'entraîner un modèle unique de détection de lignes démontrant de grandes capacités de généralisation à des documents hors échantillon.

Nous proposons également un modèle de détection à base de Transformers. La conception d'un tel modèle a nécessité de redéfinir la tâche de détection d'objets dans les images de documents et à en étudier différentes modélisations. Suite à cette étude, nous proposons une stratégie de détection d'objets consistant à prédire séquentiellement les coordonnées des rectangles englobant les objets grâce à une classification pixel. Cette stratégie permet d'obtenir un modèle comportant peu de paramètres et rapide en inférence. Les expériences préliminaires de détection de lignes de texte montrent des bonnes performances.

Enfin, dans un cadre industriel, de nouvelles données non annotées sont souvent disponibles. Ainsi, dans le cas de l'adaptation d'un modèle à ces nouvelles données, on s'attend à fournir au système le minimum de nouveaux exemples annotés. Le choix des exemples pertinents pour l'annotation manuelle est donc crucial pour permettre une adaptation réussie. Il est donc nécessaire que les systèmes effectuent la tâche finale tout en évaluant automatiquement leur confiance quant à leurs décisions. Ainsi, les décisions moins confiantes peuvent être soumises à un opérateur humain pour une annotation manuelle, tandis que les décisions plus confiantes sont conservées telles quelles pour fournir une annotation automatique.

À cet égard, nous proposons des estimateurs de confiance issus d'approches différentes pour la détection d'objets dans des images de documents. La première approche proposée est inspirée de la méthode de Monte Carlo et consiste à construire des estimations de confiance en utilisant la méthode du *dropout* au moment du test. Notre seconde proposition consiste à construire un système dédié indépendant, entraîné à prédire une estimation de confiance depuis une seule prédiction pendant l'inférence. Nous montrons que ces estimateurs permettent de réduire fortement la quantité de données annotées tout en optimisant les performances.

# ABSTRACT

Whether they are historical or modern, printed or handwritten, documents constitute a valuable collection of information that is usually difficult to access. The transformation of these documents into digital documents is now possible through their digitization and the automatic extraction of their contents. This extraction requires the detection of different elements such as text lines, which are essential to obtain the transcription of the image's textual contents. Although many methods have been proposed to detect these elements, the analysis of document structure remains a difficult problem : the proposed models suffer from difficulties in generalizing to new data and more complex structures, and they require many training examples.

In this thesis, we study multiple tasks related to document layout analysis such as the detection of text lines, the splitting into acts or the detection of the writing support (page). Thus, we propose two deep neural models following two different approaches. Neural networks have shown good learning capabilities in many application domains, and in particular in object detection. Recently, new types of neural networks have emerged, the Transformer-based networks. These systems allow processing more efficiently sequence-to-sequence tasks such as text translation. Their adaptation to vision tasks has quickly become popular thanks to their high performance and their ability to produce sequential and structured outputs.

We aim at proposing a model for object detection that considers the difficulties associated with document processing, including the limited amount of training data available. Moreover, existing systems can have long processing times that can result in significant financial costs and negative ecological impacts. In an industrial setting, the use of such systems does not seem appropriate, so it is necessary to propose more parsimonious models in terms of number of parameters to obtain reduced training and inference times.

In this respect, we propose a pixel-level detection model and a second object-level detection model. We first propose a detection model with few parameters, fast in prediction, and which can obtain accurate prediction masks from a reduced number of training data. The pre-training of this model on different annotated datasets allowed us to obtain significant performance gains. These results led us to implement a strategy of collection and uniformization of many datasets, which are used to train a single line detection model that demonstrates high generalization capabilities to out-of-sample documents.

We also propose a Transformer-based detection model. The design of such a model required redefining the task of object detection in document images and to study different approaches. Following this study, we propose an object detection strategy consisting in sequentially predicting the coordinates of the objects enclosing rectangles through a pixel classification. This strategy allows obtaining a fast model with only few parameters. Preliminary experiments on text line detection show good performances.

Finally, in an industrial setting, new non-annotated data are often available. Thus, in the case of a model adaptation to this new data, it is expected to provide the system as few new annotated samples as possible. The selection of relevant samples for manual annotation is therefore crucial to enable successful adaptation. Thus, it is necessary for the systems to perform the final task while automatically assessing their confidence about their own decisions. This way, less confident decisions can be submitted to a human operator for manual annotation, while more confident decisions are kept as is to provide an automatic annotation.

For this purpose, we propose confidence estimators from different approaches for object detection in document images. The first proposed approach is inspired by the Monte Carlo method and consists in building confidence estimates using the dropout method at test time. Our second proposal consists in building an independent dedicated system, trained to predict a confidence estimate with a single prediction during inference. We show that these estimators greatly reduce the amount of annotated data while optimizing the performances.

# TABLE DES MATIÈRES

# LISTE DES FIGURES

# LISTE DES TABLEAUX

# LISTE DES FOCUS

# LISTE DES ALGORITHMES

# PUBLICATIONS

# ACRONYMES

| | |
|---|---|
| **AP** | Average Precision |
| **CER** | Character Error Rate |
| **CNN** | Convolutional Neural Network |
| **DAP** | Dropout Average Precision |
| **DLA** | Document Layout Analysis |
| **DOV** | Dropout Object Variance |
| **FCN** | Fully Convolutional Network |
| **HTR** | Handwritten Text Recognition |
| **IoU** | Intersection-over-Union |
| **mAP** | Mean Average Precision |
| **mAP-RFR** | Mean Average Precision - Random Forest Regressor |
| **MLP** | Multi-Layer Perceptron |
| **NER** | Named Entity Recognition |
| **OCR** | Optical Character Recognition |
| **PCE** | Posterior Probability-based Confidence Estimator |
| **WER** | Word Error Rate |

1

# INTRODUCTION

## 1.1 CONTEXTE

Les documents historiques constituent un patrimoine précieux que les archives, bibliothèques et certaines entreprises cherchent à protéger, préserver et rendre accessible au plus grand nombre. Après de nombreuses années de numérisation, des millions d'images de documents sont maintenant disponibles dans le monde entier. Le contenu de ces images est cependant souvent compréhensible uniquement par des experts, qui travaillent à rendre accessible cette grande quantité de contenus afin de permettre aux chercheurs et au grand public de travailler plus facilement et efficacement. Pour cela, un long et coûteux travail de transcription manuelle des documents est souvent nécessaire. Afin de rendre cette tâche plus efficace, de nombreuses institutions cherchent à automatiser ce processus.

Grâce aux nouvelles technologies, et notamment l'amélioration majeure des méthodes d'apprentissage profond, il devient désormais possible de transformer automatiquement les documents originaux en documents digitaux, qui peuvent facilement être lus, traduits ou encore dans lesquels il est possible de faire des recherches avancées, tout en nécessitant une quantité plus raisonnable de travail de transcription manuelle. Dans le même temps, ces évolutions ouvrent de nouvelles perspectives de recherche à la communauté du traitement de document en mettant en évidence des documents complexes pour lesquels les avancées récentes restent encore insuffisantes.

Les différentes tâches liées au traitement automatique de documents numérisés telles que l'analyse de la mise en page (Document Layout Analysis (DLA)) ou la reconnaissance de texte (Handwritten Text Recognition (HTR)) sont des problématiques étudiées depuis de nombreuses années. Des solutions industrielles existent déjà mais sont souvent limitées à

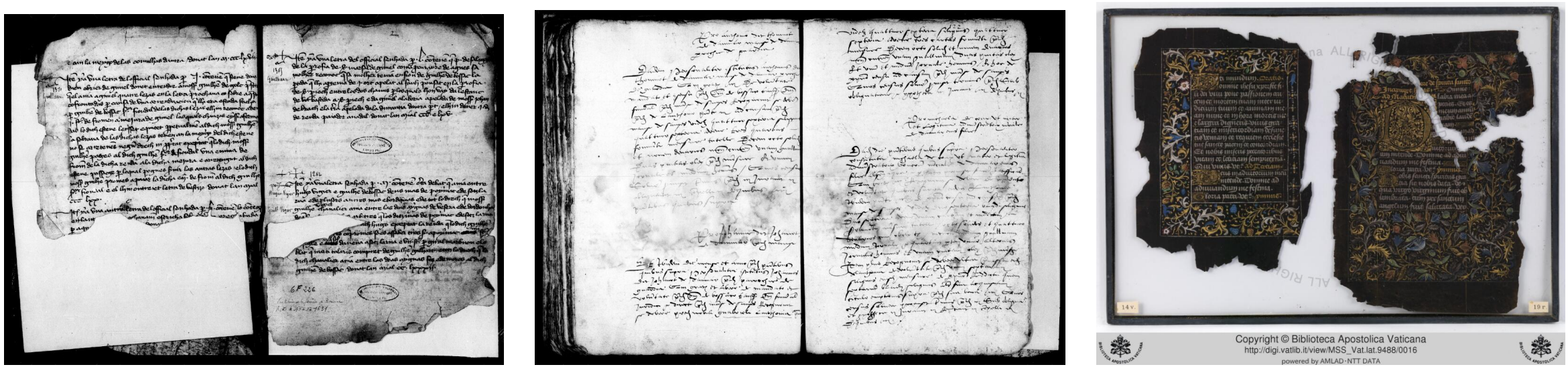

**Figure 1.1** – Pages présentant des difficultés de traitement : pages arrachées, dégradées et parties de pages manquantes. À gauche et au centre, images 4 et 141 du *Cartulaire de la famille de Boussac*[1]et, à droite, pages 14 verso et 19 recto du Livre d'heures du Vatican 09488.

1. https://bvmm.irht.cnrs.fr/resultRecherche/resultRecherche.php?COMPOSITION_ID=28605

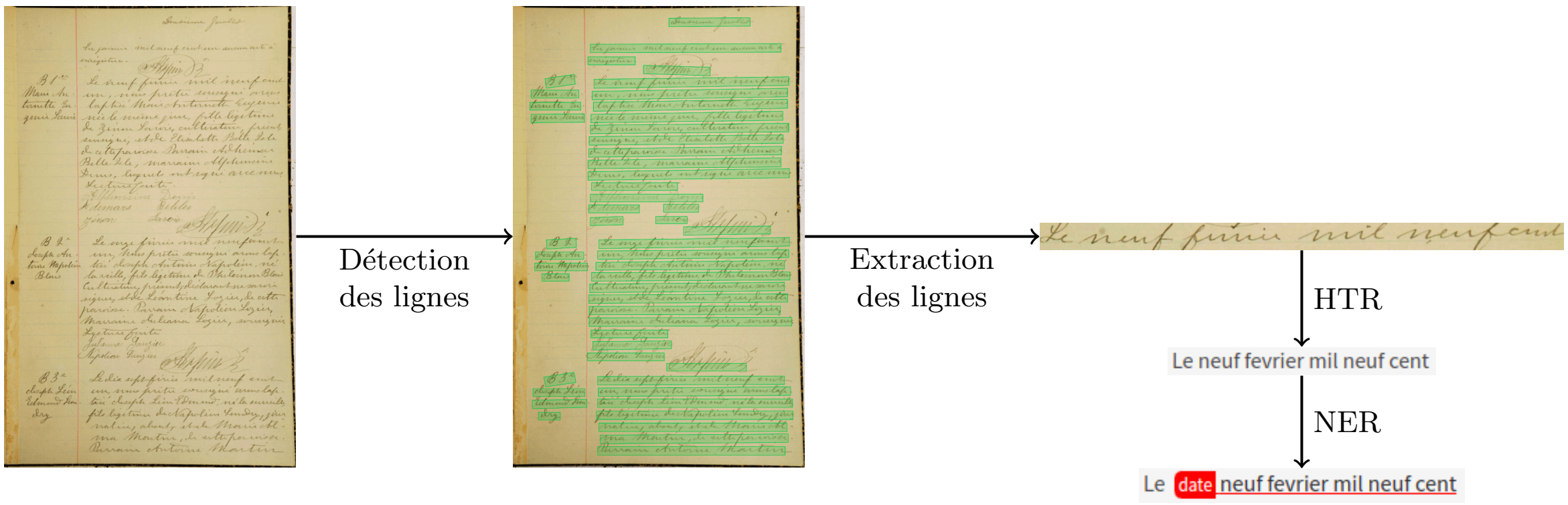

**Figure 1.2 –** Chaîne de traitement standard impliquant une détection de lignes de texte, une reconnaissance du texte manuscrit (HTR) suivi d'une détection des entités nommées (NER).

des documents modernes ou simples (mise en page simple, documents non dégradés). Les récentes avancées en *Machine Learning* et plus particulièrement en *Deep Learning* permettent désormais de lever ces limitations et d'améliorer la qualité des traitements automatiques. Ces méthodes nécessitent cependant une quantité importante de données annotées manuellement.

Pour les documents dont les mises en page sont simples et dont il est facile et rapide d'en annoter de grandes quantités, le traitement automatique a obtenu de très bons résultats. Au contraire, les jeux de données disponibles pour le traitement de documents plus complexes, tels que des documents historiques, sont très réduits. Cela est principalement dû au fait que les documents sont très variés, et donc coûteux à annoter manuellement. De plus, comme montré sur la Figure 1.1, les conditions de conservation et de numérisation peuvent mener à des manuscrits abîmés avec notamment des pages tâchées, arrachées ou dégradées. Pour toutes ces raisons, de nombreuses recherches s'intéressent à améliorer le traitement automatique de tels documents.

Avoir une version digitale d'un manuscrit historique permet, entre autres, de pouvoir faire de la recherche par mots-clés ou de retrouver des noms de personnes ou encore des dates. Pour parvenir à cela, plusieurs étapes sont appliquées à chaque page numérisée. Une chaîne de traitement utilisée dans de nombreux projets est présentée sur la Figure 1.2.

### 1.1.1 ANALYSE DE LA MISE EN PAGE

En entrée de la chaîne, nous disposons d'une image d'une page ou d'une double-page d'un document numérisé. Une première étape réalisée consiste à analyser la mise en page du document. L'objectif de ce premier module est d'identifier les diverses régions physiques d'un document et leurs caractéristiques. Cela revient donc à détecter différents éléments sur l'image tels que les blocs de texte, images, graphiques ou encore lignes de texte. Ces régions ne s'excluent pas mutuellement et une région peut contenir d'autres types de régions.

En plus de ces entités physiques, des étiquettes fonctionnelles ou logiques telles que des titres ou légendes peuvent être attribuées à certaines de ces régions. Le processus d'analyse de la structure et de la mise en page d'un document tente donc de décomposer l'image d'un document donné en ces régions et de comprendre leurs rôles fonctionnels et leurs relations.

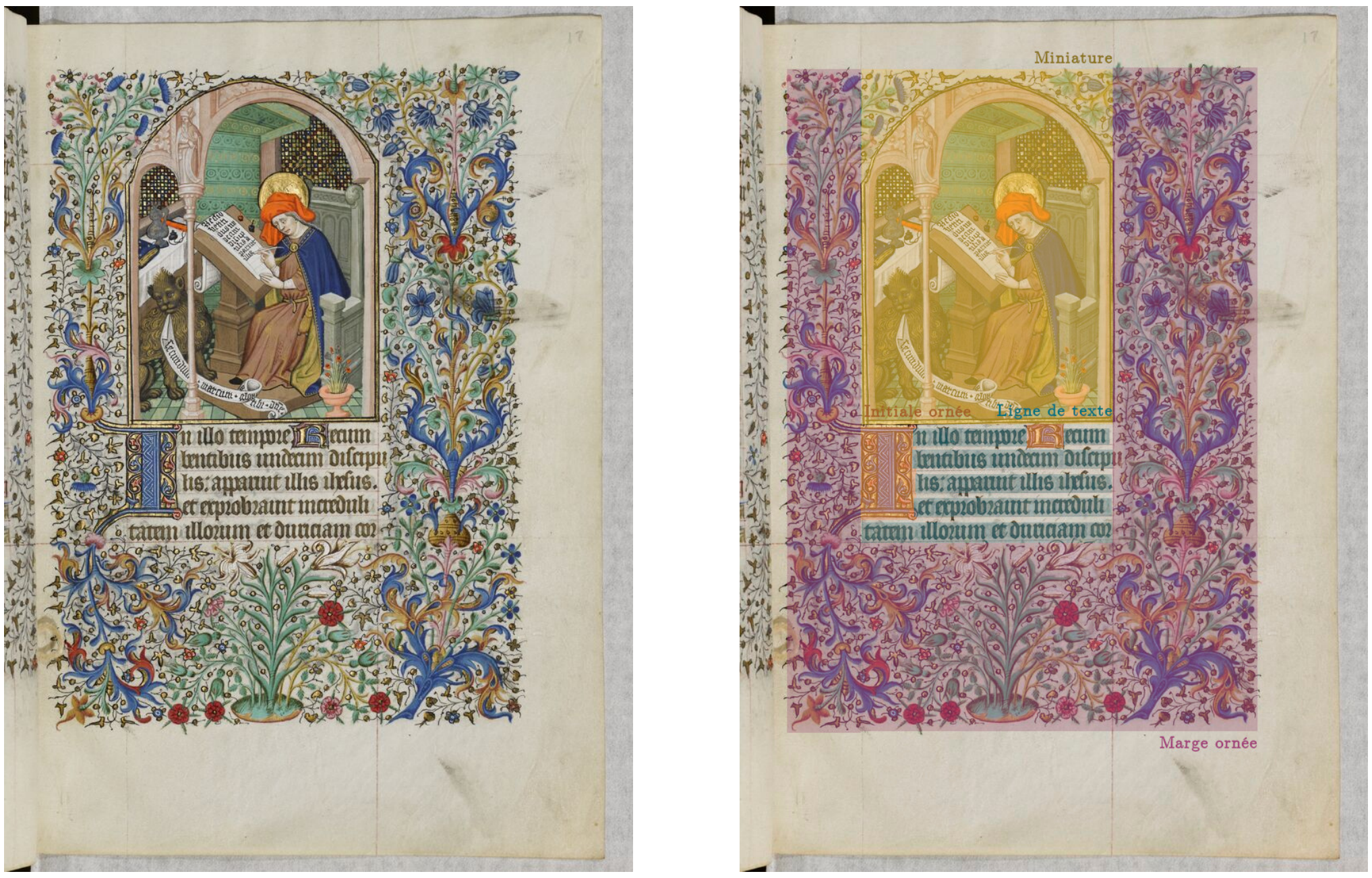

**Figure 1.3 –** Détection d'objets sur l'image de la page 17 recto du Livre d'heures *Horae ad usum Romanum*, Bibliothèque nationale de France, Département des manuscrits, NAL 3111. Source https://gallica.bnf.fr/.

Dans de nombreux cas d'usage, l'analyse de la mise en page d'un document revient à détecter les lignes de texte dans le but d'appliquer un reconnaisseur sur ces lignes. Cependant, certaines études s'intéressent également à d'autres éléments tels que les miniatures et initiales dans les livres d'heures (Boillet et al., 2019) (exemple Figure 1.3), les actes (Vézina et al., 2020) ou encore les tableaux de recensement (Constum et al., 2022). Ces tâches nécessitent des traitements plus spécifiques. En effet, la détection d'actes est souvent accompagnée d'une classification selon le type d'acte présent (baptême, mariage, décès). Il en est de même pour les tableaux, qui peuvent être traités de différentes manières : détection des lignes uniquement ou conjointement avec les colonnes ou encore détection des cellules.

### 1.1.2 Reconnaissance de texte

Une fois les lignes obtenues, elles subissent chacune un traitement menant à une version digitale du texte manuscrit (HTR). Enfin, ce texte peut être conservé tel quel, traduit dans une autre langue, ou encore traité afin d'obtenir les entités présentes dans le document (Named Entity Recognition (NER)). Des recherches récentes commencent à proposer des systèmes qui s'affranchissent de la détection des lignes de texte et permettent de transcrire le texte de l'image complète (Bluche, 2016 ; Coquenet et al., 2022 ; Singh et al., 2021 ; Yousef et al., 2020). Malgré des premiers résultats prometteurs, ces systèmes sont encore limités à des documents simples ou avec une grande régularité de mise en page.

Dans cette thèse, nous étudions les tâches liées à l'analyse de la mise en page telles que la détection de lignes de texte, d'actes ou encore de pages. Nous nous concentrons sur l'application de méthodes basées sur les réseaux de neurones profonds pour la détection d'objets dans les images de documents, principalement historiques. De nombreux systèmes permettant de résoudre ces différentes tâches ont été proposés dans la littérature (Ares Oliveira et al., 2018 ; Grüning et al., 2019 ; Mechi et al., 2021), cependant, ils sont souvent évalués uniquement sur la tâche de détection de lignes de texte, et sont difficilement généralisables à des documents aux structures plus diverses. Ainsi, dans cette thèse, nous cherchons à développer des modèles plus génériques et à réaliser des évaluations plus complètes.

De plus, les systèmes actuellement utilisés pour la détection d'objets dans les images de scènes naturelles, tels que les modèles YOLO (Redmon et al., 2016 ; 2017 ; 2018) et R-CNN (Girshick, 2015 ; Girshick et al., 2014 ; Ren et al., 2015), sont difficilement applicables aux documents historiques. Une des raisons à cela est l'importante quantité de données annotées qu'ils nécessitent pour être entraînés. Ainsi, il devient nécessaire de développer des systèmes moins complexes en termes de nombre de paramètres, de combiner plusieurs bases, de recourir au *Transfer learning* (Das et al., 2018) ou d'augmenter la quantité de données. Ces différents points ont été étudiés durant la thèse et les conclusions seront présentées dans la suite.

Enfin, des systèmes à base de Transformers (Vaswani et al., 2017) ont commencé à être proposés afin de résoudre plus efficacement les tâches liées aux problèmes séquence-à-séquence telles que la traduction de texte. À la suite de cela, certains travaux ont adapté ces systèmes aux tâches de vision et ont montré qu'ils permettent d'obtenir de très bonnes performances pour la classification d'images (Dosovitskiy et al., 2021) ou la détection d'objets (Chen et al., 2022). Nous nous sommes également intéressés à cette catégorie de systèmes, qui permettent d'avoir des sorties structurées des objets prédits.

## 1.2 CADRE DE LA THÈSE

Cette thèse a été réalisée au sein de l'entreprise Teklia [2] dans le cadre d'une collaboration avec le Laboratoire d'Informatique, de Traitement de l'Information et des Systèmes (LITIS) [3] à l'Université de Rouen Normandie.

Teklia a été fondée en 2014 et est spécialisée dans la compréhension automatique de documents. L'entreprise travaille sur diverses applications telles que le traitement automatique de documents historiques (livres d'heures, chartes), mais également le traitement de documents plus récents comme des tableaux de recensement de la population française. Les activités de recherche de l'entreprise s'inscrivent dans des projets de recherche français mais aussi internationaux comme les projets HOME [4], HuginMunin [5] et Balsac [6].

2. https://teklia.com
3. https://www.litislab.fr
4. https://www.history-of-medieval-europe.eu
5. https://hugin-munin-project.github.io/
6. https://balsac.uqac.ca

L'entreprise possède également une équipe spécialisée dans le développement, qui réalise à la fois des projets pour des clients, mais intègre également les résultats de l'équipe de recherche dans des applications [7] et facilite le travail de recherche en développant, notamment une plateforme d'annotation [8].

L'entreprise travaille sur de nombreux projets et les produits des travaux de recherche y sont directement intégrés, et donc appliqués dans de réelles conditions industrielles. Pour répondre aux demandes des projets, il est nécessaire d'avoir un détecteur d'objets robuste, performant et rapide pour traiter de grandes quantités de documents. De plus, il n'est pas rare que dans un projet il y ait peu, voire aucune donnée annotée. Il est donc également nécessaire d'avoir un détecteur assez générique afin de traiter ces documents plus facilement et d'estimer automatiquement la qualité des résultats fournis.

## 1.3 OBJECTIFS ET CONTRIBUTIONS

Les défis liés à la tâche de détection d'objets dans des images de documents sont nombreux, d'autant plus dans un cadre dans lequel de nouvelles données sont souvent disponibles, toujours plus variées et complexes. Les problématiques auxquelles nous cherchons à répondre sont les suivantes :

— Comment détecter efficacement les objets présents dans des images de documents variés, et à partir de peu d'exemples annotés manuellement ?
— Comment évaluer les modèles de détection pour représenter correctement la qualité des objets prédits ainsi que leurs impacts sur les tâches suivantes ?
— Comment estimer la confiance d'un modèle de détection quant à la qualité de ses prédictions ?

Pour répondre à toutes ces problématiques, plusieurs contributions ont été proposées durant cette thèse. Elles nous permettent de proposer une étude complète de détection d'objets allant de l'annotation manuelle à l'évaluation finale :

— Certains réseaux de neurones utilisés pour la détection d'objets fournissent un masque de prédiction où chaque pixel appartient à une classe d'objet. Nous proposons un modèle de détection possédant peu de paramètres et rapide en inférence, produisant des masques de prédiction très précis tout en nécessitant un nombre réduit de données annotées.
— D'autres systèmes plus récents permettent de générer une sortie structurée des objets détectés. Suivant cette idée, nous proposons un second modèle de détection qui montre des performances encourageantes.
— Nous montrons que malgré une grande hétérogénéité entre les documents mais aussi entre leurs annotations manuelles, l'entraînement de réseaux de neurones génériques permet d'obtenir des modèles encore plus performants et applicables à de nouvelles

7. https://arkindex.teklia.com
8. https://callico.teklia.com

données sans ré-entraînement. De plus, l'uniformisation des annotations entre les différents jeux de données permet d'entraîner des modèles de meilleure qualité.

— Nous proposons d'utiliser des métriques d'évaluation qui sont davantage en accord avec la tâche finale. En particulier, nous proposons des métriques liées à la reconnaissance de texte afin d'évaluer les modèles de détection de lignes de texte.

— Les données annotées sont souvent disponibles en faible quantité. Ainsi, nous proposons différents estimateurs de confiance et montrons, dans un cadre d'*active learning*, qu'ils permettent d'obtenir des modèles de détection d'objets plus performants avec moins d'exemples annotés.

## 1.4 ORGANISATION DU MANUSCRIT

Cette thèse est composée, outre cette introduction, de six chapitres.

**Chapitre 2 : État de l'art**

Le chapitre 2 présente un aperçu de l'état de l'art dans plusieurs domaines. Une revue des différentes approches de détection de lignes de texte et d'objets dans des images est présentée, allant des premières méthodes de traitements d'images aux plus récents systèmes établis à partir de réseaux neuronaux. De plus, les récentes méthodes combinant l'utilisation de l'image et du texte pour la détection d'objets sont décrites. Enfin, nous y présentons les techniques permettant d'estimer une confiance reflétant la qualité d'une prédiction, élément crucial lorsque les systèmes de détection sont utilisés en phase de production.

**Chapitre 3 : Entraînement et évaluation des systèmes de détection**

Nous présentons, dans le chapitre 3, une revue des différents jeux de données utilisés pour la détection d'objets dans les images de documents. Par la suite, nous proposons une étude des stratégies d'entraînement et d'évaluation utilisées par les systèmes récents avec, notamment, le détail des métriques d'évaluation basées sur les pixels et sur les objets.

**Chapitre 4 : Détection d'objets dans des images de documents**

Dans le chapitre 4, nous introduisons une architecture simple, rapide et efficace, mise au point afin de détecter des objets dans les images de documents au niveau pixel. La détection est réalisée grâce à un réseau de neurones entièrement convolutif. Ce chapitre décrit les détails d'architecture ainsi que les avantages de celle-ci par rapport aux systèmes existants. Enfin, les résultats de différentes expérimentations sur les tâches de détection de lignes de texte et d'actes y sont présentés et discutés.

**Chapitre 5 : Entraînement et évaluation d'un modèle robuste de détection d'objets**

Le chapitre 5 propose une étude avancée des techniques d'entraînement et d'évaluation des systèmes de détection d'objets. Il expose la grande hétérogénéité et les incohérences des annotations des différents jeux de données actuellement disponibles, et présente une technique d'uniformisation des annotations mise au point durant la thèse. De plus, ce chapitre met en lumière les limitations des métriques d'évaluation actuellement utilisées et détaille plusieurs métriques que nous proposons afin de lever ces limitations. Les résultats d'entraînements de modèles de détection de lignes de texte à grande capacité de généralisation sont enfin présentés.

**Chapitre 6 : Estimation de la confiance des prédictions**

Nous proposons, dans le chapitre 6, différents estimateurs de confiance. Dans un premier temps, des estimateurs basés sur le modèle de détection d'objets entraîné sont présentés. Des estimateurs basés sur un apprentissage externe au détecteur sont ensuite détaillés. Une étude comparative des différentes approches est menée sur deux tâches de détection de pages et de lignes de texte.

**Chapitre 7 : Détection séquentielle d'objets dans des images de documents**

Le chapitre 7 présente une seconde architecture de détection d'objets, celle-ci étant établie à partir de Transformers. Les détails de l'architecture sont présentés ainsi que la justification des choix de conception. Des premiers résultats d'expérimentations sont également présentés.

**Chapitre 8 : Conclusions et perspectives**

Dans le chapitre 8, nous concluons sur l'ensemble des travaux proposés et énonçons des pistes de recherche complémentaires.

# 2

# ÉTAT DE L'ART

Les recherches axées autour de la mise en place et de l'amélioration de modèles de détection d'objets sont toujours très actives, et ont motivé un nombre croissant de travaux ces dernières années du fait d'importantes avancées dans le domaine de l'apprentissage automatique. Dans ce chapitre, nous présentons une étude des travaux liés à la détection d'objets en évoquant les premiers systèmes permettant de séparer les blocs de texte du fond des images ainsi que les méthodes les plus récentes basées sur des réseaux de neurones profonds. De plus, nous passons en revue différents systèmes d'estimation de la qualité des prédictions.

En section 2.1, nous décrivons les méthodes ad hoc de détection d'objets dans les images de documents. Nous présentons ensuite les méthodes proposées à base d'apprentissage profond, avec notamment les approches pixel dans la section 2.1.2 et celles à base de Transformers en section 2.1.2. Enfin, la section 2.2 présente les travaux permettant d'estimer la qualité des prédictions, peu de travaux ayant été publiés pour la tâche de détection d'objets.

## 2.1 DÉTECTION D'OBJETS DANS DES IMAGES DE DOCUMENTS

La mise en page d'un document fait référence à la position physique et aux limites des différentes régions dans l'image du document. Le processus d'analyse de la mise en page d'un document vise à décomposer une image de document en une hiérarchie de régions, telles que les figures, l'arrière-plan, les blocs de texte, les lignes de texte, les mots, les caractères, etc. Depuis plusieurs années, différentes méthodes permettant de détecter des objets dans des images de documents ont émergé. Ces différentes approches peuvent être divisées en deux groupes : les méthodes ad hoc et les méthodes par apprentissage automatique. De plus, dans chacun de ces deux groupes, il existe des algorithmes dits ascendants et descendants (Namboodiri et al., 2007 ; Song et al., 2003).

Les algorithmes ascendants partent des plus petits composants d'un document (pixels ou composantes connexes) et les regroupent de manière itérative pour former des régions plus grandes telles que les caractères, qui sont ensuite regroupés en mots, lignes ou blocs de texte. En revanche, les algorithmes descendants partent de l'image complète du document et la divisent itérativement en sous-images pour former des régions de plus en plus petites. La procédure de découpage s'arrête lorsqu'une certaine condition est vérifiée, les sous-images obtenues à ce stade constituent les résultats finaux de la segmentation. En outre, il existe également des approches hybrides qui utilisent une combinaison de stratégies ascendantes et descendantes.

### 2.1.1 MÉTHODES AD HOC

Les approches ad hoc sont basées sur la combinaison de différentes techniques d'analyse d'image telles que le regroupement, les profils de projection ou encore le filtrage. Elles sont établies pour un type d'images donné et sont peu généralisables à un grand nombre et une grande variété d'images de documents mais sont encore aujourd'hui utilisées (Eskenazi et al., 2017).

Les premières méthodes ayant vu le jour permettaient de séparer les contenus textuels des contenus graphiques d'une image sans nécessiter d'annotations manuelles. Parmi les algorithmes descendants, le *Run-Length Smoothing Algorithm* (RLSA) (Wong et al., 1982) a été proposé pour segmenter les pages de documents. Cet algorithme fonctionne sur des images binaires dans lesquelles deux pixels noirs voisins éloignés d'une distance maximale donnée sont fusionnés en une séquence continue de pixels noirs. Le RLSA est d'abord appliqué ligne par ligne, puis colonne par colonne, et les deux bitmaps résultants sont combinés en appliquant une opération logique "ET" à chaque position de pixel. L'inconvénient de cet algorithme est qu'il ne peut être utilisé que pour extraire de petits blocs. Par la suite, la méthode du *XY-Cut* (Nagy et al., 1984) a été proposée afin de détecter les blocs de texte dans des images en niveaux de gris. Cette méthode consiste à utiliser une projection horizontale et verticale des valeurs des niveaux de gris des pixels afin de trouver les espaces interlignes et intercolonnes. Les projections sont faites de manière itérative menant à des objets homogènes. Cette technique permet d'obtenir une détection de grande qualité mais est limitée à des documents dont la mise en page est simple. En effet, elle est incapable de prédire des objets corrects sur des images dans lesquelles les lignes sont mal alignées, ou si le document est légèrement incliné. Akindele et al. (1993) ont proposé une amélioration de cette méthode afin de corriger l'inclinaison des lignes, cependant, d'autres problèmes persistent tels que la difficulté du système à traiter des documents comportant des illustrations. Ces méthodes semblent difficilement applicables à des documents historiques qui ne sont pas que textuels et qui ont des mises en page complexes. Pour résoudre le problème posé par les images de pages obliques, Pavlidis et al. (1992) ont proposé une méthode basée sur les « flux blancs ». Ils émettent l'hypothèse que les colonnes de texte d'une page contiennent un type unique de données (texte ou illustration) et qu'elles sont suffisamment espacées pour être distinguées des autres espaces tels que l'espacement entre les mots. La méthode identifie donc les larges espaces blancs afin d'estimer l'angle d'inclinaison de la page puis de localiser les objets comme étant les régions entre ces espaces. Cette méthode permet également de traiter des documents plus complexes contenant, entre autres, des illustrations.

Les systèmes présentés ci-dessus permettent de traiter des documents ayant des mises en page de type Manhattan. Il s'agit de pages ayant des composants de formes arbitraires (rectangulaires) où les segments des blocs sont parallèles ou perpendiculaires les uns par rapport aux autres. Dans cette thèse, nous nous intéressons principalement aux documents historiques. Ces méthodes semblent donc difficilement applicables à de tels documents dont les colonnes contiennent rarement des types uniques de données, comme montré sur les Figures 1.1 et 1.3, et dont les mises en page sont non-Manhattan.

Les méthodes ascendantes permettent de traiter des documents beaucoup plus variés aux mises en page complexes mais sont en général plus lentes. Une des premières méthodes, présentée par Kise et al. (1998), se base sur le diagramme de Voronoi pour la segmentation d'images de pages. Les auteurs détectent tout d'abord les points des bords des composantes connexes et construisent un diagramme de Voronoi à partir de ces points. Les arêtes de Voronoi détectées entre des caractères, mots et lignes de texte d'un même bloc sont ensuite filtrées pour garder uniquement celles qui séparent les blocs du document. Un désavantage à cette méthode est qu'elle segmente parfois les illustrations et les titres ayant des polices d'écriture larges. O'Gorman (1993) a présenté DocStrum, qui repose sur un regroupement des plus proches voisins appliqué aux composantes connexes. Par la suite, ces deux propositions ont été conjointement utilisées et améliorées par Agrawal et al. (2009) avec leur système appelé Voronoi++. Il a été mis au point pour répondre au manque d'adaptation des systèmes existants aux variations de taille, d'orientation et de distance des composants d'une page. Au lieu d'utiliser des relations linéaires entre la distance et le rapport de surfaces des composantes connexes, les auteurs montrent que la détermination dynamique de ces relations et la combinaison des caractéristiques angulaires et des caractéristiques de voisinage, ces dernières venant de l'approche de DocStrum, améliorent la précision.

Pour lever les limitations liées aux systèmes présentés ci-dessus telles que le temps de traitement, d'autres méthodes ont émergées. Celles-ci sont basées sur des algorithmes plus robustes face aux images de documents couleurs et en niveaux de gris. De plus, elles ne sont plus limitées aux documents possédant des structures et contenus simples. Dans cette optique, Coüasnon (2006) a conçu et publié un langage de grammaire de mise en page appelé DMOS. Il permet de décrire une grande variété de mises en page et l'analyseur syntaxique associé reconnaît cette disposition dans une image. La grammaire permet également d'associer une étiquette à chaque région. Par la suite, la méthode a été améliorée (Lemaitre et al., 2008) en intégrant une approche multirésolution lui permettant de segmenter des lettres manuscrites et d'identifier les lignes de texte dans des documents administratifs. Dans la même idée, Shafait et al. (2008) ont proposé un autre algorithme de grammaire basé sur une formulation probabiliste de la mise en page. L'utilisateur définit un ensemble de coupes horizontales et verticales dont la position est définie de manière approximative. Ensuite, pour chaque image, un ajustement probabiliste est effectué pour obtenir les régions finales. Cet algorithme est capable de segmenter des mises en page serrées avec de faibles marges. Bien que cette méthode ainsi que DMOS aient obtenu de très bonnes performances, les systèmes reposent sur l'hypothèse que les documents à traiter ont une mise en page homogène puisqu'ils nécessitent que l'utilisateur définisse des règles de mise en page.

D'autres systèmes ont ensuite été proposés afin de traiter des documents complexes et ne nécessitant pas de modèle de mise en page prédéfini. Par exemple, Louloudis et al. (2007) ont utilisé la transformée de Hough sur un ensemble de composantes connexes sélectionnées pour extraire les lignes de texte. Cette approche, basée sur la transformée de Hough, n'est adaptée qu'aux images de documents où les lignes ne sont pas incurvées. Journet et al.

(2008) utilisent une approche ascendante basée sur les textures des images de documents. Ils extraient cinq caractéristiques liées aux fréquences et orientations calculées à quatre résolutions, ainsi chaque pixel de l'image possède 20 valeurs. Ils utilisent ensuite un algorithme de groupement afin de regrouper les pixels correspondant à des zones homogènes. Ils ont testé leur méthode sur des documents modernes et historiques, et ont souligné l'importance d'une approche multirésolution pour réduire le bruit dans les techniques ascendantes. Dans Shi et al. (2009), les auteurs proposent une technique appelée ALCM (*Adaptive Local Connectivity Map)*. Ils utilisent des filtres directionnels orientables pour détecter les lignes de texte et appliquent des post-traitements heuristiques pour séparer les lignes connectées. Cette méthode descendante a obtenu des résultats intéressants sur la détection de lignes de texte, l'algorithme ayant été conçu pour résoudre les problèmes particulièrement complexes observés dans les documents manuscrits, notamment les lignes de texte qui fluctuent, se touchent ou se superposent. Par la suite, Erkilinc et al. (2012) ont proposé une méthode de segmentation robuste face aux fonds et aux structures complexes. Cette approche permet de résoudre un problème de détection à trois classes : texte, photographie et ligne. Tout d'abord, l'image subit une étape de prétraitement qui consiste à réaliser un filtrage, une conversion de l'espace couleur et une correction gamma. Les éléments sont ensuite détectés grâce à plusieurs techniques telles que la transformée en ondelettes et le codage par plages. Les objets détectés sont enfin combinés par un algorithme de K-moyennes. Cette méthode de classification en blocs et en pixels a montré de bons résultats. Cependant, comme la plupart des méthodes présentées ici, elle consiste en plusieurs opérations successives et est coûteuse en temps. De plus, cette méthode ne permet de résoudre qu'un problème spécifique avec trois classes très distinctes. Enfin, une autre méthode ascendante qui a obtenu de bons résultats pour détecter les lignes de texte est décrite dans Ryu et al. (2014). L'approche est basée sur les super-pixels pour obtenir des composantes connexes. Les auteurs définissent une fonction de coût pour agréger les super-pixels en une ligne de texte. Cette méthode a gagné la compétition de l'*International Conference on Document Analysis and Recognition* (ICDAR) sur la détection des lignes de texte (Murdock et al., 2015).

Concernant les méthodes hybrides, un travail récent proposé par Tran et al. (2015) utilise la méthode *Multilevel Homogeneous Structure* (MHS), et a remporté la compétition de segmentation de documents complexes en 2015 (Antonacopoulos et al., 2015). La méthode implique à la fois l'analyse en composantes connexes et l'analyse des espaces blancs (arrière-plan). Tout d'abord, l'image est binarisée puis les composantes connexes sont détectées et celles considérées de manière fiable comme étant du bruit ou des régions sans texte sont filtrées. Une classification multiniveaux est effectuée, basée sur l'analyse des régions homogènes multiniveaux et des espaces blancs, pour identifier toutes les composantes textuelles et non textuelles. Cette méthode a montré de bonnes performances sur la compétition, notamment pour sa capacité à manquer très peu de régions.

Même si la plupart de ces méthodes ont obtenu de bons résultats sur un jeu de données spécifique, elles doivent être affinées manuellement, ce qui est une tâche fastidieuse et dépend généralement de l'ensemble de données considéré. De plus, une fois mises en place, ces méthodes sont souvent difficiles à maintenir et à améliorer. Enfin, la plupart des algorithmes mentionnés ci-dessus ne créent pas de descriptions hiérarchiques ou ne permettent pas aux utilisateurs de préciser des informations sur la structure du document. En outre, à part pour les modèles à base de grammaire, ils ne fournissent pas de méthodes d'estimation des paramètres de l'algorithme à partir de données. En d'autres termes, ils ne sont pas dotés de capacités d'apprentissage.

### 2.1.2 MÉTHODES PAR APPRENTISSAGE PROFOND

Pour répondre à ces difficultés, des méthodes basées sur un apprentissage ont été proposées afin d'apprendre automatiquement la variabilité des documents à partir de données. Nos travaux se positionnent dans ce cadre.

Les méthodes par apprentissage automatique sont actuellement principalement constituées d'algorithmes de réseaux de neurones profonds. Ces algorithmes permettent à la fois d'apprendre automatiquement les caractéristiques importantes des images et d'effectuer la tâche requise. Ils ont tendance à être une « boîte noire » dont le fonctionnement est difficile à expliquer, cependant ils permettent de traiter des documents complexes que les méthodes ad hoc sont incapables de traiter.

Les approches par apprentissage profond ont obtenu de bons résultats dans de nombreux domaines d'application (LeCun et al., 2015), ainsi, de nombreux travaux ont étudié leur utilisation pour la détection d'objets dans les images. Puisque de multiples recherches s'orientent autour de la détection d'objets dans des images en général et non spécifiquement dans des images de documents, la section suivante passe en revue quelques travaux dans ces domaines connexes de détection d'objets et de textes dans des images de scènes naturelles. Dans le domaine de la vision par ordinateur, la littérature sur la détection d'objets peut être divisée en trois catégories principales : les systèmes basés sur la proposition de régions, l'estimation de la position des boîtes englobantes par régression et la détection au niveau du pixel.

#### PROPOSITION DE RÉGIONS

Pour la tâche de détection de texte dans des images de scènes naturelles, les premiers travaux basés sur des approches par apprentissage profond utilisent une méthode de fenêtre glissante (Zhu et al., 2016). Des parties d'images sont d'abord extraites à l'aide d'une fenêtre glissante, puis elles sont étiquetées grâce à un réseau de neurones profond. L'utilisation d'une fenêtre glissante induit un temps de traitement élevé et limite le contexte qui peut être utilisé pour prendre une décision. Pour limiter le temps de traitement, une solution consiste à utiliser un prétraitement pour extraire les candidats et ensuite prendre une décision pour chacun de ces candidats. C'est la méthode utilisée par Huang et al. (2014), qui extrait les candidats grâce aux *Maximally Stable Extremal Regions* (MSER) et les classe à l'aide d'un réseau de neurones convolutif (Convolutional Neural Network (CNN)) (LeCun et al., 1998). L'architecture CNN est décrite à la fin de cette section, dans le Focus 2.1.

De la même manière, l'idée d'extraire les candidats avant de les classer a été utilisée pour la détection d'objets. Ces systèmes, basés sur la proposition de régions, consistent en trois étapes consécutives. Tout d'abord, un ensemble de propositions de régions indépendantes de la catégorie est généré. Ensuite, un CNN est appliqué sur ces régions pour extraire les informations significatives, et un classificateur prédit la classe de chaque proposition de région. Cette stratégie a été proposée pour la première fois par Girshick et al. (2014) avec leur système R-CNN, détaillé dans le Focus 2.4, et appliquée aux images de scènes naturelles des jeux de données VOC 2010-2012. Malgré le développement de systèmes plus avancés (Fast R-CNN (Girshick, 2015), Faster R-CNN (Ren et al., 2015) et Zhong et al. (2017)), cette méthode a été peu adoptée par la communauté du traitement d'images de documents. En effet, ces systèmes sont bien adaptés aux images de scènes naturelles où seuls quelques objets sont présents sur les images, contrairement aux images de documents qui contiennent de nombreux objets de toutes tailles. De plus, malgré différentes améliorations qui ont permis d'accélérer ces systèmes, ils restent complexes et peu efficients, d'où l'introduction des méthodes dites « one stage » où l'on s'abstient de l'étape de proposition de régions. Certains de ces systèmes sont présentés dans les paragraphes suivants.

### Focus 2.1 – Architecture CNN

**Définition**

Un réseau de neurones profond est une succession de plusieurs couches où chaque couche est généralement composée d'une fonction paramétrée suivie d'une fonction de non-linéarité (fonction d'activation), chaque couche calculant une nouvelle représentation de l'image d'entrée. Dans le cas d'un réseau neuronal convolutif (CNN), les fonctions paramétrées sont des opérations de convolution, détaillées dans le Focus 2.2. La partie convolutive d'un CNN permet d'extraire et de compresser les caractéristiques de l'image d'entrée grâce à des couches de regroupement (*pooling*, expliqué dans le Focus 2.3).

**Avantages**

— Un CNN est capable de capturer avec succès les dépendances spatiales d'une image par l'application de filtres. L'architecture s'adapte au mieux à l'ensemble des données grâce à la réduction du nombre de paramètres impliqués et à la réutilisation des poids.

— L'architecture CNN rend possible l'apprentissage de modèles profonds avec relativement peu de paramètres grâce au partage des poids entre les couches convolutives.

— Chaque filtre d'une couche de convolution est appliqué à l'ensemble de l'image d'entrée, ainsi le traitement d'une image par un CNN est invariant par translation.

— Comparés à d'autres algorithmes de traitement d'image, les CNN utilisent relativement peu de prétraitement.

### Inconvénients

— Pour entraîner un CNN, il est souvent nécessaire d'avoir de nombreuses données annotées.

— Comme pour la plupart des systèmes à base de réseaux neuronaux, il peut être coûteux en mémoire et en temps d'entraîner un CNN.

### Exemples de systèmes de type CNN

— LeNet (LeCun et al. (1998)) : LeNet est la première architecture CNN. Il a été développé en 1998 et a été appliqué avec succès à la tâche de reconnaissance de chiffres manuscrits. L'architecture LeNet se compose de plusieurs couches de convolution et de regroupement (*pooling*), suivies d'une partie entièrement connectée. Le modèle comporte cinq couches de convolution suivies de deux couches entièrement connectées.

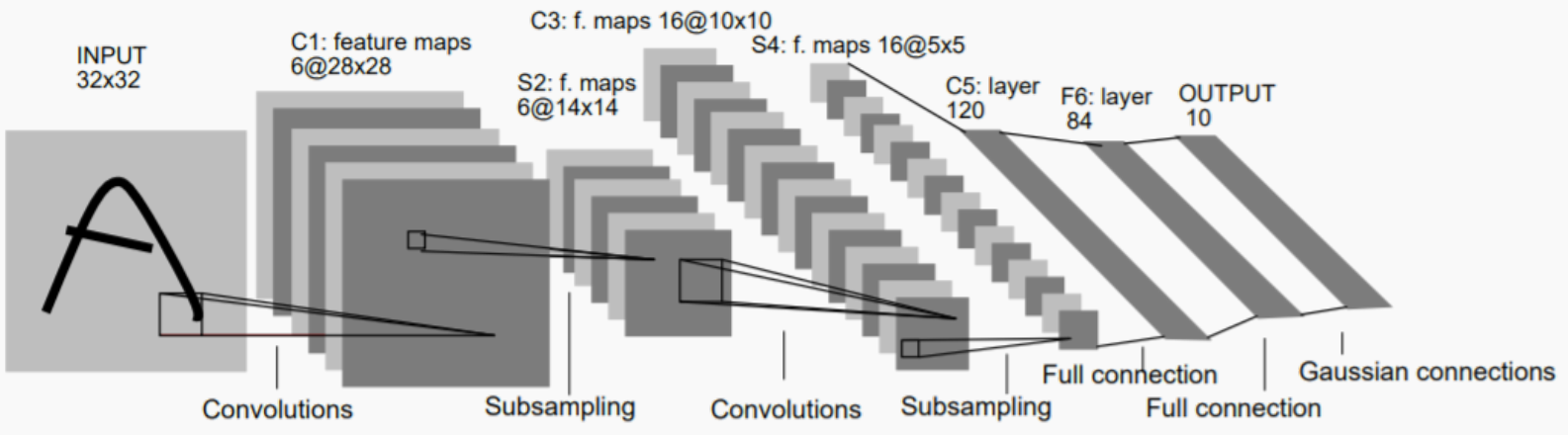

**Figure 2.1 –** Schéma de l'architecture du modèle LeNet, issu de LeCun et al. (1998), pour une image d'entrée de taille 32×32 pixels.

— AlexNet (Krizhevsky et al. (2012)) : AlexNet est l'architecture d'apprentissage profond qui a popularisé le CNN. Le réseau AlexNet a une architecture très comparable à celle de LeNet, mais est plus profond, plus grand et comporte des couches convolutives empilées les unes sur les autres. AlexNet a été utilisé pour remporter l'*ImageNet Large Scale Visual Recognition Challenge* (ILSVRC) en 2012. AlexNet est composé de cinq couches convolutives avec une combinaison de couches de *max-pooling*, de trois couches entièrement connectées et de deux couches de *dropout*. Le nombre total de paramètres dans cette architecture est d'environ 60 millions.

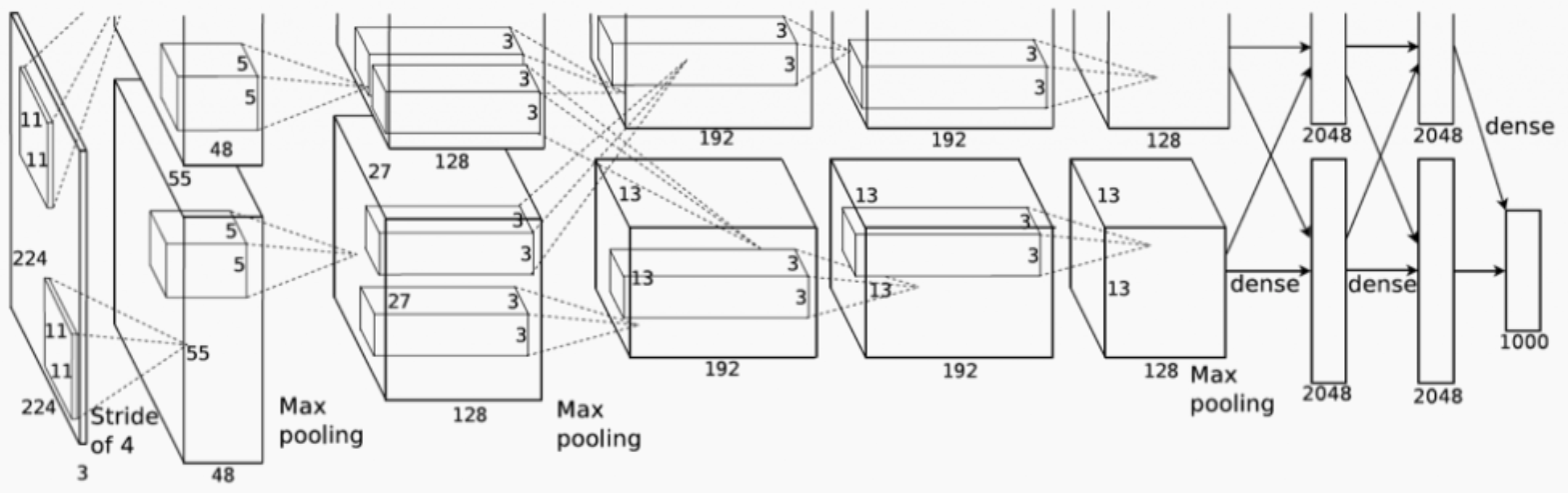

**Figure 2.2 –** Schéma de l'architecture du modèle AlexNet, issu de Krizhevsky et al. (2012), pour une image d'entrée de taille 224×224 pixels. Ici, deux cartes graphiques sont utilisées, une traite la partie haute de l'image et l'autre la partie basse.

— VGG (Simonyan et al. (2015)) : VGGNet est un réseau CNN à 16 couches comptant jusqu'à 95 millions de paramètres et entraîné sur plus d'un milliard d'images (1000 classes). Il prend des images d'entrée de taille 224×224 pixels. Il nécessite beaucoup de données d'entraînement, ce qui est la principale raison pour laquelle les architectures telles que AlexNet fonctionnent mieux pour la plupart des tâches de classification d'images où les images d'entrée ont une taille comprise entre 100×100 pixels et 350×350 pixels. Le modèle VGG est efficace et sert de base solide pour de nombreuses applications en raison de son applicabilité à de nombreuses tâches, notamment la détection d'objets. Ses représentations profondes des caractéristiques sont utilisées dans de nombreuses architectures de réseaux neuronaux telles que YOLO (Redmon et al. (2016)).

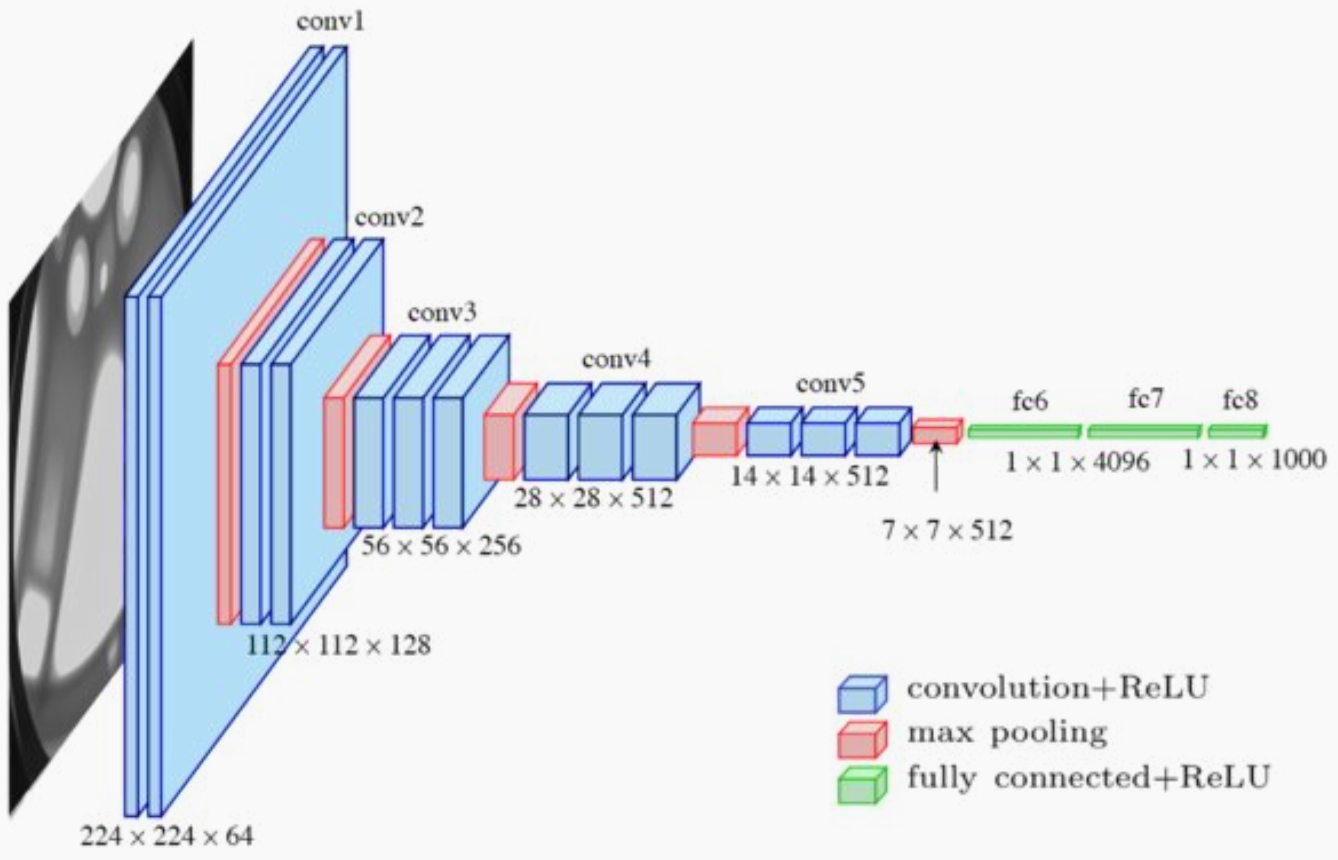

**Figure 2.3 –** Schéma de l'architecture du modèle VGG-16 (Simonyan et al. (2015)) pour une image d'entrée de taille 224×224 pixels. Schéma extrait de l'article de Ferguson et al. (2017).

— ResNet (He et al. (2016)) : ResNet a été développé dans le cadre de la compétition pour la tâche de classification de l'ILSVRC 2015. Le réseau contient des connexions résiduelles en plus des couches habituelles d'un CNN. Outre les tâches de classification d'images, ResNet a été utilisé avec succès pour résoudre des problèmes de traitement du langage naturel comme la complétion de phrases ou la compréhension automatique par l'équipe Microsoft Research Asia en 2016 et 2017 respectivement.

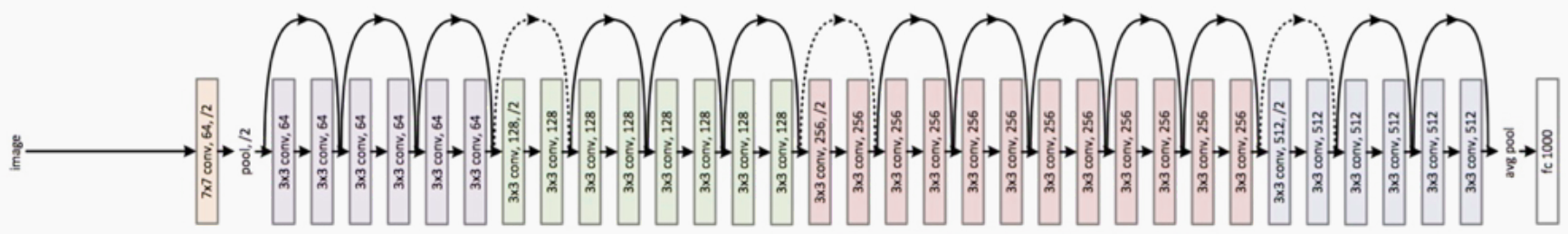

**Figure 2.4 –** Schéma de l'architecture du modèle ResNet-34, issu de He et al. (2016).

## Focus 2.2 – Convolution

### Définition

La couche de convolution est le bloc de base utilisé dans les réseaux dits convolutifs (CNN). Une couche de convolution permet de générer une nouvelle représentation de l'image d'entrée ou intermédiaire (en sortie de la couche précédente). Pour cela, elle possède un ou plusieurs filtres de convolution qui traitent une portion limitée, le champ réceptif, de l'image d'entrée. Chaque filtre est défini par un ensemble de poids appris durant l'entraînement du modèle et analyse une caractéristique de l'image d'entrée (caractéristique de couleur, de texture...). Pour cela, chaque filtre est appliqué à chaque pixel de l'image, calculant une nouvelle représentation pour chacun de ces pixels. Dans la plupart des cas, le filtre a une taille plus grande que 1 ce qui mène à utiliser du contexte, les pixels voisins, pour calculer la nouvelle représentation du pixel.

### Schéma d'une convolution 2D

La Figure 2.5 présente le schéma d'une convolution 2D avec X l'image d'entrée, W le filtre et Y la nouvelle représentation de l'image. Dans cet exemple, le filtre W a une taille 3×3, ce qui implique que, pour calculer la représentation d'un pixel, les valeurs de ses huit pixels voisins sont prises en compte.

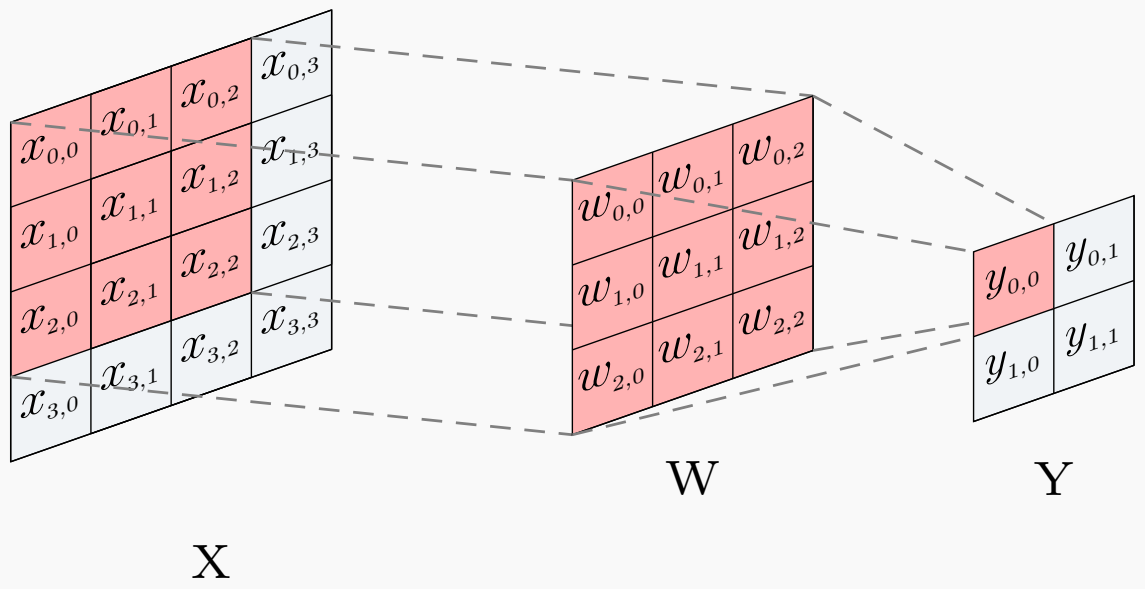

**Figure 2.5 –** Schéma d'une convolution 2D avec X l'image d'entrée, W le filtre et Y la nouvelle représentation de l'image.

## Focus 2.3 – Regroupement / Pooling

### Définition

Le *pooling* consiste à regrouper des représentations locales ou globales en résumant les valeurs de plusieurs pixels en une seule valeur unique. Les couches de regroupement réduisent les dimensions des données en combinant plusieurs entrées, et ainsi extraient les caractéristiques dominantes. Il s'agit d'opérations simples, non paramétriques, telles qu'un min (*min pooling*), un max (*max pooling*), une somme ou encore une moyenne (*average pooling*). Dans un CNN, ces couches permettent à la fois de réduire la taille des images intermédiaires en résumant les caractéristiques qu'elles contiennent, mais aussi d'avoir davantage de contexte puisque les pixels voisins sont regroupés.

**Focus 2.4 – Système R-CNN**

Le système R-CNN a été proposé par Girshick et al. (2014) et permet de réaliser de la détection d'objets sur des images à partir de propositions de régions. Un ensemble de régions (environ 2 000) est tout d'abord généré grâce à un algorithme de recherche sélective. Les caractéristiques importantes de chacune de ces régions sont ensuite extraites par un CNN. Enfin, un SVM linéaire prédit la classe de chaque région.
Bien que ce système ait été utilisé avec succès afin de détecter des objets dans les images de scènes naturelles, il reste très lent car prend en moyenne 47 secondes pour traiter une image.

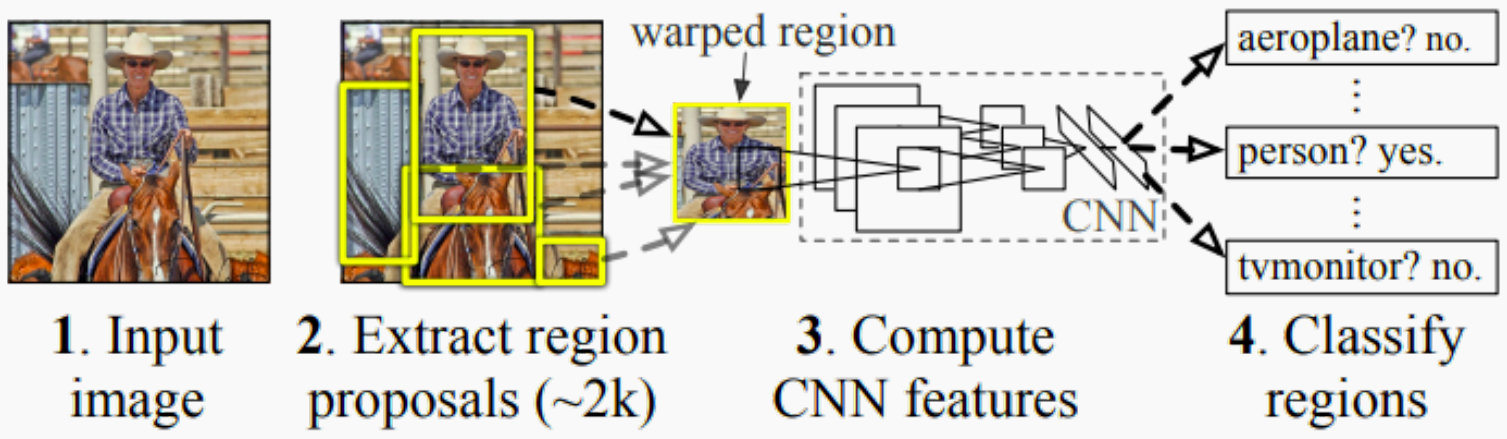

**Figure 2.6 –** Schéma du système R-CNN, issu de Girshick et al. (2014).

## RÉGRESSION DE BOÎTES ENGLOBANTES

La détection d'objets dans des images a également été réalisée à l'aide de modèles de prédiction des coordonnées des boîtes englobantes. Ces systèmes, fondés sur un algorithme de régression, ont été introduits pour la première fois par Erhan et al. (2014) qui ont proposé la méthode MultiBox. Celle-ci effectue une régression directe des positions des boîtes englobantes au lieu de s'appuyer sur des propositions d'objets. Ils utilisent un CNN comme régresseur pour directement prédire un nombre donné de coordonnées de boîtes et une confiance pour chaque boîte correspondant à sa probabilité de contenir un objet d'intérêt. Il permet de détecter un nombre variable d'objets superposés de la même classe, la taille des objets n'étant pas limitée. Mais, lorsqu'il est nécessaire de détecter un grand nombre d'objets, le nombre de paramètres du modèle augmente et une grande quantité de données est nécessaire pour l'apprentissage. YOLO et SSD peuvent être considérés comme des variantes de ce concept.

Redmon et al. (2016) ont proposé le modèle YOLO (*You Only Look Once*). L'objectif de YOLO était de détecter et de classifier les objets en un seul traitement et d'être plus rapide que les méthodes R-CNN. L'image est d'abord divisée en une grille régulière, puis chaque cellule de la grille prédit un nombre prédéfini de boîtes englobantes avec leurs confiances ainsi que les probabilités de classe grâce à un seul réseau neuronal. Les détections finales sont les boîtes ayant le score de confiance le plus important et la probabilité de la classe la plus élevée dans cette boîte. Ce système est présenté dans le Focus 2.5. De multiples méthodes ont ensuite étendu l'idée originelle de YOLO (Bochkovskiy et al., 2020 ; Redmon et al., 2017 ; 2018) mais très peu ont été appliquées aux images de documents, probablement pour la même raison que celle mentionnée ci-dessus : les images de documents contiennent trop d'objets à détecter.

D'un autre côté, LIU et al. (2016) ont proposé SSD (*Single Shot MultiBox Detector*). Le système discrétise l'espace de sortie des boîtes englobantes en un ensemble de boîtes prédéfinies par défaut avec différents rapports d'aspect et échelles par emplacement de carte de caractéristiques. Au moment de la prédiction, le réseau génère des scores reflétant la présence de chaque catégorie d'objets dans chaque boîte par défaut, et ajuste la boîte pour mieux correspondre à la forme de l'objet. De plus, le réseau combine les prédictions de plusieurs cartes de caractéristiques de différentes résolutions pour traiter des objets de différentes tailles. SSD est simple par rapport aux méthodes qui nécessitent des propositions d'objets, car il élimine la génération de propositions et les étapes ultérieures de ré-échantillonnage de pixels ou de caractéristiques. Cette méthode a montré de meilleurs résultats que Faster-RCNN et YOLO sur les données VOC 2017 tout en étant plus rapide.

Bien qu'elles aient montré de très bonnes performances sur des images de scènes naturelles où peu d'objets sont à détecter, ces méthodes sont moins adaptées au traitement d'images de documents où il y a souvent un grand nombre d'éléments à localiser. Certains travaux ont tout de même adapté ces systèmes aux images de documents.

Pour la détection de lignes de texte, les premières contributions ont été présentées par MOYSSET et al. (2016a) ; MOYSSET et al. (2016b). Dans MOYSSET et al. (2016a), les auteurs proposent une approche basée sur MultiBox pour détecter les boîtes englobantes des lignes de texte en utilisant des poids partagés afin de permettre au système d'être entraîné sur une quantité de données annotées réduite. Comme les modèles YOLO et SSD, les sorties sont attribuées à des régions locales de l'image. Cependant, le modèle est capable de prédire les objets dans sa région de support, ou en dehors.

MOYSSET et al. (2016b) proposent l'utilisation d'un réseau neuronal *Multi Dimensional Long Short Term Memory* (MDLSTM) combiné à des couches convolutives pour prédire une boîte englobante autour d'une ligne. Ils traitent la tâche de détection de lignes de texte comme étant un problème de régression, et prédisent les coordonnées des boîtes englobantes directement à partir des valeurs des pixels des images. Ils ont comparé deux stratégies de régression : prédire directement les boîtes englobantes et prédire séparément les points inférieurs gauche et supérieurs droit avant de les coupler. La seconde stratégie a montré une réelle amélioration pour la tâche de détection sur les documents du jeu de données Maurdor (OPARIN et al., 2014) mais est limitée aux lignes horizontales.

Malgré les améliorations apportées aux modèles de régression de boîtes, cette approche est toujours limitée aux éléments horizontaux et ne permet pas une détection précise des lignes de texte par exemple. C'est pour cela que les méthodes niveau pixel ont été proposées.

**Focus 2.5 – SYSTÈME YOLO**

Le système YOLO (*You Only Look Once*) a été proposé par REDMON et al. (2016) et permet de réaliser de la régression de boîtes englobantes d'objets sur des images. YOLO divise l'image d'entrée en une grille régulière. Chaque cellule de la grille prédit un nombre prédéfini de boîtes de délimitation et des scores de confiance pour chacune de ces boîtes. Enfin, les boîtes ayant les scores de confiance les plus élevés et les probabilités de classe les plus élevées dans ces boîtes sont considérées comme détections finales.
YOLO est beaucoup plus rapide que R-CNN. Il montre cependant plus de difficultés à détecter des objets proches et les petits objets.

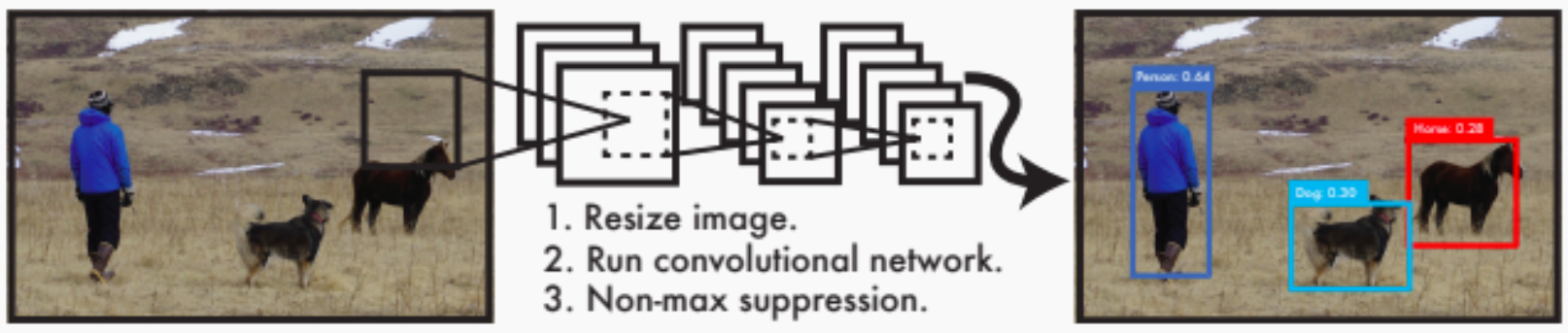

**Figure 2.7 –** Schéma du système YOLO, issu de Redmon et al. (2016).

### Détection niveau pixel

La détection d'objets au niveau pixel est actuellement l'approche la plus utilisée pour le traitement d'images de documents. De nombreux systèmes ont été proposés et c'est également dans ce cadre que se positionnent nos principaux travaux de recherche. La grande majorité de ces systèmes se base sur l'architecture Fully Convolutional Network (FCN), expliquée en détail dans le Focus 2.6, en fin de cette section.

Les premiers FCN proposés étaient composés d'une succession de convolutions et de couches de regroupement (*pooling*) permettant de résumer les caractéristiques importantes de l'image d'entrée. Ces systèmes, tels que le VGG (Simonyan et al., 2015) et le ResNet (He et al., 2016), étaient principalement utilisés pour les tâches de classification avec une classe unique en sortie.

Cependant, pour la segmentation sémantique ou la détection d'objets, il est nécessaire d'avoir également la position de la classe dans l'image, c'est-à-dire une classe pour chaque pixel de l'image d'entrée. Afin d'obtenir une telle sortie, Ciresan et al. (2012) ont entraîné un réseau utilisant une fenêtre glissante et prédisant une classe pour chaque pixel grâce à une région locale autour du pixel (un patch). Bien qu'il ait montré de très bonnes performances en gagnant notamment la compétition sur la segmentation de structures neuronales (ISBI 2012), le principal inconvénient de ce système est qu'il est très lent à traiter une image, le modèle étant appliqué à chaque patch. Pour pallier cela, Long et al. (2015) ont proposé un FCN pixel-à-pixel pour la tâche de segmentation sémantique d'images. Les auteurs ont proposé une modification de l'architecture FCN en ajoutant, après les couches standards de convolution et de regroupement (étape d'encodage), une étape de décodage constituée d'une succession de couches équivalentes à l'encodeur dans laquelle les opérations de regroupement sont remplacées par des opérations d'*upsampling*, augmentant la résolution de sortie. L'*upsampling* étant réalisé à l'aide de convolutions transposées (Focus 2.7). De plus, afin d'avoir une localisation plus précise, les auteurs proposent de combiner les caractéristiques calculées durant l'étape d'encodage à celles du décodage. Montrant de très bonnes performances et un temps d'inférence raisonnable, de nombreux autres travaux similaires ont vu le jour. Une modification de ce système a été proposée par Ronneberger et al. (2015) avec leur architecture U-Net. Les auteurs se sont concentrés sur le décodeur, l'encodeur étant comparable aux FCN que nous avons présentés plus tôt. Ils ont proposé d'utiliser des matrices de caractéristiques avec de nombreux canaux durant l'étape de décodage afin de propager davantage de contexte aux couches finales. Appliquée sur différentes tâches de segmentation d'images médicales, cette architecture a montré des gains importants de performances par rapport aux méthodes existantes.

Dans le domaine de la détection de texte dans des images de scènes naturelles, Zhang et al. (2016b) appliquent également un FCN. Tout d'abord, un FCN TextBlock est utilisé pour détecter les localisations approximatives des lignes de texte, qui sont ensuite extraites en tenant compte des informations locales des caractères. Enfin, un autre FCN est appliqué pour rejeter les fausses lignes de texte détectées.

Pour le traitement de documents, les FCN ont également été largement utilisés. En effet, l'intérêt porté à l'analyse des documents a été stimulé par les compétitions sur la détection des lignes de texte (Murdock et al., 2015), la détection des lignes de base (Diem et al., 2017 ; Diem et al., 2019) ou l'analyse de la mise en page (Antonacopoulos et al., 2011). La plupart de ces tâches ont été abordées au niveau pixel, et donc de nombreux systèmes de type FCN ont été développés. Ainsi, dhSegment (Ares Oliveira et al., 2018) a été proposé. Il s'agit d'un système complexe permettant de traiter des documents avec de nombreuses classes. C'est un réseau avec une architecture proche du U-Net où l'encodeur est pré-entraîné sur des images de scènes naturelles (ImageNet (Deng et al., 2009)). Dans la suite de cette thèse, nous comparons certains de nos modèles à dhSegment, c'est pourquoi son architecture est détaillée dans le Focus 2.9. Dans dhSegment, contrairement aux réseaux proposés par Long et al. (2015) et Ronneberger et al. (2015) où l'*upsampling* était réalisé à l'aide de convolutions transposées, la résolution de sortie est augmentée à l'aide d'interpolations bilinéaires, ce qui permet d'avoir moins de paramètres à apprendre. Cette méthode a obtenu de bons résultats sur diverses tâches de traitement de documents historiques, telles que l'analyse de mise en page ou l'extraction de lignes de base, avec peu de données d'entraînement. De plus, malgré un grand nombre de paramètres, le temps d'entraînement est considérablement réduit grâce à l'encodeur pré-entraîné. D'autres systèmes similaires ont ensuite été proposés, la principale différence entre leurs architectures étant la manière dont la résolution est augmentée dans le décodeur. Barakat et al. (2018) ont proposé un réseau entièrement convolutif pour détecter les lignes de texte. Leur proposition consiste à utiliser uniquement des cartes de caractéristiques de bas niveau pendant l'étape de décodage, en les sur-échantillonnant plusieurs fois, à l'aide de convolutions transposées, avant de les combiner. Cette architecture a donné de bons résultats sur des pages manuscrites arabes mais nécessite des images d'entrée binarisées. Mechi et al. (2019) ont présenté une architecture U-Net adaptative pour la détection de lignes de texte. Leur proposition est de réduire le nombre de filtres (deux fois moins) dans les convolutions de l'encodeur afin de diminuer la quantité de paramètres du modèle, et donc le temps d'inférence ainsi que le sur-apprentissage, leur quantité de données annotées étant faible.

Grüning et al. (2019) ont proposé un système plus complexe composé de deux étapes pour détecter les lignes de base dans les documents historiques. Tout d'abord, un réseau de neurones hiérarchique (ARU-Net) est appliqué pour détecter les lignes de texte. Cet ARU-Net est une version étendue de l'architecture U-Net (Ronneberger et al., 2015) : d'une part, un réseau d'attention spatiale est incorporé pour traiter les différentes tailles de caractères dans les pages ; d'autre part, des blocs résiduels sont ajoutés à l'architecture U-Net. Cela permet d'entraîner des réseaux neuronaux plus profonds tout en obtenant de meilleurs résultats. Ensuite, des traitements successifs sont appliqués pour regrouper les super-pixels afin de construire les lignes de base. Les auteurs ont montré que leur méthode était capable d'extraire des lignes de texte courbes. Cependant, de nombreuses étapes de post-traitement ont été introduites dans la seconde phase. Mechi et al. (2021) ont également présenté une méthode en deux étapes pour segmenter les lignes de texte dans des images de documents historiques arabes ou latins. Tout d'abord, un FCN est utilisé pour segmenter la zone centrale du texte. La seconde étape affine les résultats du FCN. Elle est basée sur une version modifiée du RLSA pour extraire les lignes complètes du texte (y compris les composantes ascendantes et descendantes). Des évaluations quantitatives et qualitatives sont rapportées sur un grand nombre d'images de documents arabes et latins collectés à partir des archives nationales tunisiennes ainsi que d'autres ensembles de données de référence. Cependant, ce système nécessite une binarisation de l'image d'entrée. Dans Tensmeyer et al. (2017), les auteurs présentent PageNet, un système mis au point pour identifier les pages dans des images de documents. Les pages détectées sont ensuite extraites, ce qui permet de supprimer le bruit induit par la numérisation des pages, et différents traitements d'analyse de la mise en page peuvent être appliqués. Dans PageNet, un réseau entièrement convolutif obtient une segmentation par pixel post-traitée afin d'extraire une région quadrilatérale. Celui-ci traite l'image d'entrée à quatre résolutions. Le système est évalué sur différents jeux de données et les auteurs montrent que PageNet peut segmenter des documents superposés à d'autres documents.

Enfin, Yang et al. (2017) ont conçu un réseau multimodal entièrement convolutif pour l'analyse de la mise en page de documents. Ils tirent parti du contenu textuel ainsi que de l'apparence visuelle pour extraire les structures sémantiques des images de documents. Cette méthode a montré des scores élevés d'Intersection-over-Union (IoU) (voir le Focus 3.2) mais nécessite des annotations de données plus complexes. En effet, pour chaque image de document, une image étiquetée pixel par pixel ainsi que son contenu textuel sont nécessaires. Ils utilisent des convolutions dilatées dans l'encodeur afin d'avoir une information contextuelle plus large et des résultats plus précis. Puisque dans la suite de cette thèse nous comparons certains de nos modèles à ce système, nous détaillons son architecture dans le Focus 2.10. Dans leurs travaux, Renton et al. (2018) ont également démontré les avantages d'utiliser de telles convolutions, détaillées dans le Focus 2.8, par rapport à des convolutions standards. Leur réseau entièrement convolutif est composé de convolutions dilatées successives qui augmentent le champ réceptif. Elles sont suivies d'une dernière convolution standard qui produit les images étiquetées. Dans notre méthode, nous tirons également profit de ces convolutions dilatées afin d'avoir un champ réceptif assez grand pour détecter correctement les objets.

### Focus 2.6 – Architecture FCN

**Définition**

Un réseau entièrement convolutif (FCN) est une extension d'un réseau neuronal convolutif (CNN) qui ne contient aucune couche dense et accepte des entrées de tailles variables. Il permet de faire de la prédiction spatiale dense, au niveau pixel, de manière rapide et précise. Pour faire de la prédiction dense, il est souvent composé d'un encodeur, résumant les caractéristiques importantes de l'image d'entrée, et d'un décodeur, augmentant la résolution des cartes de caractéristiques et prédisant des probabilités de classe pour chaque pixel d'entrée.

**Avantages**

— La suppression des couches denses d'un CNN permet de travailler avec des tailles d'entrée variables car les couches convolutives ne nécessitent pas un nombre fixe d'entrées.

— Éviter les couches denses réduit fortement le nombre de paramètres.

— Les FCN sont capables de conserver l'information spatiale et de produire une description spatiale de l'image d'entrée.

**Inconvénients**

— L'utilisation d'un réseau de neurones convolutif induit l'utilisation de couches de regroupement (*pooling*), qui réduisent la résolution d'entrée dans le but d'augmenter le champ réceptif sans augmenter le nombre de paramètres. Pour avoir un étiquetage au niveau des pixels d'une image d'entrée, la résolution de sortie du réseau doit être augmentée soit à l'aide d'une interpolation, d'une convolution transposée (Focus 2.7), d'une opération d'*unpooling* ou encore d'une convolution dilatée (Focus 2.8).

**Système de type FCN**

Un des premiers systèmes de type FCN proposé pour la détection niveau pixel est le U-Net (Ronneberger et al. (2015)). Il a montré de très bonnes performances sur différentes tâches de segmentation d'images biomédicales avec très peu de données annotées. Sur la Figure 2.8, la partie gauche constitue l'encodeur, composé de couches de convolutions et de *max pooling*. La partie droite est le décodeur, constitué de convolutions standards et transposées.

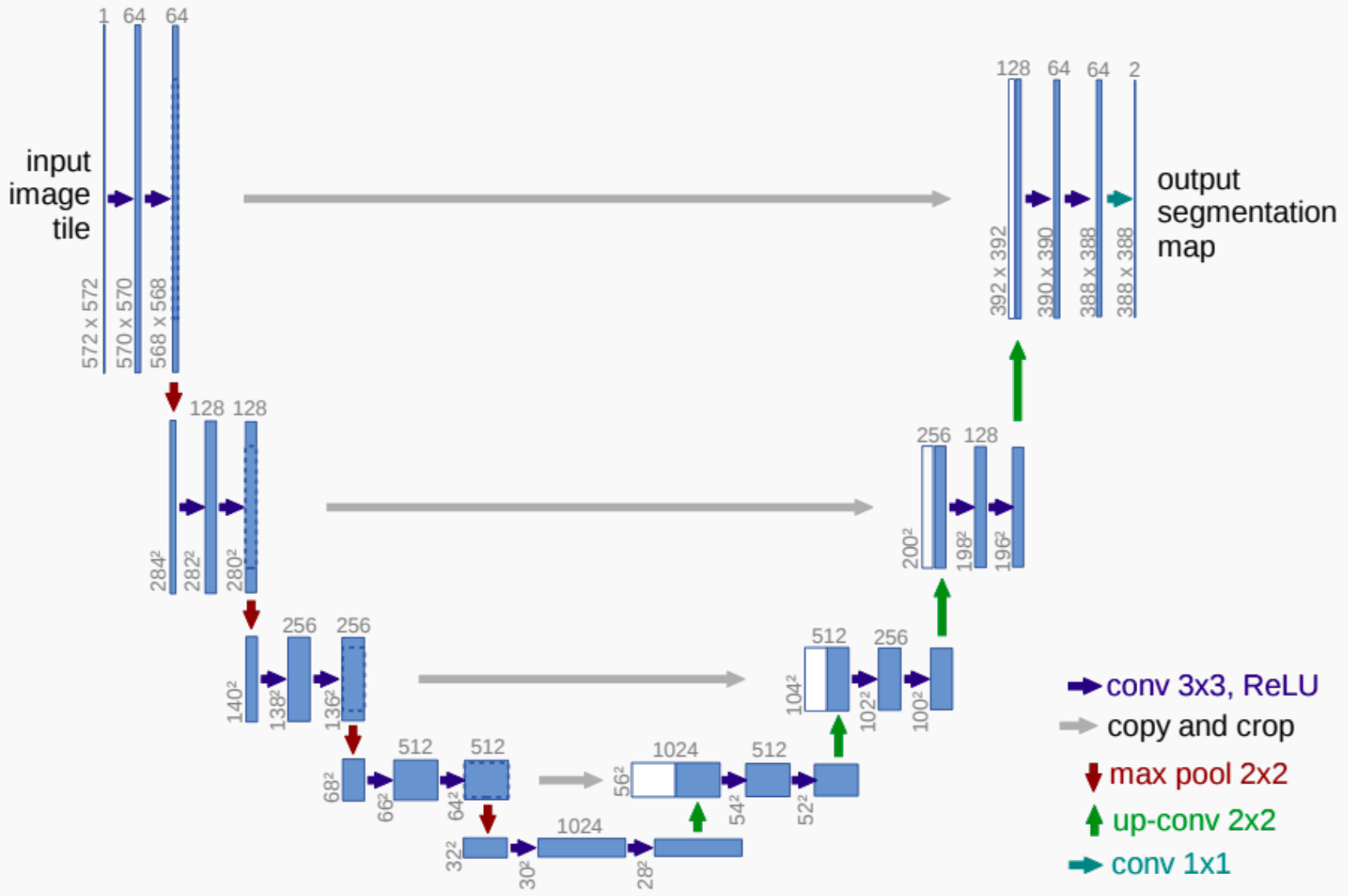

**Figure 2.8 –** Schéma de l'architecture du modèle U-Net, issu de Ronneberger et al. (2015), pour une image d'entrée de taille 572×572 pixels.

## Focus 2.7 – Convolution transposée

### Définition

La couche de convolution transposée est utilisée pour inverser le sous-échantillonnage induit par les couches de convolution standard ou de regroupement utilisées dans les réseaux convolutifs (Long et al., 2015 ; Ronneberger et al., 2015). Le principe est d'avoir la couche inverse d'une couche de convolution standard. Cette couche permet d'avoir une sortie de plus grande résolution en représentant la valeur d'un pixel d'entrée sur plusieurs pixels de sortie. Cette couche est souvent utilisée dans les réseaux suivant l'architecture encodeur-décodeur où la sortie du réseau a la même taille que l'image d'entrée.

### Avantages

— Les filtres de la couche de convolution transposée doivent être entraînés ce qui permet au réseau d'être plus expressif.

### Inconvénients

— Le réseau est plus profond et comporte plus de paramètres qu'un réseau contenant uniquement des opérations d'*upsampling* sans paramètres entraînés.

### Schéma d'une convolution transposée 2D

La Figure 2.9 présente le schéma d'une convolution transposée 2D avec X l'image d'entrée, W le filtre et Y la nouvelle représentation de l'image.

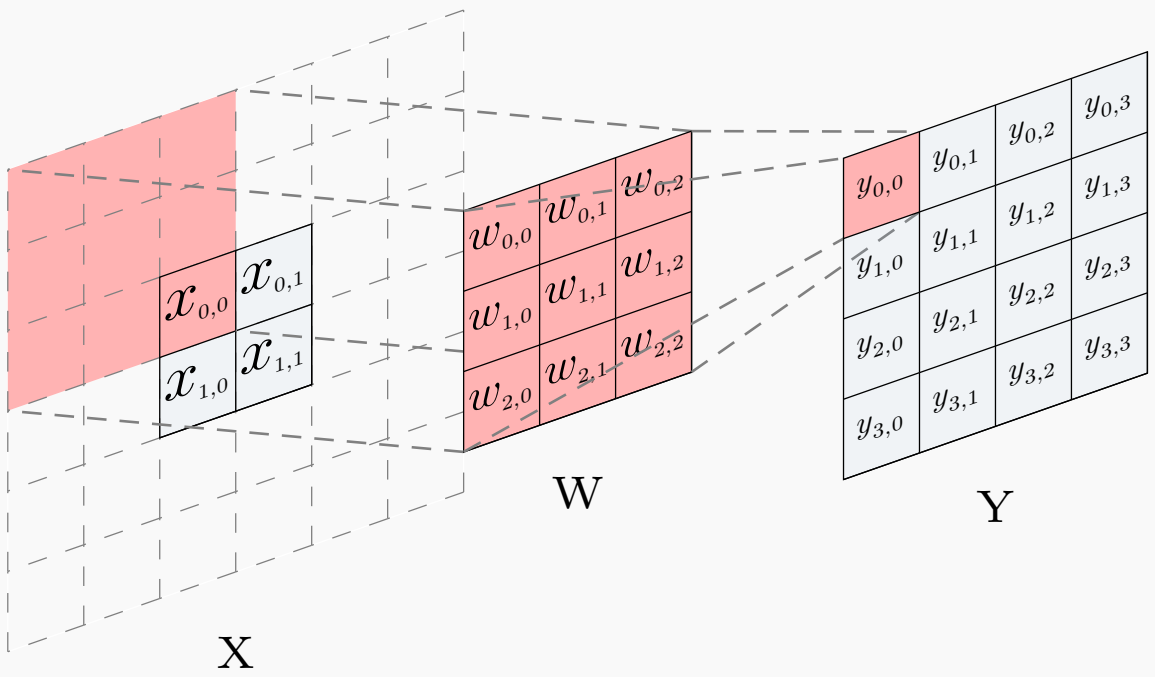

**Figure 2.9 –** Schéma d'une convolution transposée 2D avec X l'image d'entrée, W le filtre et Y la nouvelle représentation de l'image.

## Focus 2.8 – Convolution dilatée

### Définition

Une convolution dilatée suit le principe de base d'une convolution standard, mais calcule les nouvelles représentations sur une plus grande fenêtre. Les pixels considérés pour le calcul de la nouvelle représentation d'un pixel ne sont plus ses voisins directs mais des voisins plus éloignés, l'écart étant défini par le taux de dilatation.

### Avantages

- L'utilisation de convolutions dilatées permet d'avoir un champ réceptif plus grand, la nouvelle représentation d'un pixel considérant davantage de contexte, sans augmenter le nombre de paramètres.
- Elle est souvent utilisée à la place des couches de regroupement, ce qui permet de perdre moins d'informations qu'avec une fonction de min ou max.

### Schéma d'une convolution dilatée 2D

La Figure 2.10 présente le schéma d'une convolution dilatée 2D avec un taux de dilatation de 2 et X l'image d'entrée, W le filtre et Y la nouvelle représentation de l'image.

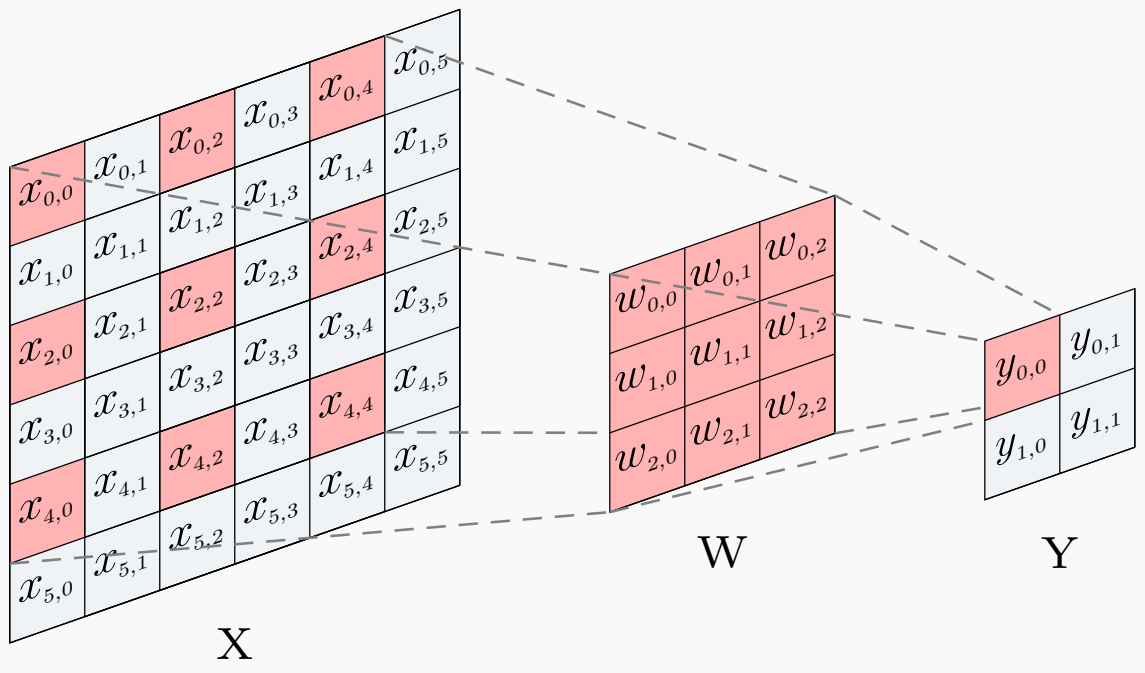

**Figure 2.10 –** Schéma d'une convolution dilatée 2D avec un taux de dilatation de 2 et X l'image d'entrée, W le filtre et Y la nouvelle représentation de l'image.

## Focus 2.9 – Système dhSegment

dhSegment (Ares Oliveira et al., 2018) est un des systèmes de référence pour les tâches d'analyse d'images de documents historiques. Il possède plusieurs avantages comme le fait de pouvoir être entraîné avec peu de données d'entraînement et un temps d'entraînement réduit. De plus, le code permettant d'entraîner et de tester le modèle est open-source [a].

C'est un modèle profond puisqu'il possède jusqu'à 2048 cartes de caractéristiques et suit l'architecture encodeur-décodeur. Dans un premier temps, l'image d'entrée est traitée par l'encodeur qui va résumer les caractéristiques importantes de l'image dans une matrice de caractéristiques. Cette matrice est ensuite traitée par le décodeur qui va générer une carte de probabilités de même taille que l'image d'entrée. Enfin, une étape de post-traitement est réalisée afin notamment de seuiller les probabilités des pixels et de supprimer les petites composantes connexes.

### Encodeur

L'encodeur est principalement constitué d'un CNN pré-entraîné sur des images de scènes naturelles de la base ImageNet (Deng et al., 2009) et représenté à gauche sur la Figure 2.11. Ce CNN pré-entraîné suit l'architecture du réseau ResNet-50 (He et al., 2016) mais a été légèrement modifié afin de réduire le nombre de paramètres, et donc la mémoire requise pendant l'entraînement. Il est également possible de remplacer ce ResNet-50 par un VGG-16 (Simonyan et al., 2015) ou un U-Net (Ronneberger et al., 2015).

Cette partie pré-entrainée présente l'avantage de réduire considérablement le nombre de paramètres à apprendre. En effet, le réseau possède 32,8 millions de paramètres au total dont la plupart proviennent du CNN. Ainsi, seuls 9,36 millions de paramètres restent à entraîner. Cela permet au réseau d'apprendre rapidement et correctement sur un nombre restreint de données annotées.

### Décodeur

Le décodeur est standard et consiste en une succession de cinq blocs de déconvolution composés d'une couche de convolution standard et d'une couche d'*upscaling*, et d'une couche finale de convolution afin de générer une carte de probabilités. Cette partie est entièrement apprise sur les données d'entrée.

### Post-traitement

En sortie du décodeur, nous disposons, pour chaque pixel, des probabilités d'appartenir aux différentes classes définies. Différentes techniques d'agrégation des résultats au niveau pixel sont possibles afin de détecter les objets pour la tâche considérée. Quatre principales techniques ont été implémentées et sont disponibles :

- Seuillage : permet d'assigner une classe aux pixels ayant une probabilité supérieure à un seuil prédéfini ;
- Opérations de morphologie mathématique : opérations d'érosion, de dilatation, d'ouverture et de fermeture afin de créer des objets plus plausibles ;
- Analyse des composantes connexes : permet de filtrer les petites composantes connexes restantes après l'étape de seuillage ;
- Vectorisation des objets : transforme les régions détectées en un ensemble de coordonnées.

Schéma de l'architecture de dhSegment

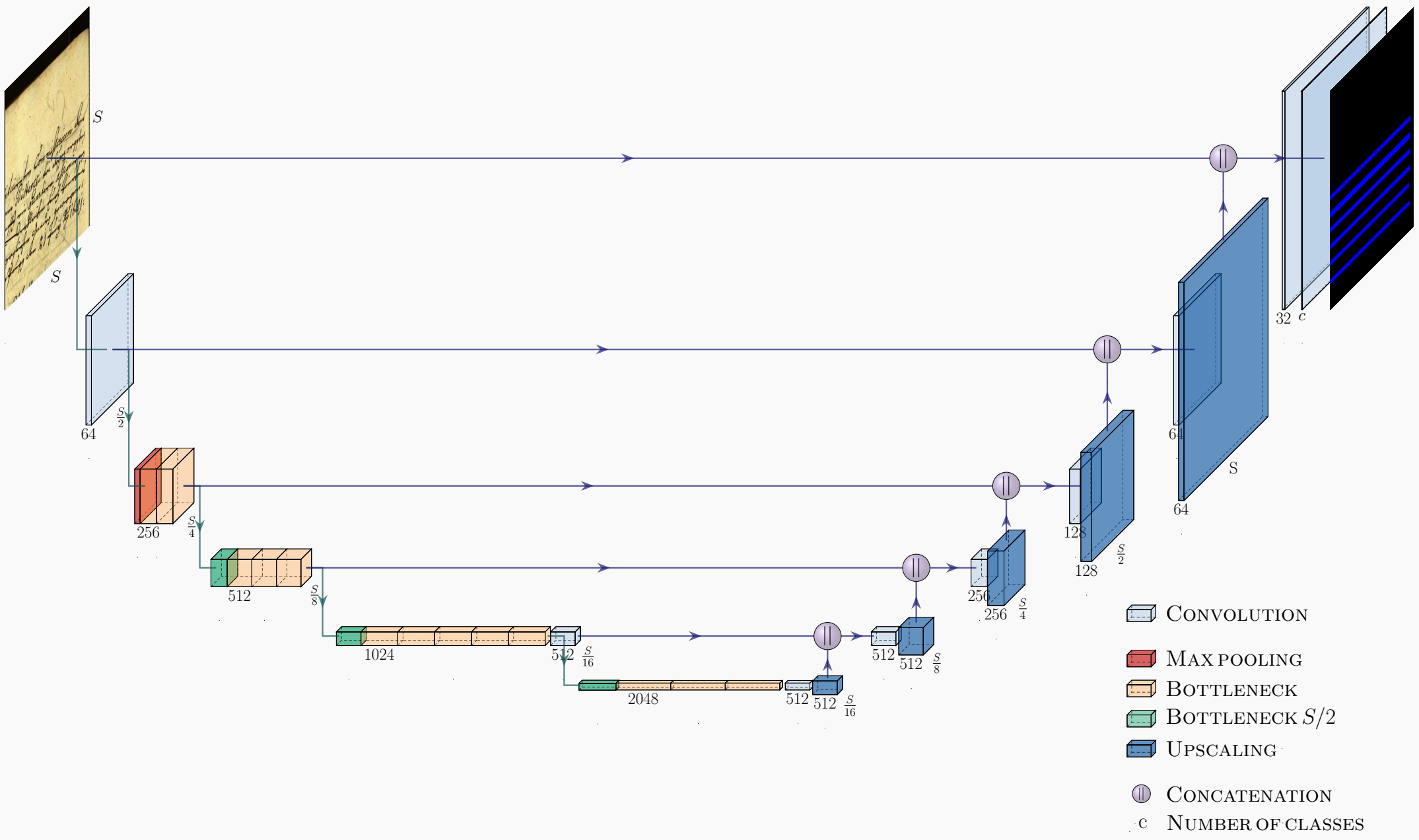

**Figure 2.11 –** Schéma de l'architecture du modèle dhSegment (Ares Oliveira et al. (2018)).

*a.* https://github.com/dhlab-epfl/dhSegment

## Focus 2.10 – Système de Yang et al.

Yang et al. (2017) ont proposé un réseau multimodal permettant de segmenter des documents en se basant sur le contenu visuel et textuel de ceux-ci. L'utilisation des textes permet d'assigner des classes spécifiques aux régions de texte en fonction de leur rôle dans le document. Ainsi, dans l'article original, les classes considérées sont les suivantes : fond, image, tableau, paragraphe, titre, liste et légende. Le système a montré de bonnes performances sur des ensembles de données synthétiques et réelles d'images de documents modernes. De plus, le code permettant d'entraîner un modèle est open-source [a].

Le modèle de Yang est un réseau multimodal entièrement convolutif (FCN) dont l'architecture est présentée sur la Figure 2.12. La base de ce modèle suit une architecture encodeur-décodeur et est constituée de quatre modules :

— Un encodeur ;
— Un décodeur ;
— Un décodeur auxiliaire ;
— Un pont (intégration du contenu textuel).

### Encodeur

L'encodeur est constitué de quatre blocs dilatés, chaque bloc comportant cinq couches de convolutions dilatées, de taux de dilatation 1, 2, 4, 8 et 16, exécutées en parallèle. L'avantage d'utiliser de telles convolutions est que le champ réceptif est plus grand, ce qui permet au modèle d'avoir davantage de contexte par rapport à une convolution standard.

DÉCODEURS

Les deux décodeurs ont la même architecture avec trois blocs contenant une couche de convolution suivie par une couche d'*unpooling*. Le premier décodeur est standard et vise à produire une carte de probabilités. Le décodeur auxiliaire est, quant à lui, utilisé uniquement durant l'entraînement et cherche à reconstruire l'image d'entrée. Il a été montré qu'une branche auxiliaire de reconstruction aide à générer de meilleures représentations de l'image d'entrée, et donc améliore les performances de la tâche principale, ici, la tâche de segmentation (ZHANG et al., 2016a).

CONTENU TEXTUEL

L'information textuelle est extraite grâce à un algorithme de reconnaissance (Optical Character Recognition (OCR) ou HTR) de textes de la manière suivante. L'algorithme de reconnaissance est appliqué au document. Pour chaque phrase extraite du document, un *embedding* moyen est calculé à partir des *embeddings* des mots de cette phrase. Enfin, une carte de caractéristiques est construite à partir de ces *embeddings* : pour chaque phrase, les pixels du document initial lui appartenant prennent la valeur de cet *embedding*. Les pixels n'appartenant à aucune phrase prennent la valeur 0. Enfin, cette carte est concaténée à la carte de caractéristiques visuelles avant la dernière convolution du décodeur principal.

SCHÉMA DE L'ARCHITECTURE DU SYSTÈME DE YANG ET AL.

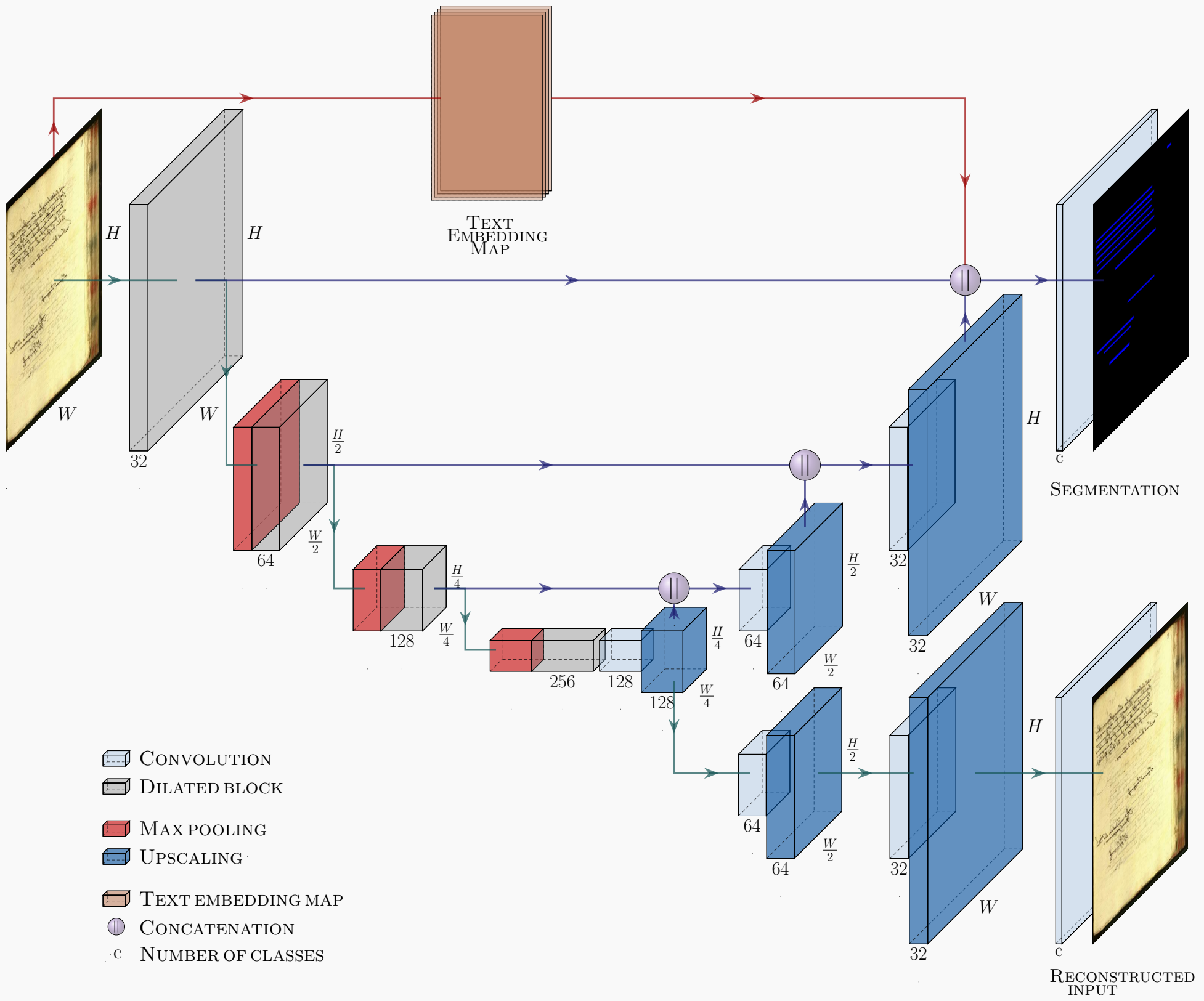

**Figure 2.12 –** Schéma de l'architecture du modèle de YANG et al. (2017).

*a.* http://personal.psu.edu/xuy111/projects/cvpr2017_doc.html

### MODÈLES SÉQUENTIELS À BASE DE TRANSFORMERS

Avant le développement des modèles à attention et des systèmes Transformers, les tâches de traitement du langage et notamment de traduction étaient réalisées à l'aide de réseaux encodeurs-décodeurs récurrents. L'encodeur est utilisé pour traiter la phrase d'entrée entière et l'encoder dans un vecteur de contexte unique. Les couches du décodeur produisent ensuite, à partir du vecteur de contexte, les mots de la phrase les uns après les autres. Le principal inconvénient de cette approche provient du traitement de la phrase d'entrée qui est résumée dans un unique vecteur de taille fixe. En effet, CHO et al. (2014) ont démontré que la performance du modèle encodeur-décodeur se dégrade rapidement lorsque la longueur de la phrase d'entrée augmente. Un autre problème est que le modèle n'a aucun moyen de donner plus d'importance à certains des mots en entrée par rapport à d'autres lors de la traduction de la phrase. C'est pour résoudre ces problèmes que l'attention (voir le Focus 2.12) a été introduite par BAHDANAU et al. (2015). Celle-ci permet de considérer tous les mots de la phrase d'entrée dans le vecteur de contexte, mais également d'accorder une importance relative à chacun d'entre eux. Ainsi, lorsque le modèle génère une phrase, il recherche un ensemble de positions dans les états cachés de l'encodeur, dans lesquels les informations les plus pertinentes sont disponibles.

Dans cette même optique, des systèmes à base de réseaux Transformers ont été proposés récemment afin de réaliser des tâches de détection en tenant compte de la séquentialité entre les éléments prédits. L'architecture de ces systèmes est présentée dans le Focus 2.11. Il s'agit de modèles reposants sur le même mécanisme d'attention, qui sélectionne les caractéristiques pertinentes à chaque itération du processus de prédiction. Le système met en œuvre une seconde attention qui tient compte des éléments précédemment prédits en sortie, pour agir comme un modèle de langage. Les premiers systèmes ont été principalement conçus pour le traitement automatique des langues sans utiliser ni récurrence ni convolutions.

Le premier système à base de Transformers (VASWANI et al., 2017) a été établi afin de résoudre plus efficacement la tâche de traduction de texte. Les auteurs ont proposé une architecture composée d'un encodeur suivi d'un décodeur, qui génère une séquence de sortie, un élément à la fois. Le modèle est auto-régressif : il intègre les éléments prédits précédemment comme entrée supplémentaire lors de la prédiction de l'élément suivant. L'encodeur extrait les caractéristiques des données d'entrée grâce à un mécanisme d'attention qui permet de considérer le contexte, ici, l'ensemble des mots de la séquence d'entrée. Cet encodeur est constitué de six blocs successifs identiques, composés de deux principaux éléments : une couche d'auto-attention et un réseau dit entièrement connecté (*feed-forward*). L'auto-attention permet de représenter l'interdépendance des mots de la séquence en entrée. Le décodeur permet de modéliser le langage de sortie. Il est également composé de six blocs successifs identiques, chacun contenant une couche d'auto-attention, un réseau entièrement connecté et une couche d'attention dite d'attention croisée. Cette dernière permet au décodeur de réaliser l'attention entre la séquence d'entrée et celle de sortie. Tous les réseaux entièrement connectés du modèle contiennent deux couches linéaires. De plus, les séquences en entrée et en sortie sont additionnées à un encodage de position, détaillé dans le Focus 2.13, avant d'être respectivement traitées par les encodeur et décodeur. Ces encodages de position permettent de garder l'ordre de la séquence durant l'ensemble des traitements. Le Transformer a rapidement été largement utilisé car il a permis de remplacer les couches récurrentes, jusqu'alors utilisées, par des couches d'attention tout en conservant des performances similaires. De plus, les couches récurrentes jusqu'ici utilisées empêchaient la parallélisation des calculs durant la phase d'entraînement. Cette récurrence ayant été remplacée par ces fameuses couches Transformer non récurrentes, entraînées par une stratégie dite de *teacher forcing*, les calculs peuvent être parallélisés et le temps d'entraînement fortement réduit. Cette architecture est détaillée dans le Focus 2.11.

Au vu des résultats obtenus par ces systèmes, certains travaux les ont adaptés à des tâches de vision. Ainsi, les Vision Transformers (ViT) ont été proposés. Les premiers travaux introduisant les ViT ont été présentés par Dosovitskiy et al. (2021) et sont détaillés dans le Focus 2.14. Ils interprètent une image en entrée de l'encodeur comme étant une séquence de patchs. Ainsi, la représentation vectorielle d'un caractère dans une tâche de traduction est ici remplacée par les valeurs des pixels d'un patch de l'image d'entrée mis à plat. Cette fois, l'encodage de position correspond à la position du patch dans l'image. Puisqu'il est appliqué à la tâche de classification d'images, qui ne nécessite pas de sortie séquentielle, seul l'encodeur Transformer est intégré au système. Ensuite, les auteurs utilisent un simple Multi-Layer Perceptron (MLP) chargé de prédire la classe de l'image. Ce système a obtenu des performances à l'état de l'art sur différents ensembles de classification d'images. Cependant, le système nécessite un pré-entraînement sur un nombre imposant de données, 303 millions d'images pour leur meilleur modèle.

Plusieurs approches ont également été présentées afin d'appliquer les Transformers à la détection d'objets. Dans le cadre d'images de scènes naturelles, DETR (*DEtection TRansformer*) a été proposé par Carion et al. (2020). Il s'agit d'un système hybride qui combine un encodeur CNN suivi d'un encodeur et d'un décodeur Transformer produisant un ensemble de boîtes englobantes. Le modèle est entrainé à prédire un nombre fixe de boîtes englobantes ainsi que leurs classes. Leur modèle obtient des résultats semblables à Faster R-CNN sur les images de scènes naturelles du jeu de données COCO [1], tout en obtenant de meilleurs résultats sur les grands objets grâce à l'auto-attention. Il ne tire cependant pas profit de la capacité de prédiction séquentielle permise par les Transformers. Chen et al. (2022) ont ensuite proposé Pix2Seq, présenté dans le Focus 2.15, afin de traiter la détection de manière séquentielle en prédisant, pour chaque objet, une séquence de coordonnées suivie de la classe de l'objet. Les auteurs comparent différents encodeurs à base de convolutions et de Transformers, suivis par un décodeur Transformer standard. Pix2Seq obtient des performances à l'état de l'art sur l'ensemble de données de référence COCO en obtenant des valeurs de précision moyenne (Average Precision (AP), voir le Focus 3.4) supérieures à celles obtenues par Faster-RCNN (Ren et al., 2015) et DETR (Carion et al., 2020) tout en nécessitant moins de paramètres. En effet, les résultats montrent que, pour tous les encodeurs, la détection est meilleure par rapport aux systèmes Faster-RCNN et DETR avec des encodeurs comparables. Les résultats

1. https://cocodataset.org/

montrent également qu'utiliser un encodeur Transformer est préférable. Cependant, davantage de données d'entraînement sont nécessaires puisque le modèle comporte beaucoup plus de paramètres.

Toujours dans le domaine de la détection dans des images naturelles, certains travaux ont été proposés afin de faire de la prédiction dense en augmentant la sortie du Transformer, et donc d'avoir une sortie de même taille que l'image d'entrée. Ainsi, ZHENG et al. (2020) ont proposé un ViT (appelé SETR) où l'encodeur Transformer est suivi d'un décodeur composé de convolutions réalisant l'augmentation d'échelle (*upsampling*). Ils ont obtenu les meilleurs résultats sur différentes bases de segmentation d'images. BISWAS et al. (2022) ont proposé un modèle hybride CNN-Transformer, très comparable à SETR mais incluant un encodeur CNN avant l'encodeur Transformer. Bien que ces méthodes aient montré des gains de performances par rapport aux systèmes existants, elles ne tirent pas pleinement profit des Transformers qui permettent d'avoir des sorties séquentielles et structurées. De plus, dans le domaine des Transformers, il n'y a pas, à notre connaissance, de travaux permettant de traiter la tâche de détection d'objets de manière séquentielle dans les images de documents. Quelques rares travaux ont appliqué les Transformers aux images de documents, principalement pour la reconnaissance de caractères niveau paragraphe ou page. Ainsi, dans les travaux de COQUENET et al. (2022) et SINGH et al. (2021), les auteurs ont proposé des modèles hybrides combinant un encodeur CNN et un décodeur Transformer afin de prédire séquentiellement les caractères du texte d'un paragraphe ou document. Le système proposé par COQUENET et al. (2022) fournit également une structuration des résultats en générant des tags de mise en page dans la séquence des caractères reconnus. Ce système est le premier à résoudre la tâche de reconnaissance de texte pleine page sans segmentation. Il a obtenu des performances de même ordre que les systèmes à l'état de l'art travaillant au niveau ligne. De leur côté, ROUHOU et al. (2022) utilisent un modèle hybride avec un encodeur CNN suivi d'un encodeur et décodeur Transformer pour traiter la tâche de reconnaissance d'entités nommées. Leur approche consiste à créer une architecture qui reconnaît les textes et les entités nommées à partir d'images de paragraphes. Ils utilisent des labels dits "visuels" correspondant aux caractères du texte présent dans les images ainsi que des labels dits "contextuels" correspondant aux entités nommées. Enfin, KIM et al. (2022) ont proposé DONUT, un modèle de compréhension de documents sans OCR composé d'un encodeur et décodeur Transformer. DONUT obtient de très bons résultats en termes de temps d'exécution et de précision sur diverses tâches telles que la classification de documents, l'extraction d'informations et le *Visual Question Answering*. Il nécessite cependant un important pré-entraînement sur des milliers de documents synthétiques.

**Focus 2.11 – ARCHITECTURE TRANSFORMER**

DÉFINITION

Un modèle à base de Transformer est un modèle permettant de réaliser un traitement séquence-à-séquence. Il s'agit d'un modèle auto-régressif prédisant séquentiellement les éléments et utilisant les éléments de la séquence d'entrée ainsi que les éléments prédits précédemment en sortie. Ce modèle repose sur un mécanisme d'attention (présenté dans le Focus 2.12), qui permet de représenter les données en utilisant le contexte et, notamment, les interdépendances entre les éléments des séquences d'entrée et de sortie.
Les modèles Transformers initialement proposés suivent une architecture encodeur-décodeur où l'encodeur génère une représentation de la séquence en entrée incluant l'interdépendance des éléments de cette séquence ainsi que leurs positions dans la séquence. Le décodeur génère une séquence de sortie grâce à la séquence d'entrée encodée et les éléments précédemment prédits.

AVANTAGES

— Dans un modèle à base de Transformer, les couches récurrentes d'un réseau à attention ont été remplacées par des couches non récurrentes pour réaliser cette attention, ce qui conduit à des temps d'entraînement réduits tout en conservant des performances similaires.
— Par rapport à un réseau récurrent, ce modèle permet de mieux représenter les dépendances entre les éléments de la séquence d'entrée grâce au mécanisme d'attention, notamment pour des séquences longues, tout en conservant un temps de traitement raisonnable.

SYSTÈME TRANSFORMER

Le premier système à base de Transformer a été proposé pour la tâche de traduction de texte (VASWANI et al., 2017). Il s'agit d'un modèle encodeur-décodeur entièrement basé sur l'attention dont l'architecture est présentée sur la Figure 2.13. Il a dépassé les résultats à l'état de l'art sur des tâches de traduction anglais-allemand et anglais-français.

ENCODEUR

Sur la Figure 2.13, la partie de gauche compose l'encodeur qui traite la séquence d'entrée. Dans l'implémentation originale, l'encodeur comporte une couche d'*embedding* de la séquence puis six blocs d'encodage Transformer. De plus, un encodage de position est additionné à la représentation de la séquence d'entrée avant les couches d'encodage. Celui-ci est réalisé à l'aide des fonctions cosinus et sinus comme détaillé dans le Focus 2.13. Les couches dites de *Multi-Head Attention* calculent les vecteurs d'auto-attention (voir Focus 2.12) et sont suivies d'un réseau entièrement connecté composé de deux couches linéaires.

DÉCODEUR

La partie de droite de la Figure 2.13 présente le décodeur. Il comporte une couche d'*embedding* de la séquence de sortie suivie de six blocs de décodage Transformer. Le même encodage de position utilisé dans l'encodeur est appliqué sur la séquence partielle de sortie courante, avant les couches de Transformer. Enfin, les couches *Multi-Head Attention* et le réseau entièrement connecté sont similaires à ceux de l'encodeur. La seule différence concerne la seconde couche d'attention qui prend en entrée la séquence d'entrée encodée ainsi que la séquence de sortie encodée pour réaliser l'attention croisée.

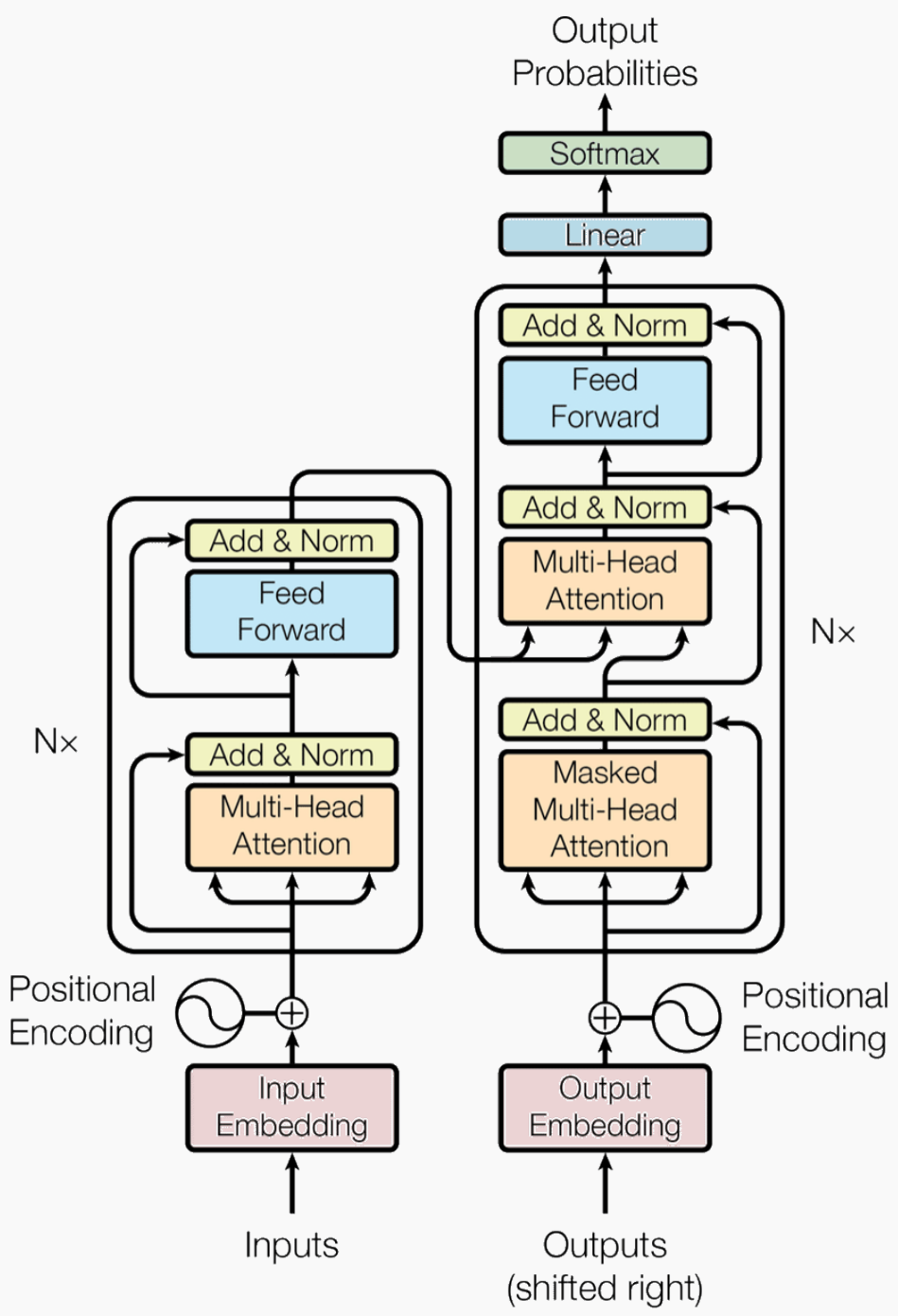

**Figure 2.13 –** Schéma de l'architecture du modèle Transformer original, issu de Vaswani et al. (2017).

## Focus 2.12 – Attention

### Définition

Le concept d'attention permet de considérer la corrélation entre les éléments de deux séquences grâce à des coefficients d'attention calculés entre chaque élément de chaque séquence. Une fonction d'attention peut être décrite comme la mise en correspondance d'une requête ($q$) et d'un ensemble de paires clé-valeur ($k$-$v$) avec une sortie. La sortie est calculée comme une somme pondérée des valeurs, le poids attribué à chaque valeur étant calculé par une fonction de compatibilité de la requête avec la clé correspondante.

Lorsque l'attention est réalisée sur une unique séquence, celle-ci est appelée auto-attention. L'attention croisée fait elle référence au mécanisme d'attention standard, appliqué sur deux séquences distinctes.

Dans le cas du traitement de la langue, le mécanisme d'attention permet de déterminer les mots sur lesquels le modèle doit porter le plus d'attention pour traiter la séquence.

### Auto-attention

L'auto-attention, appelée *self-attention*, correspond au mécanisme d'attention appliqué à une seule séquence. Elle détermine donc l'interdépendance (ou l'auto-corrélation) des éléments d'une même séquence entre-eux afin de lui associer une représentation pertinente.

### Attention multi-têtes

Dans la *multi-head attention*, le calcul d'attention est réalisé en parallèle par plusieurs blocs d'attention différents. Cela permet au modèle de considérer des informations provenant de différents sous-espaces de représentations à différentes positions. Dans le cadre d'un modèle de traitement du langage, cela permet de caractériser les mots vis-à-vis de différents points de vue ou rôles qu'ils occupent dans la phrase tels que sujet, verbe ou encore complément.
Le vecteur de sortie correspond à la concaténation des vecteurs de sortie de chaque tête.

### Mise en oeuvre

- Pour calculer une sortie (vecteur d'attention), trois vecteurs pour chaque élément de la séquence d'entrée sont considérés :
  - Vecteur requête $q$ (*query*) ;
  - Vecteur clé $k$ (*key*) de dimension $d_k$ ;
  - Vecteur valeur $v$ (*value*) de dimension $d_v$.

  Les valeurs de chacun de ces vecteurs sont apprises pendant l'entraînement du modèle Transformer.
- Pour chaque élément de la séquence d'entrée (requête $q$), les produits scalaires avec l'ensemble des éléments de la seconde séquence (clés $k$) sont calculés, puis divisés par la racine carrée de la dimension du vecteur $k$ ($d_k$). Cette division assure la stabilité du gradient.
- Une opération *softmax* est ensuite appliquée à chaque sortie puis celle-ci est multipliée par le vecteur valeur $v$ correspondant.
- Enfin, le vecteur d'attention d'une requête $q$ correspond à la somme des vecteurs ainsi calculés.

### Équation

En pratique, la fonction d'attention est calculée sur un ensemble de requêtes simultanément, regroupées dans la matrice $Q$. Les clés et valeurs sont elles aussi regroupées respectivement dans des matrices $K$ et $V$. Les sorties sont calculées comme suit :

$$\text{Attention}(Q, K, V) = \text{Softmax}\left(\frac{QK^T}{\sqrt{d_k}}\right)V \tag{2.1}$$

avec :

- $d_k$ : la dimension du vecteur clé $k$.

## Focus 2.13 – Encodage positionnel

### Définition

Dans un Transformer, chaque élément de la séquence d'entrée (ou de sortie) est traité simultanément dans la pile d'encodeurs (ou de décodeurs). Ainsi, le modèle n'a pas connaissance de la position de chaque élément dans la séquence. C'est pourquoi l'encodage positionnel est utilisé dans les réseaux à base de Transformers, afin de ne pas perdre l'ordre des éléments de la séquence d'entrée (ou de sortie) lors de la propagation des informations dans le modèle.

ÉQUATION

Le premier encodage de position a été proposé par VASWANI et al. (2017). Il s'agit d'un encodage fixe qui se base sur les fonctions cosinus et sinus, et est calculé comme suit :

$$\begin{aligned} \mathrm{PE}(pos, 2i) &= \sin(w_i \cdot pos) \quad \forall i \in \left[0, \frac{d_{model}}{2}\right] \\ \mathrm{PE}(pos, 2i+1) &= \cos(w_i \cdot pos) \quad \forall i \in \left[0, \frac{d_{model}}{2}\right] \end{aligned} \tag{2.2}$$

avec :

$$w_i = \frac{1}{10000^{\frac{2i}{d_{model}}}}$$

et :

— $pos$ : la position de l'élément dans la séquence ;

— $d_{model}$ : la dimension d'encodage de l'élément.

## Focus 2.14 – ARCHITECTURE VISION TRANSFORMER

DÉFINITION

Un Vision Transformer (ViT) est une adaptation de l'architecture Transformer standard appliquée aux images. La séquence en entrée du système correspond à une séquence de patchs de taille fixe de l'image originale, où la couche d'*embedding* est remplacée par une projection linéaire des valeurs des patchs aplanis. Pour la tâche de classification, l'encodeur est suivi d'un MLP standard produisant des probabilités pour chaque classe considérée. Pour la détection d'objets, il est suivi d'un décodeur convolutif semblable à ceux des FCN.

AVANTAGES

— Par rapport aux CNN, les performances sont au moins aussi bonnes tout en nécessitant moins de mémoire pour le traitement et en étant plus rapide en inférence.

INCONVÉNIENTS

— Le modèle nécessite un très grand nombre de données d'apprentissage ou une étape de pré-entraînement afin d'obtenir des résultats satisfaisants.

SYSTÈME VISION TRANSFORMER

Le premier Vision Transformer a été proposé pour la tâche de classification d'images (DOSOVITSKIY et al., 2021). Le modèle comporte un encodeur Transformer suivi d'un MLP. Il a obtenu des performances comparables aux systèmes CNN à l'état de l'art, tout en nécessitant beaucoup moins de ressources pour l'entraînement.

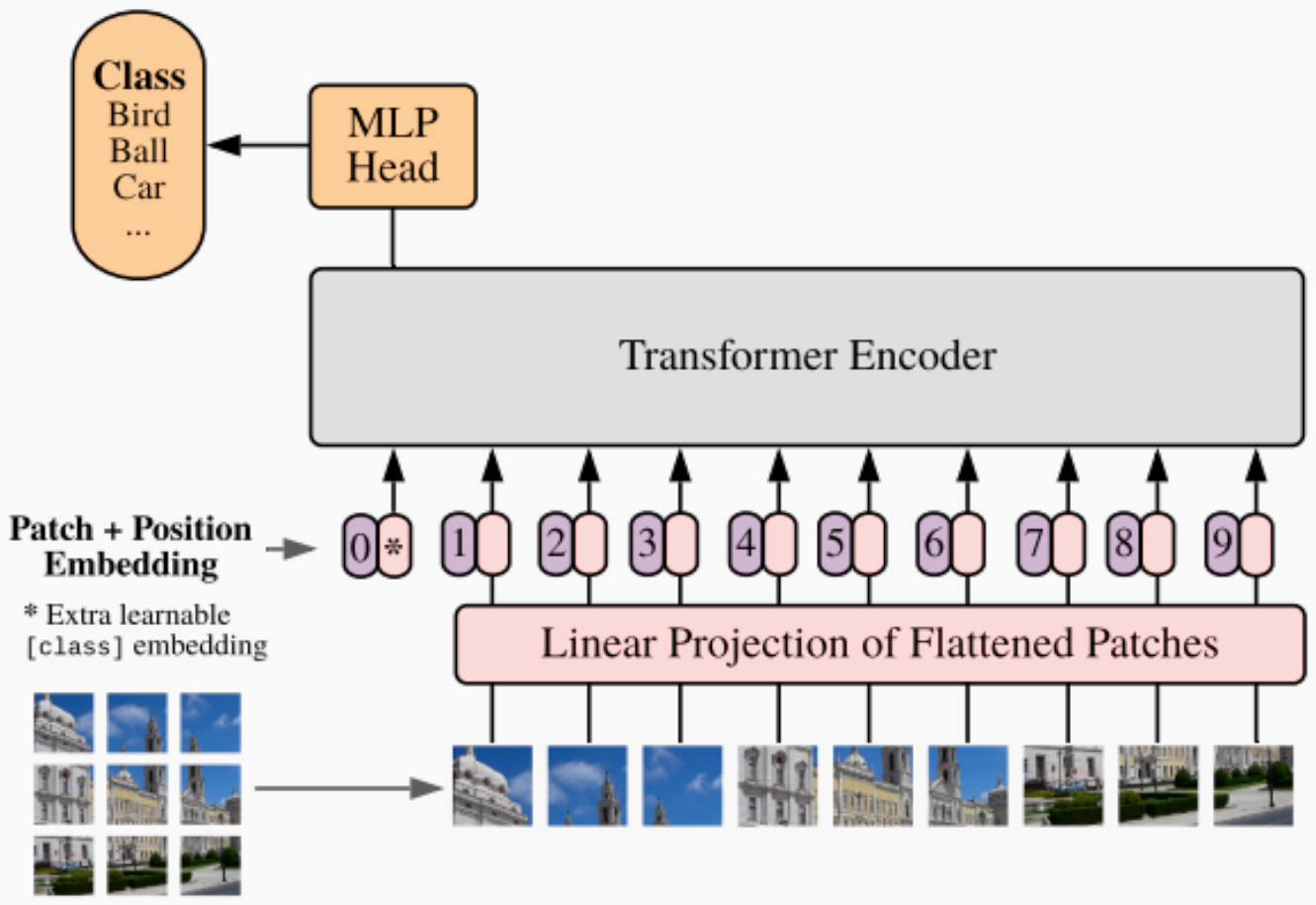

**Figure 2.14** – Schéma de l'architecture du modèle Vision Transformer original pour la classification d'images, issu de Dosovitskiy et al. (2021).

## Focus 2.15 – Système Pix2Seq

Pix2Seq (Chen et al., 2022) est un des premiers systèmes à base de Transformers proposé pour traiter la détection d'objets dans les images de scènes naturelles de manière séquentielle. Le modèle obtient des performances supérieures à celles obtenues par les systèmes à l'état de l'art, tels que Faster-RCNN (Ren et al., 2015) et DETR (Carion et al., 2020), tout en nécessitant moins de paramètres.
Le modèle est composé d'un encodeur suivi d'un décodeur Transformer. Les auteurs ont comparé différents encodeurs à base de convolutions, de Transformers ou des encodeurs hybrides, leurs expériences montrant les meilleures performances avec un encodeur Transformer. Le décodeur est standard et comporte six couches de décodeur Transformer. Celui-ci produit une séquence de coordonnées et de classes représentant les objets détectés ainsi que leurs classes.

### Modélisation de la détection

Le modèle Pix2Seq est entraîné à prédire séquentiellement chaque objet, une coordonnée à la fois, de la manière suivante : ordonnée du point supérieur gauche, abscisse du point supérieur gauche, ordonnée du point inférieur droit, abscisse du point inférieur droit et classe de l'objet. Ainsi, un objet et sa classe sont détectés par cinq valeurs prédites. De plus, les auteurs considèrent la détection comme une tâche de classification en considérant une classe pour chaque valeur possible en ordonnée et en abscisse.

SCHÉMA DE L'ARCHITECTURE DE PIX2SEQ

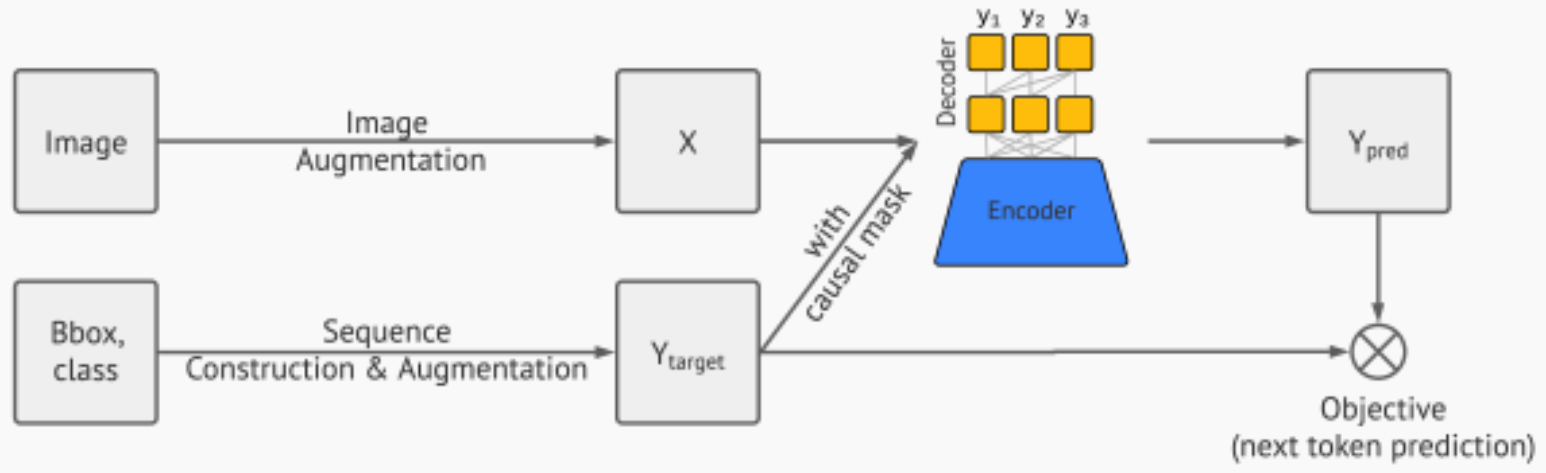

**Figure 2.15 –** Schéma du système Pix2Seq, issu de CHEN et al. (2022).

### APPROCHES COMBINANT IMAGE ET TEXTE

Les systèmes présentés en section 2.1.2 sont actuellement les plus utilisés pour la détection d'objets dans les images de documents. Certaines recherches se sont également orientées vers des approches combinant l'image et le texte du document afin de réaliser la tâche de détection. Dans le cas de documents complexes, l'ajout du texte dans le processus de détection peut aider à détecter et à classifier des éléments de plus haut niveau tels que des actes (PRIETO et al., 2020). Dans la tâche de *Visual Document Understanding* (DELTEIL et al., 2022), les informations sont extraites à l'aide d'une combinaison des caractéristiques textuelles et visuelles de l'image d'un document. Certaines propositions sont présentées dans cette section pour les tâches de détection, mais aussi de pré-entraînement pour différentes tâches de compréhension d'images de documents. Il est important de noter que pour la plupart des systèmes présentés dans cette section, un reconnaisseur (HTR ou OCR) a été entraîné au préalable afin d'extraire le contenu textuel des images de documents.

YANG et al. (2017) ont été parmi les premiers à proposer un réseau entièrement convolutif multimodal pour extraire des structures sémantiques de documents modernes. Pour aider à distinguer des classes similaires comme les paragraphes et les listes, ils incorporent des informations textuelles à l'aide d'une carte d'intégration de texte concaténée avant la dernière convolution du modèle. L'ajout de cette carte n'a pas montré d'amélioration significative dans des conditions réelles d'utilisation. Suivant cette idée, BARMAN et al. (2021) ont proposé un système capable de segmenter finement les journaux historiques et de gérer les variations de mise en page dans le temps. Ils utilisent la même représentation textuelle que YANG et al. (2017) mais sur des jetons produits par un processus OCR au lieu de phrases, ce qui est plus réaliste. Ils ont montré que l'ajout des cartes d'intégration du texte au début du réseau donne de meilleures performances. Certains travaux ont également étudié la combinaison de caractéristiques textuelles et visuelles pour classifier les pages des documents. Dans WIEDEMANN et al. (2018), une combinaison de deux réseaux neuronaux convolutifs (CNN), l'un basé sur des données textuelles et l'autre sur des numérisations d'images, est utilisée pour classifier les pages. Les paramètres sont ensuite combinés et transmis à un perceptron multicouche pour la classification finale. Cette combinaison a permis d'augmenter les performances par rapport à un seul CNN basé uniquement sur le texte ou l'image.

Pour les documents modernes, LayoutLM (Xu et al., 2020) a été proposé. Il s'agit d'une méthode de pré-entraînement simple pour les tâches de compréhension d'images de documents qui permet de modéliser conjointement les interactions entre le texte et les informations de mise en page dans les documents numérisés. Tout d'abord, un processus de reconnaissance complet est appliqué à l'image d'entrée afin de détecter les objets textuels et de reconnaître l'ensemble des textes. Ensuite, les auteurs utilisent une combinaison de BERT (Devlin et al., 2019), où l'information textuelle d'entrée est principalement représentée par des plongements de mots, et des caractéristiques d'image données par Faster-RCNN (Ren et al., 2015). LayoutLM permet de modéliser conjointement les interactions entre le texte et les informations de mise en page dans les documents numérisés et est ensuite utilisé comme pré-entraînement pour un grand nombre de tâches de compréhension d'images de documents. Ce système a montré des performances à l'état de l'art sur des documents commerciaux numérisés, mais nécessite un nombre important de données d'apprentissage. Dans Li et al. (2021), les auteurs présentent VTLayout, un système qui fusionne les caractéristiques visuelles profondes, superficielles et textuelles des documents pour localiser et identifier les différents blocs. Dans la première étape, le modèle Cascade Mask R-CNN est appliqué directement sur l'image pour localiser tous les blocs du document. Dans la seconde étape, les caractéristiques visuelles profondes, superficielles et textuelles sont extraites et fusionnées afin d'identifier les classes de chaque bloc. Les caractéristiques textuelles sont extraites par PaddleOCR (Du et al., 2020) puis transformées par une application de TF-IDF. Ce modèle a montré un gain de performances de détection par rapport aux systèmes standards manquant, notamment, de précision sur la classe de titre.

Prieto et al. (2020) ont également étudié le cas où l'aspect graphique des images n'est pas suffisant pour segmenter les chartes médiévales en actes. Ils ne visent pas seulement à détecter les actes mais cherchent également à les classifier comme *début*, *milieu*, *fin* d'acte ou *acte complet*. Ils utilisent une carte d'indexation probabiliste pour construire des caractéristiques supplémentaires basées sur le contenu textuel, puis les caractéristiques graphiques et textuelles sont fusionnées afin d'obtenir une seule entrée pour le système de segmentation. Ils montrent que l'ajout de contenu textuel peut faciliter la segmentation des actes, et que l'ajout de connaissances préalables permet d'améliorer encore les performances, cependant, leur méthode reste complexe à mettre en place.

## 2.2 ESTIMATION DE LA CONFIANCE DES OBJETS DÉTECTÉS

Les réseaux de neurones obtiennent désormais des performances remarquables dans de nombreux domaines d'application. Cependant, leur utilisation pour des applications industrielles exige qu'ils soient à la fois capables de fournir le résultat attendu tout en évaluant leur propre certitude, ou incertitude, quant à cette décision. Ceci est particulièrement important pour les applications critiques telles que celles liées aux images médicales ou à la conduite autonome par exemple.

L'apprentissage actif (*active learning*, détaillé dans le Focus 2.16) (Lewis et al., 1995) est une méthode d'apprentissage automatique itératif dans lequel l'algorithme d'apprentissage demande des données d'entraînement, celles jugées les plus pertinentes. Ces données sont sélectionnées en fonction de la confiance de l'algorithme quant à ses propres décisions. Les premières propositions consistaient à utiliser directement les probabilités *a posteriori* du classifieur afin de sélectionner les exemples à annoter. Ainsi, les exemples ayant une probabilité proche de 0,5 (*uncertainty sampling*) étaient sélectionnés pour l'itération suivante. Les réseaux neuronaux de détection d'objets produisent également des probabilités qui pourraient directement être utilisées comme estimations de confiance. Cependant, il a été démontré que ces probabilités sont souvent des estimateurs trop confiants qui donnent une confiance élevée même sur des prédictions erronées (Nguyen et al., 2015). Pour résoudre ce problème, plusieurs études ont été menées afin de concevoir de meilleurs estimateurs.

Ainsi, toujours dans le cadre de l'apprentissage actif, l'une des premières approches proposées pour sélectionner les échantillons à annoter manuellement était basée sur les machines à vecteurs de support (SVM) linéaires. Dans cette optique, Tong et al. (2002) ont proposé SVM Min Margin qui consiste à entraîner un SVM linéaire et à choisir les échantillons étant les plus proches de la limite de décision. Une autre approche populaire est l'échantillonnage d'incertitude (*uncertainty sampling*) (Settles et al., 2008) où les échantillons menant à des prédictions avec une grande incertitude sont sélectionnés. Pour quantifier l'incertitude, plusieurs mesures basées sur les probabilités *a posteriori* ont été proposées, comme l'entropie ou le score de moindre confiance (Brust et al., 2019).

Pour modéliser l'incertitude des décisions des réseaux neuronaux, d'autres approches ont été proposées, comme le *dropout* de Monte Carlo (Gal et al., 2016). Le *dropout* (Srivastava et al., 2014) est une méthode de régularisation utilisée dans les réseaux neuronaux afin de lutter contre le manque de généralisation des modèles. Il consiste à désactiver (mettre à 0) des valeurs, choisies aléatoirement, de l'image en entrée d'une couche. Il est appliqué uniquement durant la phase d'apprentissage et permet d'éviter le sur-apprentissage et la coadaptation, chaque neurone devant apprendre indépendamment des autres. Dans le MC *dropout*, au lieu de calculer une seule prédiction au moment du test, il est demandé au réseau de fournir plusieurs prédictions avec *dropout*, dont la distribution est ensuite analysée pour dériver une estimation de la confiance de la prédiction sans *dropout*. Cette technique, qui se rapproche des modèles bayésiens par apprentissage profond, a été utilisée pour de nombreuses tâches. Elle s'est souvent révélée efficace pour la classification afin de choisir les données à étiqueter (Gal et al., 2017). Dans Dechesne et al. (2021), le MC *dropout* est utilisé pour estimer l'incertitude de résultats de segmentation sémantique d'images. De plus, Moon et al. (2020) utilisent le MC *dropout* comme technique de régularisation des probabilités de classe pour obtenir un meilleur classement ordinal des prédictions.

D'autres travaux font appel à des modèles d'estimation de confiance profonds indépendants du modèle de détection. Dans Granell et al. (2021), un réseau adversaire est entraîné en parallèle du modèle de détection. Celui-ci est entraîné pour estimer la proximité des prédictions avec la vérité du terrain.

La plupart des travaux présentés ici se concentrent sur la tâche de classification. En effet, malgré les nombreux travaux présentant de nouveaux systèmes de détection d'objets, il y a très peu de travaux, dans la littérature, discutant l'estimation de la confiance pour cette tâche.

**Focus 2.16 – Apprentissage actif / Active learning**

Définition

L'*Active learning* (ou apprentissage actif) (Lewis et al., 1995) est une méthode d'apprentissage automatique qui permet à un algorithme d'interagir avec un oracle durant le processus. Dans un cadre d'apprentissage classique, les données sont choisies au préalable et imposées. En apprentissage actif, c'est l'algorithme d'apprentissage qui demande les données jugées les plus pertinentes.
Le processus est itératif et s'arrête lorsqu'un critère de performances ou un nombre défini de données annotées ou d'itérations est atteint.

Avantages

— L'utilisation de l'apprentissage actif permet de réduire fortement le coût d'annotation manuelle.
— Les performances du modèle final sont améliorées en comparaison avec un entraînement classique puisque les données sont choisies afin d'optimiser les résultats.

Inconvénients

— Une fonction d'acquisition est nécessaire. Les fonctions d'acquisition permettent d'associer une donnée à une valeur qui encode soit l'incertitude du modèle sur cet exemple soit sa contribution dans l'ajustement du modèle. Plusieurs fonctions ont été proposées dans le domaine de l'*uncertainty sampling* telles que l'entropie, se basant directement sur les probabilités a posteriori, ou la distance des exemples par rapport à la limite de décision dans les SVM. D'autres approches se basent sur la différence entre les résultats donnés par plusieurs modèles (*Query By Committee*).
— Il est également nécessaire de définir une stratégie de sélection : quels exemples seront utilisés pour l'itération suivante ? Certains travaux choisissent les exemples où les probabilités *a posteriori* sont les plus hautes ou les plus basses. Dans le cas des SVM, certains choisissent les exemples les plus proches, d'autres les plus éloignés de la limite de décision. Il n'y a, à notre connaissance, pas de consensus sur la stratégie à utiliser.

Exemple d'apprentissage actif

La Figure 2.16 présente un exemple d'apprentissage actif. Les paramètres du modèle sont initialisés ou pré-entraînés sur l'ensemble d'apprentissage annoté. Le modèle est appliqué aux exemples non annotés qui sont ensuite sélectionnés selon la stratégie choisie. Les exemples sélectionnés sont annotés par un opérateur puis ajoutés à l'ensemble d'entraînement. Un nouveau modèle est entraîné sur ce nouvel ensemble. Le processus est répété jusqu'à ce que les conditions d'arrêt prédéfinies soient atteintes.

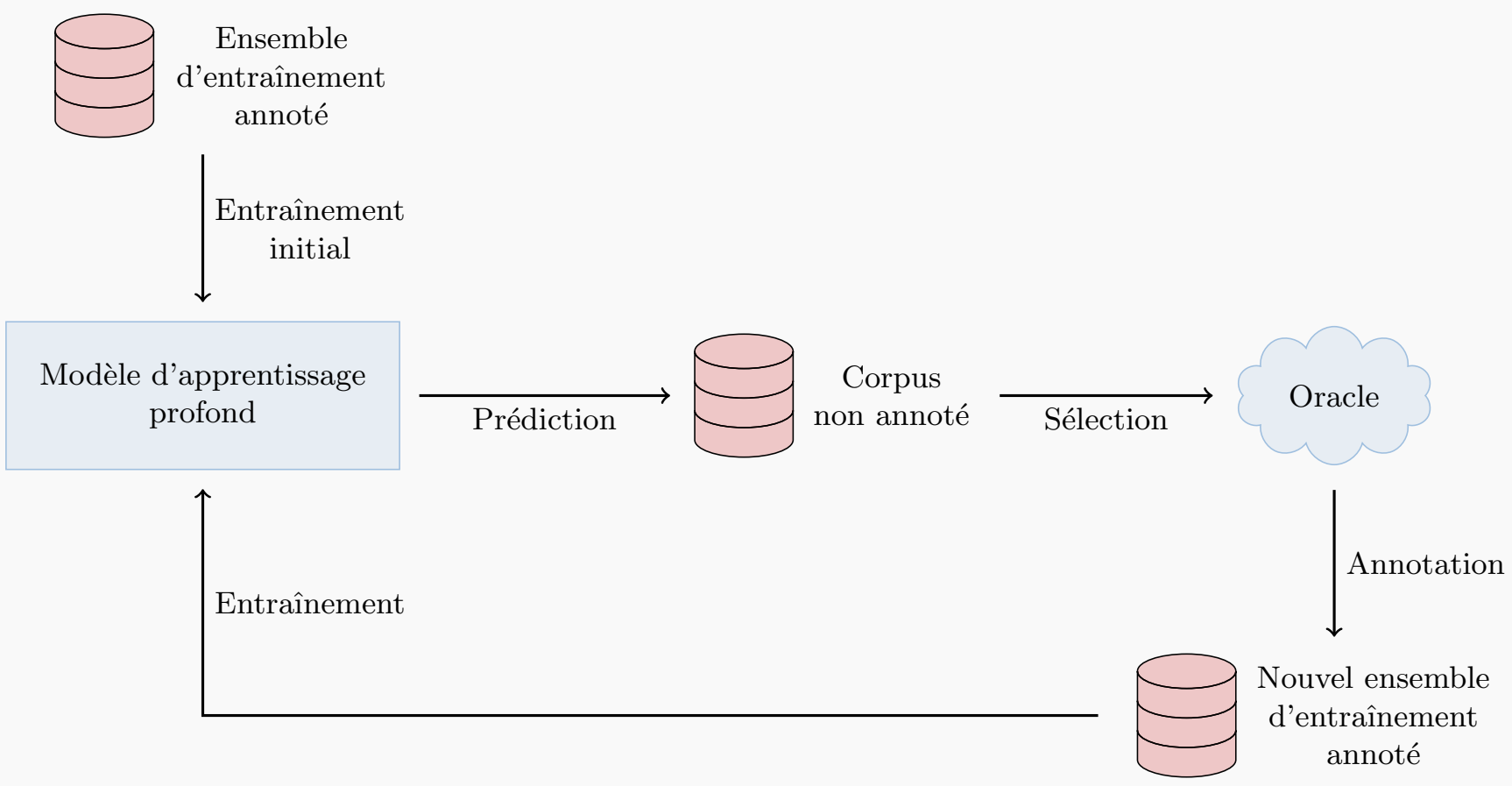

**Figure 2.16 –** Schéma présentant le processus d'apprentissage actif.

# 3

# ENTRAÎNEMENT ET ÉVALUATION DES SYSTÈMES DE DÉTECTION

La mise en place et l'amélioration de modèles de détection d'objets conduisent à explorer différents axes de recherche. Bien que la majorité des travaux dans la littérature se concentrent uniquement sur la proposition de nouvelles architectures, nous avons souhaité nous intéresser à des études et des solutions plus complètes, en évoquant notamment les problématiques liées aux annotations et évaluations. En effet, l'évaluation de la qualité des algorithmes de détection ou de reconnaissance est cruciale dans la mise au point de systèmes et leurs comparaisons. Elle nécessite donc l'utilisation de métriques appropriées. Cependant, il faut également étudier les données annotées utilisées pendant l'entraînement et l'évaluation. Si les annotations des données ne sont pas cohérentes avec la métrique utilisée, la métrique ne peut pas refléter les performances réelles du modèle.

Dans ce chapitre, nous mettons tout d'abord en avant les problèmes liés aux annotations des jeux de données. Dans une première section 3.1, nous présentons une étude des récents jeux de données utilisés dans les systèmes à l'état de l'art, principalement pour la détection de lignes de texte. Nous mettons ensuite en évidence, en section 3.2, les différentes règles d'annotation manuelle, ainsi que les défis liés et les solutions proposées dans la littérature. Par la suite, nous discutons, en section 3.3, des différentes métriques proposées et utilisées dans la littérature afin d'évaluer et de comparer les systèmes de détection d'objets dans les images de documents.

## 3.1 JEUX DE DONNÉES

Dans cette partie, nous présentons les jeux de données utilisés dans les systèmes récemment proposés, notamment pour la détection de lignes de texte. Nous présentons également les jeux de données privés que nous avons utilisés durant la thèse. Ces jeux de données sont détaillés dans les paragraphes suivants, résumés dans la Table 3.1, et un exemple est montré sur la Figure 3.2.

Nous nous focalisons sur la tâche de détection de lignes de texte car c'est une étape centrale de l'analyse de la mise en page des documents puisqu'elle est nécessaire à la reconnaissance de texte et qu'elle a un fort impact sur la qualité de la reconnaissance. De plus, c'est une des tâches pour laquelle le type d'annotation et la définition même de la tâche peuvent être très variables. Une étude plus générale sur les jeux de données historiques est proposée par Nikolaidou et al. (2022).

**Table 3.1 –** Tableau récapitulatif des différents jeux de données utilisés pour la détection de lignes de texte. Le symbole "–" indique une résolution ou date non disponible. Pour chaque jeu de données, la colonne TAILLE indique la taille moyenne des images, calculée sur l'ensemble d'entraînement.

| JEU DE DONNÉES | DATE | IMAGES | LIGNES | LANGUE(S) | RÉSOLUTION (DPI) | TAILLE (PIXELS) |
|---|---|---|---|---|---|---|
| AN-INDEX† | – | 34 | 666 | Français | – | [1 949, 1 338] ± [796, 607] |
| BALSAC<br>VÉZINA et al. (2020) | 1850 – 1916 | 913 | 45 685 | Anglais<br>Français | – | [3 746, 2 671] ± [1 141, 627] |
| BNPP† | 19e – 20e siècle | 12 | 1 281 | Français | – | [3 710, 5 103] ± [21, 86] |
| BOZEN<br>SÁNCHEZ et al. (2016) | 1470 – 1805 | 450 | 10 550 | Allemand | – | [3 524, 2 398] ± [22, 62] |
| cBAD2019<br>DIEM et al. (2019) | – | 3 021 | 193 858 | Diverses européennes | Variable | [3 268, 2 751] ± [1 364, 1 504] |
| DIVA-HISDB<br>SIMISTIRA et al. (2016) | 11e / 14e siècles | 150 | 12 808 | Italien, Latin<br>Allemand, Grec | 600 | [5 493, 3 843] ± [709, 728] |
| HOME-ALCAR<br>STUTZMANN et al. (2021) | 12e – 14e siècle | 1 845 | 136 206 | Latin | Variable | [3 850, 4 506] ± [949, 1 820] |
| HOME-NACR<br>BOROS et al. (2020) | 1145 – 1491 | 496 | 7 614 | Allemand<br>Latin<br>Tchèque | – | [4 499, 6 206] ± [1 639, 2 292] |
| HUGIN-MUNIN<br>MAARAND et al. (2022) | 19e – 20e siècle | 849 | 23 732 | Norvégien | – | [3 998, 3 740] ± [1 405, 1 403] |
| HORAE<br>BOILLET et al. (2019) | 14e – 15e siècle | 573 | 13 796 | Latin | Variable | [4 200, 4 648] ± [1 361, 2 112] |
| IAM<br>MARTI et al. (2002) | 1999 | 1 539 | 13 353 | Anglais | 300 | [3 542, 2 479] ± [0.45, 0.33] |
| RASM<br>CLAUSNER et al. (2018) | 10e – 19e siècle | 120 | 2 619 | Arabe | 400 | [7 674, 5 408] ± [1 384, 914] |
| READ<br>GRÜNING et al. (2017) | 1470 – 1930 | 2 035 | 132 124 | Diverses européennes | Variable | [3 966, 3 121] ± [1 350, 1 268] |
| SCRIBBLELENS<br>DOLFING et al. (2020) | 16e – 18e siècle | 1 000 | 28 255 | Néerlandais | Variable<br>150 – 300 | [3 519, 2 375] ± [665, 459] |

† Jeux de données privés utilisés durant la thèse.

AN-INDEX – Ce premier jeu de données est composé de 34 images de documents des instruments de recherche numérisés des Archives nationales françaises. Il s'agit d'une base privée dont les documents sont rédigés en français.

BALSAC – Depuis 50 ans, le projet BALSAC[1] construit une importante base de données sur la population du Québec. Pour entraîner des modèles de traitement automatique et ainsi aider l'intégration de millions d'enregistrements, un échantillon du corpus contenant les actes de naissance, de mariage et de décès de la population québécoise de 1850 à 1916 a été annoté. Le jeu de données Balsac (VÉZINA et al., 2020) consiste donc en 913 images (pages simples ou doubles) extraites de 74 registres manuscrits. Elles ont été annotées au niveau des actes et des lignes avec leurs transcriptions et entités nommées.

BNPP – Ce jeu de données privé a été fourni par les Archives historiques de la banque BNP Paribas[2]. Il consiste en un échantillon de 12 images manuscrites extraites de cinq registres scannés de procès-verbaux du Comptoir National d'Escompte de Paris. Elles ont été sélectionnées parmi une centaine de registres rédigés en français entre le 19e et le 20e siècle.

BOZEN – Ce jeu de données (SÁNCHEZ et al., 2016) fait partie du projet READ et consiste en 450 pages manuscrites annotées. Les pages sont extraites de documents de la collection Ratsprotokolle écrits entre 1470 et 1805. Il est annoté au niveau des lignes de texte avec leurs transcriptions.

cBAD2019 – Le jeu de données cBAD (DIEM et al., 2019) est constitué de 3 021 images de documents collectées dans sept archives européennes. Il a été utilisé lors de la compétition cBAD à ICDAR2019 pour la détection des lignes de base.

DIVA-HISDB – DIVA-HisDB (SIMISTIRA et al., 2016) est une base de données qui contient 150 images extraites de trois manuscrits médiévaux des 11e et 14e siècles. Ces manuscrits ont été choisis pour la complexité de leurs mises en page avec du texte principal, et des commentaires dans les marges et entre les lignes. Pour chaque manuscrit, 50 images ont été sélectionnées et réparties en 20 images d'entraînement, 10 images de validation, 10 images de test et 10 autres images de test « privées ». Chaque image a été annotée manuellement au niveau du pixel pour les classes de corps de texte, décorations et commentaires.

HOME-ALCAR – Le jeu de données HOME-Alcar (STUTZMANN et al., 2021) contient 17 cartulaires, recueils des chartes et des actes juridiques produits entre le 12e et le 14e siècle et écrits en latin. Les images ont été annotées au niveau des lignes de texte avec leurs transcriptions.

1. https://balsac.uqac.ca/
2. https://history.bnpparibas/

HOME-NACR – Le jeu de données HOME-NACR (Boros et al., 2020) est composé de 496 chartes médiévales sélectionnées parmi 43 000 chartes numérisées provenant des archives de la Couronne de Bohême et des archives des monastères. Elles ont été rédigées de 1145 à 1491 en allemand, latin et tchèque du début de l'ère moderne. Les chartes ont été annotées au niveau des lignes avec leurs transcriptions et entités nommées.

Hugin-Munin – La base de données Hugin-Munin (Maarand et al., 2022) est constituée de pages provenant de correspondances et de journaux intimes de 12 artistes norvégiens écrits de 1820 à 1950. Les documents ont été annotés au niveau des lignes avec leurs transcriptions correspondantes. La base comporte 691 images d'entraînement, 85 de validation et 73 de test.

Horae – Durant le projet de recherche *Hours : Recognition, Analysis, Edition* (HORAE) (Stutzmann et al., 2019), un jeu de données a été créé (Boillet et al., 2019) et consiste en 573 images annotées de livres d'heures. Elles ont été sélectionnées parmi 500 manuscrits car elles représentent la variété des mises en page et des contenus. Les images ont été annotées à différents niveaux et avec différentes classes : page, paragraphe, ligne, miniature, initiale (simple, ornée ou illustrée), marge (ornée ou illustrée), ornementations et rubriques.

IAM – Le jeu de données IAM (Marti et al., 2002) a été créé en 1999 et contient 1 539 images de documents. Chaque image comporte une page avec un texte imprimé extrait du corpus Lancaster - Oslo/Bergen corpus (LOB), puis ce même texte écrit à la main. Les textes datent de 1961 et sont très divers : fictions, écrits scientifiques ou encore textes traitant de religion.

RASM – La compétition RASM 2018 (Clausner et al., 2018) visait la reconnaissance de manuscrits historiques en arabe. Un ensemble de 15 images de pages à une colonne a été utilisé pour l'entraînement et 85 pour évaluer les tâches de détection de lignes de texte, segmentation de pages et reconnaissance d'écriture manuscrite. Au total, le jeu contient 120 images extraites parmi une collection de manuscrits scientifiques arabes.

READ-BAD – Ce jeu de données (Grüning et al., 2017) contient de 2 035 images de documents écrits entre 1470 et 1930 et extraits de neuf archives européennes. Les données sont très variées avec des registres paroissiaux, des procès-verbaux ou encore des tables de recensement. Ce jeu a été utilisé lors de la compétition sur la détection de lignes de base cBAD : ICDAR2017 (Diem et al., 2017). Le jeu de données est divisé en sous-ensembles simples et complexes dépendant de la complexité de mise en page des documents.

SCRIBBLELENS – Le jeu de données ScribbleLens (DOLFING et al., 2020) contient 1 000 images de pages de documents néerlandais datant du début de l'ère moderne, tels que des journaux de bord de navires et des journaux de bord quotidiens produits entre le 16e et le 18e siècle. Les manuscrits consistent en des voyages écrits par des capitaines et des commerçants de la Vereenigde Oost-indische Company (VOC). L'ensemble de test est composé de 21 images annotées et transcrites au niveau de la ligne.

Nous observons que de nombreux jeux de données ont été présentés pour la détection de lignes de texte. Il est important de noter que ces jeux de données ont été annotés à l'aide de différents outils et pour différentes tâches : détection de lignes de texte ou de lignes de base. Nous détaillons les différents types d'annotations dans la section suivante 3.2.

## 3.2 ANNOTATION DES DONNÉES

Bien que la tâche de détection de lignes de texte soit assez triviale dans le cas de documents imprimés, dans le cas de documents manuscrits de nombreux aspects peuvent venir perturber la bonne détection des lignes. En effet, il n'est pas rare que des lignes se chevauchent ou que des initiales soient de taille très différente comparé au corps de texte. De plus, la qualité de numérisation et les possibles dégradations liées à la conservation peuvent rendre cette tâche d'autant plus complexe.

Un autre défi avec la détection de lignes de texte concerne la définition de ce qu'est une ligne de texte. Dans la littérature, une ligne de texte a été définie de plusieurs manières, comme présenté sur la Figure 3.1 (MECHI et al., 2021 ; RENTON et al., 2018). Tout d'abord, elle peut être définie uniquement par sa ligne de base (GRÜNING et al., 2017), qui correspond à une ligne virtuelle soulignant la plupart des caractères tandis que les descendants restent en dessous. Dans ce cas il est nécessaire d'estimer la hauteur de la ligne pour appliquer un reconnaisseur d'écriture. Elle a également été définie comme étant une boîte englobante (rectangle ou polygone) incluant tous les ascendants et descendants (Moysset et al., 2015), comme un ensemble de pixels appartenant au contenu textuel (Simistira et al., 2016 ; Vo et al., 2016) ou encore s'appuyant sur la hauteur en X (X-height). Il s'agit de la bande de base de la ligne sans les ascendants et descendants.

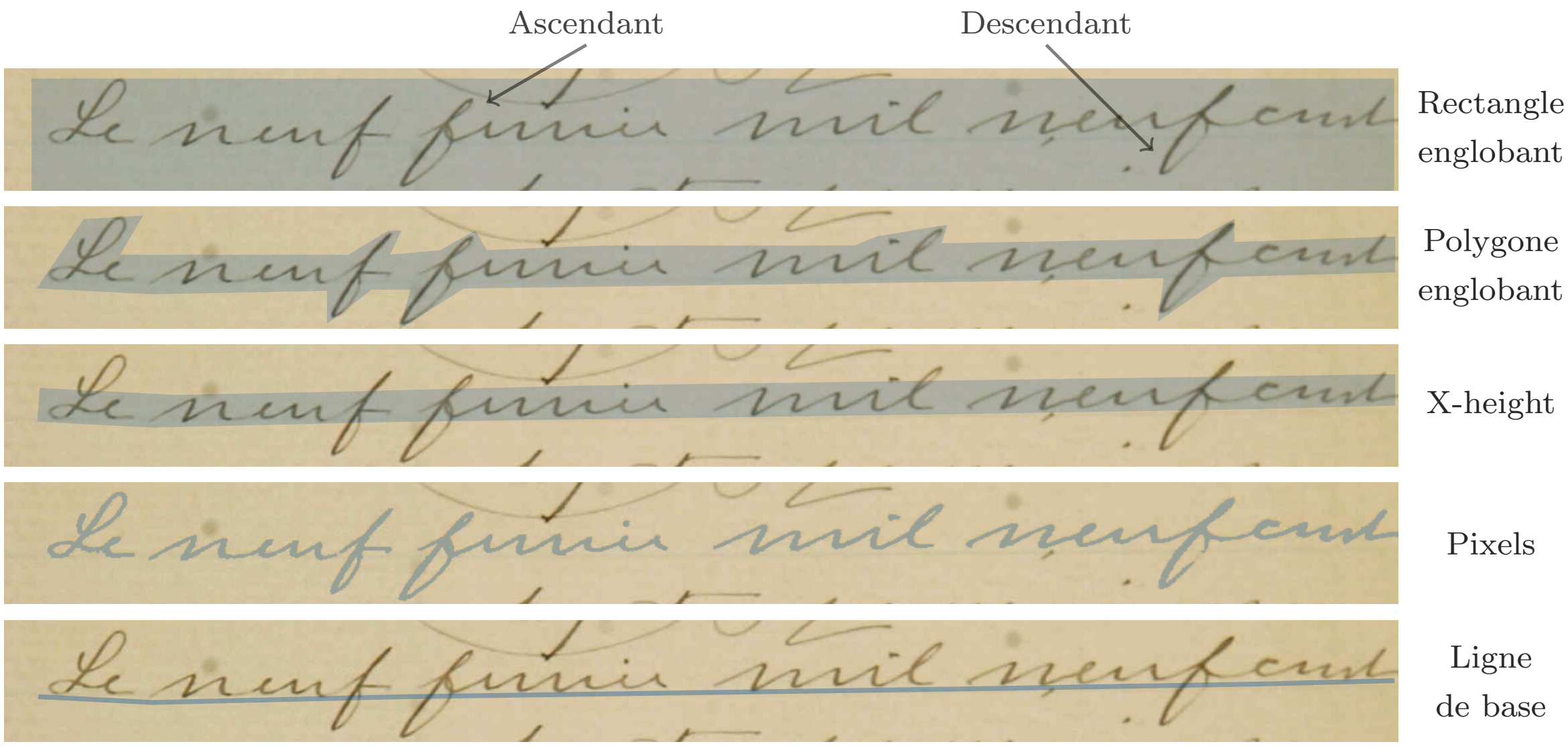

**Figure 3.1 –** Représentation des modélisations d'une ligne de texte proposées dans la littérature.

Définir une ligne de texte par sa bande de base présente de nombreux avantages par rapport aux autres représentations. Tout d'abord, elle représente bien les interlignes même lorsque les lignes se chevauchent en raison des ascendants ou des descendants, contrairement à une représentation par boîte englobante incapable de séparer les lignes qui se chevauchent. De plus, l'utilisation de la représentation pixel ou de la ligne de base nécessite un post-traitement avant de pouvoir être transmise à un reconnaisseur, contrairement à la bande de base qui semble plus appropriée à fournir une entrée convenable pour les reconnaisseurs de texte.

Nous résumons, dans la Table 3.2, les détails d'annotations de chaque jeu de données présenté en section 3.1. Dans cette Table, nous indiquons comment les lignes ont été annotées : ligne de base, bande de base, polygone englobant, quadrilatère (ou polygone simple) englobant et rectangle englobant. Nous indiquons également le taux de relâchement des annotations par rapport aux pixels de texte. Nous définissons ce taux comme étant la quantité de fond présent autour des pixels de texte dans les annotations (autres que la ligne de base). Les colonnes Intersections et Source présentent respectivement la quantité de chevauchements entre les annotations, et la manière dont les annotations ont été obtenues (manuellement, semi-automatiquement ou automatiquement). La dernière colonne indique la présence de transcription des lignes de texte. Cette Table montre explicitement que les annotations entre les différents jeux de données sont très variables. En effet, il n'y a aucun consensus sur la définition même d'une ligne de texte ni sur la forme que les annotations doivent avoir ou la quantité de fond à intégrer dans les polygones et boîtes englobants. De plus, il n'y a aucune étude, à notre connaissance, comparant les différentes annotations possibles et évaluant leurs impacts sur les résultats de reconnaissance de texte finaux.

La Figure 3.2 présente une image de chaque jeu de données associée à son taux de relâchement.

Les problèmes liés aux annotations des jeux de données ont également été peu étudiés dans la littérature. En effet, Barakat et al. (2018) montrent les problèmes liés aux lignes de texte qui se touchent et se superposent dans leur jeu de données mais ne traitent pas ces différents problèmes. Quelques rares travaux en discutent et proposent quelques solutions. Par exemple, face à des boîtes englobantes qui se touchent, Melnikov et al. (2020) ont suggéré de supprimer les ascendants et les descendants des lignes de texte en réduisant la hauteur des boîtes annotées de 30 % en haut et en bas. Ils ont ensuite redimensionné les polygones à la résolution d'entrée du modèle pour entraîner le système. Même si cette méthode s'est avérée efficace pour réduire le biais d'étiquetage de l'annotation, certains problèmes subsistent lorsqu'il s'agit de lignes verticales et inclinées. Dans le même esprit, Peskin et al. (2020) ont proposé différents masques d'annotation (voir Figure 3.3) pour la détection et la classification de formes géométriques à partir d'images en niveaux de gris. Ils ont suggéré d'annoter les objets (cercles, rectangles et triangles) de quatre façons : avec de petites marques centrales, avec de grandes marques centrales, avec de petites marques centrales et un contour d'un pixel, et avec de petites marques centrales et un contour de deux pixels. Concernant les problèmes de localisation, ils ont montré que les petites marques centrales donnent les meilleures performances. Cependant, de meilleurs résultats de classification sont obtenus avec les petites marques centrales avec contours. Par conséquent, trouver l'annotation la plus adaptée n'est pas un problème trivial et les discussions sont toujours ouvertes, mais nous proposons, dans le chapitre 5, une solution pour l'unification des différents types d'annotations pour entraîner des modèles de détection de lignes de texte.

**Table 3.2 –** Tableau récapitulatif du type d'annotation des différents jeux de données utilisés pour la détection de lignes de texte. La colonne RELÂCHEMENT indique la quantité de fond présent dans les annotations (important, moyen ou faible). La colonne INTERSECTIONS indique si des lignes se chevauchent. La colonne SOURCE indique comment les annotations ont été obtenues : manuellement, semi-automatiquement ou automatiquement. La colonne TEXTE indique la présence de transcriptions des lignes de texte.

| JEU DE DONNÉES | Ligne de base | Bande de base | Polygone englobant | Quadrilatère englobant | Rectangle englobant | RELÂCHEMENT | INTERSECTIONS | SOURCE | TEXTE |
|---|---|---|---|---|---|---|---|---|---|
| AN-INDEX† | | | | ✓ | ✓ | Important | Rares | Manuelle | |
| BALSAC<br>VÉZINA et al. (2020) | | | | ✓ | ✓ | Moyen | Rares | Manuelle | ✓ |
| BNPP† | | | | ✓ | | Moyen | Non | Manuelle | ✓ |
| BOZEN<br>SÁNCHEZ et al. (2016) | ✓ | | | ✓ | ✓ | Important | Oui | Semi-automatique | ✓ |
| cBAD2019<br>DIEM et al. (2019) | ✓ | ✓ | | ✓ | | Variable | Oui | Manuelle | |
| DIVA-HISDB<br>SIMISTIRA et al. (2016) | ✓ | | ✓ | | | Faible | Non | Semi-automatique | |
| HOME-ALCAR<br>STUTZMANN et al. (2021) | | | ✓ | | | Important | Oui | Automatique | ✓ |
| HOME-NACR<br>BOROS et al. (2020) | | ✓ | | | | Important | Rares | Manuelle | ✓ |
| HUGIN-MUNIN<br>MAARAND et al. (2022) | | ✓ | ✓ | | | Variable | Oui | Semi automatique | ✓ |
| HORAE<br>BOILLET et al. (2019) | | ✓ | | | | Moyen | Non | Manuelle | ✓ |
| IAM<br>MARTI et al. (2002) | | | | | ✓ | Moyen | Rares | Automatique | ✓ |
| RASM<br>CLAUSNER et al. (2018) | | | ✓ | | | Faible | Rares | Manuelle | ✓ |
| READ<br>GRÜNING et al. (2017) | ✓ | ✓ | | ✓ | | Variable | Oui | Manuelle | |
| SCRIBBLELENS<br>DOLFING et al. (2020) | | | | | ✓ | Important | Oui | Automatique | ✓ |

† Jeux de données privés utilisés durant la thèse.

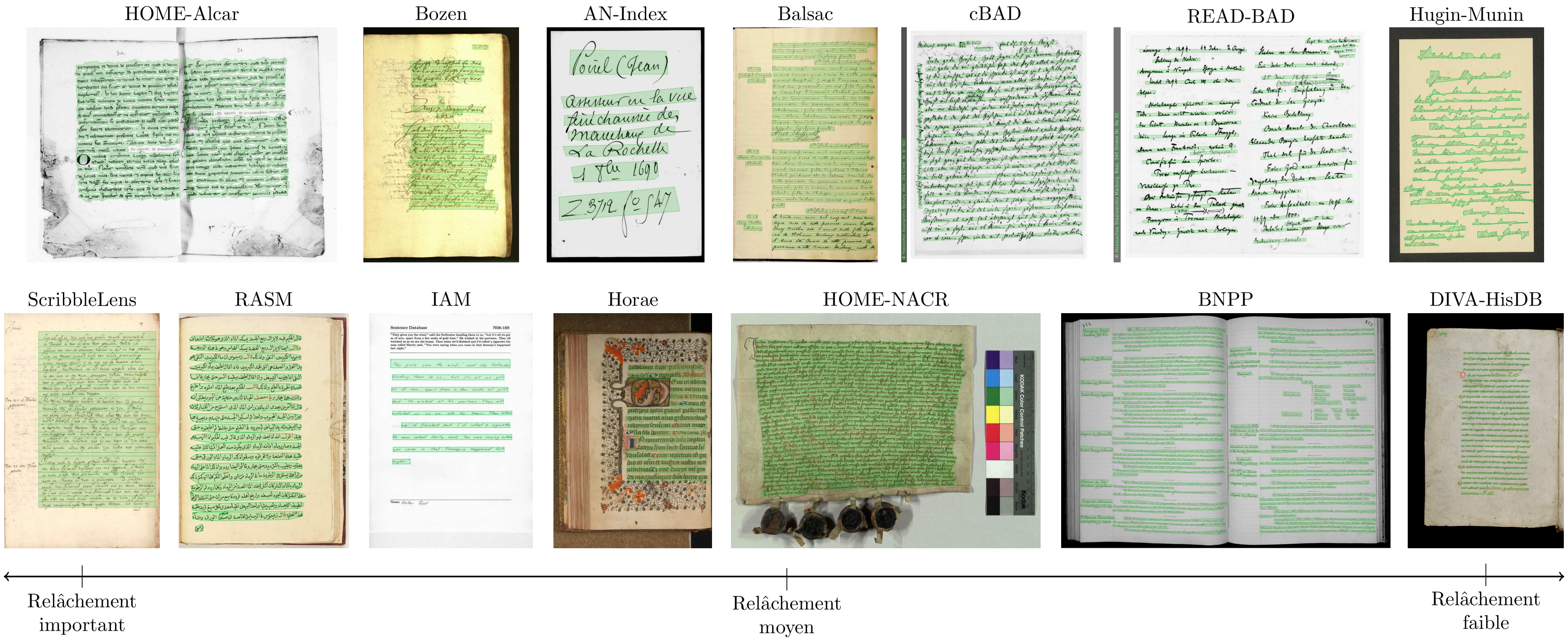

**Figure 3.2 –** Visualisation des différents taux de relâchement détectés dans les jeux de données. Les taux de relâchement indiquent la quantité de fond présent autour des pixels de texte dans les annotations.

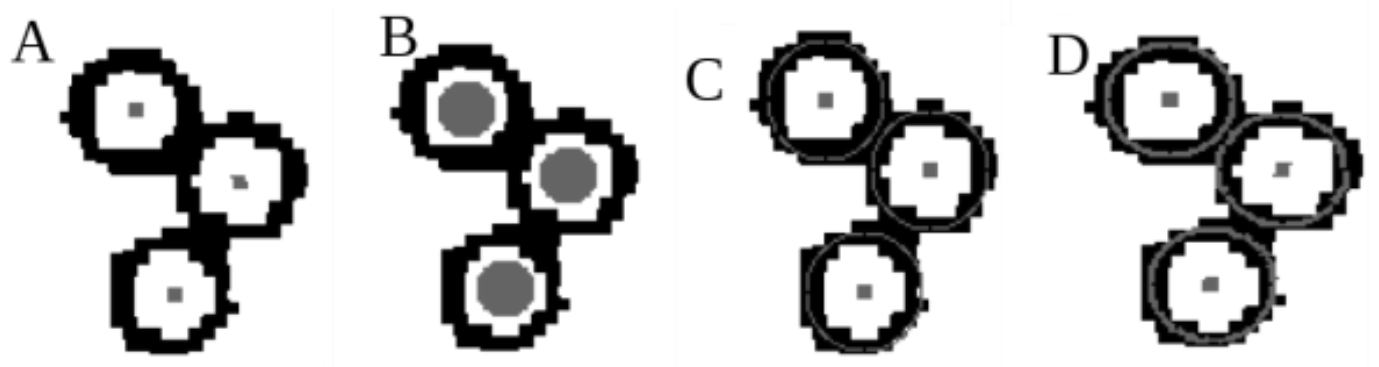

**Figure 3.3 –** Masques de segmentation comparés par PESKIN et al. (2020) : A. petites marques centrales, B. marques centrales plus grandes, C. petite marque centrale avec un contour d'un pixel, D. petite marque centrale avec un contour de deux pixels. Schéma issu de PESKIN et al. (2020).

## 3.3 MÉTRIQUES D'ÉVALUATION

En plus des problèmes liés aux différents types d'annotations se pose le problème de recherche de métriques appropriées pour évaluer et comparer correctement les résultats de détection. En effet, afin d'évaluer la qualité d'un modèle, une métrique d'évaluation doit être définie et calculée. Cette métrique doit avoir un sens vis-à-vis des données utilisées et de l'application future des résultats obtenus. Les différentes métriques permettant d'évaluer un modèle de détection d'objets peuvent être regroupées en plusieurs catégories : les métriques basées sur les pixels, les métriques objets et les métriques orientées vers la tâche finale. Celles-ci sont détaillées dans les paragraphes suivants.

De plus, HEMERY et al. (2010) ont étudié les propriétés clés qu'une métrique doit avoir pour une tâche de localisation. À partir de l'analyse de 33 métriques existantes, ils ont établi les plus appropriées pour cette tâche. En suivant cette idée, nous montrons, dans cette section, que les principales métriques actuellement utilisées ne sont pas suffisantes pour évaluer et comparer les modèles de détection d'objets.

### 3.3.1 MÉTRIQUES BASÉES SUR LES PIXELS

Les métriques calculées au niveau des pixels sont principalement basées sur l'intersection entre les pixels d'une région prédite et ceux d'une région annotée manuellement.

Comme le montre la Table 3.3, la majorité des systèmes de détection existants sont évalués à l'aide de métriques pixel. Les mesures de précision et de rappel, détaillées dans le Focus 3.1, sont largement utilisées, ainsi que l'Intersection-sur-Union (IoU) (Focus 3.2) et le score F1 (Focus 3.3). Dans le cas d'une détection à plusieurs classes et afin de calculer une unique valeur d'évaluation, les métriques sont calculées pour chaque classe et la moyenne arithmétique des valeurs obtenues est calculée. Alberti et al. (2017) ont d'ailleurs développé un outil permettant d'évaluer des modèles à partir des images de vérité terrain et des prédictions. Cet outil permet de calculer différentes valeurs dont l'IoU, mais également de visualiser les résultats.

**Table 3.3 –** Métriques d'évaluation utilisées dans les récents travaux liés à la détection d'objets dans les images de documents. P et R représentent respectivement les métriques précision et rappel. Les métriques R@.85 et mAP@.65 représentent respectivement le rappel pour un seuil d'IoU de 0,85 et la précision moyenne pour un seuil d'IoU de 0,65.

| Système | Pixel | | | Objet | | | |
|---|---|---|---|---|---|---|---|
| | IoU | P/R | F1 | P/R | R@.85/.95 | mAP@.65 | mAP |
| Tensmeyer et al. (2017) | ✓ | | | | | | |
| Yang et al. (2017) | ✓ | | ✓ | | | | |
| Barakat et al. (2018) | | ✓ | ✓ | | | | |
| Renton et al. (2018) | | ✓ | ✓ | | | | |
| dhSegment Ares Oliveira et al. (2018) | ✓ | ✓ | ✓ | | ✓ | | |
| Mechi et al. (2019) | ✓ | ✓ | ✓ | | | | |
| Tarride et al. (2019) | | ✓ | ✓ | ✓ | | ✓ | |
| Soullard et al. (2020) | | | | | | | ✓ |
| Melnikov et al. (2020) | | ✓ | ✓ | | | | |
| Mechi et al. (2021) | | ✓ | ✓ | | | | |

Ces métriques sont basées sur le nombre de pixels correctement prédits. Cependant, elles ne donnent aucune information sur le nombre d'objets correctement prédits et manqués ou divisés. Ces métriques sont également biaisées. En effet, comme le montre la Figure 3.4, plusieurs prédictions de qualités différentes peuvent être caractérisées par les mêmes valeurs d'IoU et de F1. En effet, les Figures 3.4b et 3.4c présentent deux prédictions pour la même image, en haut, et leur superposition, en bas, avec l'image de vérité terrain présentée sur la Figure 3.4a. La première prédiction montre des lignes divisées et fusionnées, une ligne manquante (en rouge) et quelques faux positifs (en cyan). Au contraire, la seconde prédiction montre des lignes plus épaisses mais pas de ligne manquante ni de faux positifs. Ainsi, la seconde prédiction semble meilleure. Cependant, les valeurs d'IoU et de score F1 sont égales

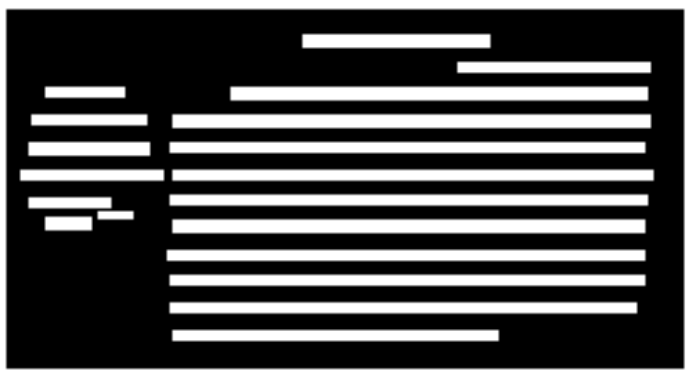

**(a)** Image de label.

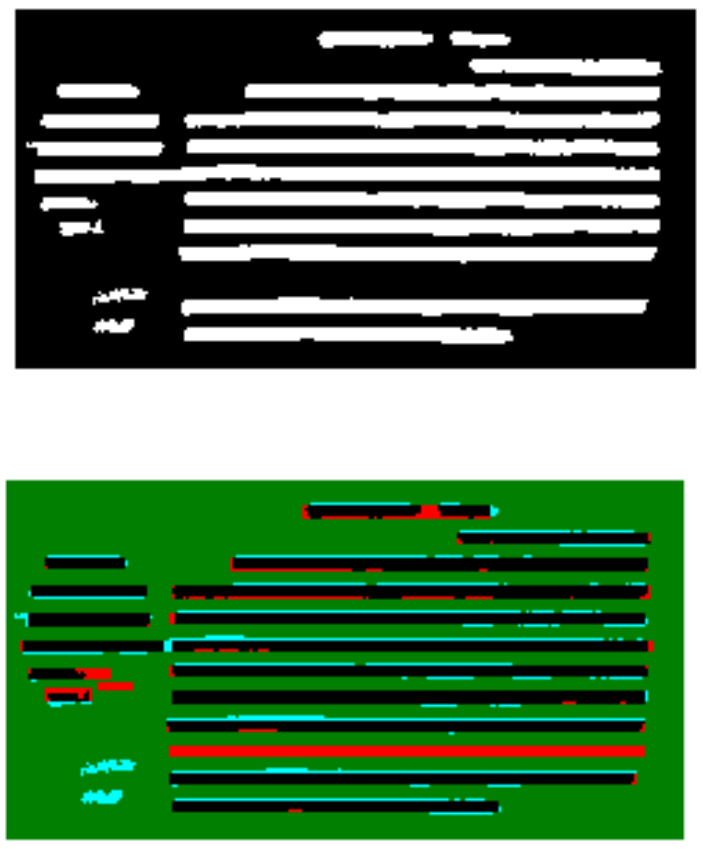

**(b)** Première prédiction :
IoU = 0,72 F1 = 0,84
P = 0,81 R = 0,87
AP@.5 = **0,68**

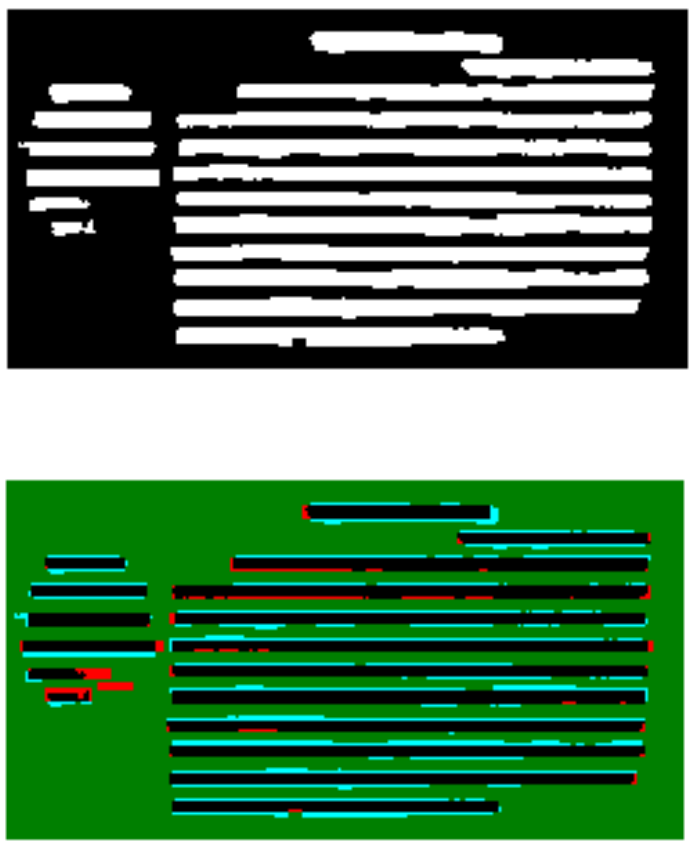

**(c)** Seconde prédiction :
IoU = 0,72 F1 = 0,84
P = 0,75 R = 0,95
AP@.5 = **0,94**

**Figure 3.4 –** Deux détections de lignes différentes obtenues pour une même image et obtenant les mêmes scores d'IoU et de F1. Les superpositions sont générées avec l'outil DIVA (Alberti et al. (2017)). Le vert et le noir correspondent respectivement aux pixels d'arrière-plan et d'avant-plan correctement prédits. Le cyan représente les pixels faussement positifs et le rouge les pixels faussement négatifs. Ici, seul le score AP au niveau objet (avec un seuil IoU de 50 %) permet d'évaluer et de comparer les prédictions avec précision.

pour les deux prédictions. Cela illustre le fait que ces métriques ne sont pas les plus adaptées pour évaluer les systèmes de détection d'objets, d'où le développement de nouvelles métriques basées sur les objets. Cependant, le calcul de ces valeurs de précision et de rappel au niveau de la ligne ou de l'objet n'est pas directement applicable car la décision qu'un objet soit bien ou mal détecté est plus complexe.

Un autre aspect des faiblesses des métriques pixel a été étudié par Cheng et al. (2021). Ils mettent en évidence le fait que la métrique IoU ne se concentre pas suffisamment sur les contours des objets qui sont les positions les plus importantes à détecter dans des tâches de détection d'objets. Ils présentent *Boundary IoU*, une nouvelle mesure d'évaluation de la segmentation axée sur la qualité des frontières. Ils analysent différents types d'erreurs sur différentes tailles d'objets et montrent que leur métrique est significativement plus sensible aux erreurs de frontière pour les objets de grande taille, sans pénaliser les erreurs sur les petits objets. De la même manière, ils proposent une adaptation de la précision moyenne se basant sur le *Boundary IoU*.

De plus, lorsqu'un objet prédit ne possède pas d'intersection avec un objet réel, leur IoU est égale à zéro. Cependant, la prédiction peut-être plus ou moins proche de la vérité, or l'IoU ne peut pas refléter ce phénomène. Ainsi, Rezatofighi et al. (2019) ont proposé la GIoU (*Generalized IoU*) qui permet de considérer l'espace entre les deux objets dans l'évaluation.

### Focus 3.1 – Précision et rappel

Définition

Dans le cadre de la détection d'objets, la précision est le nombre d'éléments (pixels ou objets) pertinents détectés, d'une classe considérée, rapporté au nombre total d'éléments détectés de cette classe. Elle tente de répondre à la question « Quelle proportion d'identifications positives est correcte ? ».
Le rappel est défini par le nombre d'éléments pertinents détectés rapporté au nombre total d'éléments annotés de la classe considérée. Il montre donc la proportion de positifs réels qui ont été correctement identifiés.

Équations

$$\mathrm{P} = \frac{TP}{TP + FP} \qquad \mathrm{R} = \frac{TP}{TP + FN} \tag{3.1}$$

avec :
— $TP$ : nombre de pixels ou d'objets positifs correctement prédits ;
— $FP$ : nombre de pixels ou d'objets négatifs prédits comme positifs ;
— $FN$ : nombre de pixels ou d'objets positifs prédits comme négatifs.

### Focus 3.2 – Intersection-sur-Union

Définition

La métrique Intersection-sur-Union (IoU) évalue la division entre la zone de chevauchement et la zone d'union entre deux régions. En d'autres termes, elle évalue le degré de chevauchement entre la vérité terrain et les prédictions. Elle est comprise entre 0 et 1, où 1 correspond à un chevauchement parfait entre la vérité terrain et la prédiction.

Équation

$$\mathrm{IoU} = \frac{TP}{TP + FP + FN} \tag{3.2}$$

avec :
— $TP$ : nombre de pixels positifs correctement prédits ;
— $FP$ : nombre de pixels négatifs prédits comme positifs ;
— $FN$ : nombre de pixels positifs prédits comme négatifs.

### Focus 3.3 – F1-score

Définition

La F-mesure, aussi appelé F1-score, est la moyenne harmonique entre la précision et le rappel.

ÉQUATION

$$\begin{aligned} \text{F1-score} &= \frac{2 \times TP}{2 \times TP + FP + FN} \\ &= 2 \times \frac{P \times R}{P + R} \end{aligned} \tag{3.3}$$

avec :

— $TP$ : nombre de pixels ou d'objets positifs correctement prédits ;
— $FP$ : nombre de pixels ou d'objets négatifs prédits comme positifs ;
— $FN$ : nombre de pixels ou d'objets positifs prédits comme négatifs.

### 3.3.2 MÉTRIQUES ORIENTÉES OBJETS

Bien que des évolutions des métriques usuelles au niveau pixel aient été proposées, elles ne fournissent toujours pas d'évaluation au niveau objet et ne permettent pas d'indiquer le nombre d'objets correctement détectés. Un des problèmes pour la mise en place d'une métrique objet est la difficulté à déterminer si un objet est correctement détecté ou non. Pour résoudre ce problème, des métriques conçues à l'origine dans la communauté de la recherche d'information (Information Retrieval) ont été adaptées aux images, et utilisées lors du PASCAL VOC Challenge 2012 pour calculer la précision au niveau des objets. Lors de cette compétition, la tâche de détection a été évaluée sur la base de la courbe Précision-Rappel au niveau objet, où les détections sont considérées comme de vrais ou de faux positifs en fonction de leur zone de recouvrement avec les objets de vérité terrain.

Suivant cette approche, TARRIDE et al. (2019) associent d'abord les objets prédits à ceux annotés et considèrent une prédiction comme un vrai positif si son IoU est supérieure à un seuil choisi $t = 0{,}65$. Ainsi, ils peuvent calculer la précision (P@.65), le rappel (R@.65) et la précision moyenne (mAP@.65) au niveau des objets. Les calculs de précision moyenne sont détaillés dans le Focus 3.4. SOULLARD et al. (2020) utilisent la mean Average Precision (mAP), c'est-à-dire la précision moyenne calculée pour différents seuils d'IoU, afin d'évaluer leur modèle de détection.

WOLF et al. (2006) ont montré l'importance de la qualité de détection (précision des objets détectés) et de la quantité de détections (nombre d'objets) lors de l'évaluation d'un système. La mesure mAP, qui est l'aire sous la courbe Précision-Rappel, permet d'évaluer la quantité de détections en fonction d'un critère de qualité donné : le seuil d'IoU. Afin de pouvoir mesurer autant la qualité que la quantité de détections, l'utilisation de la mAP moyennée sur une gamme de seuils d'IoU a émergé pour la détection d'objets. Ainsi, SOULLARD et al. (2020) ont utilisé cette moyenne mAP pour évaluer leur modèle de détection appliqué aux journaux historiques.

La métrique ZoneMap proposée par GALIBERT et al. (2015) évalue également les systèmes de détection au niveau objet et ne repose sur aucun seuil. Elle est basée sur les liens entre les zones d'hypothèse et de référence. Les forces des liens sont d'abord calculées : si une zone prédite est correcte, alors la force avec une zone de référence sera élevée. Au contraire, toutes les forces pour une zone faussement positive seront faibles. Ensuite, les zones sont regroupées en fonction de ces liens et chaque groupe reçoit une erreur de segmentation et une erreur de classification, calculées en fonction du type de groupe (*match*, *miss*, *false alarm*, *merge* ou *split*). Ces deux erreurs sont ensuite combinées pour donner une seule valeur. Même si cette métrique s'est avérée complémentaire de la métrique IoU dans l'évaluation du projet Maurdor (Oparin et al., 2014), elle n'est pas réellement utilisée à l'heure actuelle en raison de la complexité de ses calculs et de sa difficile applicabilité aux images comportant de nombreux objets.

### Focus 3.4 – Précision moyenne / Average precision

**Définition**

Le concept de la métrique d'évaluation de la précision moyenne est principalement lié aux compétitions PASCAL VOC. Basé sur un seuil défini d'IoU, elle considère les objets prédits comme vrais ou faux positifs, et calcule la précision moyenne grâce à l'aire sous la courbe de la précision par rapport au rappel.

La *mean Average Precision* (mAP) est définie de plusieurs manières selon la compétition. Dans le cas d'un problème à plusieurs classes, la mAP est définie comme étant la valeur moyenne des AP calculées pour chaque classe.

Elle peut aussi correspondre à la moyenne arithmétique réalisée sur plusieurs seuils d'IoU. En effet, pour s'abstenir d'un seuil prédéfini, la mAP est la moyenne des AP calculées pour plusieurs seuils. Dans le cas des compétitions PASCAL VOC, la moyenne est calculée sur des valeurs de seuil allant de 0,5 à 0,95 avec un pas de 0,05 (mAP@[.5 :.05 :.95] ou mAP@[.5,.95]).

**Mise en oeuvre**

- Toutes les prédictions sont ordonnées par leur confiance moyenne décroissante ;
- Les prédictions dont l'IoU est supérieur ou égal à un seuil $t$ sont considérées comme vrais positifs ;
- La courbe Précision-Rappel est construite à partir des prédictions ordonnées. Cette courbe Précision-Rappel permet d'évaluer les performances d'un détecteur d'objets en fonction d'un seuil sur la confiance associée à la prédiction. Il existe une courbe pour chaque classe d'objets. Un détecteur d'objets d'une classe particulière est considéré comme bon si sa précision reste élevée alors que le rappel augmente, ce qui signifie que si le seuil de confiance varie, la précision et le rappel resteront élevés. Habituellement, la courbe Précision-Rappel commence par des valeurs de précision élevées, qui diminuent à mesure que le rappel augmente ;
- La courbe est interpolée de telle sorte que la précision $p$ pour un rappel $r$ prenne la valeur de la précision maximale des rappels supérieurs à $r$ :

$$p_{interp}(r) = \max_{\tilde{r} \geqslant r} p(\tilde{r}) \tag{3.4}$$

- La précision moyenne (AP@$t$) est égale à l'aire sous la courbe Précision-Rappel interpolée.

ÉQUATIONS

Pour une classe donnée $c$ et un seuil d'IoU $t$, nous avons :

$$\text{AP@}t^c = \int_0^1 p_t^c(r_t^c)dr_t^c \tag{3.5}$$

Pour un seuil d'IoU $t$, la AP moyennée sur toutes les classes est calculée comme suit :

$$\text{mAP@}t = \frac{\sum_{c=1}^{C} AP@t^c}{C} \tag{3.6}$$

Pour une classe donnée $c$, la AP moyennée sur plusieurs seuils est calculée comme suit :

$$\text{mAP@[.5,.95]}^c = \frac{\sum_{t=0.5}^{0.95} AP@t^c}{10} \tag{3.7}$$

Enfin, la AP moyennée sur plusieurs seuils et l'ensemble des classes est calculée comme suit :

$$\text{mAP} = \frac{\sum_{c=1}^{C} mAP@[.5,.95]^c}{C} \quad \text{ou} \quad \text{mAP} = \frac{\sum_{t=0.5}^{0.95} \text{mAP@}t}{10} \tag{3.8}$$

avec :
- $C$ : le nombre de classes ;
- $p$ : la précision ;
- $r$ : le rappel.

### 3.3.3 MÉTRIQUES ORIENTÉES VERS LA TÂCHE FINALE

TRIER et al. (1995) ont montré l'importance d'une évaluation orientée vers la tâche finale pour les méthodes de binarisation, puisque l'évaluation par un expert humain dépend de ses critères visuels. Ils ont appliqué onze méthodes de binarisations adaptatives locales à des images de test avant de transmettre les résultats à un module de reconnaissance OCR. Les méthodes de binarisation ont ensuite été comparées avec les taux de reconnaissance, d'erreur et de rejet. Ils ont montré que les classements des méthodes en fonction de la qualité de binarisation et des taux d'erreurs obtenus après le module OCR étaient différents, insistant sur l'importance d'utiliser des métriques orientées vers la tâche finale. De même, WOLF et al. (2006) ont montré l'importance de ces mesures pour la détection d'objets. Dans un cadre de prédiction pixel, les objets sont reconstruits en regroupant les pixels connectés. Or, les objets extraits sont similaires même si quelques pixels ont été mal prédits. Dans ce sens, une de nos propositions vise à utiliser des métriques de reconnaissance de texte (HTR) pour évaluer les systèmes de détection de lignes de texte. A notre connaissance, il n'existe pas de travaux antérieurs dans la littérature à ce sujet.

# 4 DÉTECTION D'OBJETS DANS DES IMAGES DE DOCUMENTS

La compréhension automatique de documents, et plus particulièrement l'analyse de la mise en page de documents historiques, est toujours un domaine de recherche actif. Cette tâche consiste à diviser un document en différentes régions en fonction de leur contenu. La grande variété des documents existants rend cette tâche très complexe. Pour détecter des objets dans des images de documents, de nombreux systèmes ont été proposés, la plupart assignant une classe à chaque pixel de l'image donnée en entrée. Bien que ces systèmes aient montré des performances intéressantes, ils nécessitent un grand nombre de données d'apprentissage annotées et présentent des temps d'inférence longs. Dans ce chapitre, nous présentons deux architectures à l'état de l'art et comparons leurs avantages et inconvénients. Par la suite, nous proposons, en section 4.3, un système appelé Doc-UFCN mis au point afin de dépasser les limitations mises en évidence par la comparaison des systèmes à l'état de l'art. Nous présentons enfin une comparaison expérimentale des systèmes pour la détection des lignes de texte, en section 4.4, et la détection d'actes, en section 4.5.

## 4.1 PRÉSENTATION DU PROBLÈME

L'entraînement de modèles de détection d'objets dans les images de documents requiert un grand nombre de données annotées. Cependant, dans le cas de documents historiques, ces données annotées sont rarement disponibles. Pour pallier ce problème, des systèmes utilisant des poids pré-entraînés tels que dhSegment (Ares Oliveira et al., 2018) ont été proposés. Cela permet d'utiliser des réseaux avec plus de paramètres sur des jeux de données de tailles réduites. De plus, l'utilisation du pré-entraînement a montré de nombreux avantages tels que la diminution du temps d'apprentissage et l'amélioration de la précision du modèle. Cependant, ces poids sont souvent appris sur des images de scènes naturelles (ImageNet (Deng et al., 2009)), puis appliqués à des images de documents, ce qui pose un problème d'adaptation des modèles à un nouveau domaine. De plus, bien qu'ils obtiennent des performances élevées, les systèmes actuellement à l'état de l'art peuvent présenter des temps d'inférence longs qui peuvent avoir de grands impacts financiers et écologiques. Dans un cadre industriel, l'utilisation de tels systèmes ne semble pas appropriée. C'est pourquoi, nous proposons un nouveau modèle, appelé Doc-UFCN, dans l'optique de répondre à ces problématiques de temps de traitement. Les contraintes étant que ce système possède un nombre réduit de paramètres et présente un temps d'inférence réduit par rapport aux systèmes existants, tout en obtenant des performances à l'état de l'art.

## 4.2 SYSTÈMES À L'ÉTAT DE L'ART

De nombreux modèles (Barakat et al., 2018 ; Mechi et al., 2019 ; Renton et al., 2018) ont été proposés pour la détection d'objets dans les images de documents, notamment pour la détection de lignes de texte. Ces modèles ont des architectures similaires, suivant une architecture U-Net (Ronneberger et al., 2015). Le modèle dhSegment suit également l'architecture U-Net, mais, contrairement aux autres systèmes, il intègre une partie pré-entraînée et a été testé sur de nombreuses tâches telles que la détection de pages, de décorations ou encore de photographies. C'est pourquoi, nous avons choisi de nous comparer à ce système. Il en est de même pour le modèle de Yang et al. (2017) qui intègre l'information textuelle afin d'aider la détection des objets et qui a obtenu de bonnes performances sur des images de documents modernes. Ces deux systèmes sont détaillés dans les Focus 2.9 et 2.10.

## 4.3 ARCHITECTURE DU SYSTÈME PROPOSÉ : DOC-UFCN

Nous présentons, dans cette section, l'architecture du modèle que nous proposons Doc-UFCN. Notre objectif est de concevoir un modèle comportant peu de paramètres afin d'être entraîné sur des jeux de données restreints. De plus, celui-ci devra montrer des temps d'inférence réduits par rapport aux modèles à l'état de l'art. Le développement de ce modèle étant réalisé dans un cadre industriel, il est nécessaire d'avoir un modèle rapide en inférence, capable de traiter des millions d'images de documents dans des délais raisonnables. Enfin, ce modèle doit répondre à ces différents points tout en obtenant des performances élevées.

Pour concevoir notre système, Doc-UFCN, nous avons choisi d'utiliser le cœur du réseau de Yang et al. (2017) car il possède un nombre réduit de paramètres et ne contient pas de parties pré-entraînées. Pour réduire davantage le nombre de paramètres et être capable d'entraîner notre modèle sur peu de données, seul le contenu visuel est utilisé. Par conséquent, la carte d'incorporation de texte, le pont et le second décodeur pour la tâche de reconstruction ne sont pas intégrés. Notre architecture est donc un réseau entièrement convolutif (FCN) en forme de U composé d'un encodeur (blocs rouges sur la Figure 4.1) suivi d'un décodeur (blocs bleus) et d'une couche de convolution finale. L'encodeur de Doc-UFCN diffère de celui de dhSegment puisqu'il ne suit pas l'architecture ResNet-50. Notre encodeur possède beaucoup moins de paramètres et est entièrement entraîné sur des images de documents. Cependant, les deux systèmes ont des décodeurs similaires avec des connexions résiduelles où les cartes de caractéristiques calculées pendant l'encodage, à une échelle donnée, sont utilisées pendant l'étape de décodage de cette même échelle.

L'utilisation d'un FCN sans couche dense présente de nombreux avantages. Tout d'abord, il réduit fortement le nombre de paramètres puisqu'il n'y a aucune connexion dense. De plus, cela permet au réseau de traiter des images de taille variable et de conserver les informations spatiales telles quelles.

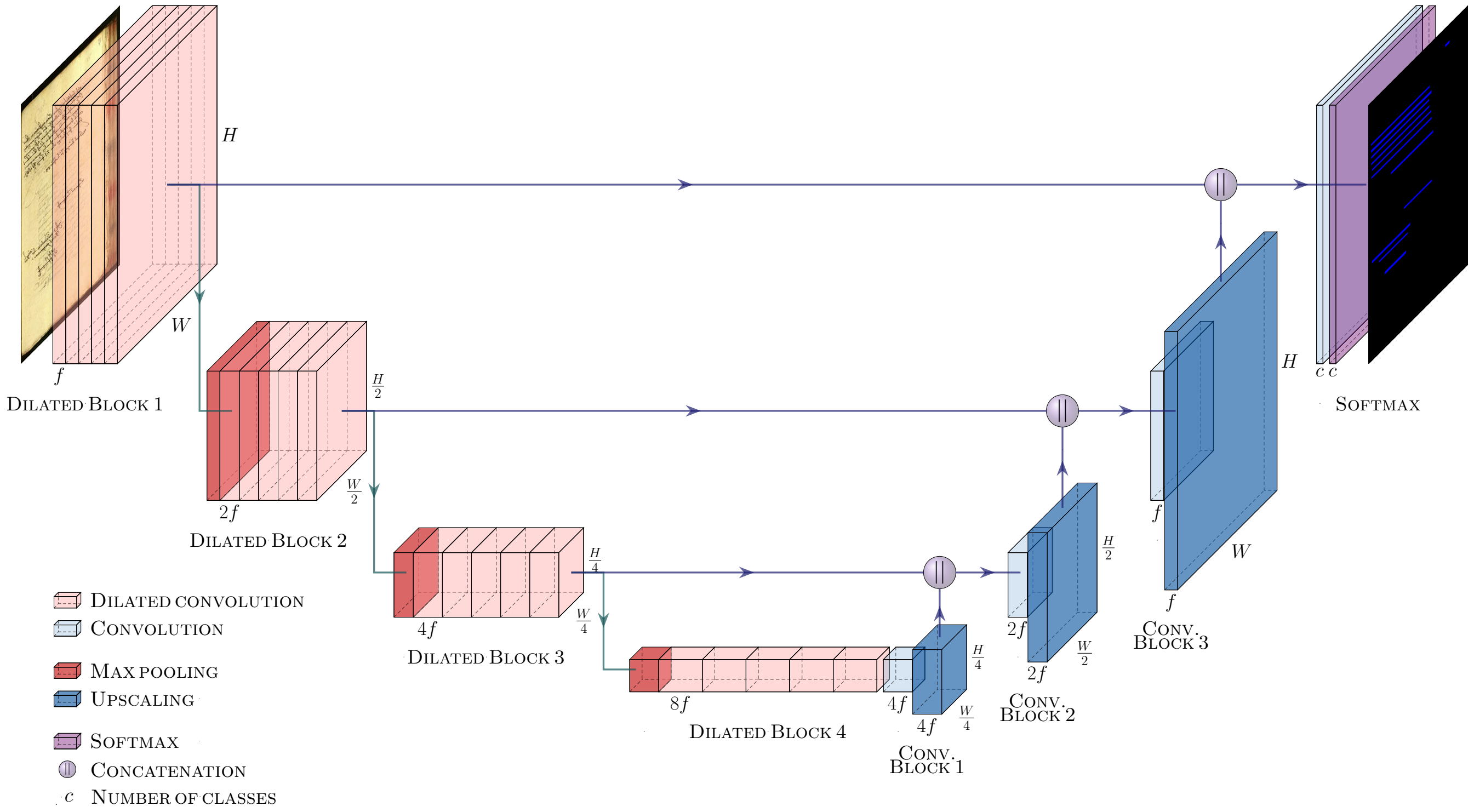

**Figure 4.1 –** Schéma de l'architecture du modèle Doc-UFCN. L'encodeur est représenté en rouge et le décodeur en bleu avec respectivement $H$ et $W$ les hauteur et largeur de l'image d'entrée et $f$ le nombre de cartes de caractéristiques.

### 4.3.1 ENCODEUR

L'encodeur vise à extraire les caractéristiques importantes de l'image d'entrée. Il consiste en quatre blocs dilatés. Ceux-ci sont légèrement différents de ceux présentés par Yang et al. (2017) puisqu'ils consistent en cinq convolutions dilatées consécutives. L'utilisation de convolutions dilatées au lieu de convolutions standards permet au champ réceptif d'être plus grand et au réseau d'avoir plus d'informations contextuelles. De plus, exécuter ces convolutions successivement plutôt qu'en parallèle permet d'agrandir le champ réceptif. Chaque bloc est suivi d'une couche de *max-pooling*, sauf le dernier.

### 4.3.2 DÉCODEUR

L'objectif du décodeur est de reconstruire l'image d'entrée avec un étiquetage pixel par pixel à la résolution de l'image d'entrée originale. Cette déconvolution est généralement effectuée à l'aide de convolutions transposées ou d'une mise à l'échelle. Comme suggéré par Mechi et al. (2019), nous avons décidé de remplacer les couches de déconvolution du système de Yang par des convolutions transposées afin de conserver la même résolution en entrée et en sortie. Par conséquent, le chemin de décodage est composé de trois blocs convolutifs, chacun consistant en une convolution standard suivie d'une convolution transposée. De plus, les caractéristiques calculées lors de l'étape d'encodage sont concaténées avec celles de l'étape de décodage (flèches violettes sur la Figure 4.1).

La dernière couche convolutive produit des cartes de caractéristiques en pleine résolution. Elle renvoie $C$ cartes de caractéristiques de la même taille que l'image d'entrée, $C$ étant le nombre de classes concernées dans l'expérience. Une couche *softmax* est ensuite appliquée pour transformer ces cartes de caractéristiques en cartes de probabilités.

### 4.3.3 DÉTAILS D'IMPLÉMENTATION

Nous donnons, dans cette section, des détails techniques sur l'implémentation utilisée lors de nos expériences.

#### TAILLE DES IMAGES EN ENTRÉE

Nous avons décidé d'utiliser la même taille d'image d'entrée que celle de YANG et al. (2017). Les images d'entrée ainsi que leurs vérités terrain sont donc redimensionnées en images plus petites telles que la plus grande dimension de l'image soit égale à 384 pixels tout en conservant le ratio de l'image originale. Cela permet de réduire le temps d'apprentissage et de traitement sans perdre trop d'informations. Nous avons également entraîné le modèle sur une taille d'entrée plus grande, de 768 pixels, pour voir l'impact de ce choix (voir en section 4.4.4).

#### BLOCS DILATÉS

Comme indiqué précédemment, tous les blocs dilatés sont composés de cinq convolutions dilatées consécutives avec des taux de dilatation de 1, 2, 4, 8 et 16. Les blocs comportent respectivement 32, 64, 128 et 256 filtres. Chaque convolution a un noyau de taille 3×3, un stride de 1 et un padding adapté pour garder la même taille de tenseur dans tout le bloc. Toutes les convolutions des blocs sont suivies d'une couche de *batch normalization*, d'une activation ReLU et d'une couche de *dropout*. Le modèle comportant peu de paramètres, les couches de *dropout* permettent d'éviter le sur-apprentissage.

#### BLOCS CONVOLUTIFS

Les blocs convolutifs sont utilisés pendant l'étape de décodage. Le décodeur est composé de trois blocs convolutifs et chaque bloc est composé d'une convolution standard suivie d'une convolution transposée. Les blocs ont respectivement 128, 64 et 32 filtres. Chaque convolution standard a un noyau de taille 3×3, un stride et un padding de 1. Chaque convolution transposée a un noyau de taille 2×2 et un stride de 2. Comme pour les blocs dilatés, toutes les convolutions standards et transposées sont suivies d'une couche de normalisation, d'une activation ReLU et d'une couche de *dropout*.

La dernière couche de convolution est paramétrée comme suit : $C$ (nombre de classes) filtres, noyau 3×3, stride et padding de 1. Elle est suivie d'une couche *softmax* qui calcule les probabilités de chaque pixel d'appartenir aux $C$ classes.

Table 4.1 – Statistiques des jeux de données utilisés pour la détection de lignes de texte.

| Jeu de données | Images | | | Lignes | | |
|---|---|---|---|---|---|---|
| | train | valid | test | train | valid | test |
| Balsac<br>Vézina et al. (2020) | 730 | 92 | 91 | 36 941 | 4 592 | 4 323 |
| DIVA-HisDB<br>Simistira et al. (2016) | 60 | 30 | 30 | 6 037 | 2 999 | 2 897 |
| Horae<br>Boillet et al. (2019) | 522 | 20 | 30 | 12 568 | 270 | 958 |
| READ-BAD<br>Grüning et al. (2017) | 388 | 49 | 49 | 22 885 | 2 699 | 2 481 |

#### POST-TRAITEMENT

Comme étape de post-traitement, nous appliquons les mêmes opérations que celles appliquées par dhSegment : les pixels ayant une probabilité supérieure à un seuil $t$ sont conservés comme appartenant à la classe correspondante, les autres étant assignés au fond. Les composantes connexes composées de moins de $min_cc$ pixels sont supprimées.

## 4.4 EXPÉRIENCES DE DÉTECTION DE LIGNES DE TEXTE

Dans cette section, nous comparons Doc-UFCN et dhSegment sur une tâche de détection de lignes de texte. Nous montrons que Doc-UFCN obtient de meilleures performances tout en ayant moins de paramètres et un temps de prédiction réduit. Nous présentons tout d'abord les données utilisées, puis nous discutons des entraînements et des résultats obtenus.

### 4.4.1 JEUX DE DONNÉES

Les systèmes sont comparés sur quatre jeux de données annotés pour la détection de lignes de texte : Balsac (Vézina et al., 2020), DIVA-HisDB (Simistira et al., 2016), Horae (Boillet et al., 2019) et READ-BAD (Grüning et al., 2017). La Table 3.1 présente les détails de ces bases et la Table 4.1 en présente les statistiques.

Ces jeux de données sont très différents, ce qui permet de tester les systèmes sur des tâches à complexité variable. En effet, la base Balsac contient des pages avec uniquement du texte réparti en actes. Il s'agit de documents structurés semblables les uns aux autres. Les pages de la base DIVA-HisDB ne contiennent également que du texte, mais les mises en page sont plus complexes avec des commentaires dans les marges et entre les lignes. Les images de la base Horae présentent des pages hétérogènes qui peuvent contenir des illustrations, une quantité variable de lignes de texte et d'initiales. Enfin, READ-BAD comporte deux sous-ensembles, l'un dit "simple" et l'autre "complexe", qui permettent d'évaluer et de comparer les systèmes sur une grande diversité de données.

### 4.4.2 RÉSULTATS ET DISCUSSION

Nous avons entraîné Doc-UFCN et dhSegment dans les mêmes conditions sur les quatre jeux de données. Cette section détaille les entraînements et présente les résultats obtenus.

#### DÉTAILS DES ENTRAÎNEMENTS

Doc-UFCN est implémenté à l'aide du framework PyTorch. Nous l'avons entraîné avec un *learning rate* initial de $5e-3$, l'optimiseur Adam (Kingma et al., 2015) et la fonction de coût d'entropie croisée. Les poids sont initialisés grâce à l'initialisation Glorot (Glorot et al., 2010). De plus, nous avons utilisé des mini-batchs de taille 4 pour réduire le temps d'apprentissage. Nous avons testé différentes probabilités de *dropout* et décidé de conserver une probabilité de 0,4, car elle correspond au modèle ayant donné les meilleures performances, en moyenne, sur l'ensemble de validation. Le modèle est entraîné sur un maximum de 200 époques et nous gardons le meilleur modèle en validation.

Nous avons également entraîné dhSegment sur ces mêmes données pour un maximum de 60 époques puisque le modèle est pré-entraîné et converge plus rapidement que le nôtre. Nous avons utilisé des mini-batchs de taille 4 et des patchs de taille 400×400 pixels. Le *learning rate* initial est de $5e-5$ et nous avons choisi d'utiliser un ResNet-50 (He et al., 2016) comme encodeur pré-entraîné puisque les bonnes performances présentées dans Ares Oliveira et al. (2018) ont été obtenues avec ResNet. Comme pour Doc-UFCN, le meilleur modèle obtenu pendant l'apprentissage sur l'ensemble de validation est sélectionné.

Les deux modèles ont la même étape de post-traitement avec les mêmes hyper-paramètres. Après avoir comparé des valeurs de seuil allant de 0,5 à 0,9, nous avons conservé le seuil $t=$ 0,7 qui permet d'obtenir les meilleurs résultats sur l'ensemble de validation, avec une bonne capacité d'acceptation des pixels attendus comme des lignes de texte et de rejet des pixels appartenant au fond. Enfin, les composantes connexes de moins de $min_cc = 50$ pixels sont écartées. Plusieurs valeurs ont également été comparées pour ce paramètre, cependant, il n'a que peu d'impact sur les résultats.

#### ÉVALUATION DES MODÈLES

La plupart des méthodes existantes sont évaluées avec la métrique IoU. Cette métrique mesure la similarité moyenne entre les pixels prédits et les pixels de la vérité terrain. Alberti et al. (2017) ont conçu un outil pour évaluer la performance d'un modèle en calculant l'IoU, la précision, le rappel et la F-mesure. Nous avons utilisé cet outil pour évaluer les modèles car il permet d'obtenir plus d'informations concernant les performances du modèle au niveau du pixel que l'IoU seule.

Par conséquent, pour évaluer les modèles, nous avons calculé différentes métriques au niveau du pixel. Nous rapportons l'IoU ainsi que la précision (P), le rappel (R) et le score F1 dans la Table 4.2. Pour être comparables, les images prédites par dhSegment sont redimensionnées de sorte que leur plus grande dimension soit égale à 384 pixels avant de calculer les métriques. De plus, les valeurs ne sont présentées que pour la classe des lignes de texte (le fond n'est pas considéré ici).

**Table 4.2** – Résultats obtenus par Doc-UFCN et dhSegment au niveau pixel. Résultats donnés sur les ensembles de test pour la tâche de détection de lignes de texte.

| Jeu de données | Système | IoU | P | R | F1-score |
|---|---|---|---|---|---|
| Balsac | Doc-UFCN | **0,84** | **0,95** | **0,88** | **0,91** |
| | dhSegment | 0,74 | 0,92 | 0,79 | 0,85 |
| DIVA-HisDB | Doc-UFCN | **0,76** | **0,92** | **0,81** | **0,86** |
| | dhSegment | 0,74 | **0,92** | 0,79 | 0,85 |
| Horae | Doc-UFCN | 0,64 | **0,78** | 0,80 | **0,85** |
| | dhSegment | **0,65** | 0,72 | **0,89** | 0,82 |
| READ-Simple | Doc-UFCN | 0,64 | 0,82 | **0,76** | **0,77** |
| | dhSegment | **0,65** | **0,85** | 0,72 | **0,77** |
| READ-Complex | Doc-UFCN | **0,54** | **0,84** | **0,62** | **0,73** |
| | dhSegment | 0,53 | 0,79 | 0,59 | 0,69 |

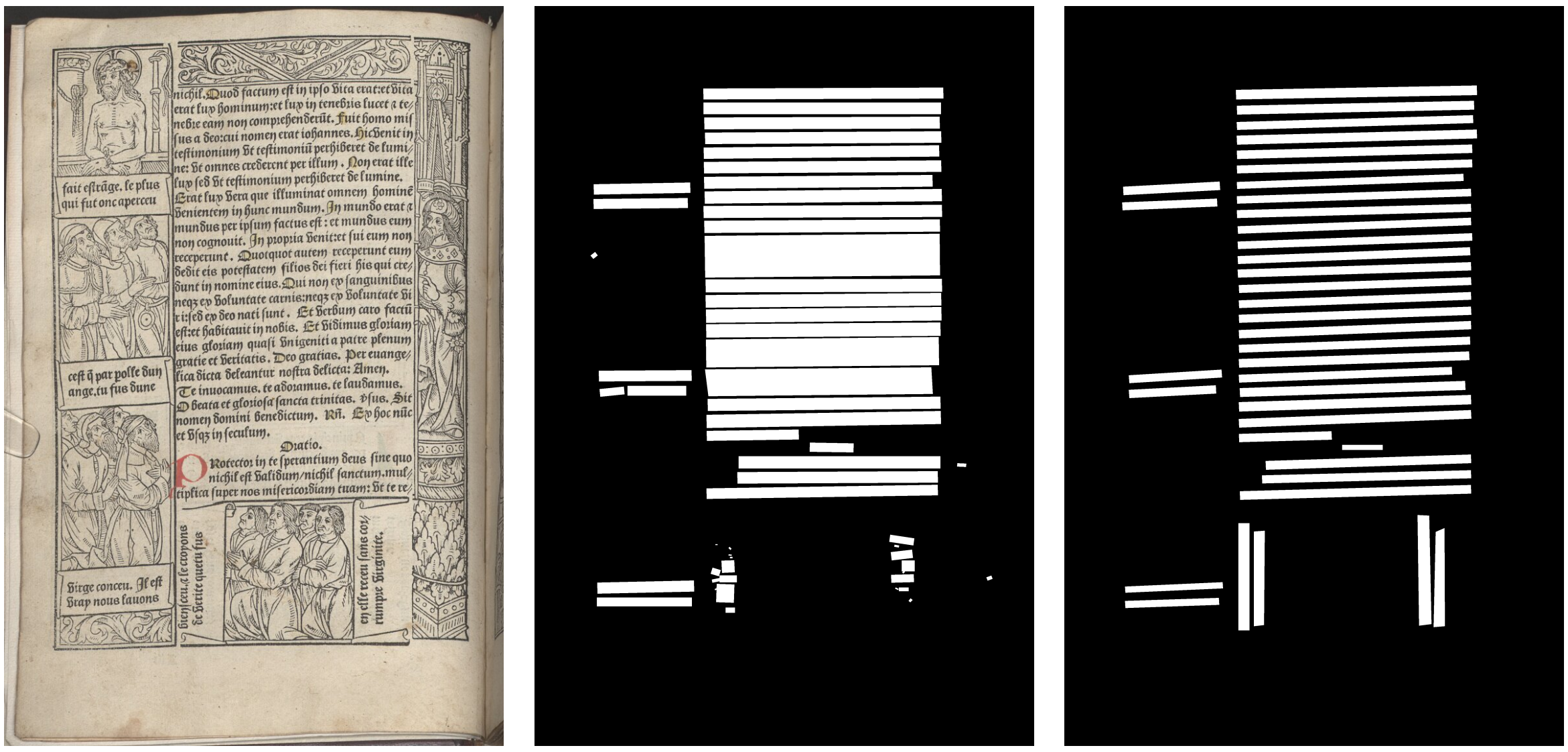

**Figure 4.2** – Détections de lignes produites par dhSegment, au centre, et notre modèle Doc-UFCN, à droite, sur l'image de la page 5 verso du Livre d'heures *Horae*[1].

Les résultats obtenus par notre méthode sont en moyenne supérieurs à ceux obtenus par dhSegment. Sur le jeu de données Balsac, notre modèle surpasse dhSegment jusqu'à 6 points pour la métrique F1-score. Nous observons également des gains respectifs de 3 et 4 points de F1-score pour les bases Horae et READ-Complex. Ces résultats s'expliquent par une meilleure séparation des lignes de texte proches qui sont souvent fusionnées par dhSegment (Figure 4.2). Notre modèle aide à séparer ces lignes là où dhSegment échoue, mais permet également d'obtenir des contours plus lisses et plus précis. Enfin, dhSegment semble avoir plus de difficultés à détecter les lignes verticales et prédit parfois de très petites lignes dans les miniatures, contrairement à Doc-UFCN qui semble plus robuste sur ce type d'images.

1. https://www.digitale-sammlungen.de/en/view/bsb00070331

Table 4.3 – Temps d'inférence, en secondes par image, rapportés pour Doc-UFCN et dhSegment, et calculés sur les ensembles de test. La colonne RATIO contient les ratios d'amélioration (dhSegment/Doc-UFCN).

| JEU DE DONNÉES | TEMPS D'INFÉRENCE† | | RATIO |
|---|---|---|---|
| | Doc-UFCN | dhSegment | |
| BALSAC | **0,41** | 2,95 | 7,20 |
| DIVA-HISDB | **0,80** | 12,90 | 16,13 |
| HORAE | **0,97** | 7,87 | 8,11 |
| READ-SIMPLE | **0,45** | 3,73 | 8,29 |
| READ-COMPLEX | **0,59** | 4,70 | 7,97 |

† Prédictions faites sur une carte graphique GeForce RTX 2070 8G.

#### COMPARAISON DES MODÈLES

Jusqu'à présent, notre modèle a obtenu, en moyenne, de meilleures performances que dhSegment bien qu'il ne bénéficie pas d'un encodeur pré-entraîné. Un autre point intéressant est que notre modèle comporte moins de paramètres à apprendre que dhSegment : 4,1 millions pour Doc-UFCN contre 32,8 millions pour dhSegment, dont 9,36 millions, correspondant au décodeur non pré-entraîné, qui doivent être entièrement entraînés. Cette diminution du nombre de paramètres conduit à une réduction significative du temps de prédiction : Doc-UFCN est jusqu'à 16 fois plus rapide que dhSegment, comme illustré dans la Table 4.3. Cette réduction du temps d'inférence peut également s'expliquer par le fait que dhSegment utilise des patchs de taille 400×400 pixels. Ainsi, pour la base DIVA-HisDB, il devra prédire en moyenne 117 patchs de cette taille, là où notre modèle ne fera qu'une seule prédiction de taille moyenne 768×512 pixels.

Grâce à ces premières expériences, nous avons montré que notre modèle Doc-UFCN obtient de meilleures performances que dhSegment tout en étant plus rapide en inférence. Nous allons maintenant étudier l'impact du pré-entraînement sur les résultats finaux.

### 4.4.3 IMPACT DU PRÉ-ENTRAÎNEMENT

Nous analysons maintenant l'impact du pré-entraînement sur des images de documents. dhSegment est pré-entraîné sur des images de scènes naturelles (DENG et al., 2009), ce qui lui permet d'obtenir des résultats satisfaisants, même avec peu de données annotées. Nous nous questionnons donc sur l'intérêt que pourrait avoir un pré-entraînement de Doc-UFCN sur des images de documents. Pour cela, nous avons entraîné Doc-UFCN ainsi que dhSegment sur l'ensemble des quatre jeux de données présentés en section 4.4.1. Ces modèles ont été entraînés avec 1700 images d'entraînement et 191 de validation, dans les mêmes conditions que les expériences précédentes, puis évalués sur chaque base indépendamment. Les deux modèles ainsi obtenus sont dits "génériques" dans la suite de cette section. Les résultats obtenus sont résumés dans la Table 4.4. Pour plus de lisibilité, seules les valeurs d'IoU et de F1 sont présentées dans la Table. En plus des résultats des modèles pré-entraînés (génériques), nous montrons, dans cette table, les résultats après adaptation (*fine tuning*) des modèles génériques sur chaque jeu de données.

Table 4.4 – Résultats obtenus par Doc-UFCN et dhSegment au niveau pixel pour la tâche de détection de lignes de texte. Les résultats montrent les performances des modèles génériques sur les ensembles de test avec et sans adaptation.

| Jeu de données | Système | IoU | | F1-score | |
|---|---|---|---|---|---|
| | | Générique | Adapté | Générique | Adapté |
| Balsac | Doc-UFCN | 0,85 | **0,86** | **0,92** | **0,92** |
| | dhSegment | 0,74 | 0,75 | 0,85 | 0,85 |
| DIVA-HisDB | Doc-UFCN | **0,75** | **0,75** | **0,85** | **0,85** |
| | dhSegment | 0,73 | 0,74 | 0,84 | **0,85** |
| Horae | Doc-UFCN | **0,69** | 0,68 | **0,89** | 0,88 |
| | dhSegment | 0,61 | 0,63 | 0,82 | 0,80 |
| READ-Simple | Doc-UFCN | **0,68** | **0,68** | 0,79 | 0,79 |
| | dhSegment | 0,65 | 0,64 | **0,81** | 0,77 |
| READ-Complex | Doc-UFCN | **0,60** | **0,60** | **0,78** | **0,78** |
| | dhSegment | 0,53 | 0,53 | 0,68 | 0,69 |

### COMPARAISON DES SYSTÈMES

Les résultats obtenus par les modèles Doc-UFCN et dhSegment génériques confirment ceux obtenus par les modèles spécifiques présentés en section 4.4.2. En effet, avec et sans adaptation, Doc-UFCN obtient quasiment toujours de meilleurs résultats que dhSegment puisqu'il obtient des valeurs d'IoU et de F1 plus élevées.

### PRÉ-ENTRAÎNEMENT

La Figure 4.3 compare les résultats obtenus par le modèle Doc-UFCN générique par rapport aux modèles spécifiques présentés précédemment. Nous pouvons noter que, pour les deux métriques, le modèle générique obtient de meilleures performances sur toutes les bases sauf sur DIVA-HisDB où il perd un point d'IoU et de F1. Sur les autres jeux de données, nous observons des gains allant jusqu'à 6 points d'IoU, ce qui indique que le modèle a réussi à apprendre des caractéristiques suffisamment génériques pour représenter des données très variées, et qu'il a donc réussi à tirer profit de chaque jeu de données, malgré des données de pré-entraînement non équilibrées.

### ADAPTATION

Nous observons également sur la Figure 4.3 qu'adapter le modèle générique à chaque jeu de données n'apporte que très peu voire aucun gain de performance. Notre hypothèse est que le modèle générique avait déjà appris au mieux sur ces données, l'adaptation n'apportant pas de nouvelles données. Ce comportement peut être expliqué par le faible nombre de paramètres que comporte le modèle et la taille réduite des jeux de données annotés.

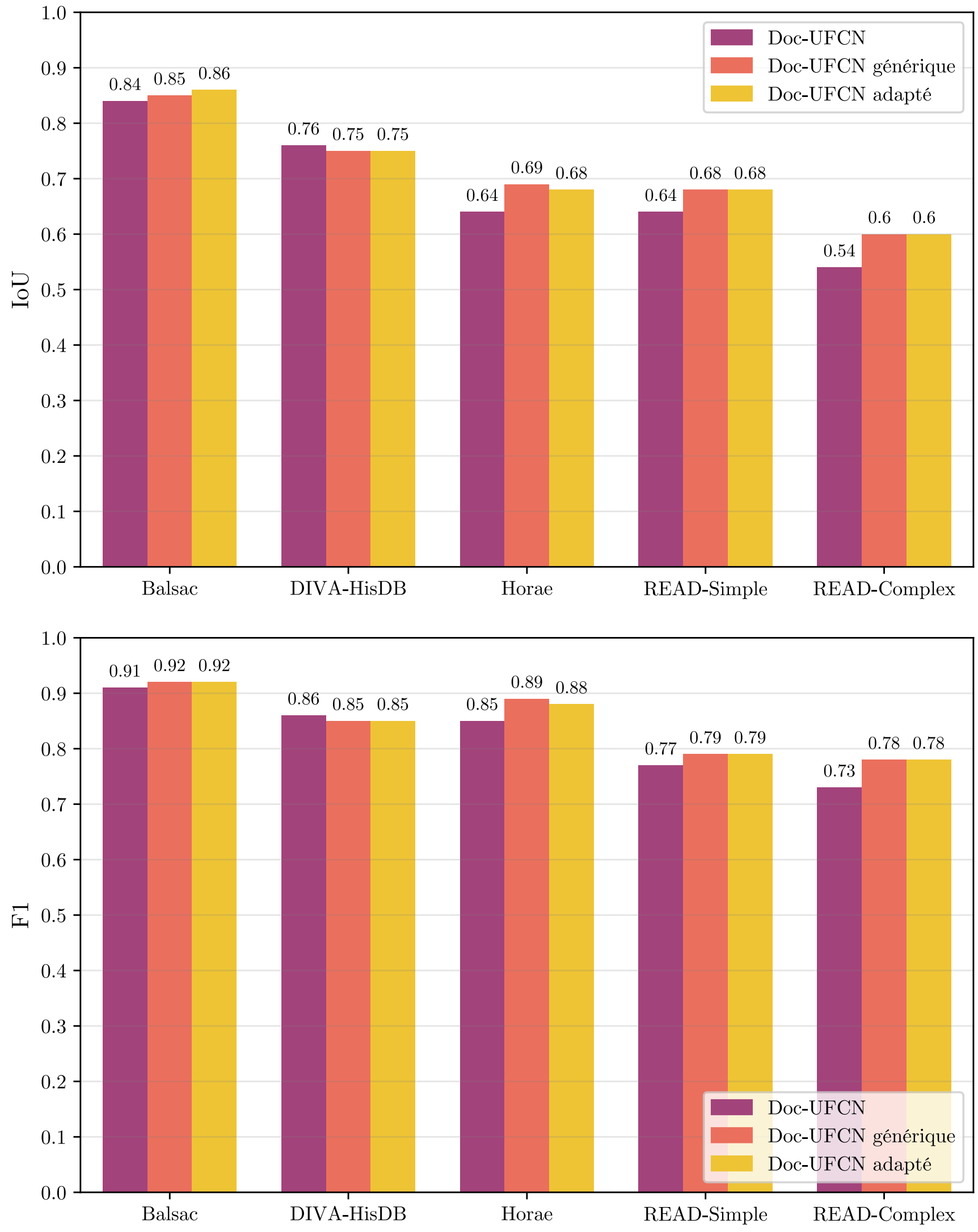

**Figure 4.3 –** Impact du pré-entraînement de Doc-UFCN, évalué sur les ensembles de test. Les résultats montrent les performances des modèles génériques avec et sans adaptation.

Pour conclure, nous avons montré qu'utiliser un modèle générique permet d'améliorer la qualité des détections par rapport à un modèle spécifique, même avec une quantité limitée de données. Dans le chapitre 5, nous cherchons à aller plus loin dans cette idée en analysant et levant les défis liés à l'obtention d'un modèle encore plus générique et robuste, obtenant de bonnes performances sur de nouvelles données sans aucune adaptation.

### 4.4.4 ÉTUDE ABLATIVE

Après avoir démontré l'intérêt de notre modèle Doc-UFCN pour la détection de lignes de texte, nous synthétisons ici les expérimentations réalisées afin de valider nos choix d'architecture et d'évaluer l'impact de certains composants et hyper-paramètres sur les performances finales. Les paramètres étudiés dans les paragraphes suivants sont l'utilisation de la normalisation des batchs, l'utilisation du *dropout*, les taux de dilatation dans les blocs dilatés et la taille des images en entrée. Les Tables 4.5, 4.6 et 4.7 regroupent les résultats de cette étude.

**Table 4.5 –** Étude ablative de Doc-UFCN sur la détection de lignes de texte. Résultats donnés pour les ensembles de test. "BN" indique l'utilisation de la *batch normalization* durant l'entraînement.

| Jeu de données | Version | IoU | F1-score |
|---|---|---|---|
| Balsac | ∅ | 0,79 | 0,88 |
| | BN | 0,81 | 0,89 |
| | BN + dropout | **0,84** | **0,91** |
| DIVA-HisDB | ∅ | 0,41 | 0,56 |
| | BN | 0,74 | 0,85 |
| | BN + dropout | **0,76** | **0,86** |
| Horae | ∅ | 0,56 | 0,78 |
| | BN | **0,64** | 0,81 |
| | BN + dropout | **0,64** | **0,85** |
| READ-Simple | ∅ | 0,59 | 0,73 |
| | BN | 0,58 | 0,72 |
| | BN + dropout | **0,64** | **0,77** |
| READ-Complex | ∅ | 0,39 | 0,56 |
| | BN | 0,50 | 0,69 |
| | BN + dropout | **0,54** | **0,73** |

### Batch normalization

La normalisation des batchs appliquée durant l'entraînement de modèles neuronaux est souvent utilisée, car elle a un impact positif sur la vitesse de convergence, mais parfois aussi sur les performances (Ioffe et al., 2015). L'entraînement de Doc-UFCN avec cette normalisation confirme ces résultats puisque le modèle a convergé plus de deux fois plus rapidement par rapport au modèle sans normalisation. D'après la Table 4.5, la normalisation a également un réel impact sur les valeurs de F1, en particulier pour Diva-HisDB, Horae et READ-Complex. En plus de ces valeurs quantitatives, nous avons noté que les résultats visuels sont meilleurs avec normalisation. Elle aide à séparer les objets proches mais aussi à joindre les parties d'objets qui, sans normalisation, étaient sur-segmentés. En outre, les contours des objets prédits sont souvent plus précis et plus lisses.

### Dropout

Le *dropout* (Srivastava et al., 2014) est également souvent utilisé dans les réseaux de neurones profonds, car il permet notamment de limiter le sur-apprentissage. Nos expériences, présentées en Table 4.5, montrent qu'entraîner avec *dropout* permet, en effet, d'obtenir de meilleures performances sur toutes les bases, d'autant plus que notre modèle comporte assez peu de paramètres.

### Taux de dilatation

Lors de la conception de notre modèle, nous avons choisi d'utiliser une version modifiée du bloc dilaté proposé par Yang et al. (2017) dans le but de prendre en compte davantage de contexte pour prédire les lignes de texte. Pour justifier nos choix de taux de dilatation, nous avons testé quatre configurations sur le jeu de données Balsac. Ainsi, nous avons entraîné des modèles avec des blocs dilatés configurés comme suit :

- 1 convolution et un taux de dilatation de 1 : [1] ;
- 1 convolution et un taux de dilatation de 16 : [16] ;
- 5 convolutions et des taux de dilatation de 1 : [1, 1, 1, 1, 1] ;
- 5 convolutions et des taux de dilatation de 1, 2, 4, 8 et 16 : [1, 2, 4, 8, 16].

**Table 4.6 –** Impact du taux de dilatation dans les blocs d'encodeur de Doc-UFCN sur la détection de lignes de texte. Résultats donnés pour l'ensemble de test du jeu de données Balsac.

| Dilatation | IoU | F1-score |
|---|---|---|
| [1] | 0,76 | 0,86 |
| [1, 1, 1, 1, 1] | 0,80 | 0,89 |
| [16] | 0,77 | 0,87 |
| [1, 2, 4, 8, 16] | **0,84** | **0,91** |

**Table 4.7 –** Impact de la taille des images en entrée de Doc-UFCN sur la détection des lignes de texte. Résultats donnés pour les ensembles de test.

| Jeu de données | Taille | IoU | F1-score |
|---|---|---|---|
| Balsac | 384 | 0,84 | 0,91 |
| | 768 | **0,87** | **0,93** |
| DIVA-HisDB | 384 | 0,76 | 0,86 |
| | 768 | **0,77** | **0,87** |

Les résultats obtenus sont présentés dans la Table 4.6. Les résultats de la dernière configuration sont meilleurs que ceux des autres puisque le champ réceptif est beaucoup plus grand et que le modèle considère davantage de contexte pour prédire les lignes de texte. Le fait d'avoir des convolutions dilatées au lieu de convolutions standards a un réel impact sur la taille du champ réceptif, ce qui permet d'utiliser plus de contexte pour prédire les lignes de texte et d'obtenir de meilleures performances.

### Taille des images d'entrée

Comme indiqué précédemment, nous avons choisi, pour Doc-UFCN, d'utiliser la même taille d'image en entrée que celle utilisée par Yang et al. (2017). C'est pourquoi, pour tous les résultats présentés jusqu'à maintenant, les images étaient redimensionnées de sorte que leur plus grande dimension soit égale à 384 pixels, tout en conservant l'aspect de l'image. Cependant, il est important de connaître l'impact de ce choix sur les résultats du modèle. Dans cette optique, nous avons entraîné un modèle sur les données du jeu Balsac et un autre sur celles de DIVA-HisDB sur des images redimensionnées à 768 pixels.

La Table 4.7 montre que l'entraînement sur des images plus grandes améliore les résultats. Les lignes étant souvent de petite hauteur sur ces bases, agrandir les images d'entrée permet au modèle de mieux séparer les lignes proches et de les prédire avec une plus grande précision.

Grâce à ces premières expériences, nous avons montré que notre modèle Doc-UFCN obtient de meilleures performances sur la tâche de détection de lignes de texte dans des images de document que dhSegment, tout en comportant moins de paramètres et en étant plus rapide en inférence. De plus, nous avons montré qu'entraîner un modèle sur plusieurs bases permet d'obtenir un modèle plus générique et d'améliorer la qualité des prédictions par rapport à des modèles spécifiques. Nous évaluons, dans la section suivante, le modèle Doc-UFCN sur une tâche plus complexe de détection et de classification d'actes.

## 4.5 EXPÉRIENCES DE DÉTECTION D'ACTES

Les registres sont des types très courants de documents historiques qui contiennent des listes d'enregistrements, appelés "actes", se rapportant à des personnes, des objets ou des événements. Ils peuvent être présentés sous forme de tableaux ou de séquences de textes. Dans le cas des registres royaux, des cartulaires médiévaux ou des registres paroissiaux et civils, les actes sont des segments textuels composés d'un ou plusieurs paragraphes. Pour traiter le problème de la détection d'actes, la plupart des méthodes existantes utilisent soit le contenu textuel des documents, soit le contenu visuel. Les systèmes récents basés sur des règles heuristiques ou des réseaux neuronaux se basent uniquement sur les caractéristiques visuelles des images pour détecter les actes, en ignorant le texte des documents. En effet, Tarride et al. (2019) combinent des règles et un réseau neuronal pour segmenter les registres paroissiaux français en actes. Ils détectent d'abord les signatures des prêtres situées à la fin de chaque acte à l'aide d'un réseau neuronal (dhSegment (Ares Oliveira et al., 2018) ou ARU-Net (Grüning et al., 2019)) avant d'utiliser un système à base de règles pour générer les actes. Même s'ils ont obtenu un système avec 80 % de rappel au niveau des actes, leur méthode repose principalement sur l'hypothèse que chaque acte se termine par une signature, ce qui n'est pas toujours le cas, et si elle est effectivement présente, celle-ci n'est pas toujours détectée par le système automatique. La méthode que nous proposons comprend également des caractéristiques basées sur des règles, mais combinées avec l'image originale.

De plus, les documents historiques peuvent avoir un contenu textuel riche qui peut permettre un meilleur processus de détection. Ainsi, Prieto et al. (2020) ont étudié le cas où l'aspect graphique des images n'est pas suffisant pour segmenter les chartes médiévales en actes. Ils ne visent pas seulement à détecter les actes mais cherchent également à les classer comme *début*, *milieu*, *fin* d'acte ou *acte complet*. Pour cela, ils utilisent une carte d'indexation probabiliste pour construire des caractéristiques supplémentaires fondées sur le contenu textuel, puis les caractéristiques graphiques et textuelles sont fusionnées afin d'obtenir une seule entrée pour le système de détection. Ils montrent que l'ajout de contenu textuel peut faciliter la détection des actes et que l'ajout de connaissances *a priori* permet d'améliorer encore les performances (73 % à 88 % de la F-mesure). Inspiré par cette idée et celle proposée par Yang et al. (2017), notre travail se concentre sur l'utilisation des deux modalités en entrée d'un système de détection par apprentissage profond pour améliorer la détection des actes présents dans des documents historiques. Dans cette section, nous présentons tout d'abord les données utilisées, notre approche pour résoudre la tâche de détection d'actes, puis nous discutons des entraînements et des résultats obtenus.

**Table 4.8 –** Statistiques des jeux de données utilisés pour la détection d'actes : nombre de pages, lignes transcrites et actes par type.

| Jeu de données | | Images | Lignes | Actes | | | |
|---|---|---|---|---|---|---|---|
| | | | | *complet* | *début* | *milieu* | *fin* |
| Balsac Vézina et al. (2020) | train | 730 | 36 941 | 1 474 | 503 | 2 | 487 |
| | valid | 92 | 4 592 | 181 | 66 | 1 | 58 |
| | test | 91 | 4 323 | 173 | 62 | 0 | 52 |
| Himanis-Act Bluche et al. (2017) | train | 132 | – | 144 | 46 | 21 | 40 |
| | valid | 19 | – | 29 | 3 | 3 | 2 |
| | test | 411 | – | 172 | 203 | 115 | 196 |
| Himanis-GMV | train | – | 18 504 | – | – | – | – |
| | valid | – | 2 367 | – | – | – | – |
| | test | – | 2 241 | – | – | – | – |

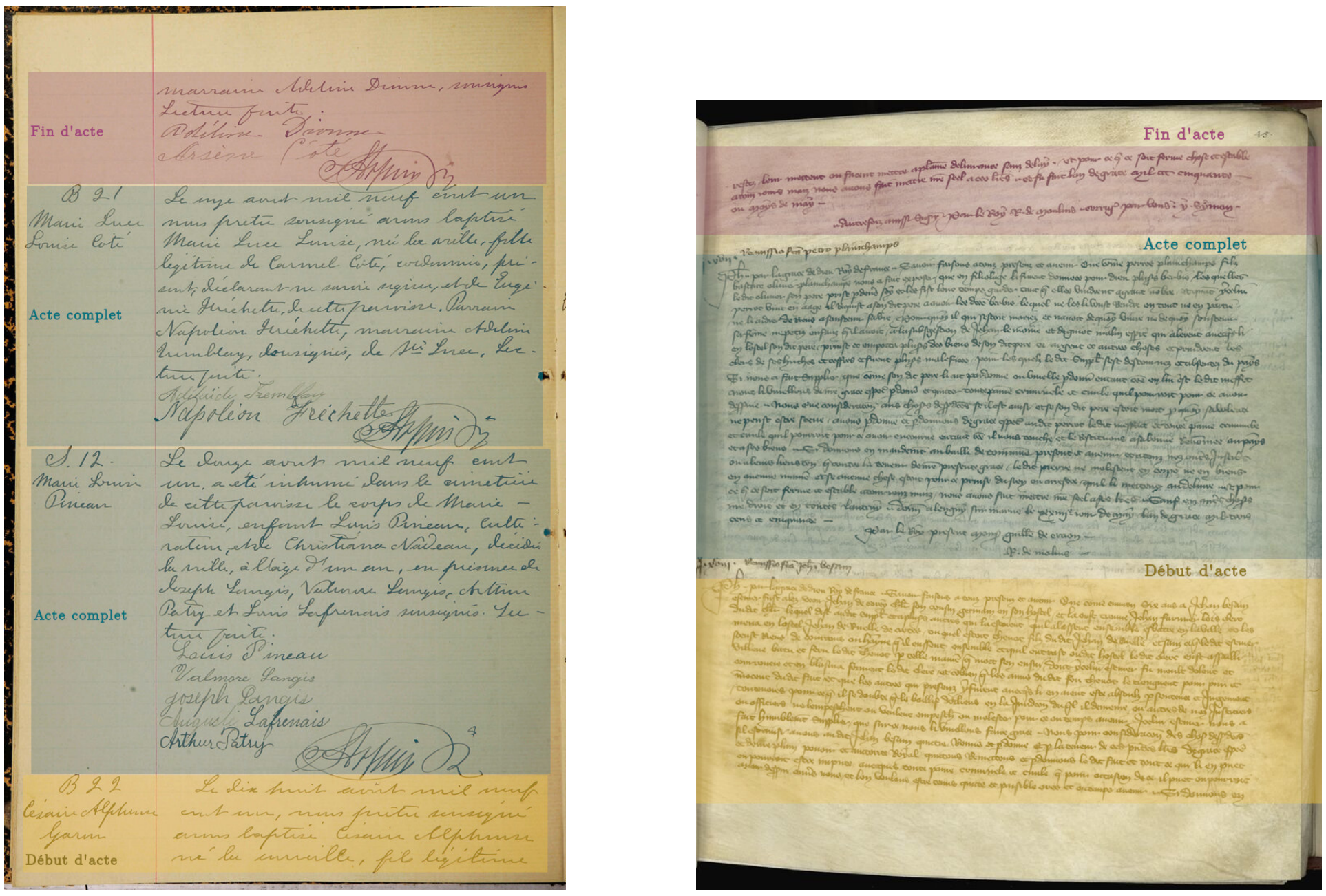

**Figure 4.4 –** Annotations manuelles pour la détection et la classification d'actes sur une image du jeu de données Balsac, à gauche, et Himanis-Act, à droite.

### 4.5.1 JEUX DE DONNÉES

Pour nos expériences, nous avons utilisé deux jeux de données, Balsac (Vézina et al., 2020) (présenté en section 3.1) et Himanis-Act (Prieto et al., 2020). Pour traiter les actes répartis sur plusieurs pages, les actes sont annotés avec quatre classes : *acte complet*, *début d'acte*, *milieu d'acte* et *fin d'acte*. Les statistiques de ces ensembles de données sont présentées dans la Table 4.8. La Figure 4.4 présente un exemple d'annotation pour chaque jeu de données.

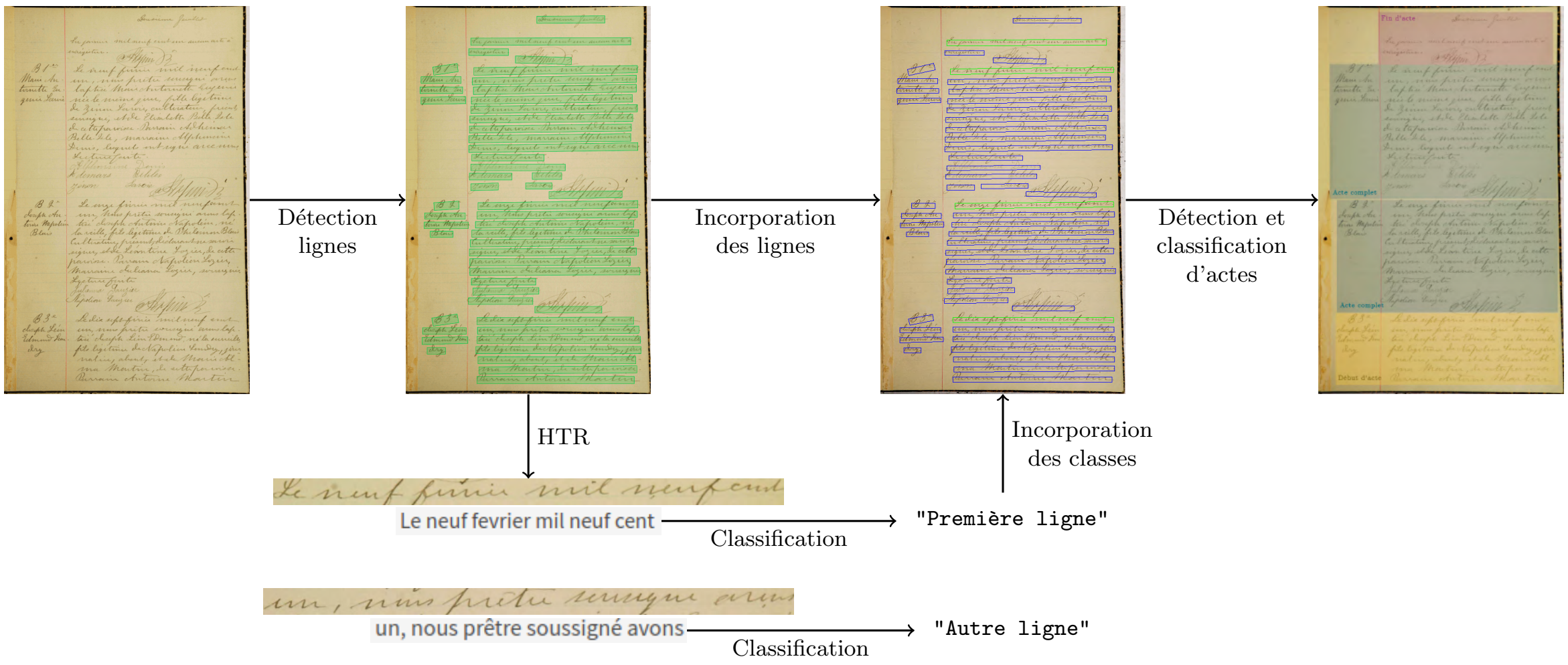

**Figure 4.5 –** Chaîne de traitement proposée pour la détection et la classification d'actes avec l'utilisation du contenu textuel.

Le jeu de données Himanis est extrait du corpus *Chancery*, une grande collection de registres produits par la chancellerie royale française. Il est composé de 80 000 images contenant des chartes promulguées par les rois de France aux 14$^{e}$ et 15$^{e}$ siècles. Ces documents consignent les décisions royales comme les donations ou les grâces, et sont organisés en actes.

Le jeu de données Himanis-Act consiste en un échantillon de 739 images extraites des données Himanis et est annoté au niveau des actes. Pour réaliser nos expériences, nous avons utilisé la répartition proposée par Prieto et al. (2020), obtenue après avoir éliminé les pages ne contenant aucune information telles que les pages blanches. Ce jeu de données est uniquement utilisé pour entraîner et évaluer le système de détection d'actes puisque les annotations de lignes de texte et de transcription ne sont pas disponibles. La répartition finale est présentée dans la Table 4.8.

Le jeu de données annoté Himanis-GMV est composé de 1 435 images extraites des données Himanis, alignées automatiquement au niveau des lignes, des éditions imprimées (Guérin et al., 1881-1958 ; Guyotjeannin et al., 2005 ; Viard, Jules, 1899) avec Transkribus (Kahle et al., 2017). Après le processus d'alignement, 23 112 lignes de texte sont disponibles pour l'entraînement et l'évaluation des systèmes de reconnaissance de l'écriture manuscrite, dont la répartition est présentée dans la Table 4.8. Ce jeu de données est uniquement utilisé pour entraîner un système de reconnaissance HTR puisque la segmentation en actes n'est pas disponible.

### 4.5.2 Approche proposée

La Figure 4.5 détaille l'approche proposée pour résoudre la tâche de détection et de classification d'actes. L'approche consiste en plusieurs traitements successifs de l'image d'entrée :

1. L'image d'entrée est d'abord traitée par un détecteur de lignes de texte ;
2. Les lignes prédites sont extraites et traitées par un reconnaisseur de texte manuscrit ;

**Table 4.9** – Résultats du modèle générique de détection de lignes de texte sur l'ensemble de test du jeu de données Balsac.

| Jeu de données | Système | IoU | AP@.5 | AP@.75 | mAP |
|---|---|---|---|---|---|
| Balsac | Doc-UFCN | **0,87** | **0,98** | **0,91** | **0,76** |
| | dhSegment | 0,74 | 0,94 | 0,54 | 0,51 |

3. Chaque ligne est classifiée selon le texte qu'elle contient en "première ligne" (s'il est probable que la ligne soit la première de l'acte) et "autre ligne" ;
4. L'image d'entrée est enrichie par ces lignes classifiées en dessinant les lignes de texte de couleurs différentes selon leurs classes ;
5. Enfin, les actes sont détectés et classifiés à partir de cette image enrichie.

Les paragraphes suivants présentent et discutent chacune de ces étapes en détail.

### Détection des lignes de texte

La première étape du traitement des images consiste à détecter les lignes de texte sur les images. Pour cette tâche, nous avons utilisé Doc-UFCN. Afin de créer un modèle générique pouvant être utilisé sur différents jeux de données, le modèle a été entraîné sur neuf jeux de données historiques dont Balsac et à l'exception d'Himanis, puisqu'il ne contient aucune annotation au niveau des lignes. Les images ont été redimensionnées de manière à ce que leur plus grande dimension soit égale à 768 pixels, tout en conservant le rapport d'aspect original. Les annotations originales ont également été uniformisées grâce au processus détaillé dans la section 5.1.1. dhSegment (Ares Oliveira et al., 2018) a également été entraîné afin de fournir une comparaison de référence avec Doc-UFCN.

Les deux modèles de détection de lignes de texte ont été évalués à l'aide de l'IoU. Les modèles ont également été évalués à l'aide de l'AP, qui quantifie le nombre d'objets correctement détectés, alors que la mesure IoU considère uniquement le nombre de pixels correctement prédits. Plus de détails sur le calcul de la mAP sont donnés dans le Focus 3.4.

Les résultats sont présentés dans la Table 4.9, pour le jeu de données Balsac uniquement, car aucune annotation manuelle n'est disponible pour les jeux de données Himanis. Les résultats montrent que Doc-UFCN surpasse dhSegment pour toutes les métriques et obtient des performances très satisfaisantes.

### Reconnaissance de texte

La reconnaissance de texte manuscrit (HTR) est appliquée aux lignes de texte détectées et produit le texte correspondant. Le modèle de reconnaissance est construit avec la librairie Kaldi[2], basée sur un modèle DNN-HMM (Deep Neural Network - Hidden Markov Model). Notre modèle est comparable à celui décrit dans Arora et al. (2019).

2. https://kaldi-asr.org/

**Table 4.10 –** Résultats de reconnaissance de textes manuscrits sur les jeux de données Balsac et Himanis-GMV. Les résultats d'un modèle HTR+ entraîné avec Transkribus sont donnés à titre de référence mais ne sont pas directement comparables à notre système puisque les séparations train/valid ne sont pas identiques.

| Jeu de données | Système | CER (%) | | | WER (%) | | |
|---|---|---|---|---|---|---|---|
| | | train | valid | test | train | valid | test |
| Balsac | Kaldi | 4,1 | 6,2 | 6,4 | 12,4 | 17,1 | 17,4 |
| | Transkribus | 12,2 | 9,5 | – | – | – | – |
| Himanis-GMV | Kaldi | 5,4 | 9,4 | 8,0 | 11,9 | 19,3 | 18,1 |
| | Transkribus | 9,5 | 5,3 | – | – | – | – |

Nous avons entraîné un modèle Kaldi sur le jeu de données Balsac en suivant la répartition présentée dans la Table 4.1. Aucune donnée supplémentaire n'a été utilisée pour le modèle linguistique ni pour le modèle optique. Pour Himanis, nous avons entraîné un modèle sur le jeu de données annoté Himanis-GMV présenté ci-dessus. Pour les deux modèles, les lignes d'entrée ont été redimensionnées à une hauteur de 40 pixels tout en conservant le rapport d'aspect. De plus, les lignes ayant des largeurs similaires ont été regroupées pour un entraînement plus efficace. À titre de comparaison, un modèle HTR+ a été entraîné à l'aide de la plateforme Transkribus (Kahle et al., 2017).

Les performances des modèles HTR sont décrites dans la Table 4.10. Concernant notre système HTR (Kaldi), les résultats obtenus pour les deux jeux de données sont similaires. Même si les modèles présentent des taux d'erreurs de mots (Word Error Rate (WER)) relativement élevés, nous pensons qu'ils sont capables de prédire une transcription suffisamment correcte pour la détection de mots-clés et la détection des actes. L'interface d'entraînement de Transkribus ne fournit qu'une évaluation pour le Character Error Rate (CER) et ne peut pas être directement comparée à la nôtre puisque les répartitions train/valid/test ne sont pas identiques. Cependant, les résultats sont du même ordre de grandeur. Le modèle de Transkribus présente un CER plus élevé sur l'ensemble d'entraînement en raison de son processus d'augmentation des données, processus non utilisé dans notre système Kaldi.

### Classification des lignes de texte

La classification des lignes de texte est effectuée à l'aide de règles. Le modèle utilise les transcriptions prédites en entrée et prédit si une ligne de texte est la première ligne d'un acte en se basant sur la présence de phrases clés définies *a priori*.

Pour le jeu de données Balsac, la plupart des actes commencent par une date telle que `"Le trente un janvier, mil neuf"`, et il n'y a souvent pas d'autres dates dans la suite des actes. Ainsi, la règle est donc de compter le nombre de mots qui sont des chiffres ou des mois pour que la ligne soit considérée comme une date. Expérimentalement, trois mots semblent être suffisants pour que la ligne soit considérée comme contenant une date.

Pour les actes de Himanis-Act, la tâche est plus complexe car ils ne commencent pas toujours par les mêmes mots. Nous avons donc analysé les premières lignes des actes d'entraînement et avons conservé les phrases clés les plus fréquentes (par exemple `"dei gratia francorum rex"` ou `"par la grâce de dieu roys de france"`). Si une phrase clé est incluse dans une ligne, elle est considérée comme se trouvant au début d'un acte.

Table 4.11 – Résultats de classification des lignes de texte en première ligne d'un acte / autre ligne sur les jeux de données Balsac et Himanis-Act.

| Jeu de données | | Précision | Rappel | F1-score |
|---|---|---|---|---|
| Balsac | train | 0,69 | 0,87 | 0,77 |
| | valid | 0,72 | 0,87 | 0,79 |
| | test | 0,68 | 0,86 | 0,76 |
| Himanis-Act | train | 0,79 | 0,65 | 0,71 |
| | valid | 0,81 | 0,53 | 0,64 |
| | test | 0,68 | 0,86 | 0,76 |

La Table 4.11 montre la précision, le rappel et le F1-score de la classe "première ligne". Seule la classe "première ligne" est donnée car c'est la seule classe apportant des informations à la détection des actes, d'autant plus que la distribution des classes est très déséquilibrée. Pour le jeu de données Balsac, les résultats sont stables entre les trois ensembles et le rappel est élevé, ce qui est favorable à l'inclusion de cette information dans le système visuel. Pour le jeu de données Himanis-Act, le rappel est plus faible et les résultats varient entre les ensembles, ce qui montre bien que la tâche est plus complexe et que la détection des phrases clés est plus difficile, et donc moins fiable.

### 4.5.3 Résultats et discussion

La Table 4.12 présente les résultats des différents modèles de détection des actes : Doc-UFCN entraîné sur des images brutes (visuel) et Doc-UFCN entraîné sur les images brutes avec les polygones des lignes de texte dessinés de deux couleurs dépendant de la classe de la ligne (visuel + textuel).

Pour le jeu de données Balsac, nous ne présentons pas les résultats de la classe *milieu d'acte* car il n'y en a pas dans le jeu de test. D'après la Table 4.12, les résultats sont en moyenne meilleurs pour le modèle utilisant les contenus visuels et textuels. Pour les classes de *début* et de *fin* d'acte, les deux systèmes sont presque équivalents pour toutes les métriques. En revanche, nous constatons que l'ajout de la date directement à l'image d'entrée améliore les performances de la classe d'actes complets de 36 points de pourcentage de mAP. Cela conduit à une meilleure séparation des actes complets consécutifs dans les prédictions, ce qui était l'un de nos principaux objectifs.

Pour le jeu de données Himanis-Act, les résultats sans contenu textuel sont significativement meilleurs que ceux l'incorporant. Nous pensons que ces résultats sont dûs aux raisons suivantes. Tout d'abord, le contenu textuel des documents de la base Himanis-Act est plus complexe et diversifié que celui de la base de données Balsac, qui est très standardisé. De plus, nous avons pu trouver de nombreux actes imbriqués dans le jeu de données (*Vidimus*), ce qui peut ajouter de la confusion au système. La définition des phrases clés s'est avérée plus complexe et la Table 4.11 montre que le rappel est faible, même pour l'ensemble d'entraînement, ce qui conduit à des caractéristiques textuelles peu fiables pour entraîner le modèle de détection d'actes. De plus, les modèles de détection de lignes et d'HTR n'ont pas été entraînés sur l'ensemble de données Himanis-Act. Le modèle de détection de lignes de texte a été entraîné sur des données similaires mais sans images venant du jeu de données Himanis-Act. Il en est de même pour le système HTR qui n'a pas été entraîné directement sur les mêmes volumes, ce qui peut créer un décalage entre les conditions d'entraînement et de test.

Table 4.12 – Résultats de détection d'actes sur les ensemble de test des jeux de données Balsac et Himanis-Act.

| Jeu de données | Système | Actes | IoU | AP@.5 | AP@.75 | mAP |
|---|---|---|---|---|---|---|
| Balsac | Visuel | *complet* | **0,84** | 0,57 | 0,37 | 0,38 |
| | | *début* | **0,58** | 0,86 | 0,85 | 0,76 |
| | | *fin* | **0,58** | 0,85 | 0,64 | 0,59 |
| | Visuel + textuel | *complet* | 0,82 | **0,89** | **0,81** | **0,74** |
| | | *début* | **0,58** | **0,90** | **0,87** | **0,78** |
| | | *fin* | 0,54 | **0,86** | **0,73** | **0,63** |
| Himanis-Act | Visuel | *complet* | 0,61 | **0,75** | **0,73** | **0,70** |
| | | *début* | **0,76** | **0,84** | **0,82** | **0,77** |
| | | *milieu* | **0,88** | **0,84** | **0,83** | **0,83** |
| | | *fin* | **0,73** | **0,73** | **0,65** | **0,62** |
| | Visuel + textuel | *complet* | **0,64** | 0,54 | 0,51 | 0,49 |
| | | *début* | 0,68 | 0,69 | 0,64 | 0,60 |
| | | *milieu* | 0,84 | 0,80 | 0,80 | 0,80 |
| | | *fin* | 0,70 | 0,64 | 0,63 | 0,58 |

En plus de ces expériences, nous avons comparé nos résultats avec ceux de l'état de l'art de Prieto et al. (2020). Ils ont testé différentes configurations avec et sans le contenu textuel pour détecter où se terminent les actes. Pour obtenir une comparaison juste, nous avons utilisé le même protocole d'évaluation. L'évaluation est effectuée à l'aide du *Transkribus Baseline Evaluation Scheme* (TBES) (Diem et al., 2017). Cet outil a été conçu pour évaluer la détection de la ligne de base. Ainsi, pour l'utiliser, la ligne de base des actes est définie comme étant la ligne droite horizontale à la fin d'un acte complet ou d'une fin d'acte. Pour être en accord avec leurs résultats, nous avons utilisé la même valeur de tolérance de 128 pixels.

D'après la Table 4.13, nous pouvons voir que Doc-UFCN utilisant uniquement l'image améliore les résultats de l'état de l'art. En effet, en comparaison au système visuel de Prieto et al. (2020), notre méthode obtient des performances supérieures de 10 points de pourcentage. En outre, nous constatons que les deux modèles utilisant le contenu textuel se comportent de la même manière et sont moins performants que notre système basé uniquement sur les informations visuelles. Pour ce jeu de données, il semble préférable de se concentrer sur les caractéristiques visuelles avec un système d'apprentissage profond robuste, plutôt que d'ajouter un contenu textuel trop peu fiable.

**Table 4.13** – Résultats obtenus par Doc-UFCN et Prieto et al. (2020) sur le jeu de données Himanis-Act avec et sans l'information textuelle.

| Système | | Visuel | Visuel + textuel |
|---|---|---|---|
| Doc-UFCN | train | 0,96 | 0,96 |
| | valid | 0,96 | 0,91 |
| | test | **0,90** | **0,88** |
| Prieto et al. (2020) | test | 0,80 | **0,88** |

Dans cette partie, nous avons présenté une chaîne de traitement simple permettant d'enrichir des images d'entrée avec le contenu textuel de documents. Ces images enrichies permettent d'effectuer une tâche de détection d'actes en utilisant simultanément le contenu visuel et la position des lignes de texte contenant des phrases clés définies manuellement. Nous avons montré que l'utilisation de ces images peut améliorer la détection des actes, en particulier des actes consécutifs. Sur le jeu de données Balsac, pour lequel des règles de détection de phrases clés ont pu être définies de manière fiable, l'utilisation de ces images augmente la détection d'actes de 38 % à 74 % de mAP par rapport à un modèle standard se basant uniquement sur le contenu visuel.

## 4.6 CONCLUSION

Dans ce chapitre, nous avons présenté Doc-UFCN, un nouveau système de détection d'objets dans les images de documents. Nous avons montré que ce système permet d'entraîner des modèles plus performants, plus rapides et comportant moins de paramètres que ceux de l'état de l'art pour la détection de lignes de texte. Le code de ce système est disponible publiquement [3]. De plus, les expérimentations décrites dans ce chapitre ont permis d'amorcer une analyse sur l'intérêt des modèles génériques, qui seront l'objet du prochain chapitre.

Enfin, nous nous sommes intéressés aux méthodes combinant l'image et le texte pour la détection et la classification d'actes. Dans un cadre dans lequel la séparation visuelle des actes suit la séparation logique du texte, nous avons montré que l'incorporation du contenu textuel dans l'image d'entrée permet de réellement améliorer la détection des actes.

3. https://pypi.org/project/doc-ufcn/

# 5
# ENTRAÎNEMENT ET ÉVALUATION D'UN MODÈLE ROBUSTE DE DÉTECTION D'OBJETS

La littérature montre que des systèmes compétitifs et robustes ont été développés pour résoudre le problème de la détection des lignes de texte, obtenant des performances satisfaisantes sur des jeux de données individuels. Cependant, leurs performances sont souvent insuffisantes sur d'autres documents hors échantillon, car ils manquent de capacités de généralisation. Or, dans un cadre industriel avec de nombreux projets et des données très différentes, il est nécessaire de développer des modèles plus génériques, obtenant des performances élevées sur des documents très variés provenant de différents projets.

L'entraînement de systèmes sur des données très diverses permettrait d'obtenir de tels modèles, montrant de meilleures capacités de généralisation. Pour cela, il est nécessaire de combiner plusieurs ensembles de données. Cependant, les schémas d'annotation ne sont pas toujours compatibles entre les jeux de données publics (comme décrit en section 3.2), ce qui rend difficile leur combinaison dans un seul ensemble d'entraînement unifié. Ces différents schémas ne permettent pas une comparaison équitable des approches de détection d'un jeu de données à l'autre. Par conséquent, dans la littérature, aucune comparaison de systèmes n'a été effectuée sur un jeu de données large et diversifié, tant en entraînement qu'en évaluation.

De plus, dans la littérature, les études sur la détection des lignes de texte manuscrites se concentrent généralement sur le développement d'une architecture de réseau neuronal spécifique, ainsi que sur une bonne stratégie d'entraînement. Cependant, elles omettent souvent d'analyser les annotations utilisées lors de l'entraînement et de l'évaluation, alors qu'elles sont aussi importantes que l'architecture du réseau elle-même, puisqu'elles guident la phase d'entraînement et les résultats finaux. En effet, le biais d'annotation est particulièrement important lorsque nous voulons analyser l'impact de l'étape de détection sur l'étape de reconnaissance. Cependant, il est rarement étudié dans les études se concentrant sur la détection de lignes de texte. Il n'est pas non plus considéré dans les études portant sur la reconnaissance de l'écriture manuscrite, car le processus de détection n'entre pas dans leur champ d'application. Par exemple, la première lettre dans les documents historiques est parfois ornée, l'ajouter ou non dans les lignes de texte pendant le processus d'annotation peut avoir un réel impact sur les résultats de la reconnaissance finale, d'où l'importance de créer et d'analyser soigneusement les annotations. Un autre problème se pose lors de la détection des lignes de texte lorsque deux boîtes de délimitation annotées se touchent. Dans une telle situation, le réseau fournit généralement des lignes fusionnées qui ne seront pas reconnues par le système de reconnaissance HTR, alors que généralement les métriques d'évaluation ne tiennent pas compte de ces problèmes. En effet, la métrique IoU, très souvent utilisée, est incapable de détecter ces situations et de compter correctement les séparations de lignes correctes et incorrectes.

Dans ce chapitre, nous abordons ce problème, en section 5.1.1, en introduisant une stratégie d'uniformisation d'étiquetage des jeux de données, qui réduit les chevauchements des polygones englobants afin d'obtenir un résultat cohérent avec l'entrée requise des systèmes de reconnaissance (HTR ou OCR).

Toujours dans un cadre industriel où l'objectif est la reconnaissance du texte des documents, et pas uniquement la détection des lignes, il est nécessaire d'avoir une bonne évaluation des modèles de détection. Cependant, l'évaluation de tels modèles est complexe. En effet, les métriques d'évaluation au niveau pixel ne sont pas toujours représentatives de l'impact réel de l'étape de détection de lignes de texte sur l'étape de reconnaissance de texte. De plus, les comparaisons des systèmes de détection de lignes de texte en termes de taux d'erreur de reconnaissance sont rarement rapportées en raison de la complexité de cette évaluation.

La plupart des méthodes de détection de lignes de texte existantes sont évaluées et comparées à l'aide de métriques au niveau pixel telles que l'IoU, la précision, le rappel et le score F1. Même si ces mesures indiquent les performances du modèle, elles ne donnent aucune information réelle sur la quantité d'informations détectées, comme le nombre de lignes. Certaines mesures au niveau de l'objet, telles que la précision moyenne (AP), ont été proposées pour surmonter ce problème, mais elles reposent toujours sur un seuil d'IoU fixe. Dans la section 5.3.2, nous analysons donc ces limitations et introduisons la métrique mAP déjà utilisée dans les défis de détection COCO en vision, qui ne nécessite pas de seuil de détection pour être mise en œuvre. De plus, comme la détection des lignes manuscrites est la première étape de l'ensemble du processus de reconnaissance, il devrait être plus réaliste d'évaluer son véritable impact sur les résultats de reconnaissance finaux (taux d'erreur caractères et mots), en effectuant une évaluation orientée vers la tâche finale. À cet égard, nous proposons, en section 5.4.1, une stratégie d'évaluation fondée sur les résultats de reconnaissance obtenus après un système de reconnaissance de texte (HTR).

Dans cette partie, nous fournissons une évaluation juste et approfondie de trois approches pour détecter les lignes de texte, Doc-UFCN (Boillet et al., 2021a), dhSegment (Ares Oliveira et al., 2018) et ARU-Net (Grüning et al., 2019), sur une large collection de jeux de données historiques et avec plusieurs métriques, y compris une métrique orientée reconnaissance de texte (HTR). Nous analysons les métriques de détection de lignes de texte de la littérature par rapport à de multiples jeux de données publiquement disponibles et montrons certaines incohérences entre les jeux de données. Nous proposons, en section 5.1.1, une stratégie d'uniformisation des annotations des jeux de données qui évite le biais d'étiquetage pour la tâche de détection de lignes de texte. Cet étiquetage modifié permet de considérer la variabilité des annotations et d'entraîner des modèles avec une plus grande capacité de généralisation. Dans un second temps, nous effectuons une évaluation de l'état de l'art grâce à différentes métriques et protocoles d'entraînement. Ces protocoles permettent de construire des modèles de détection plus génériques qui considèrent les limitations mentionnées ci-dessus et obtiennent des résultats similaires, voire meilleurs, que les modèles entraînés sur des ensembles de données uniques.

**Table 5.1 –** Statistiques des jeux de données utilisés pour la détection de lignes de texte.

| Jeu de données | Images | | | Lignes | | |
|---|---|---|---|---|---|---|
| | train | valid | test | train | valid | test |
| AN-Index† | 19 | 3 | 12 | 433 | 62 | 171 |
| Balsac<br>Vézina et al. (2020) | 730 | 92 | 91 | 36 941 | 4 592 | 4 323 |
| BNPP† | 7 | 2 | 3 | 705 | 218 | 358 |
| Bozen<br>Sánchez et al. (2016) | 350 | 50 | 50 | 8 366 | 1 040 | 1 138 |
| cBAD2019<br>Diem et al. (2019) | 720 | 716 | | 45 266 | 42 672 | |
| DIVA-HisDB<br>Simistira et al. (2016) | 60 | 30 | 30 | 6 037 | 2 999 | 2 897 |
| HOME-NACR<br>Boros et al. (2020) | 338 | 40 | 42 | 6 590 | 602 | 909 |
| Horae<br>Boillet et al. (2019) | 522 | 20 | 30 | 12 568 | 270 | 958 |
| READ-Complex<br>Grüning et al. (2017) | 216 | 27 | 27 | 17 768 | 2 160 | 1 758 |
| READ-Simple<br>Grüning et al. (2017) | 172 | 22 | 22 | 5 117 | 539 | 723 |
| HOME-Alcar<br>Stutzmann et al. (2021) | – | – | 55 | – | – | 2 727 |
| ScribbleLens<br>Dolfing et al. (2020) | – | – | 21 | – | – | 563 |

† Jeux de données privés utilisés durant la thèse.

## 5.1 UNIFORMISATION DES ANNOTATIONS

L'un de nos objectifs étant de développer un détecteur générique de lignes de texte sur des documents historiques, nous avons collecté neuf jeux de données historiques dont sept jeux publics pour mener les expérimentations. Ces jeux de données sont présentés en section 3.2 et une description est donnée dans la Table 5.1. De plus, comme nous souhaitons évaluer la capacité de généralisation des modèles obtenus, nous avons collecté deux jeux de données supplémentaires : ScribbleLens (Dolfing et al., 2020) et HOME-Alcar (Stutzmann et al., 2021) utilisés uniquement pendant l'étape d'évaluation.

Tous ces ensembles de données ont été choisis pour leur diversité en termes de tailles, d'écritures et de mises en page. La Figure 3.2 présente la variété des documents en montrant un exemple d'image de chaque ensemble de données avec ses annotations lignes. La répartition utilisée pour entraîner les modèles a été obtenue en regroupant simplement les données d'entraînement et de validation respectives des sous-ensembles. De plus, puisque le jeu de données HOME-Alcar ne contenait pas d'ensembles d'entraînement, de validation et de test officiels au moment des expériences présentées ci-après nous avons rassemblé, pour générer un ensemble de test, 55 pages annotées au niveau ligne avec leurs transcriptions correspondantes parmi tous les manuscrits.

### 5.1.1 ANALYSE DES ANNOTATIONS

Tous les ensembles de données cités ci-dessus ont été utilisés pour entraîner des modèles génériques de détection de lignes de texte avec les modèles Doc-UFCN (Boillet et al., 2021a), dhSegment (Ares Oliveira et al., 2018) et ARU-Net (Grüning et al., 2019). Ces modèles nécessitent des images annotées au niveau pixel pour l'entraînement. Nous présentons les défis rencontrés pour générer un ensemble d'entraînement annoté unifié.

#### DIVERSITÉ DANS LES ANNOTATIONS

Pour générer les images d'annotations au niveau pixel, les polygones englobants sont extraits de la vérité terrain et dessinés sur une image de fond noir de même taille que l'image originale. Comme nous pouvons le voir sur la Figure 3.2, les annotations sont très variées parmi les ensembles de données :

- Dans la plupart des jeux de données (AN-Index, Balsac, Bozen, BNPP, HOME et Horae), les images sont annotées à l'aide de simples polygones incluant les ascendants et descendants ;
- Dans les bases cBAD2019, READ-Simple et READ-Complex, les ascendants et descendants ne sont généralement pas inclus dans les polygones, qui sont très fins par rapport au premier cas ;
- Dans les images de la base ScribbleLens, les annotations sont de larges rectangles qui incluent de nombreux pixels appartenant au fond ;
- Dans la base DIVA-HisDB, les lignes de texte sont annotées à l'aide de polygones plus complexes qui suivent précisément le contour de chaque lettre. Les polygones des lignes de texte du jeu HOME-Alcar sont également précis mais légèrement moins que ceux de DIVA-HisDB.

Cette diversité dans les annotations nous empêche d'entraîner directement un modèle générique qui pourrait être appliqué à de nouveaux jeux de données, car l'annotation serait parfois incohérente entre deux exemples provenant de deux jeux de données différents. De telles incohérences dégraderaient considérablement les performances du système. Corriger les incohérences d'annotation entre les ensembles de données est donc une nécessité pour permettre l'unification des ensembles d'entraînement.

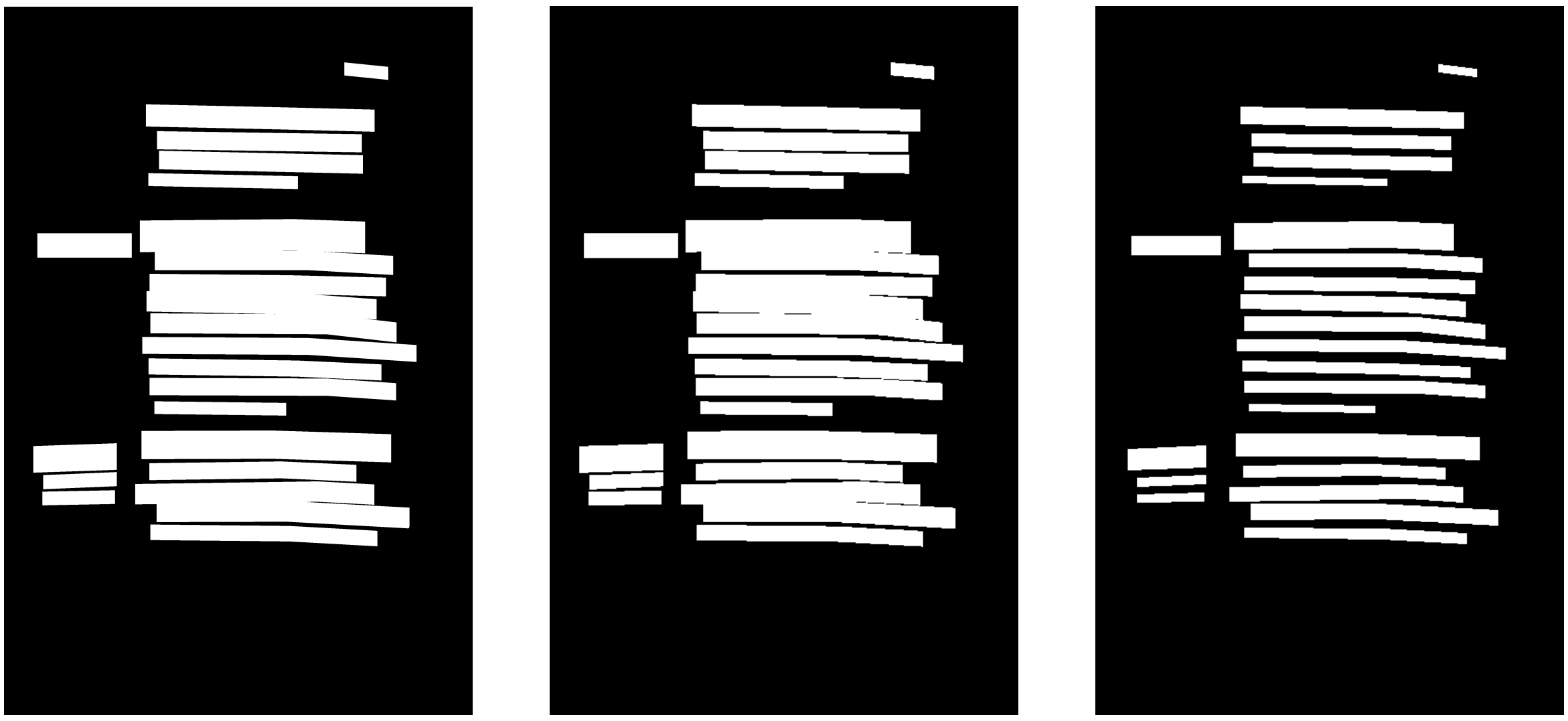

**Figure 5.1** – Processus de génération d'annotations pour une image du jeu de données de Bozen. À gauche : génération d'annotations à la taille originale de l'image. Au centre : redimensionnement de l'image à la taille de l'entrée du réseau (768 pixels). À droite : redimensionnement des polygones englobants à la taille d'entrée du réseau, atténuation des chevauchements et génération de l'image d'annotations à la taille d'entrée du réseau (768 pixels).

### CHEVAUCHEMENT DES POLYGONES

Un autre problème qui accentue les incohérences des annotations est le chevauchement des polygones englobants annotés. Même si certains ensembles de données ont été annotés de telle sorte que jamais les polygones ne se touchent ni ne se chevauchent, d'autres ont été annotés moins précisément, ce qui entraîne des polygones qui se touchent et se superposent (page ScribbleLens sur la Figure 3.2), comme le montre l'image de gauche de la Figure 5.1. Évidemment, une telle vérité terrain ne peut pas servir à une évaluation précise de la capacité d'un système à détecter chaque ligne de texte. De plus, Doc-UFCN et ARU-Net utilisent des sous-résolutions des images d'entrée, ce qui peut entraîner des fusions supplémentaires dans les images d'annotations. La Figure 5.1 présente cet effet indésirable en montrant l'image d'annotation originale, à gauche, et sa version redimensionnée à 768 pixels, au centre.

Dans la littérature, la plupart des études utilisent la vérité terrain telle quelle, sans prêter beaucoup d'attention à ce problème de fusion. La raison principale à cela est probablement que les mesures d'évaluation utilisées comptent uniquement la précision des pixels, sans évaluer plus précisément le processus de détection des lignes. Cependant, un nombre important de fusions dans l'ensemble de données d'entraînement va certainement faire dévier le réseau vers la fusion de plus de lignes que souhaité, avec un effet induit négatif sur le système HTR incapable de reconnaître les lignes fusionnées verticalement.

Lors de nos expériences, nous avons atténué ce problème en supprimant, autant que possible, les chevauchements entre les lignes, tout en perdant le moins d'informations possible, comme on peut le voir sur l'image de droite de la Figure 5.1.

#### STRATÉGIE D'UNIFORMISATION

Pour unifier les annotations, nous avons choisi d'utiliser uniquement des polygones simples pendant l'étape d'entraînement. Par conséquent, les rectangles englobant les polygones de DIVA-HisDB ont été utilisés pendant l'apprentissage au lieu des polygones complexes originaux. De plus, pour résoudre le problème du chevauchement des polygones englobants de certains jeux de données, nous avons utilisé la stratégie suivante. Pour une image donnée, nous recherchons les paires de lignes qui se touchent et se chevauchent. Ensuite, trois cas ont été identifiés pour chaque paire :

— Les polygones se touchent : nous les érodons de 1 pixel ;
— Les polygones se chevauchent de moins de 20 % chacun : nous appliquons le processus de scission des polygones décrit ci-dessous ;
— Les polygones se chevauchent de plus de 20 % : nous les gardons tels quels car la scission peut entraîner la perte de trop d'informations (perte d'un polygone ou séparation en deux polygones).

Dans le cas d'un petit chevauchement (moins de 20 % de l'aire des polygones), le processus suivant est appliqué : la ligne ayant le plus petit ratio de surface de chevauchement par rapport à sa surface totale est détectée, et son intersection avec l'autre ligne est supprimée, l'autre ligne étant conservée telle quelle. Enfin, tous les polygones sont dessinés sur une image de fond noir. Ce seuil de 20 % a été choisi car il correspond, à peu près, à la hauteur des ascendants et descendants des lignes de texte.

Comme le redimensionnement de l'image d'annotation peut provoquer des fusions indésirables, les polygones englobants sont d'abord redimensionnés à la taille de l'image d'entrée du réseau, puis la scission est appliquée à cette échelle. Ainsi, l'image d'annotation est directement générée à la résolution souhaitée de l'entrée du réseau, ce qui empêche la fusion de certaines lignes. L'image de droite de la Figure 5.1 présente le résultat de ce processus. Comme nous pouvons le voir, l'image d'annotation produite contient des polygones mieux séparés. Même si certaines lignes se chevauchent encore sur certaines pages, nous espérons avoir généré une vérité terrain plus appropriée qui aidera à entraîner le modèle de détection et à améliorer sa capacité à prédire des lignes de texte séparées. Le code permettant de générer ces annotations modifiées et les images de labels utilisées dans les expériences sont accessibles publiquement [1].

## 5.2 COMPARAISON DES APPROCHES DE DÉTECTION

Pour nos expériences, nous avons choisi d'étudier trois systèmes à l'état de l'art : Doc-UFCN (Boillet et al., 2021a), dhSegment (Ares Oliveira et al., 2018) et ARU-Net (Grüning et al., 2019). Doc-UFCN a été choisi pour ses bonnes performances sur des jeux de données historiques et son nombre réduit de paramètres. De plus, nous avons sélectionné les systèmes dhSegment et ARU-Net car ils sont open-source, faciles à entraîner et ont obtenu des performances satisfaisantes sur des tâches de détection sur des documents historiques. ARU-Net est

1. https://gitlab.com/teklia/dla/arkindex_document_layout_training_label_normalization

également le modèle de détection de lignes de texte utilisé dans Transkribus (Kahle et al., 2017). Nous présentons maintenant les systèmes et les détails d'entraînements puisque nous les avons tous entraînés afin de pouvoir comparer équitablement leurs performances.

### 5.2.1 DOC-UFCN

Le système Doc-UFCN est le même que celui présenté en section 4.3. La seule différence est la taille d'entrée du réseau. En effet, nous avons montré dans le chapitre précédent, que de meilleures performances sont obtenues avec des images d'entrée plus grandes. Ainsi, les images sont redimensionnées telles que leur plus grande dimension soit de 768 pixels tout en conservant leur rapport d'aspect original. Par conséquent, pour entraîner Doc-UFCN, les annotations sont directement générées à 768 pixels grâce au processus présenté dans la section 5.1.1. Pour les expériences suivantes, Doc-UFCN est entraîné avec un taux d'apprentissage initial de $5e-3$, des mini-batchs de taille 2, l'optimiseur Adam, la fonction de coût Dice et l'arrêt anticipé (*early stopping*).

### 5.2.2 DHSEGMENT

L'encodeur pré-entraîné présent dans le système dhSegment nécessite que les images d'entrée soient de taille fixe de 400×400 pixels. Ainsi, contrairement aux deux autres systèmes, dhSegment est entraîné sur des patchs de 400×400 pixels d'images en taille réelle. Le processus de scission est donc appliqué sur les polygones de taille originale. Le modèle est entraîné avec un arrêt anticipé et des mini-batchs de taille 4. De plus, nous avons conservé le post-traitement proposé dans Ares Oliveira et al. (2018) en seuillant les probabilités avec $t = 0{,}7$. Différentes valeurs ont été testées pour ce paramètre et le seuil de 0,7 a donné les meilleurs résultats sur l'ensemble de validation.

### 5.2.3 ARU-NET

Le système ARU-Net (Grüning et al., 2019) est une version étendue du modèle U-Net (Ronneberger et al., 2015) standard. Deux concepts ont été ajoutés : une attention spatiale et une structure résiduelle. L'attention spatiale (A-Net) est un CNN multicouche et est utilisée pour gérer différentes tailles de police sur une même page. Les blocs résiduels sont introduits pour permettre la rétro propagation des erreurs sur les couches basses du réseau.

Pour les annotations ARU-Net, nous utilisons le même processus que pour Doc-UFCN mais sur des polygones redimensionnés à 33 % de leur taille originale. Le modèle est entraîné en utilisant l'arrêt anticipé. Nous avons utilisé la fonction de coût d'entropie croisée et un taux d'apprentissage initial de $1e-3$. Comme pour dhSegment, il faut seuiller les probabilités pour obtenir les prédictions finales. Cependant, le choix du seuil pour ARU-Net n'a pas été une tâche facile car il a un impact réel sur les résultats. Finalement, nous avons choisi un seuil bas de $t = 0{,}3$ car une valeur plus élevée éliminerait une quantité importante de pixels de lignes de texte.

Table 5.2 – Comparaison des systèmes Doc-UFCN, dhSegment et ARU-Net : nombre de paramètres, en millions, et temps d'inférence moyen, en secondes par image, mesuré sur l'ensemble de test du jeu de données Balsac. dhSegment possède 32,8 millions de paramètres mais comme l'encodeur est pré-entraîné, seulement 9,36 millions doivent être entraînés.

| Système | Temps d'inférence† | Paramètres |
|---|---|---|
| Doc-UFCN | **0,41** | **4,09** |
| dhSegment | 2,95 | 32,8 (9,36) |
| ARU-Net | 1,39 | 4,14 |

† Prédictions faites sur une carte graphique GeForce RTX 2070 8G.

Pour toutes les architectures, nous conservons les meilleurs modèles en validation. En outre, les éléments détectés de taille inférieur à 50 pixels sont supprimés. Cette paramétrisation est optimisée sur l'ensemble de validation.

La Table 5.2 montre le nombre de paramètres et les temps d'inférence des trois systèmes. Doc-UFCN et ARU-Net ont des poids similaires en nombre de paramètres tout en étant beaucoup plus légers que dhSegment. Pour les temps d'inférence, dhSegment et ARU-Net sont peu compétitifs, étant bien plus lents que Doc-UFCN.

## 5.3 ÉVALUATION DES DÉTECTIONS

Les trois systèmes ont été entraînés sur toutes les images d'entraînement afin de disposer de modèles génériques. Dans cette section, nous présentons les résultats au niveau des pixels et des objets. Pour une comparaison équitable, toutes les prédictions sont d'abord redimensionnées à la taille de l'image originale avant de procéder à l'évaluation. En outre, pour être comparables à d'autres résultats publiés dans la littérature, les modèles sont évalués avec les lignes de vérité terrain originales. Il est intéressant de noter que, malgré des résultats visuels solides, cette évaluation basée sur les annotations originales est en défaveur des systèmes testés puisque les polygones d'entraînement sont beaucoup plus fins que ceux de la vérité terrain. Puisque notre objectif est de développer un modèle historique générique obtenant des performances satisfaisantes sur des jeux de données hors échantillon, nous rapportons également les résultats sur les jeux de données ScribbleLens (Dolfing et al., 2020) et HOME-Alcar (Stutzmann et al., 2021), deux jeux de données qui n'ont pas été utilisés pour l'entraînement. Enfin, nous montrons l'impact de l'unification des annotations sur les résultats de détection lors de l'entraînement du système Doc-UFCN.

### 5.3.1 MÉTRIQUES NIVEAU PIXEL

Dans cette section, nous présentons les résultats d'évaluation des systèmes par les métriques pixel IoU et F1-score, souvent utilisées dans la littérature pour évaluer les systèmes de détection d'objets.

Table 5.3 – Résultats au niveau pixel obtenus par les systèmes Doc-UFCN, dhSegment et ARU-Net sur les ensembles de test. Les résultats présentent les performances des modèles génériques sans adaptation. ScribbleLens* rapporte les résultats des modèles spécifiques.

| Jeu de données | IoU | | | F1-score | | |
|---|---|---|---|---|---|---|
| | Doc-UFCN | dhSegment | ARU-Net | Doc-UFCN | dhSegment | ARU-Net |
| AN-Index | **0,69** | 0,68 | 0,68 | **0,82** | 0,81 | 0,74 |
| Balsac | 0,87 | 0,74 | **0,98** | **0,93** | 0,85 | 0,84 |
| BNPP | 0,65 | 0,60 | **0,90** | **0,78** | 0,75 | 0,75 |
| Bozen | 0,82 | 0,70 | **0,99** | **0,90** | 0,82 | 0,74 |
| cBAD2019 | 0,66 | 0,62 | **0,89** | **0,79** | 0,76 | 0,74 |
| DIVA-HisDB | 0,67 | 0,46 | **0,96** | **0,80** | 0,60 | 0,60 |
| HOME-NACR | 0,60 | 0,55 | **0,94** | **0,77** | 0,73 | 0,67 |
| Horae | 0,64 | 0,63 | **0,87** | **0,79** | **0,79** | 0,75 |
| READ-Complex | 0,49 | 0,58 | **0,81** | 0,70 | **0,73** | **0,73** |
| READ-Simple | 0,60 | 0,57 | **0,88** | **0,73** | 0,71 | 0,71 |
| HOME-Alcar | 0,35 | 0,49 | **0,58** | 0,49 | 0,60 | **0,70** |
| ScribbleLens | 0,35 | 0,36 | **0,41** | 0,51 | 0,51 | **0,58** |
| ScribbleLens* | 0,80 | **0,95** | – | 0,89 | **0,97** | – |

#### Comparaison des systèmes sur les ensembles de test des jeux d'entraînement

Les résultats obtenus par les trois réseaux sur les ensembles de test des jeux d'entraînement sont présentés en haut de la Table 5.3. Le réseau ARU-Net semble plus performant en termes d'IoU, alors qu'il l'est moins que les autres systèmes si nous considérons le score F1. En effet, le score F1 repose réellement sur les mesures de précision et de rappel (non présentées ici), en les résumant de manière précise. Ceci peut expliquer les faibles résultats obtenus par ARU-Net puisque ses scores en précision ne sont jamais supérieurs à 60 % (sauf pour Balsac). Au contraire, le score IoU étant moins focalisé sur les pixels correctement prédits (TP est considéré deux fois pour le score F1 et seulement une fois pour IoU), les scores IoU sont plus élevés, ce qui conduit à un meilleur classement dans le tableau des résultats.

Ces valeurs de précision faibles mais de rappel élevées obtenues par ARU-Net suggèrent que le modèle a correctement prédit la majorité des pixels de lignes de texte alors que, par ailleurs, beaucoup de pixels d'arrière-plan ont été classés comme lignes de texte. Cela reflète la présence de fusions dans les lignes détectées. La Figure 5.2 montre les prédictions obtenues par les modèles sur une image tirée au hasard dans l'ensemble de test Horae (Boillet et al., 2019). Elle confirme notre hypothèse selon laquelle ARU-Net a fusionné certaines lignes. Cependant, comme indiqué précédemment, l'utilisation d'un seuil plus élevé aurait conduit à manquer un grand nombre de pixels de lignes de texte. Nous pensons qu'ARU-Net n'est peut-être pas le système le plus approprié pour détecter des objets proches. En effet, il a souvent obtenu de très bonnes performances lorsqu'il était entraîné avec des lignes de base, où les objets sont plus espacés et plus fins que les polygones englobants des lignes de texte.

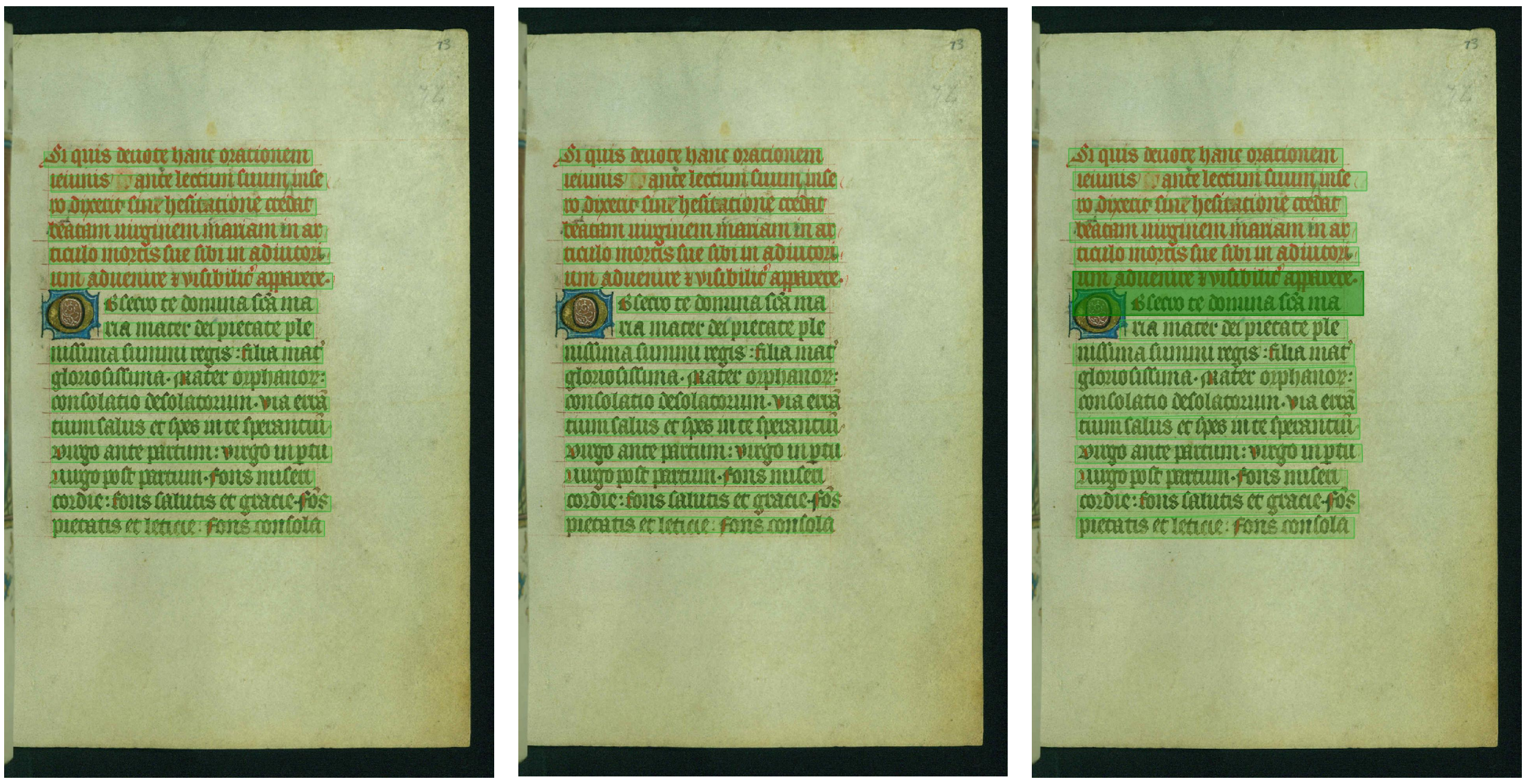

**Figure 5.2** – Détections de lignes produites sur une image du jeu de données Horae : Doc-UFCN à gauche, dhSegment au centre et ARU-Net à droite. Doc-UFCN et dhSegment produisent des résultats similaires, tandis que ARU-Net surestime l'épaisseur des lignes et fusionne plusieurs lignes (l'une d'elles est mise en évidence en vert foncé).

Comparer Doc-UFCN et dhSegment est un peu plus facile car ils se comportent de la même manière pour les scores IoU et F1. Doc-UFCN surpasse dhSegment sur la majorité des jeux de données pour les deux mesures. Il est cependant moins bon sur le jeu de données READ-Complex. Nous supposons que cela est dû au nombre élevé de petits objets dans les images des documents de ce jeu qui peuvent avoir été manqués par Doc-UFCN puisqu'il travaille à une faible résolution, contrairement à dhSegment et ARU-Net.

L'évaluation et la comparaison des trois modèles sur la base de l'IoU uniquement conduiraient à choisir ARU-Net comme étant le meilleur modèle. Or, nous avons montré que ses faibles précisions peuvent conduire à une faible capacité à distinguer des lignes proches. Les mesures au niveau objet, qui peuvent rendre compte des lignes fusionnées, devraient être utilisées en complément de ces valeurs de pixels.

### ÉVALUATION HORS ÉCHANTILLON

La Table 5.3 présente également les résultats des modèles génériques appliqués aux jeux de données ScribbleLens et HOME-Alcar. Nous avons également entraîné Doc-UFCN et dhSegment sur ScribbleLens afin de disposer de modèles spécifiques pour la comparaison. Pour le jeu de données HOME-Alcar, nous ne disposons pas d'images d'entraînement pour la détection de lignes, donc seuls les résultats génériques sont présentés.

Les performances obtenues par les trois systèmes sur les jeux de données ScribbleLens et HOME-Alcar sont bien inférieures à celles obtenues sur les jeux de données d'entraînement, et également à celles obtenues par les modèles spécifiques de ScribbleLens*. Pour le jeu de test ScribbleLens, la précision est égale à 97 % pour les modèles génériques dhSegment et Doc-UFCN alors qu'elle se situe entre 82 et 85 % pour HOME-Alcar. Cela suggère que presque tous les pixels prédits étaient corrects, alors qu'un grand nombre de pixels de la vérité terrain n'ont pas été détectés. Notre hypothèse est que les modèles ont prédit de bons polygones mais très fins par rapport aux polygones annotés très larges des pages ScribbleLens, ce qui a conduit à des valeurs d'IoU dégradées. Il en est de même pour les images HOME-Alcar, où des polygones fins rectangulaires comprenant uniquement quelques pixels d'arrière-plan ont probablement été prédits.

**Table 5.4 –** Résultats au niveau pixel obtenus par Doc-UFCN avec et sans uniformisation des labels. Les résultats montrent les performances des modèles génériques sans adaptation.

| Jeu de données | IoU | | F1-score | |
|---|---|---|---|---|
| | Originaux | Uniformes | Originaux | Uniformes |
| AN-Index | 0,67 | **0,69** | 0,80 | **0,82** |
| Balsac | 0,71 | **0,87** | 0,83 | **0,93** |
| BNPP | 0,63 | **0,65** | 0,77 | **0,78** |
| Bozen | 0,67 | **0,82** | 0,80 | **0,90** |
| cBAD2019 | 0,61 | **0,66** | 0,75 | **0,79** |
| DIVA-HisDB | 0,65 | **0,67** | 0,78 | **0,80** |
| HOME-NACR | 0,56 | **0,60** | 0,74 | **0,77** |
| Horae | **0,64** | **0,64** | 0,79 | **0,79** |
| READ-Complex | **0,53** | 0,49 | **0,74** | 0,70 |
| READ-Simple | 0,58 | **0,60** | 0,72 | **0,73** |
| HOME-Alcar | **0,51** | 0,35 | **0,63** | 0,49 |
| ScribbleLens | **0,42** | 0,35 | **0,59** | 0,51 |

Sur la base de ces métriques, nous ne pouvons pas être certains que les systèmes ne parviennent pas à généraliser sur les deux nouveaux ensembles de données. D'autres mesures pourraient donner un meilleur aperçu des capacités de généralisation réelles des modèles.

### impact de l'unification des annotations

Pour évaluer l'impact de l'unification des annotations sur les résultats, nous avons entraîné Doc-UFCN sur tous les jeux de données avec des annotations non uniformisées. Selon la Table 5.4, l'entraînement avec les annotations uniformisées améliore les performances au niveau pixel jusqu'à +16 points de pourcentage d'IoU sur Balsac. Cependant, comme expliqué pour ARU-Net dans la section 5.3.1, les mesures de rappel sont plus élevées sans le processus d'unification. En effet, les pixels entre les lignes consécutives et ceux le long des bords des lignes sont plus souvent prédits comme des lignes de texte, ce qui augmente les valeurs de rappel. Cependant, certains de ces pixels ne sont pas censés faire partie des lignes de texte (puisqu'ils créent des fusions), ce qui diminue les valeurs de précision. Sur la base de ces métriques, la scission des lignes proches semble être nécessaire pour aider le modèle à les distinguer.

#### LIMITATION DES MÉTRIQUES PIXEL

Même si ces mesures au niveau pixel peuvent donner une première idée de la performance d'un modèle, nous présentons, sur la Figure 3.4, deux exemples prouvant qu'elles peuvent ne pas être suffisantes. En effet, deux prédictions différentes peuvent être qualifiées par les mêmes valeurs d'IoU et de score F1 malgré une différence importante de qualité. Or, dans la littérature, les systèmes sont souvent comparés par leurs valeurs d'IoU et de F1-score, nous montrons donc ici que ces métriques ne sont pas appropriées pour choisir le meilleur modèle car elles ne prennent pas en compte le nombre d'objets détectés. En conclusion, ces métriques ne nous permettent pas de déterminer la capacité de généralisation des modèles entraînés. Pour surmonter ces problèmes, la section suivante présente et analyse les résultats des métriques au niveau objet.

### 5.3.2 MÉTRIQUES NIVEAU OBJET

Nous avons montré, dans la section précédente, que les métriques au niveau pixel peuvent ne pas être suffisantes pour une évaluation et une comparaison approfondies des modèles. Nous présentons maintenant les métriques au niveau objet et montrons qu'elles sont complémentaires aux métriques précédentes, et peuvent donner des informations plus parlantes sur la qualité d'un résultat de détection.

Comme indiqué précédemment, déterminer si un objet doit être considéré comme positif ou négatif est complexe. En se basant sur l'idée proposée dans les compétitions PASCAL VOC, il est possible de calculer la précision, le rappel et la précision moyenne (AP) au niveau de l'objet. Pour ce faire, les objets prédits et les objets annotés sont d'abord appariés en fonction de leurs scores IoU, de sorte qu'un seul objet prédit soit apparié à un objet annoté et inversement.

Ensuite, les objets appariés sont classés par score de confiance décroissant. Pour chaque objet prédit $i$, les mesures de précision $\mathrm{P}_i$ et de rappel $\mathrm{R}_i$ sont calculées en considérant uniquement les objets ayant des scores de confiance supérieurs ou égaux à celui de l'objet courant $i$. Ces mesures sont calculées en fonction d'un seuil d'IoU choisi $t$, à l'aide des équations 5.1 suivantes.

$$\mathrm{P}_i = \frac{TP_i}{Total_i} \qquad \mathrm{R}_i = \frac{TP_i}{Total_{GT}} \tag{5.1}$$

Ces équations s'appliquent avec :

— $TP_i$ : nombre d'objets positifs correctement prédits ayant une confiance supérieure ou égale à celle de l'objet $i$ ;

— $Total_i$ : nombre d'objets prédits ayant une confiance supérieure ou égale à celle de l'objet $i$ ;

— $Total_{GT}$ : nombre d'objets annotés à retrouver ;

où un objet est considéré comme positif si son IoU est supérieur au seuil choisi $t$.

Table 5.5 – Résultats au niveau ligne obtenus par les systèmes Doc-UFCN, dhSegment et ARU-Net sur les ensembles de test. Les résultats présentent les performances des modèles génériques sans adaptation. ScribbleLens* rapporte les résultats des modèles spécifiques.

| Jeu de données | AP@.5 | | | AP@[.5, .95] | | |
|---|---|---|---|---|---|---|
| | Doc-UFCN | dhSegment | ARU-Net | Doc-UFCN | dhSegment | ARU-Net |
| AN-Index | 0,75 | **0,76** | 0,51 | 0,34 | **0,35** | 0,17 |
| Balsac | **0,98** | 0,94 | 0,76 | **0,76** | 0,51 | 0,34 |
| BNPP | **0,83** | 0,78 | 0,50 | **0,31** | 0,27 | 0,13 |
| Bozen | **0,99** | 0,74 | 0,01 | **0,69** | 0,35 | 0,0 |
| cBAD2019 | **0,86** | 0,71 | 0,29 | **0,48** | 0,24 | 0,07 |
| DIVA-HisDB | **0,77** | 0,39 | 0,10 | **0,36** | 0,17 | 0,04 |
| HOME-NACR | **0,85** | 0,78 | 0,19 | **0,46** | 0,28 | 0,04 |
| Horae | 0,83 | **0,85** | 0,56 | **0,38** | 0,34 | 0,17 |
| READ-Complex | 0,60 | **0,62** | 0,22 | 0,23 | **0,24** | 0,08 |
| READ-Simple | **0,69** | 0,58 | 0,21 | **0,28** | 0,21 | 0,05 |
| HOME-Alcar | 0,16 | **0,76** | 0,0 | 0,03 | **0,26** | 0,0 |
| ScribbleLens | **0,06** | 0,02 | 0,0 | **0,02** | **0,02** | 0,0 |
| ScribbleLens* | **0,94** | 0,0 | – | **0,61** | 0,0 | – |

La courbe Précision-Rappel est ensuite calculée et interpolée et la précision moyenne (AP) est définie comme l'aire sous cette courbe. Cette AP est calculée pour toutes les classes d'une expérience, puis la moyenne est calculée pour toutes les classes, ce qui donne la précision moyenne (mAP). Pour la détection des lignes de texte, nous n'avons qu'une seule classe d'objets, la mAP est donc égale à l'AP et est notée AP@$t$ dans la suite, $t$ étant toujours le seuil IoU.

#### comparaison des systèmes sur les ensembles de test des jeux d'entraînement

La Table 5.5 présente les résultats d'AP obtenus sur les ensembles de test pour un seuil d'IoU de 50 % (AP@.5). En outre, et afin de s'affranchir de tout seuil, la moyenne des AP sur une plage de valeurs d'IoU (50 % – 95 %) est également calculée et présentée comme AP@[.5,.95].

Les résultats présentés ici renforcent notre hypothèse précédente selon laquelle ARU-Net ne parvient pas à séparer les objets proches. En effet, tous les résultats de ARU-Net sont très inférieurs à ceux des deux autres systèmes, sauf pour le jeu de données Balsac où les polygones des lignes de texte sont vraiment espacés dans les annotations. De plus, nous observons que, pour un seuil bas de 50 %, Doc-UFCN surpasse légèrement dhSegment. En passant de 50 % à la moyenne des AP, nous constatons que les résultats des deux modèles se dégradent, ce qui signifie que, avec des seuils plus élevés, certaines lignes deviennent considérées comme fausses positives car leur localisation n'est pas assez précise. Cependant, cette dégradation est plus faible pour Doc-UFCN que pour dhSegment, ce qui signifie que la localisation des objets par dhSegment est moins précise que celle de Doc-UFCN. La Figure 5.3 présente les résultats des trois modèles sur une image tirée au hasard dans le jeu de données Bozen. Ces prédictions confirment l'intérêt des mesures AP pour évaluer les prédictions de détection puisqu'elles mettent en évidence les mauvais comportements, comme ceux montrés par ARU-Net.

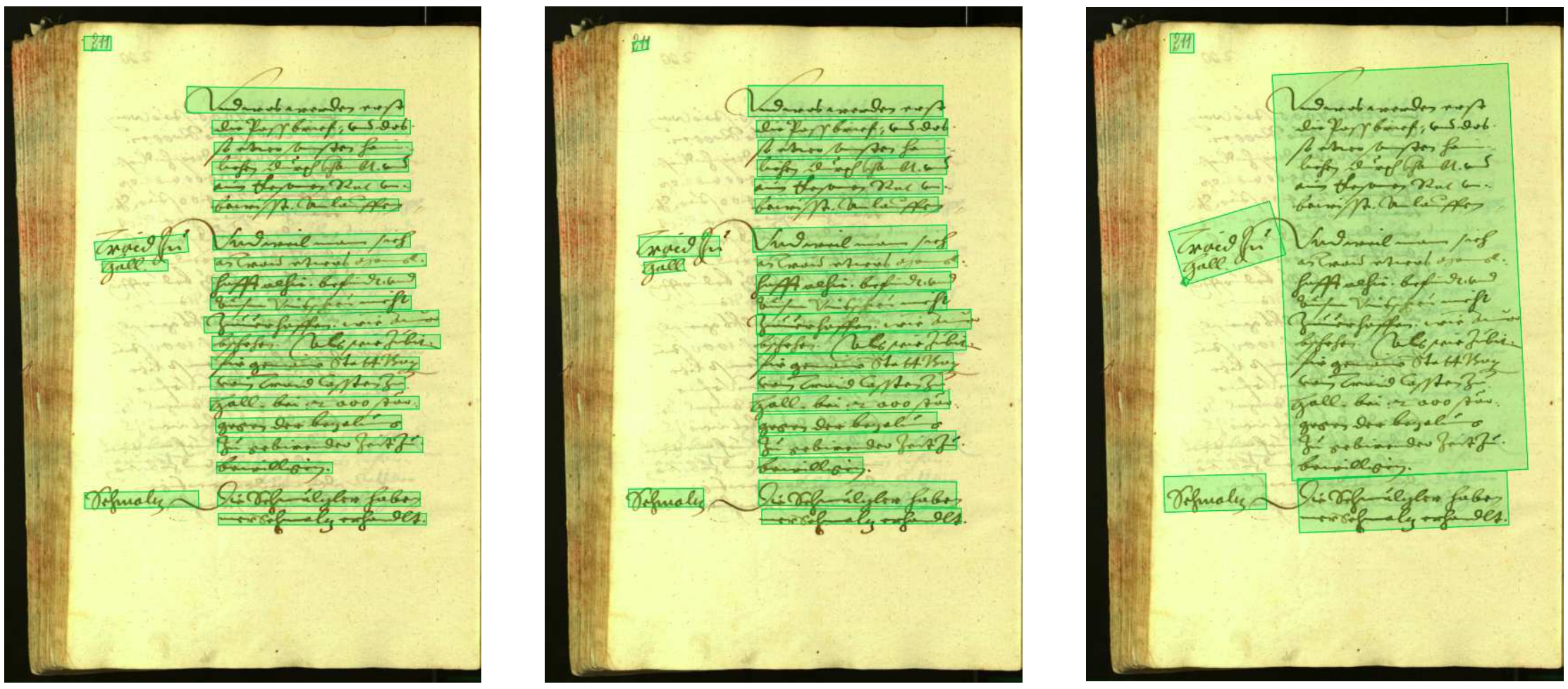

**Figure 5.3 –** Détections de lignes produites sur une image du jeu de données Bozen : Doc-UFCN à gauche, dhSegment au centre et ARU-Net à droite. Doc-UFCN prédit des lignes bien séparées alors que ARU-Net prédit des lignes fusionnées. dhSegment ne produit pas de lignes fusionnées mais elles sont plus proches que celles produites par Doc-UFCN. De plus, les polygones de dhSegment incluent plus d'espace en haut des lignes, ce qui peut avoir un impact négatif sur la reconnaissance du texte.

### ÉVALUATION HORS ÉCHANTILLON

La Table 5.5 présente également les résultats des modèles génériques appliqués à ScribbleLens et HOME-Alcar et les modèles spécifiques de ScribbleLens. Les résultats obtenus par les modèles spécifiques au niveau objet sont totalement opposés à ceux obtenus au niveau pixel. Les résultats obtenus par Doc-UFCN confirment que le modèle fonctionne bien lorsqu'il est entraîné directement sur ScribbleLens, sauf sur quelques images comme montré sur la Figure 5.4. Au contraire, alors que dhSegment a obtenu de bonnes mesures au niveau des pixels, ses valeurs d'objet sont toutes à 0. En effet, comme pour les résultats précédents obtenus par ARU-Net, nous observons de nombreuses lignes fusionnées prédites par le modèle spécifique dhSegment, ce qui signifie que ce dernier n'a pas réussi à apprendre directement à partir des images ScribbleLens.

Les faibles scores AP des modèles génériques peuvent être expliqués par la façon dont le jeu de données ScribbleLens a été annoté : des rectangles englobants très larges. Les modèles ayant été entraînés sur des polygones bien divisés et beaucoup plus fins, seuls quelques polygones réels ont été appariés aux polygones prédits lors du calcul de l'AP. La même observation s'applique aux résultats du jeu HOME-Alcar. Les Figures 5.4 et 5.5 présentent une visualisation des résultats obtenus sur les images des bases ScribbleLens et HOME-Alcar. Malgré des valeurs de métriques peu élevées, les modèles génériques semblent nettement surpasser les modèles spécifiques, d'où l'importance de développer des modèles génériques. De plus, il est nécessaire de les évaluer sur des annotations cohérentes avec celles de l'ensemble d'entraînement des modèles.

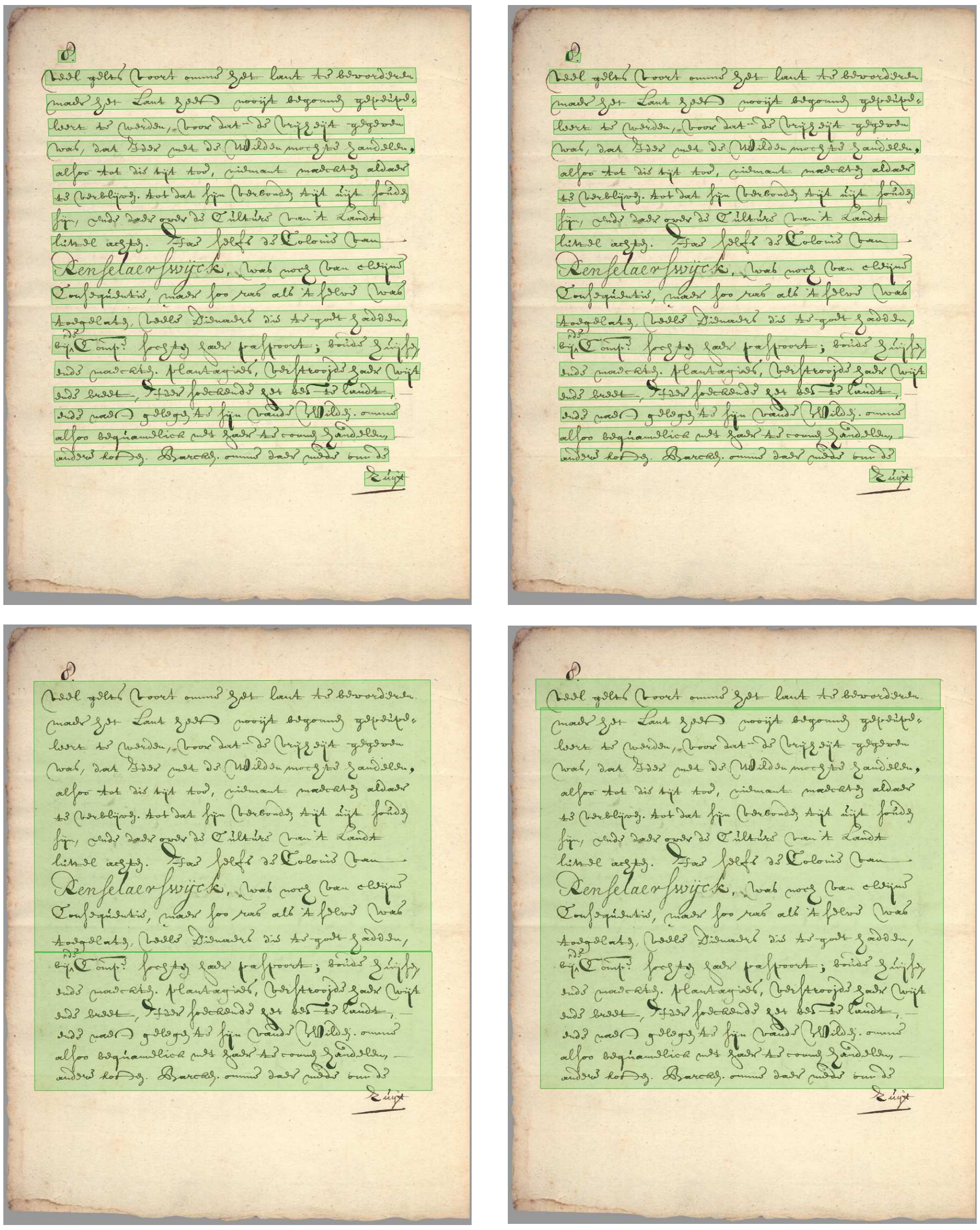

**Figure 5.4 –** Détections de lignes produites par les modèles génériques, en haut, et spécifiques, en bas, sur une image du jeu de données ScribbleLens. Les images de gauche montrent les résultats produits par Doc-UFCN et celles de droite par dhSegment.

### IMPACT DE L'UNIFICATION DES ANNOTATIONS

Sans surprise, selon la Table 5.6, presque toutes les valeurs sont meilleures lorsque nous utilisons le modèle entraîné sur les annotations unifiées, parfois avec une marge assez importante (+33 points de pourcentage pour Balsac et +37 pour Bozen). Pour le jeu de données DIVA-HisDB, les résultats sont mitigés. Nous supposons que cela est dû au processus d'unification qui peut considérablement modifier les annotations en réduisant la hauteur de la ligne.

Ces métriques au niveau objet ont souligné la nécessité de les utiliser avec celles au niveau du pixel pour évaluer et comparer les modèles. Cependant, il est encore difficile de voir l'avantage d'utiliser des modèles génériques sur des documents hors échantillon. Les métriques orientées vers les objectifs, décrites dans la section suivante, permettront une meilleure comparaison des objets prédits et des objets réels.

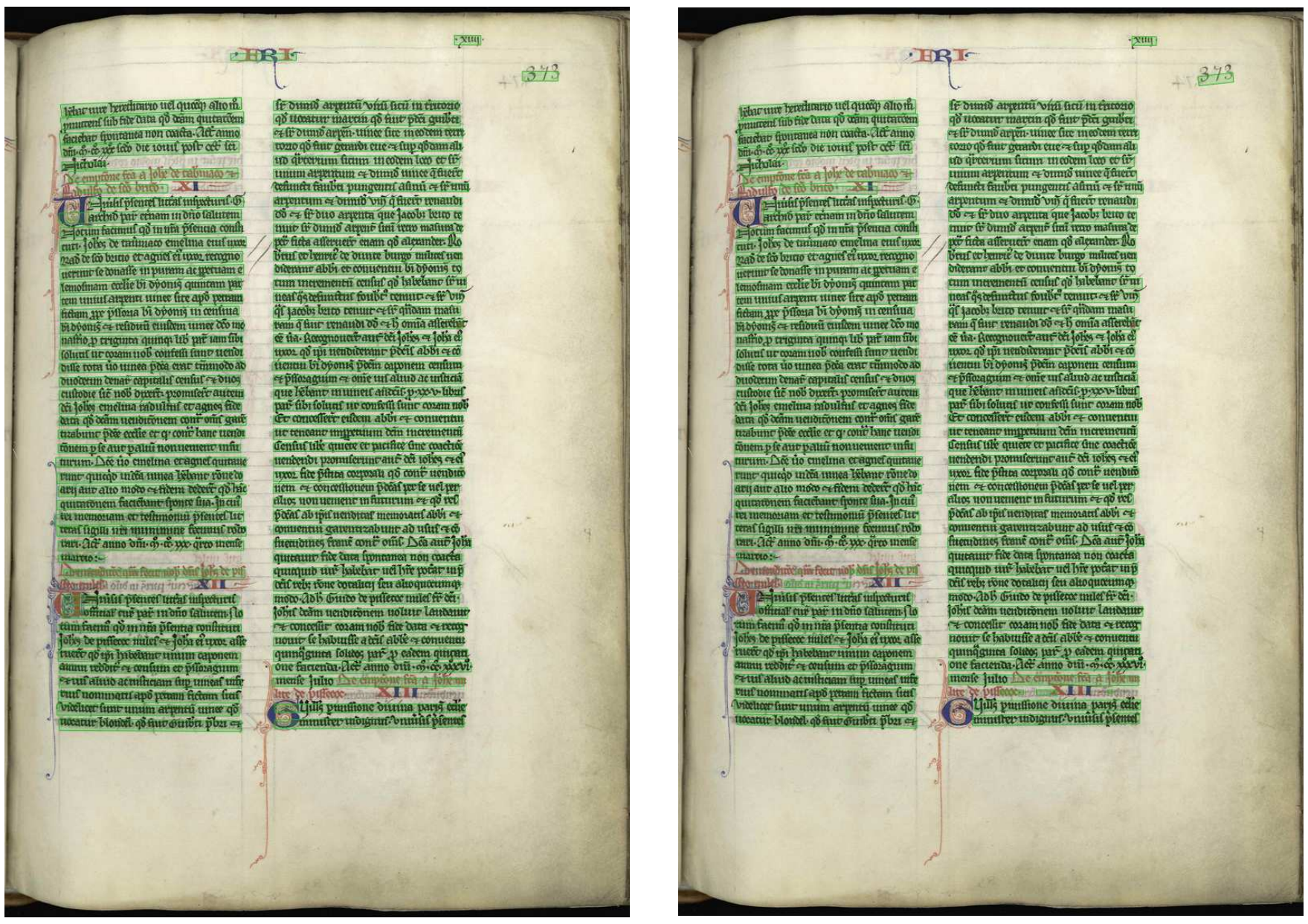

**Figure 5.5 –** Détections de lignes produites par les modèles génériques Doc-UFCN, à gauche, et dh-Segment, à droite, sur une image du jeu de données HOME-Alcar.

**Table 5.6 –** Résultats au niveau ligne obtenus par Doc-UFCN avec et sans uniformisation des labels. Les résultats montrent les performances des modèles génériques sans adaptation.

| Jeu de données | AP@.5 | | AP@[.5,.95] | |
|---|---|---|---|---|
| | Originaux | Uniformes | Originaux | Uniformes |
| AN-Index | 0,69 | **0,75** | 0,28 | **0,34** |
| Balsac | 0,95 | **0,98** | 0,44 | **0,76** |
| BNPP | 0,81 | **0,83** | 0,30 | **0,31** |
| Bozen | 0,77 | **0,99** | 0,31 | **0,69** |
| cBAD2019 | 0,71 | **0,86** | 0,25 | **0,48** |
| DIVA-HisDB | **0,86** | 0,77 | **0,40** | 0,36 |
| HOME-NACR | 0,82 | **0,85** | 0,33 | **0,46** |
| Horae | **0,84** | 0,83 | 0,34 | **0,38** |
| READ-Complex | **0,61** | 0,60 | **0,24** | 0,23 |
| READ-Simple | 0,60 | **0,69** | 0,19 | **0,28** |
| HOME-Alcar | **0,86** | 0,16 | **0,27** | 0,03 |
| ScribbleLens | **0,41** | 0,06 | **0,08** | 0,02 |

## 5.4 ÉVALUATION ORIENTÉE VERS LA TÂCHE DE RECONNAISSANCE

Dans les sections précédentes, nous avons discuté des résultats aux niveaux pixel et objet pour les trois modèles. Nous avons montré que les mesures d'objets donnent plus d'informations sur la qualité et les performances d'un modèle de détection de lignes de texte que les mesures au niveau du pixel. Cependant, elles sont toujours limitées lorsque les objets prédits sont trop fins par rapport aux objets réels. Les mesures orientées vers la tâche finale peuvent aider à déterminer les capacités réelles des modèles dans ce cas.

Nous avons effectué une évaluation orientée vers la reconnaissance de texte sur les cinq ensembles de données pour lesquels la transcription des lignes de texte est disponible, en calculant le taux d'erreur de caractère (CER) et le taux d'erreur de mot (WER). Par souci de clarté, dans les tables suivantes, seuls les CER sont présentés car les WER y sont fortement corrélés.

Pour réaliser cette évaluation axée sur la reconnaissance du texte, nous avons utilisé un reconnaisseur de texte manuscrit (Boros et al., 2020) basé sur la bibliothèque Kaldi (Arora et al., 2019). Le modèle est composé de deux éléments principaux : un modèle optique utilisant un modèle hybride Deep Neural Network-Hidden Markov Model et un modèle de langue fondé sur un modèle $n$-gram entraîné sur des sous-mots générés par la méthode *Byte Pair Encoding* (BPE). Contrairement au modèle de détection de lignes de texte entraîné sur tous les jeux de données, nous avons entraîné un modèle de reconnaissance spécifique pour chaque jeu de données et utilisé ces modèles pour l'évaluation.

Les paragraphes suivants présentent et analysent les résultats de la détection à l'aide de deux métriques basées sur le CER au niveau des pages et des lignes.

### 5.4.1 CER NIVEAU PAGE

Pour commencer l'évaluation, nous avons d'abord choisi de calculer le CER au niveau de la page. Les calculs sont détaillés dans l'Algorithme 5.1. Tout d'abord, tous les polygones de lignes prédits et annotés d'une image sont triés de haut en bas et de gauche à droite de l'image. En suivant cet ordre, toutes les transcriptions sont concaténées en une seule ligne de texte et le CER@page est calculé. La Table 5.7 présente les CER@page obtenus par les systèmes. En outre, nous avons calculé le CER obtenu par le système HTR lors de la transcription des polygones annotés manuellement. Par conséquent, la colonne "Manuel" des tables suivantes correspond au meilleur CER réalisable avec le système de détection idéal. Il s'agit du CER que nous aurions si nous avions 100 % pour toutes les métriques pixel et objet.

**Algorithme 5.1** Calcul du CER@page

**Entrée:** $HTR \leftarrow$ modèle de reconnaissance entraîné
**Entrée:** $DLA \leftarrow$ modèle de détection de lignes de texte entraîné
**Entrée:** $image \leftarrow$ image à évaluer
**Entrée:** $transcription \leftarrow$ transcription manuelle de l'image, texte ordonné de haut en bas, gauche à droite

$lignes \leftarrow DLA(image)$
$ord(lignes)$ {ordonne les lignes de haut en bas, gauche à droite}
$prediction \leftarrow$ ""
**pour chaque** $ligne \in lignes$ **faire**
    $prediction \leftarrow concat(prediction, HTR(ligne))$
**fin pour**
$cer \leftarrow CER(prediction, transcription)$
**Sortie:** $cer$

Table 5.7 – Résultats de reconnaissance niveau page obtenus par les systèmes Doc-UFCN, dhSegment et ARU-Net sur les ensembles de test. Les résultats présentent les performances des modèles génériques sans adaptation. ScribbleLens* rapporte les résultats des modèles spécifiques.

| Jeu de données | CER@page (%) | | | |
|---|---|---|---|---|
| | Manuel | Doc-UFCN | dhSegment | ARU-Net |
| Balsac | 4,3 | **14,9** | 15,8 | 31,5 |
| BNPP | 15,5 | **37,2** | 38,2 | 46,5 |
| Bozen | 5,8 | **11,7** | 13,2 | 74,9 |
| HOME-NACR | 11,9 | 38,6 | **22,3** | 75,2 |
| Horae | 10,3 | 14,8 | **12,1** | 31,5 |
| HOME-Alcar | 12,5 | **37,4** | 43,5 | 43,3 |
| ScribbleLens | 4,4 | **9,5** | 21,9 | 15,4 |
| ScribbleLens* | 4,4 | **25,2** | 92,6 | – |

### Comparaison des systèmes sur les ensembles de test des jeux d'entraînement

Les résultats de la Table 5.7 montrent la faible performance de la reconnaissance de texte sur les lignes détectées par ARU-Net. Ce taux d'erreur élevé est la conséquence de la fusion de nombreuses lignes de texte détectées qui ne peuvent pas être correctement reconnues, comme cela a déjà été mis en évidence avec l'évaluation au niveau de l'objet.

Doc-UFCN est plus performant que dhSegment sur trois des cinq jeux de données et l'est légèrement moins que dhSegment pour le jeu de données Horae. Ces résultats confirment ceux obtenus avec les évaluations au niveau du pixel et de l'objet. Doc-UFCN est cependant loin derrière dhSegment sur le jeu de données HOME-NACR, contrairement aux résultats obtenus avec les métriques pixel et objet.

Si nous analysons davantage les résultats de détection obtenus par Doc-UFCN sur le jeu de données HOME-NACR, nous constatons qu'environ la moitié des pages ont été parfaitement segmentées sans aucune fusion, ce qui augmente considérablement les scores AP. Cependant, les autres pages contiennent des lignes prédites qui sont des fusions de deux, trois lignes, ou même des fusions de lignes de paragraphes entiers. Cela conduit à une légère diminution des scores AP mais à une dégradation drastique des performances en termes de CER. En effet, si une seule fusion n'a qu'un faible impact sur le score AP, elle a un impact direct sur le CER par deux types d'erreurs :

— Le CER entre la prédiction et sa ligne annotée correspondante (qui est souvent élevé dans le cas d'une fusion) ;
— Le CER des lignes annotées non appariées, égal à la longueur de chaque ligne.

Cette seconde erreur n'est pas significative lorsque seules quelques lignes annotées ne sont pas appariées. C'est le cas pour les quatre premiers ensembles de données où le nombre de fusions est négligeable. Elle est encore moins significative lorsque les lignes non appariées ont un petit nombre de caractères. Cependant, HOME-NACR est le jeu de données avec la plus grande densité de caractères par ligne (jusqu'à six fois plus que les autres jeux de données).

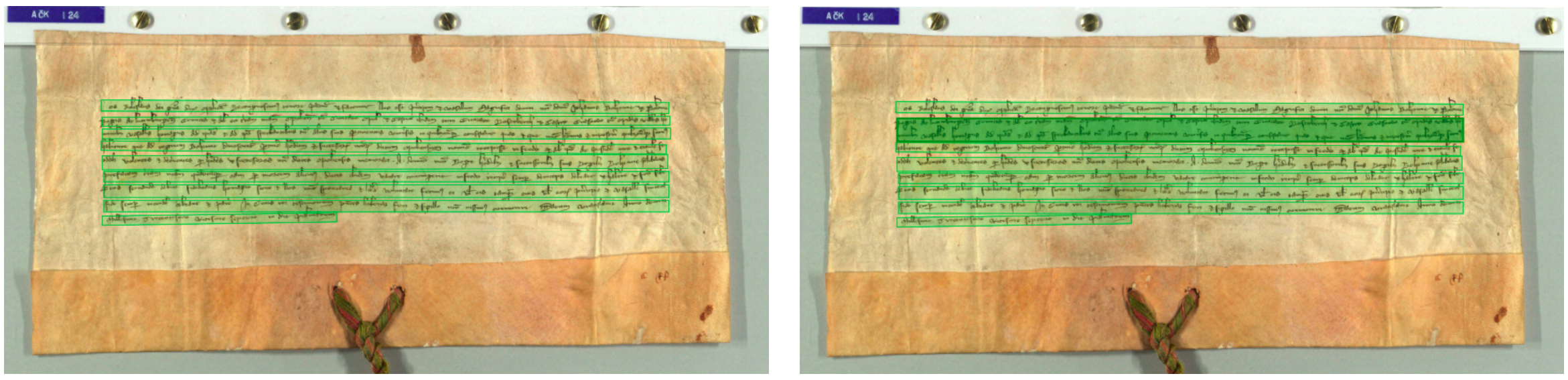

**Figure 5.6 –** Simulation des scores lorsque deux lignes sont bien séparées, à gauche, et fusionnées, à droite, sur une image du jeu de données HOME-NACR. À gauche, AP[.5,.95]=60 % et CER@page=7.3 % ; à droite, AP[.5,.95]=51.3 % et CER@page=20.4 %.

C'est pourquoi, cette seconde erreur a un réel impact sur le CER final de l'ensemble de données HOME-NACR. C'est également la raison pour laquelle les scores AP ne révèlent pas le phénomène.

La Figure 5.6 illustre ce point : l'image de gauche est correctement segmentée alors que sur celle de droite, deux lignes sont fusionnées. Dans ce cas, l'introduction d'une fusion dans les prédictions entraîne une diminution relative de la moyenne AP@[.5,.95] de 15 % (60 % à 51,3 %) tandis que le CER@page se dégrade de 179 % (7,3 % à 20,4 %). Cela prouve qu'une fusion n'affecte pas les différentes métriques de la même manière.

Toujours sur le jeu de données HOME-NACR, dhSegment montre une localisation moins précise des lignes de texte (scores AP plus faibles) par rapport à Doc-UFCN mais très peu de fusions, ce qui conduit à de meilleures performances de reconnaissance. En effet, HOME-NACR est le jeu de données où l'écriture est la plus dense et où les lignes sont les plus proches les unes des autres parmi les dix jeux de données. En raison de la mince hauteur des lignes de texte et du redimensionnement à 768 pixels, nous pensons que Doc-UFCN n'est pas l'architecture la plus adaptée pour travailler avec ces pages, contrairement à dhSegment qui présente une meilleure détection puisqu'il est appliqué sur les images dans leur taille originale.

Cette métrique supplémentaire donne de nouveau un aperçu des performances des modèles, en étant complémentaire aux métriques vues précédemment. Elle peut, en effet, détecter des comportements qui ne sont pas mis en évidence par les mesures de pixels ou d'objets.

### ÉVALUATION HORS ÉCHANTILLON

Les résultats de la généralisation sont également présentés dans la Table 5.7. Pour le jeu ScribbleLens, nous constatons l'avantage d'utiliser un modèle générique : les résultats des modèles spécifiques (Doc-UFCN 25,2 % de CER, dhSegment 92,5 de % CER) sont nettement moins bons que ceux des modèles génériques (Doc-UFCN 9,5 % de CER, dhSegment 21,9 % de CER). Les valeurs de CER du jeu HOME-Alcar sont élevées pour tous les systèmes, ce qui peut être dû à la complexité de certaines images de documents : mauvaise qualité de la numérisation, mauvaises conditions de conservation (certaines pages ont été déchirées, par exemple). Cependant, ces résultats mettent en évidence les capacités de généralisation du modèle générique Doc-UFCN, donnant de meilleurs résultats sur ScribbleLens que le modèle spécifique.

Table 5.8 – Résultats de reconnaissance niveau page obtenus par Doc-UFCN avec et sans uniformisation des labels. Les résultats présentent les performances des modèles génériques sans adaptation.

| Jeu de données | CER@page (%) | |
|---|---|---|
| | Originaux | Uniformes |
| Balsac | **14,4** | 14,9 |
| BNPP | **34,4** | 37,2 |
| Bozen | 27,6 | **11,7** |
| HOME-NACR | **33,5** | 38,6 |
| Horae | 15,1 | **14,8** |
| HOME-Alcar | 43,2 | **37,4** |
| ScribbleLens | 12,9 | **9,5** |

#### impact de l'unification des annotations

Comme présenté dans la Table 5.8, les deux Doc-UFCN avec et sans le processus d'unification ont des résultats assez similaires sur quatre jeux de données sans aucune dégradation significative. Cependant, l'impact de l'uniformisation des annotations est plus important sur la base de données Bozen. Pour la même raison que ARU-Net, le modèle entraîné avec les annotations originales prédit un grand nombre de lignes fusionnées, ce qui conduit à un taux d'erreur caractères très élevé par rapport au modèle entraîné avec les annotations uniformes. Le processus d'unification n'a pas amélioré les résultats sur Balsac et BNPP, car les annotations originales étaient déjà fines et constituaient une entrée correcte pour le système de reconnaissance HTR.

Même si l'entraînement avec les annotations uniformisées n'a pas montré d'amélioration significative des valeurs de CER pour quatre ensembles de données, il a eu un réel impact sur les prédictions de Bozen. Concernant les jeux de données hors échantillon, l'unification des annotations a également un impact positif. Le modèle entraîné avec les annotations uniformisées donne un CER de 9,5 % pour ScribbleLens et 37,4 % pour HOME-Alcar, ce qui correspond respectivement à 26 % et à 13 % de diminution relative de l'erreur caractère.

### 5.4.2 cer niveau ligne

Cette dernière mesure est étroitement liée au CER au niveau de la page. Ici, le CER n'est pas calculé sur le texte complet de la page, mais sur chaque ligne de texte prédite. À cet égard, les lignes prédites et les lignes annotées doivent d'abord être appariées. Dans la littérature, elles sont souvent appariées sur la base d'un seuil IoU de $t = 50$ %. Comme pour l'AP, nous avons calculé le CER pour ce seuil d'IoU de 50 % (CER@.5) ainsi qu'une moyenne sur la plage 50 % – 95 % d'IoU (CER@[.5,.95]). Les lignes prédites sont appariées avec celles annotées manuellement qui ont l'IoU la plus élevée de sorte qu'une seule prédiction soit appariée avec une annotation et inversement. Une fois les lignes appariées, nous calculons le CER pour tous les couples dont l'IoU est supérieur au seuil. En outre, le CER final est pénalisé par toutes les lignes qui ne sont pas appariées.

**Table 5.9** – Résultats de reconnaissance niveau ligne obtenus par les systèmes Doc-UFCN, dhSegment et ARU-Net sur les ensembles de test. Les résultats présentent les performances des modèles génériques sans adaptation. ScribbleLens* rapporte les résultats des modèles spécifiques.

| Jeu de données | CER@.5† (%) | | | CER@[.5, .95] (%) | | |
|---|---|---|---|---|---|---|
| | Doc-UFCN | dhSegment | ARU-Net | Doc-UFCN | dhSegment | ARU-Net |
| Balsac | **7,2/0,95** | 8,2/0,95 | 29,7/0,64 | **14,2** | 14,9 | 52,0 |
| BNPP | **22,4/0,93** | 21,5/0,83 | 32,1/0,73 | 44,2 | **42,8** | 53,2 |
| Bozen | **8,8/0,94** | 10,0/0,93 | 86,5/0,12 | 20,7 | **18,2** | 93,3 |
| HOME-NACR | 36,1/0,61 | **23,3/0,79** | 80,1/0,18 | 61,8 | **42,0** | 91,1 |
| Horae | 15,2/0,98 | **12,0/0,97** | 30,3/0,90 | **22,6** | 29,0 | 58,0 |
| HOME-Alcar | **22,9/0,92** | 27,6/0,73 | 30,0/0,72 | **46,6** | 49,3 | 54,0 |
| ScribbleLens | **9,8/0,80** | 18,2/0,37 | 15,8/0,81 | **40,3** | 60,0 | 45,5 |
| ScribbleLens* | **24,3/0,76** | 90,9/0,10 | – / – | **32,6** | 93,3 | – |

† CER@.5 / Proportion de caractères des lignes annotées appariés à une ligne de prédiction.
1 signifie que 100 % des caractères de l'annotation ont été appariés à une ligne de prédiction.

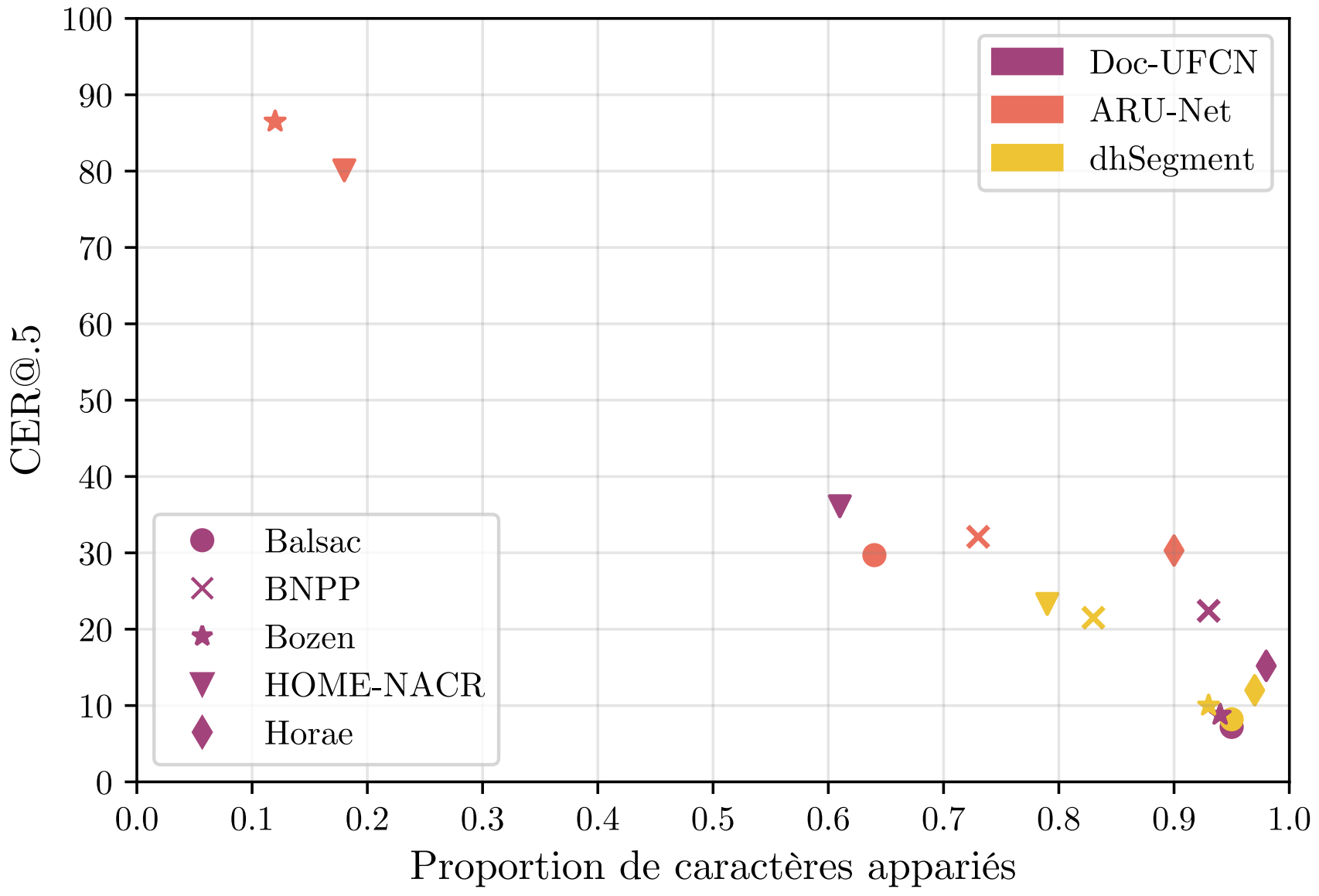

**Figure 5.7** – Résultats de reconnaissance niveau ligne obtenus, sur les ensembles de test, par les modèles génériques Doc-UFCN, dhSegment et ARU-Net sans adaptation.

La Table 5.9 et la Figure 5.7 présentent les résultats obtenus après le système HTR au niveau ligne. Pour les résultats de CER@.5, la Table montre la proportion de caractères des lignes annotées appariés à une ligne prédite pour calculer les valeurs de CER. Il aurait été possible de faire cela au niveau de la ligne (proportion de lignes appariées) pour voir sur quelle quantité de lignes les CER ont été calculés. Cependant, comme les lignes peuvent contenir un nombre variable de caractères, cela ne refléterait pas précisément le nombre réel de correspondances.

Table 5.10 – Résultats de reconnaissance niveau ligne obtenus par Doc-UFCN avec et sans uniformisation des labels. Les résultats montrent les performances des modèles génériques sans adaptation.

| Jeu de données | CER@.5 (%) | | CER@[.5,.95] (%) | |
|---|---|---|---|---|
| | Originaux | Uniformes | Originaux | Uniformes |
| Balsac | **7,2/0,97** | 7,2/0,95 | 17,1 | **14,2** |
| BNPP | **18,8/0,87** | 22,4/0,93 | **36,0** | 44,2 |
| Bozen | 27,3/0,67 | **8,8/0,94** | 47,3 | **20,7** |
| HOME-NACR | **29,4/0,73** | 36,1/0,61 | **46,6** | 61,8 |
| Horae | 15,7/0,98 | **15,2/0,98** | **20,6** | 22,6 |
| HOME-Alcar | 31,7/0,79 | **22,9/0,92** | 61,3 | **46,6** |
| ScribbleLens | 14,3/0,71 | **9,8/0,80** | 54,2 | **40,3** |

Comme pour les métriques précédentes, ARU-Net n'est pas compétitif : pas assez de caractères appariés et des valeurs de CER très élevées. Pour comparer Doc-UFCN et dhSegment, il est nécessaire d'étudier le CER et la proportion d'appariement dans leur ensemble. En effet, avoir un CER très bas lorsqu'il n'est calculé que sur une petite partie des lignes prédites n'est pas significatif puisque certaines lignes peuvent être plus faciles à reconnaître. Il est préférable d'avoir un bon compromis entre le nombre de caractères appariés et le taux d'erreur.

### Comparaison des systèmes sur les ensembles de test des jeux d'entraînement

Les résultats obtenus au niveau des lignes reflètent réellement ceux obtenus avec les scores AP et CER au niveau des pages. À 50 % d'IoU, Doc-UFCN semble meilleur pour les jeux de données Balsac, BNPP et Bozen. En outre, si nous considérons les résultats moyens CER@[.5,.95], dhSegment et Doc-UFCN ont tous deux des performances similaires sur les jeux de données Balsac, BNPP et Bozen. Dans le but d'avoir un modèle historique générique, les deux architectures semblent appropriées, obtenant de bons résultats au niveau des pixels et des objets, et des taux d'erreurs caractères acceptables au niveau des pages et des lignes.

### Évaluation hors échantillon

Les résultats de l'évaluation hors échantillon sont également présentés dans la Table 5.9. Les résultats pour ScribbleLens confirment ceux obtenus au niveau page. Les modèles génériques sont, en effet, meilleurs que les modèles spécifiques puisqu'ils montrent des scores de CER plus bas (Doc-UFCN 9,8 % de CER par rapport à 24,3 %, dhSegment 18,2 % de CER par rapport à 90,9 %) et des proportions d'appariement supérieures. Les performances sur les données HOME-Alcar sont comparables à celles obtenues au niveau page avec des taux d'erreurs élevés, malgré de hautes proportions d'appariement.

#### IMPACT DE L'UNIFICATION DES ANNOTATIONS

La Table 5.10 présente les résultats, au niveau ligne, de reconnaissance obtenus par Doc-UFCN avec et sans uniformisation des annotations. Les résultats présentés dans cette table sont semblables à ceux présentés au niveau page, à savoir des résultats assez similaires sur quatre jeux de données, avec et sans uniformisation, sans aucune dégradation significative. L'impact est cependant très important sur le jeu de données Bozen puisqu'il y a un gain de 18,5 points de pourcentage de CER@.5 en uniformisant les labels, impact expliqué par les mêmes raisons que celles exposées dans la section 5.4.1.

Concernant les jeux de données hors échantillon, l'unification des annotations a également un impact très positif avec des diminutions de CER@.5 de 8,8 et 4,5 points de pourcentages respectivement sur les bases HOME-Alcar et ScribbleLens, par rapport aux labels originaux.

## 5.5 CONCLUSION

Dans ce chapitre, nous avons montré qu'il est possible d'entraîner un modèle générique pour détecter les lignes de texte dans les documents historiques. Nous avons entraîné trois systèmes à l'état de l'art qui ont obtenu de bonnes performances sur différents ensembles de données. Ceci a été rendu possible par la création d'un large jeu de données d'entraînement, qui est, à notre connaissance, le plus grand et le plus diversifié des jeux de données historiques utilisés pour comparer les systèmes de segmentation de documents. Nous avons également montré que, lors de l'agrégation de différents ensembles de données, l'uniformisation des polygones englobants annotés réduit les incohérences d'annotation entre les jeux annotés et permet d'entraîner de meilleurs modèles. En outre, les modèles génériques entraînés sur plusieurs ensembles de données peuvent être meilleurs, non seulement sur les ensembles de données individuels, mais également sur les documents hors échantillon, ce qui prouve leurs capacités de généralisation.

Pour une évaluation pertinente des performances des trois systèmes, ce chapitre compare et analyse également plusieurs métriques de détection d'objets. Nous avons montré que les métriques standards au niveau pixel ne sont pas suffisantes car elles ne tiennent pas compte de la qualité des objets prédits. Pour pallier cet inconvénient, des métriques au niveau des lignes ont été introduites. Celles-ci ont montré que le système ARU-Net n'est pas approprié pour la tâche de détection de lignes de texte lorsqu'il est entraîné avec de telles annotations, le nombre de lignes fusionnées étant important par rapport aux deux autres approches. Ce système est, en effet, souvent utilisé pour détecter les lignes de base des documents, qui sont plus fines et plus espacées que les polygones englobants des lignes. Ces mesures ont également confirmé les bonnes performances de Doc-UFCN et dhSegment sur la plupart des jeux de données, fournissant une détection précise et exacte des objets. Ces résultats n'auraient pas été possibles en utilisant uniquement des mesures au niveau pixel. Nous sommes convaincus que l'utilisation des scores de précision moyenne est nécessaire pour évaluer correctement les modèles de détection de lignes de texte. Notre bibliothèque d'évaluation a été rendue publique [2], elle peut être utilisée sur n'importe quel jeu de données.

Enfin, ce chapitre fournit une évaluation orientée vers la tâche de reconnaissance de texte qui, à notre connaissance, n'a jamais été réalisée jusqu'à présent. Les métriques d'évaluation de reconnaissance HTR donnent encore davantage d'informations sur les objets prédits, étant complémentaires aux métriques au niveau des objets. De plus, elles permettent d'explorer l'impact de la qualité des lignes détectées sur les résultats finaux de reconnaissance.

2. https://gitlab.com/teklia/dla/document_image_segmentation_scoring

# 6 ESTIMATION DE LA CONFIANCE DES PRÉDICTIONS

Malgré les performances remarquables des réseaux de neurones profonds obtenus dans les travaux scientifiques, leur utilisation dans des applications réelles exige qu'ils soient, non seulement performants, mais aussi capables d'évaluer la confiance de leurs décisions. Ceci est particulièrement important pour les applications liées aux images médicales ou à la conduite autonome, par exemple. Le problème se pose également dans le cas de l'adaptation d'un modèle à un nouveau domaine, où nous souhaitons fournir au système le minimum de nouveaux exemples étiquetés pour réaliser l'adaptation. Le choix des exemples pertinents à soumettre à un annotateur humain est crucial pour optimiser le processus d'adaptation. Ce cadre, connu sous le nom d'apprentissage actif (*active learning)*, exige qu'un premier système effectue la tâche finale tout en évaluant automatiquement sa confiance sur de nouvelles données non vues, de sorte que les décisions moins confiantes puissent être soumises à un opérateur humain pour une annotation manuelle, tandis que les décisions plus confiantes prises par le système seraient conservées telles quelles pour fournir un étiquetage automatique. Dans ce chapitre, nous visons à développer des mesures de confiance pour l'adaptation d'un modèle de détection d'objets dans un cadre d'apprentissage actif, afin de réduire au minimum l'effort d'annotation humaine.

Pour cela, notre objectif est de construire un estimateur de confiance pour la détection d'objets dans des images de documents dans un scénario d'apprentissage actif. Dans ce but, nous étudions trois approches afin d'estimer la confiance. La première consiste à utiliser les probabilités de classe *a posteriori* du modèle de détection pour estimer la confiance. La seconde approche proposée est inspirée de la méthode de Monte Carlo (Gal et al., 2016) et consiste à construire des estimations de confiance en utilisant la méthode de *dropout* au moment du test. Le principal avantage de cette approche est qu'aucun entraînement supplémentaire n'est nécessaire pourvu que le modèle ait été entraîné avec des couches de *dropout*. Elle peut être appliquée à des modèles déjà entraînés sans aucune modification. Cette approche est cependant coûteuse en calculs, c'est pourquoi notre dernière proposition consiste à construire un système dédié qui peut prédire une estimation de confiance avec une seule prédiction pendant l'inférence. Indépendant du système de prédiction, ce système nécessite cependant une phase d'entraînement spécifique.

Ce chapitre présente tout d'abord, en section 6.1, les estimateurs de confiance que nous proposons. La configuration utilisée pour les expériences (données, détails de l'entraînement des modèles de détection et ceux des estimateurs de confiance) est ensuite détaillée dans la section 6.2. Enfin, dans la section 6.3, nous présentons et discutons les résultats obtenus.

## 6.1 MÉTHODES D'ESTIMATION DE LA CONFIANCE

Comme énoncé dans la section 2.2, très peu de travaux ont été proposés afin d'estimer la confiance des objets prédits par un modèle de détection d'objets dans les images. Certains travaux utilisent le *dropout* de Monte Carlo (Gal et al. (2016)) et analysent la distribution des prédictions afin d'estimer la confiance de la prédiction sans *dropout*. Dans d'autres travaux, un réseau adverse est entraîné pour estimer la proximité des prédictions avec la vérité terrain.

Dans ce qui suit, nous proposons quatre estimateurs de confiance. Le premier se base sur les probabilités *a posteriori* des classes données par le modèle de détection. Les deux suivants s'inspirent des travaux réalisés sur le *dropout* de Monte Carlo et sont déduits de la variance des prédictions calculées avec *dropout*. Enfin, le dernier se base sur des statistiques descriptives des objets attendus et prédits.

Dans la suite de ce chapitre, les objets prédits font référence aux composantes obtenues après l'application d'un modèle de détection au niveau pixel suivi d'un seuillage. Le seuillage assigne à chaque pixel la classe (ou fond) de plus grande probabilité.

### 6.1.1 ESTIMATEUR BASÉ SUR LES PROBABILITÉS *a posteriori*

Les réseaux neuronaux de détection d'objets produisent des probabilités au niveau pixel qui sont ensuite seuillées afin de créer des objets. Le premier estimateur que nous proposons, dénoté Posterior probability-based Confidence Estimator (PCE), utilise directement ces probabilités *a posteriori* afin d'estimer la confiance des prédictions.

Tout d'abord, le modèle de détection est appliqué à une image d'entrée, les probabilités $p_j$ obtenues pour chaque pixel sont ensuite seuillées afin d'en extraire les objets. Pour chaque objet prédit sur une image, nous calculons d'abord la moyenne des probabilités des pixels prédits par le modèle de détection. Ensuite, le score PCE de l'image est déduit en calculant la moyenne de toutes les probabilités des objets. Le calcul du score PCE est détaillé dans l'équation 6.1. Les valeurs calculées par cet estimateur sont comprises entre 0 et 1, une valeur de 1 étant interprétée comme un indicateur d'une détection correcte.

$$PCE = \frac{1}{N} \times \sum_{i=1}^{N} \left( \frac{1}{N_i} \times \sum_{j=1}^{N_i} p_j \right) \tag{6.1}$$

avec :

— $N$ : le nombre d'objets prédits sur l'image d'entrée ;

— $N_i$ : le nombre de pixels composant l'objet $i$ ;

— $p_j$ : la probabilité du pixel $j$ d'appartenir à la classe d'objet.

Cet estimateur présente les avantages d'être simple et rapide à calculer. De plus, il ne nécessite aucun entraînement supplémentaire autre que le modèle de détection, et peut ainsi être utilisé pour n'importe quel modèle de détection produisant des probabilités en sortie.

### 6.1.2 ESTIMATEURS BASÉS SUR LE DROPOUT DE MONTE CARLO

L'estimation de la confiance d'une prédiction avec le *dropout* de Monte Carlo consiste à calculer $N$ prédictions de la même observation et à analyser la distribution des prédictions. La variance entre les $N$ prédictions est un indicateur de l'incertitude du modèle et peut donc être considérée comme une estimation de la confiance. Dans cette partie, nous proposons deux scores résumant la variance des prédictions : la précision moyenne (Dropout Average Precision (DAP)) et la variance du nombre d'objets (Dropout Object Variance (DOV)).

#### DROPOUT AVERAGE PRECISION

Comme démontré dans le chapitre 5, la précision moyenne (mAP) utilisée dans les défis PASCAL VOC et décrite dans Boillet et al. (2022b) permet d'évaluer une prédiction au niveau objet par rapport à une annotation manuelle. L'avantage de cette métrique est qu'elle considère la taille et la position des objets prédits puisqu'elle s'appuie sur une correspondance des objets basée sur l'IoU. Inspirés de cette métrique, nous dérivons l'estimateur DAP qui est calculé en considérant chaque paire de prédictions ($(p_i, p_j)$ où $p_i$ et $p_j$ sont deux prédictions distinctes de la même image avec $i, j \in [1, N]$ et $i \neq j$) et en calculant la mAP (voir Focus 3.4) pour chaque paire, une des deux prédictions étant considérée comme vérité terrain arbitrairement. Enfin, le DAP est la moyenne de tous les scores mAP (voir l'équation 6.2). Les valeurs calculées par cet estimateur sont comprises entre 0 et 1, un score DAP élevé indique que les $N$ prédictions sont très similaires et est interprété comme un indicateur d'une détection correcte.

$$DAP = \frac{1}{N^2 - N} \times \sum_{i=1, j=1, i \neq j}^{N} mAP(p_i, p_j) \tag{6.2}$$

#### DROPOUT OBJECT VARIANCE

Le second estimateur que nous proposons est basé uniquement sur la variance du nombre d'objets prédits parmi les $N$ prédictions avec *dropout*. Lorsque le modèle est peu confiant, nous avons observé qu'un nombre très variable d'objets est prédit avec de nombreux petits objets autour de l'objet principal (comme le montre l'image de droite de la Figure 6.1). Pour obtenir une valeur unique, nous calculons la variance du nombre d'objets dans les prédictions avec *dropout* comme indiqué dans l'équation 6.3, où $n_i$ est le nombre d'objets dans la prédiction $p_i$. Les valeurs calculées par cet estimateur sont comprises entre 0 et 1, un score DOV de 0 indique que toutes les prédictions ont le même nombre d'objets et est interprété comme un indicateur d'une détection correcte.

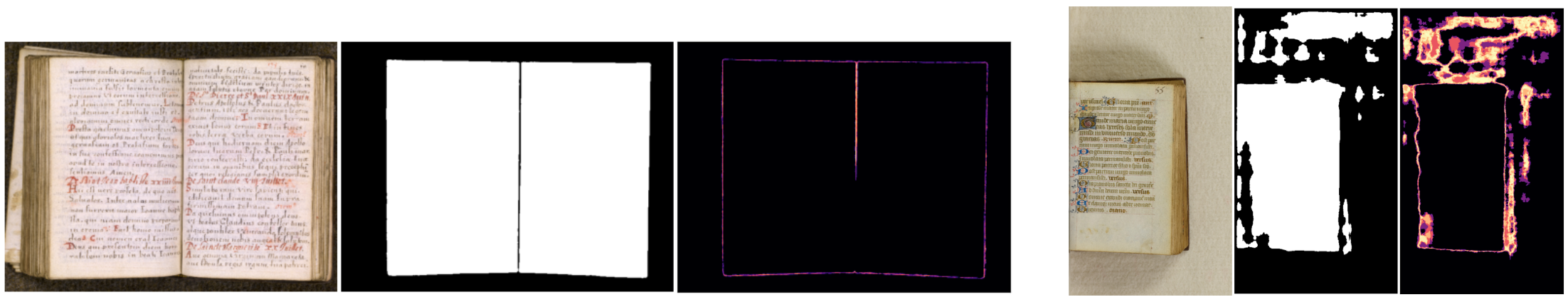

**Figure 6.1 –** Deux images issues du jeu de données Horae, à gauche, avec leurs prédictions, au centre et la variance pour $N$=10 prédictions avec *dropout*, à droite. Une variance élevée est représentée en jaune alors que les zones sans variance sont en noir. L'image de gauche a des estimations de confiance de DOV=0,0, DAP=1,0 et mAP-RFR=1,0 et celle de droite DOV=17,36, DAP=0,0993 et mAP-RFR=0,5553.

$$DOV = \frac{1}{N-1} \times \sum_{i=1}^{N} (n_i - \overline{n})^2 \quad \text{avec} \quad \overline{n} = \frac{1}{N} \times \sum_{i=1}^{N} n_i \tag{6.3}$$

Comme l'estimateur PCE, le calcul des scores DAP et DOV ne nécessitent pas d'autre entraînement que celui du modèle de détection. En effet, ces estimateurs peuvent être utilisés pour tout modèle de détection possédant des couches de *dropout*.

### 6.1.3 ESTIMATEUR BASÉ SUR LES STATISTIQUES D'OBJETS

Pour ce dernier estimateur, nous adoptons une approche basée sur une extraction de caractéristiques pour estimer la confiance. Nous concevons un système qui analyse les caractéristiques des objets détectés et estime la mAP, car aucune vérité terrain n'est disponible au moment du test. Contrairement à nos premières propositions, le système étant indépendant du détecteur, cette approche peut être appliquée à tout type de détecteur.

#### STATISTIQUES DESCRIPTIVES D'OBJETS

Les modèles de détection développés dans nos travaux fournissent, pour chaque pixel, les probabilités d'appartenir à une classe d'objet ou d'arrière-plan. Les pixels sont d'abord affectés à la classe ayant la plus forte probabilité, puis nous détectons les éléments constitués des pixels connexes, ce qui conduit à plusieurs objets prédits pour une image donnée. Ensuite, nous extrayons les polygones englobants des éléments détectés ainsi que leurs rectangles englobants. À partir de ces informations, nous calculons les huit caractéristiques d'objets suivantes. Pour chaque image et chaque objet, nous calculons :

1. Ratio entre la hauteur du rectangle englobant et la hauteur de l'image ;
2. Ratio entre la largeur du rectangle englobant et la largeur de l'image ;
3. Ratio entre la hauteur et la largeur du rectangle englobant ;
4. Ratio entre l'aire du polygone et l'aire de l'image ;
5. Ratio entre l'aire du polygone et l'aire du rectangle englobant ;
6. Ratio entre l'aire du rectangle englobant et l'aire de l'image ;

et pour chaque image :

7. Distances en y (hauteur) entre les centroïdes de tous les rectangles englobants, normalisées par la hauteur de l'image ;
8. Distances en x (largeur) entre les centroïdes de tous les rectangles englobants, normalisées par la largeur de l'image.

Les distances sont calculées en considérant chaque paire de rectangles englobants. Les caractéristiques permettent de décrire les tailles, les formes et les positions des objets détectés dans les images de documents. Pour une image donnée et chacune des huit caractéristiques, les ratios sont calculés pour chaque objet détecté dont les valeurs résultantes sont regroupées en $B$ intervalles pour fournir un histogramme. Les histogrammes de caractéristiques sont ensuite concaténés pour constituer un vecteur de statistiques d'objets de taille $8 \times B$. Ces statistiques sont ensuite utilisées pour entraîner un modèle de régression.

#### MEAN AVERAGE PRECISION - RANDOM FOREST REGRESSOR

Pour construire l'estimateur de confiance, nous avons choisi d'estimer la mAP des prédictions, car nous avons montré, en section 5.3.2, qu'elle est plus significative que l'IoU (Boillet et al., 2022b). Pour estimer la mAP d'une prédiction, plusieurs méthodes de régression peuvent être utilisées telles que la régression par vecteur de support (SVR) ou le régresseur Random Forest (RFR). Dans nos expériences, nous avons utilisé RFR, car il a obtenu les meilleurs résultats dans nos travaux préliminaires. Après l'application du modèle de régression, aucun traitement supplémentaire n'est nécessaire puisqu'il fournit directement un score unique considéré comme l'estimation de confiance. Dans ce qui suit, cet estimateur est appelé mean Average Precision - Random Forest Regressor (mAP-RFR).

## 6.2 CADRE EXPÉRIMENTAL

Nous avons évalué et comparé les estimateurs présentés en 6.1 sur deux tâches de difficultés différentes : la détection de pages et la détection de lignes de texte manuscrites. La détection de pages correspond au détourage des pages dans des prises de vues de doubles ou de simples pages dont les dimensions ne correspondent pas exactement aux dimensions des images produites par l'imageur (scanner ou caméra). Il s'agit d'une tâche assez simple puisqu'il y a souvent un ou deux objets sur une image. La détection de lignes de texte manuscrites est une tâche plus complexe car les pages de documents peuvent contenir un nombre variable, parfois important, de lignes de texte qui ont des formes et des positions très différentes.

### 6.2.1 JEUX DE DONNÉES

Pour les expériences de détection de pages, nous avons utilisé les jeux de données READ-BAD (Grüning et al., 2017) et Horae (Boillet et al., 2019). Notre objectif est d'adapter le modèle de détection pré-entraîné sur les données READ-BAD aux images de documents de la base Horae en annotant le moins de données possible. Pour la tâche de détection de lignes de texte, notre objectif est d'adapter un modèle générique pré-entraîné à un nouvel ensemble de documents hors échantillon d'apprentissage, à savoir le jeu de données Hugin-Munin (Maarand et al., 2022), détaillé en section 3.1. Les statistiques de ces jeux de données sont présentés dans la Table 6.1.

**Table 6.1 –** Statistiques des jeux de données utilisés pour la détection de pages.

| Jeu de données | | Images | Pages | | |
|---|---|---|---|---|---|
| | | | *simple* | *double* | *anormal* |
| READ-BAD Grüning et al. (2017) | train | 1 635 | 1 459 | 171 | 5 |
| | valid | 200 | 179 | 21 | – |
| | test | 200 | 179 | 20 | 1 |
| READ-BAD* Grüning et al. (2017) | train | 1 630 | 1 801 | – | – |
| | valid | 200 | 221 | – | – |
| | test | 199 | 219 | – | – |
| Horae Boillet et al. (2019) | train | 522 | 789 | – | – |
| | valid | 20 | 27 | – | – |
| | test | 30 | 51 | – | – |
| | test-300 | 300 | 364 | – | – |

### jeu de données read-bad

Le jeu de données READ-BAD (Grüning et al., 2017), présenté en section 3.1, contient 2 035 images de documents manuscrits utilisées lors des compétitions READ-BAD pour la détection des lignes de base. Le jeu de données a été annoté aux niveaux simple et double pages [1]. Dans nos expériences, nous prédisons au niveau simple page, ce qui conduit à détecter deux objets sur les images qui présentent un document en double-page. De plus, les images ayant été annotées comme "anormales" dans la base ont été supprimées car leurs annotations n'étaient pas assez précises. Dans ce qui suit, cette version du jeu de données est appelée READ-BAD* et comprend 1 630 images d'entraînement, 200 images de validation et 199 images de test avec respectivement 1 801, 221 et 219 pages simples.

### jeu de données horae

Le jeu de données Horae (Boillet et al., 2019) est semblable à celui présenté en section 3.1. Afin d'avoir des résultats plus significatifs, nous avons étendu l'ensemble de test original qui ne contenait que 30 images en annotant 300 images supplémentaires choisies au hasard parmi les 1 158 livres d'heures, ce qui représente 364 pages simples. Cet ensemble de test est dénommé Horae-test-300 dans la suite.

Le corpus complet Horae est composé de 1 158 livres d'heures présentant une grande diversité d'images de documents non annotés en termes de types de numérisations, de fonds et de formes. Ce corpus est utilisé pour comparer les différents estimateurs lorsqu'ils sont utilisés dans un cadre d'apprentissage actif.

1. https://github.com/ctensmeyer/pagenet

### 6.2.2 ENTRAÎNEMENT DES SYSTÈMES DE DÉTECTION

Pour nos expériences, nous avons utilisé le système Doc-UFCN comme détecteur d'objets, car, comme détaillé dans les deux chapitres précédents 4 et 5, il a montré de bonnes performances pour la détection d'objets sur des documents historiques tout en ayant un temps d'inférence réduit par rapport aux autres systèmes.

Pour les deux tâches, les modèles Doc-UFCN pré-entraînés (désignés par "référence" dans la suite) sont entraînés avec les images redimensionnées de telle sorte que leur plus grande dimension soit égale à 768 pixels, en conservant leur rapport d'aspect. Un prétraitement est appliqué aux labels d'entraînement afin d'éviter que les zones annotées ne se touchent lors du redimensionnement des images (prétraitement détaillé dans la section 5.1.1). Les modèles sont entraînés pendant 150 époques avec un taux d'apprentissage de $5e-3$ et l'optimiseur Adam. La configuration (poids) qui minimise la fonction de perte sur l'ensemble de validation est conservée à l'issue de l'apprentissage.

#### MÉTRIQUES D'ÉVALUATION

Outre les métriques de détection standards, les modèles de lignes de texte sont également évalués à l'aide de métriques orientées vers la tâche finale, notamment le CER et le WER au niveau page. À cette fin, un reconnaisseur de texte manuscrit (HTR) basé sur Kaldi (Arora et al., 2019) a été entraîné sur les lignes transcrites Hugin-Munin. Nous avons choisi cet HTR parce qu'il s'agit d'un outil prêt à l'emploi qui fonctionne généralement assez bien dans la plupart des cas d'utilisation et qui a obtenu des performances compétitives sur les documents Hugin-Munin (Maarand et al., 2022).

Le modèle de reconnaissance entraîné est appliqué à toutes les lignes prédites par Doc-UFCN, ordonnées par leur centroïde du coin supérieur gauche de la page au coin inférieur droit. Les textes prédits sont concaténés dans ce même ordre pour fournir une transcription unique au niveau de la page. Les transcriptions manuelles sont ordonnées de la même manière et les CER et WER au niveau de la page sont calculés. Le modèle de détection de référence obtient environ 24 % de CER sur les images Hugin-Munin. En outre, nous calculons la WordCountFMeasure (WCFM) (Pletschacher et al., 2015) qui évalue les modèles HTR sur la base du nombre de mots correctement prédits, indépendamment de leur position. Nous avons utilisé le *PRIMA Text Evaluation Toolkit*[2] pour calculer les scores WCFM. Kaldi obtient un WCFM de 59 % par rapport aux transcriptions manuelles. Ces valeurs de CER relativement élevées et de WCFM faibles indiquent que les lignes détectées par le modèle de référence ne sont pas de très bonne qualité pour le modèle de reconnaissance. Elles peuvent refléter des lignes détectées mal placées (pas de texte), des lignes trop fines (texte coupé) ou des lignes manquées.

**Table 6.2** – Résultats de détection de pages obtenus par le modèle de référence entraîné sur le jeu de données READ-BAD* et évalué sur les jeux de données READ-BAD* et Horae-test-300.

| Jeu de données | | IoU | F1-score | mAP |
|---|---|---|---|---|
| READ-BAD* | train | 0,97 | 0,98 | 0,92 |
| | valid | 0,97 | 0,98 | 0,91 |
| | test | 0,97 | 0,98 | 0,94 |
| Horae | test-300 | 0,90 | 0,94 | 0,60 |

**Table 6.3** – Résultats de détection de lignes de texte obtenus par le modèle de référence entraîné sur 19 jeux de données et évalué l'ensemble de test du jeu de données Hugin-Munin.

| Jeu de données | | IoU | F1-score | mAP | CER (%) | WCFM |
|---|---|---|---|---|---|---|
| Hugin-Munin | test | 0,48 | 0,63 | 0,21 | 24,37 | 0,59 |

#### Résultats des systèmes de détection

Pour la tâche de détection de pages, le modèle de référence est entraîné sur des images READ-BAD* dont les résultats sont présentés dans la Table 6.2. Il obtient une IoU de 97 % et une mAP de 94 % sur READ-BAD*. Cependant, la mAP sur les images de l'ensemble Horae-test-300 est d'environ 60 %, ce qui laisse une marge d'amélioration importante. Dans ce qui suit, les images ayant les plus faibles scores de confiance estimés dans le corpus Horae sont annotées afin d'améliorer la détection sur Horae-test-300.

Pour la détection des lignes de texte, nous avons entraîné un modèle générique de détection des lignes de texte, différent de celui présenté dans le chapitre 5, ainsi que les estimateurs de confiance sur de nombreux jeux de données. À cet égard, nous avons rassemblé 19 bases de données principalement publiques, comprenant des documents historiques et modernes. Au total, ce jeu de données contient 9 432 images d'entraînement, 1 907 images de validation et 6 669 images de test, ce qui correspond à 374 316 lignes annotées d'entraînement, 85 208 lignes de validation et 190 502 lignes de test. Ce modèle générique appliqué à l'ensemble de test Hugin-Munin a été évalué à 48 % d'IoU et 21 % de mAP (Table 6.3). Ces résultats assez faibles étaient attendus puisque les documents sont beaucoup plus complexes que ceux utilisés lors du pré-entraînement.

### 6.2.3 Entraînement des estimateurs de confiance

Aucun apprentissage supplémentaire n'est requis pour les estimateurs basés sur le *dropout* de Monte Carlo, puisque seuls les modèles de détection d'objets sont utilisés pour estimer la confiance. En revanche, les régresseurs doivent être entraînés sur les statistiques d'objets décrites en section 6.1.3. Tout d'abord, le modèle de détection d'objets est appliqué à toutes les images (READ-BAD* pour la détection de pages et les 19 jeux de données pour la détection

2. https://www.primaresearch.org/tools/PerformanceEvaluation

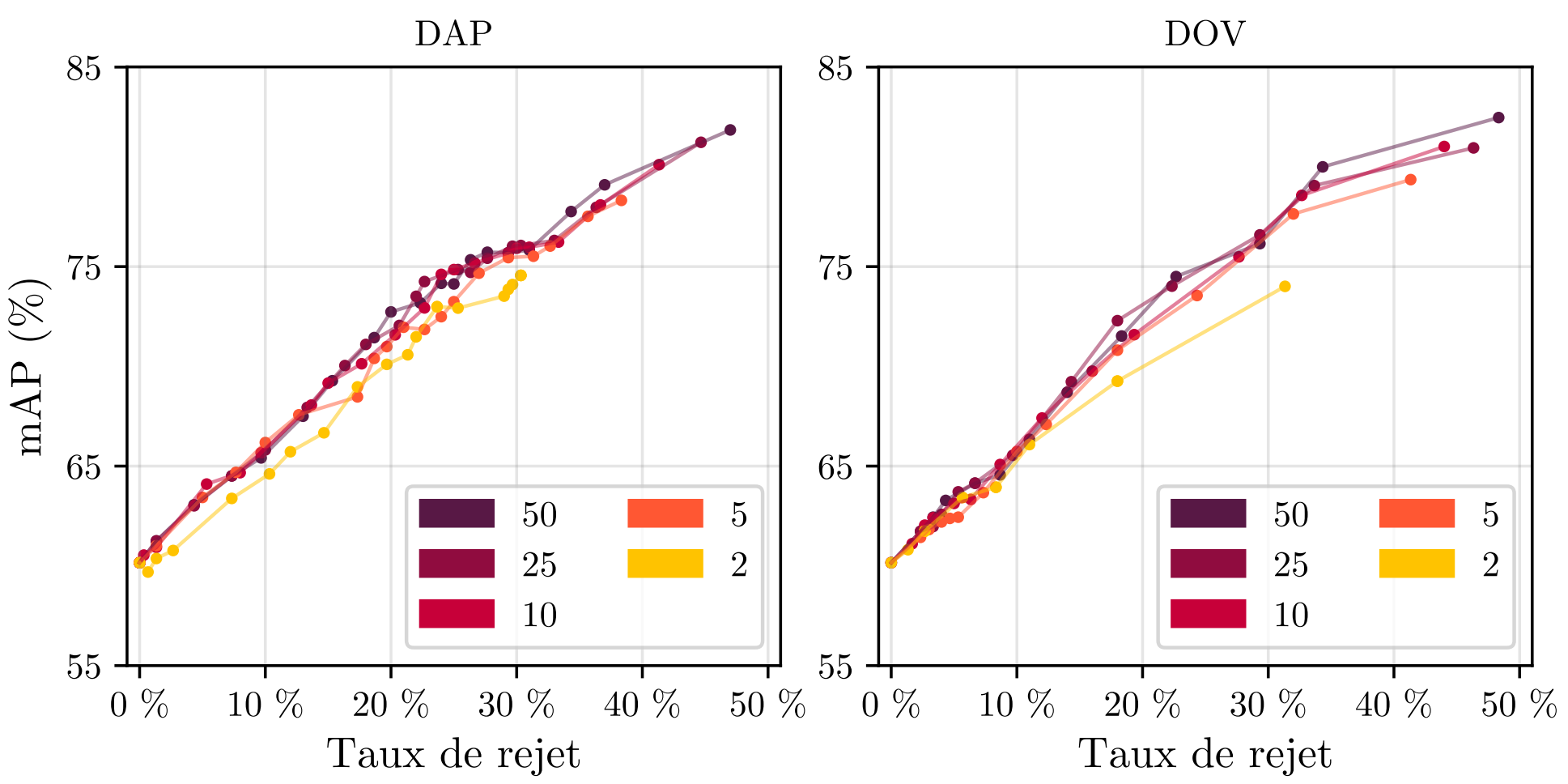

**Figure 6.2** – Courbes de rejet présentant l'évolution des performances du modèle de détection de pages de référence sur l'ensemble de test Horae-test-300. Courbes présentées pour les estimateurs DAP, à gauche, et DOV, à droite, en fonction du nombre de prédictions avec *dropout* $N$.

de lignes de texte), ce qui permet de calculer les statistiques. Comme les jeux de données sont annotés, le modèle de détection est ensuite évalué sur chaque image séparément, ce qui fournit un IoU et une mAP pour chaque image. Ces valeurs de mAP sont utilisées comme cible pour l'entraînement des régresseurs.

Pour entraîner les modèles de régression, nous avons utilisé le *RandomForestRegressor* de *scikit-learn* avec les paramètres par défaut. Les modèles de régression présentent de faibles erreurs quadratiques moyennes (MSE) sur les ensembles de données d'entraînement (0,0164 MSE sur l'ensemble de test de READ-BAD*).

## 6.3 RÉSULTATS ET DISCUSSION

Dans cette section, nous évaluons et comparons les estimateurs de confiance à l'aide de courbes de rejet, puis nous comparons leurs performances lorsqu'ils sont intégrés dans un cadre d'apprentissage actif.

### 6.3.1 NOMBRE DE PRÉDICTIONS AVEC DROPOUT

Pour les expérimentations avec le *dropout* de Monte Carlo (DAP et DOV), nous devons définir le nombre de prédictions $N$ à calculer pour estimer la qualité des prédictions. La Figure 6.2 montre la mAP en fonction du taux de rejet pour les estimateurs DAP et DOV calculés pour différentes valeurs de $N$ (2, 5, 10, 25 et 50). Nous avons choisi ces valeurs car nous recherchons un ordre de grandeur de $N$ plutôt qu'une valeur précise. L'idée est de savoir si nous avons besoin d'un nombre important de prédictions pour obtenir une variance suffisamment fiable, ou si quelques prédictions suffisent. De plus, nous ne sommes pas allés au-delà de 50 prédictions car nous voulons garder un temps de calcul raisonnable.

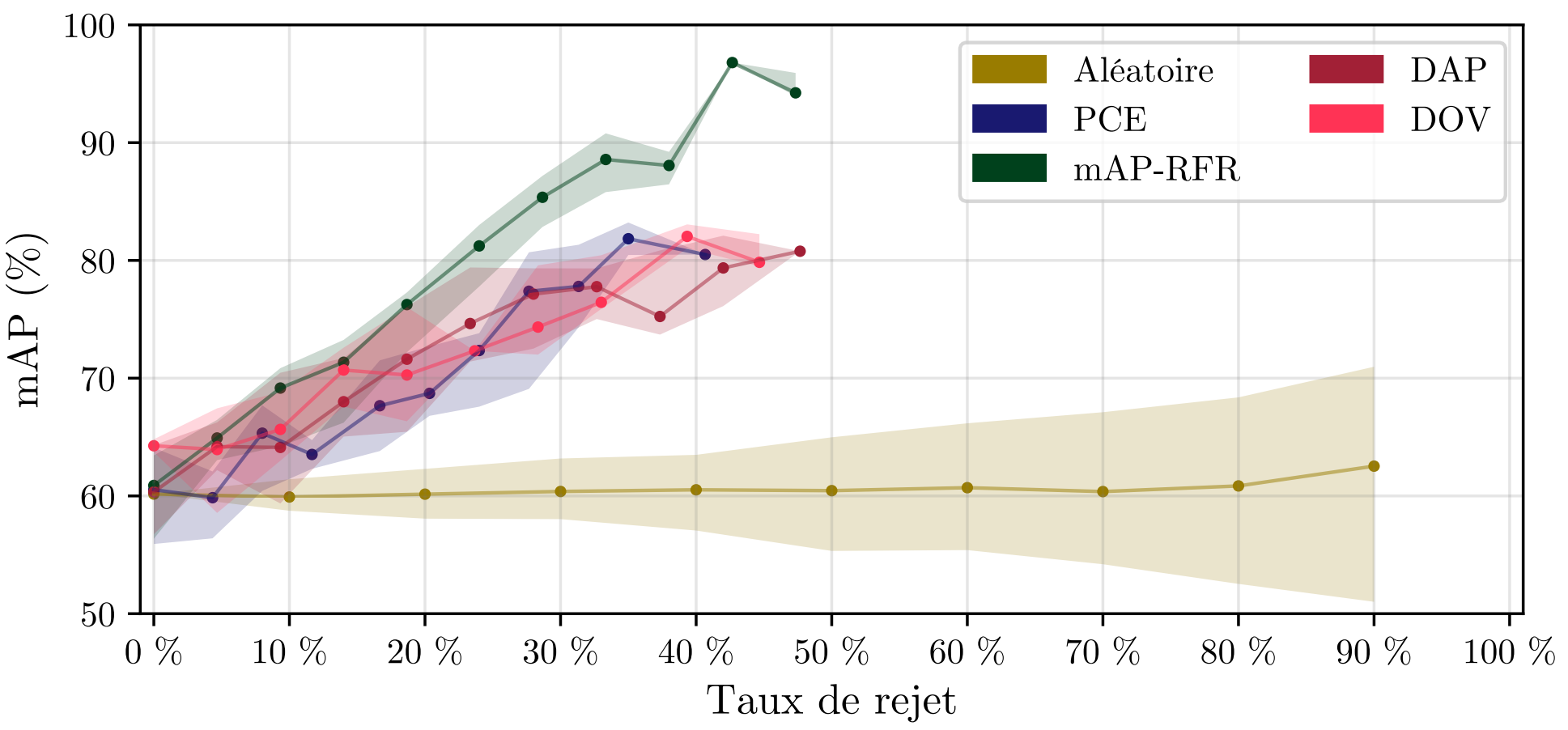

**Figure 6.3 –** Courbes de rejet présentant l'évolution du score mAP en fonction du taux de rejet. Les courbes présentent les résultats du modèle de détection de pages de référence sur l'ensemble de test Horae-test-300.

Les résultats sont présentés sur Horae-test-300 pour la détection de pages. Les courbes de rejet sont construites en ordonnant les images en fonction de leur confiance estimée, les exemples ayant une valeur DAP inférieure (ou une valeur DOV supérieure) à un seuil prédéfini sont retirés de l'ensemble d'évaluation et la mAP est calculée sur les exemples restants. Pour DAP, le seuil varie de 0 à 1 avec un pas de 0,05. Pour DOV, les valeurs ne sont pas bornées, le seuil varie donc de 10 à 0 avec un pas de -1.

Ces graphiques montrent que l'utilisation de $N$=10 prédictions pour l'estimation avec *dropout* est suffisante et qu'aucune amélioration n'est observée avec $N$=25 ou $N$=50. De plus, le coût de calcul est réduit avec seulement 10 prédictions. Sur la base de cette observation, nous avons utilisé $N$=10 prédictions avec *dropout* pour estimer les scores de confiance dans le reste des expériences.

### 6.3.2 PERFORMANCES DES ESTIMATEURS EN REJET

Dans une première expérience, nous évaluons la capacité des estimateurs de confiance à détecter les exemples mal prédits. Pour ce faire, nous évaluons les performances du modèle de détection lorsque les images ayant le score de confiance estimé le plus faible sont retirées de l'ensemble d'évaluation. Cette évaluation est réalisée grâce à des courbes de rejet. Sur les courbes de rejet, chaque point correspond à un seuil pour lequel les images dont le score estimé est inférieur à ce seuil sont retirées de l'évaluation. Les courbes n'atteignent pas 100 % car, au-dessus d'un seuil donné, il reste uniquement des images ayant le même score, de sorte qu'elles ne peuvent plus être retirées sans que l'ensemble d'évaluation soit vide. Par souci de clarté, nous montrons seulement l'évolution de la mAP, les résultats d'IoU suivant la même tendance.

La Figure 6.3 montre l'évolution des performances du modèle de référence sur Horae-test-300 pour la tâche de détection de pages par rapport au taux de rejet pour différents estimateurs de confiance. Nous montrons les courbes médianes ainsi que les intervalles de confiance ($10^{e}$et $90^{e}$ percentiles) obtenus en calculant 100 courbes de rejet générées par 100 ré-échantillonnages avec remplacement à partir de l'ensemble de test original. La courbe aléatoire montre les résultats obtenus pour 100 échantillonnages aléatoires.

Notre objectif est d'avoir un modèle avec une mAP élevée et un faible taux de rejet. Nous pouvons constater que les estimateurs basés sur le *dropout* ne sont pas compétitifs par rapport au régresseur basé sur les statistiques. De plus, comme mAP-RFR ne nécessite qu'une seule prédiction en inférence, ce premier résultat montre que l'utilisation de mAP-RFR au lieu du *dropout* de Monte Carlo est plus intéressante. Les résultats de PCE étant comparables à ceux de DAP et DOV, il semble que les estimateurs *dropout* de Monte Carlo ne fassent pas de meilleurs indicateurs que les probabilités *a posteriori* des détecteurs. Cela peut s'expliquer par le fait qu'aucune information supplémentaire à part les prédictions du réseau neuronal ne soit fournie à ces trois estimateurs. Cette première expérience montre que notre proposition mAP-RFR a une grande capacité à estimer la confiance des pages prédites. Il surpasse DAP et DOV qui sont eux-mêmes à peine meilleurs que PCE.

Sur la Figure 6.1, nous montrons deux prédictions obtenues par le modèle de référence pour la tâche de détection de pages. À gauche, nous montrons une bonne prédiction où la variance est faible, sauf sur les bords des objets. Les estimations de confiance DOV=0,0, DAP=1,0 et mAP-RFR=1,0 reflètent bien la bonne qualité de la détection de l'image de gauche tandis que les estimations de confiance DOV=17,36, DAP=0,0993 et mAP-RFR=0,5553 de l'image de droite reflètent également la très mauvaise qualité de la détection, qui contient un nombre élevé de petits objets prédits autour du principal.

### 6.3.3 APPRENTISSAGE ACTIF

Dans un cadre d'apprentissage actif, l'objectif est d'entraîner un bon détecteur d'objets tout en minimisant la quantité d'exemples à annoter manuellement. Pour y parvenir, il est crucial de bien choisir les données à annoter.

Dans nos expériences, nous suivons une configuration standard d'apprentissage actif (Cohn et al., 1996). Tout d'abord, un modèle Doc-UFCN de référence est entraîné, puis appliqué à des documents non vus et non annotés provenant d'un nouveau jeu de données. Ensuite, ces images sont classées en fonction de leur confiance estimée, celles dont la confiance est la plus faible sont sélectionnées pour une annotation manuelle et utilisées pour entraîner un nouveau modèle. Bien que de nombreuses stratégies de sélection des données à annoter aient été proposées pour améliorer au mieux les modèles (Settles et al., 2008), nous nous concentrons, dans cette section, sur la sélection des images ayant la plus faible confiance. Ainsi, les images ayant une confiance inférieure à un seuil prédéfini sont sélectionnées. Le seuil varie d'une itération à l'autre en fonction de la distribution des confiances estimées. Nous présentons une analyse de deux stratégies de sélection dans la section 6.4.

**Table 6.4 –** Résultats des modèles de détection de pages sur l'ensemble de test Horae-test-300 après apprentissage actif. La colonne ITÉRATION indique le nombre d'itérations réalisées afin d'obtenir le meilleur modèle. Le nombre d'images annotées est indiqué dans la colonne IMAGES.

| ESTIMATEUR | ITÉRATION | IMAGES | IoU | mAP |
|---|---|---|---|---|
| Référence | – | 0 | 0,90 | 0,60 |
| Aléatoire | 5 | 300 | 0,93 | 0,86 |
| PCE | 9 | 90 | 0,93 | 0,86 |
| mAP-RFR | 8 | 107 | 0,94 | 0,89 |
| DAP | 9 | 129 | 0,94 | 0,91 |
| DOV | 9 | 168 | 0,95 | 0,92 |

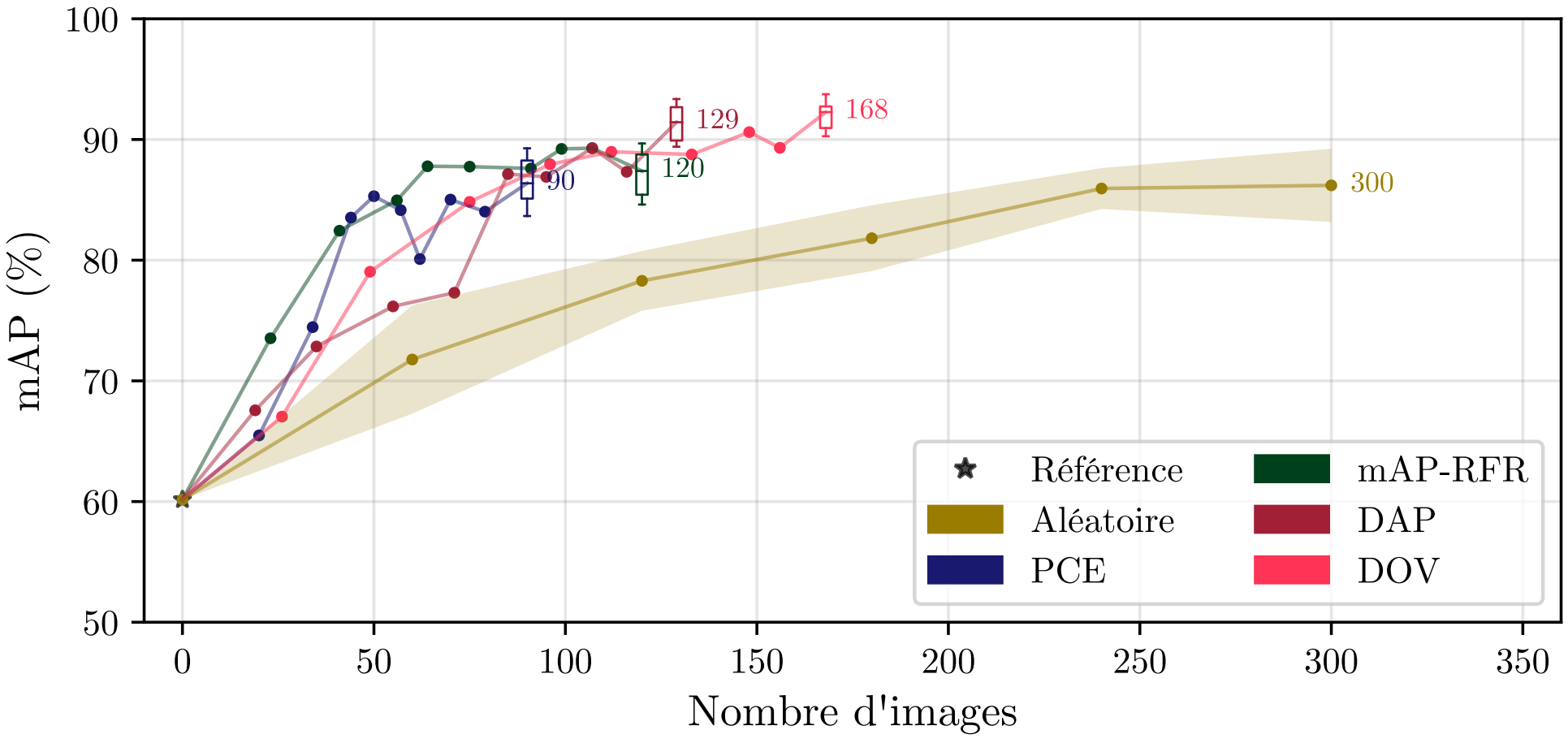

**Figure 6.4 –** Évolution des performances de détection de pages (mAP) sur l'ensemble de test Horae-test-300 pendant les itérations d'apprentissage actif.

Chaque modèle de détection est entraîné dans la même configuration que les modèles de référence décrits dans la section 6.2.2. Pendant les itérations d'apprentissage actif, plusieurs stratégies d'initialisation des poids des modèles peuvent être envisagées. Ils peuvent être initialisés avec les poids des derniers modèles entraînés, ceux du modèle de référence (à chaque itération) ou encore ceux du meilleur modèle entraîné durant les itérations précédentes. Pour nos expériences, nous initialisons les poids avec ceux des derniers modèles entraînés.

Enfin, pour les expériences suivantes, nous avons calculé un intervalle de confiance sur les modèles de la dernière itération. Pour cela, nous avons utilisé le *bootstrapping* empirique (WASSERMAN, 2004) avec 100 ré-échantillonnages avec remplacement. En outre, les expériences avec la sélection aléatoire sont répétées cinq fois et les valeurs moyennes et les écarts types sont présentés.

### DÉTECTION DE PAGES

La Figure 6.4 et la Table 6.4 présentent les résultats obtenus pour la tâche de détection de pages. À chaque itération, le modèle courant est appliqué aux images du corpus Horae. Les images dont le score de confiance estimé est inférieur à un seuil sont annotées manuellement et ajoutées à l'ensemble d'entraînement de l'itération précédente afin d'entraîner un nouveau modèle. Comme pour les courbes de rejet, ces graphiques montrent que les estimateurs sont capables de détecter les mauvaises prédictions afin d'entraîner des modèles plus performants, avec seulement une petite quantité de données annotées. En effet, les estimateurs sont meilleurs qu'une sélection aléatoire puisqu'avec deux fois moins de données, les modèles présentent des augmentations relatives de 6 % de mAP (+5 points de pourcentage) pour DAP, 7 % (+6 points de pourcentage) pour DOV et presque 3,5 % (+3 points de pourcentage) pour mAP-RFR. Sur la Figure 6.4, nous observons également que la courbe correspondant à mAP-RFR est presque toujours supérieure à celles des autres estimateurs, ce qui indique des modèles plus performants avec moins de données annotées.

Ces résultats montrent que l'estimateur mAP-RFR est plus performant que les estimateurs basés sur le *dropout* de Monte Carlo puisqu'il présente une mAP plus élevée tout en ne nécessitant qu'une seule prédiction pendant l'inférence et moins de données annotées. Une explication possible à ces résultats, que nous avons déjà formulée précédemment, est que les estimateurs DAP et DOV sont non supervisés : ils n'ont aucune connaissance préalable de ce qu'est une prédiction correcte. Au contraire, mAP-RFR est entraîné avec les mAPs réelles calculées sur les données annotées.

### DÉTECTION DE LIGNES DE TEXTE

La Figure 6.5 et la Table 6.5 montrent les résultats obtenus avec les estimateurs mAP-RFR, DAP et PCE pour l'apprentissage actif. Nous ne montrons pas les résultats de DOV car ils sont équivalents à ceux de DAP. De plus, le WER n'est pas rapporté ici puisqu'il est fortement corrélé au CER.

D'après la Figure 6.5, il apparaît que la sélection aléatoire donne de bons résultats avec seulement 50 images. Cependant, ces résultats dépendent fortement des données choisies, ce qui conduit à des performances très variables d'une sélection à l'autre. Par conséquent, au vu de cette grande variabilité des résultats, nous pensons qu'il est préférable de se concentrer sur un estimateur plus robuste et moins aléatoire qui peut obtenir des résultats tout aussi satisfaisants.

D'après la Table 6.5, l'estimateur DAP se distingue des autres en obtenant des valeurs d'IoU et de mAP plus faibles que les autres estimateurs mais un CER bien moins élevé. mAP-RFR ne semble pas ici faire un meilleur estimateur que PCE ou que la sélection aléatoire. Malgré des résultats bien moins bons en termes d'IoU et de mAP, DAP présente de meilleures valeurs de CER et de WCFM par rapport à mAP-RFR. En effet, mAP-RFR a été conçu pour estimer la mAP de chaque prédiction et ainsi maximiser la mAP des modèles. Cependant, nous avons montré, en section 5.4.1, que la maximisation de la mAP ne signifie pas nécessairement l'amélioration de l'entrée pour le reconnaisseur.

Pour cette tâche, il serait intéressant de sélectionner les images en fonction d'un score de confiance lié à la reconnaissance de texte. Le modèle de détection s'adapterait pour améliorer directement la reconnaissance du texte.

**Table 6.5** – Résultats des modèles de détection de lignes de texte sur l'ensemble de test du jeu de données Hugin-Munin après apprentissage actif. La colonne Itération indique le nombre d'itérations réalisées afin d'obtenir le meilleur modèle. Le nombre d'images annotées est indiqué dans la colonne Images.

| Estimateur | Itération | Images | IoU | mAP | CER (%) | WCFM |
|---|---|---|---|---|---|---|
| Référence | – | 0 | 0,48 | 0,21 | 24,37 | 0,59 |
| Aléatoire | 1 | 50 | 0,63 | 0,45 | 22,18 | 0,64 |
| PCE | 6 | 83 | 0,67 | 0,46 | 22,79 | 0,66 |
| mAP-RFR | 9 | 139 | 0,64 | 0,44 | 22,50 | 0,66 |
| DAP | 6 | 110 | 0,63 | 0,40 | 20,23 | 0,68 |

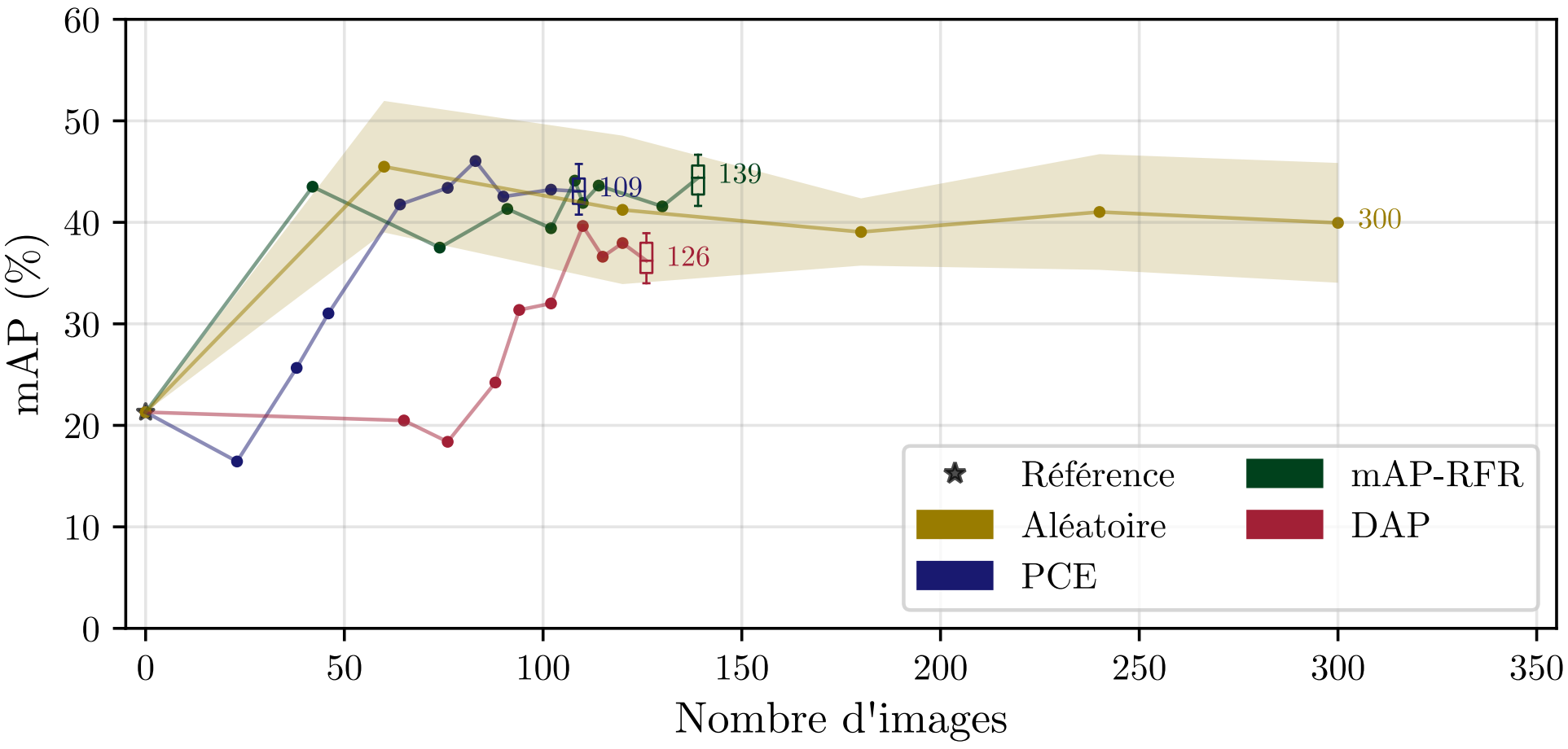

**Figure 6.5** – Évolution des performances de détection de lignes de texte (mAP) sur l'ensemble de test du jeu de données Hugin-Munin pendant les itérations d'apprentissage actif.

## 6.4 STRATÉGIE D'ENTRAÎNEMENT : SÉLECTION ET ANNOTATION DES DONNÉES

Dans le cadre de l'apprentissage actif, de nombreuses stratégies d'entraînement peuvent être exploitées. Bien que de nombreuses méthodes aient été proposées dans la littérature, aucune ne semble réellement surpasser les autres. C'est pourquoi, dans cette section, nous étudions deux stratégies d'entraînement qui concernent la sélection des données ainsi que leur annotation.

La première est la même que celle utilisée dans les expériences précédentes : les exemples avec les confiances estimées les plus faibles sont annotés manuellement puis ajoutés à l'ensemble d'entraînement. La seconde stratégie sélectionne les exemples avec les confiances les plus élevées et utilise les prédictions du modèle de détection comme labels pour les entraînements suivants. Cette stratégie de sélection permet de réduire le coût d'annotation manuelle au minimum puisqu'aucune donnée n'est annotée manuellement. Nous présentons tout d'abord les résultats pour la détection de pages, en section 6.4.1, puis pour la détection de lignes de texte, en section 6.4.2. Pour l'ensemble des résultats présentés dans ce qui suit, les modèles sont entraînés dans les mêmes conditions que dans les expériences précédentes.

**Table 6.6 –** Résultats des modèles de détection de pages sur l'ensemble de test Horae-test-300 après apprentissage actif et pour différentes stratégies de sélection de données. La colonne ITÉRATION indique le nombre d'itérations réalisées afin d'obtenir le meilleur modèle. La sélection "Faible" correspond à la sélection des images avec les confiances les plus faibles. La sélection "Élevée" correspond à la sélection des images avec les confiances les plus élevées où leurs prédictions sont directement utilisées comme labels d'entraînement. Les colonnes "Manuelle" et "Auto." indiquent respectivement les nombres d'images d'entraînement avec annotations manuelles et automatiques permettant d'obtenir le meilleur modèle.

| ESTIMATEUR | SÉLECTION | ITÉRATION | IMAGES | | IOU | MAP |
|---|---|---|---|---|---|---|
| | | | Manuelle | Auto. | | |
| Référence | – | – | – | – | 0,90 | 0,60 |
| mAP-RFR | Faible | 8 | 107 | – | 0,94 | 0,89 |
| | Élevée | 9 | – | 444 | 0,90 | 0,84 |
| DAP | Faible | 9 | 129 | – | 0,94 | 0,91 |
| | Élevée | 8 | – | 475 | 0,90 | 0,72 |
| DOV | Faible | 9 | 168 | – | 0,95 | 0,92 |
| | Élevée | 3 | – | 163 | 0,90 | 0,64 |

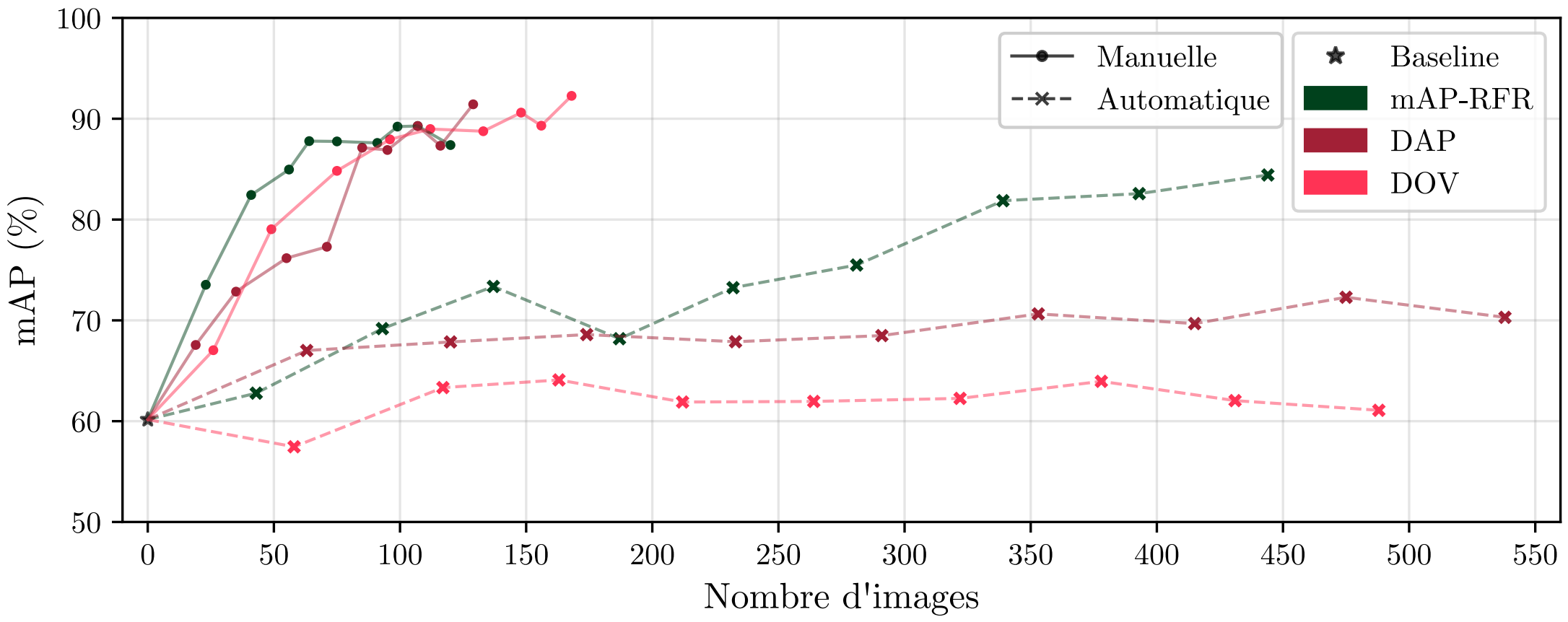

**Figure 6.6 –** Évolution des performances de détection de pages (mAP) sur l'ensemble de test Horae-test-300 pendant les itérations d'apprentissage actif pour différentes stratégies de sélection de données. Les courbes "Manuelle" correspondent à la sélection des exemples avec les confiances les plus faibles et une annotation manuelle de ces exemples. Les courbes "Automatique" correspondent à la sélection des exemples avec les confiances les plus élevées et l'utilisation des prédictions comme labels d'entraînement.

### 6.4.1 DÉTECTION DE PAGES

Les Figure 6.6 et Table 6.6 présentent les résultats obtenus pour les deux stratégies de sélection pour la tâche de détection de pages. Les courbes et valeurs correspondant à la sélection basée sur les faibles confiances sont les mêmes que celles présentées en section 6.3.3. D'après la Figure 6.6, l'utilisation des prédictions comme labels d'entraînement pour l'estimateur DOV ne permet pas réellement d'améliorer le modèle de détection par rapport au modèle de référence. Au contraire, pour les estimateurs DAP et mAP-RFR, l'utilisation des prédictions permet une importante amélioration des performances par rapport au modèle de référence. En effet, le modèle mAP-RFR permet une amélioration de 40 % de mAP (+24 points de pourcentage) et le modèle DAP de 20 % (+12 points de pourcentage) par rapport au modèle de référence, sans aucune donnée annotée manuellement.

**Table 6.7 –** Résultats des modèles de détection de lignes de texte sur l'ensemble de test du jeu de données Hugin-Munin après apprentissage actif et pour différentes stratégies de sélection de données. La colonne ITERATION indique le nombre d'itérations réalisées afin d'obtenir le meilleur modèle. La sélection "Faible" correspond à la sélection des images avec les confiances les plus faibles. La sélection "Élevée" correspond à la sélection des images avec les confiances les plus élevées où leurs prédictions sont directement utilisées comme labels d'entraînement. Les colonnes "Manuelle" et "Auto." indiquent respectivement les nombres d'images d'entraînement avec annotations manuelles et automatiques permettant d'obtenir le meilleur modèle.

| ESTIMATEUR | SÉLECTION | ITERATION | IMAGES | | IOU | MAP | CER (%) | WCFM |
|---|---|---|---|---|---|---|---|---|
| | | | Manuelle | Auto. | | | | |
| Référence | – | – | – | – | 0,48 | 0,21 | 24,37 | 0,59 |
| mAP-RFR | Faible | 9 | 139 | – | 0,64 | 0,44 | 22,50 | 0,66 |
| | Élevée | 4 | – | 54 | 0,53 | 0,28 | 22,89 | 0,62 |
| DAP | Faible | 6 | 110 | – | 0,63 | 0,40 | 20,23 | 0,68 |
| | Élevée | 1 | – | 38 | 0,51 | 0,26 | 21,98 | 0,63 |

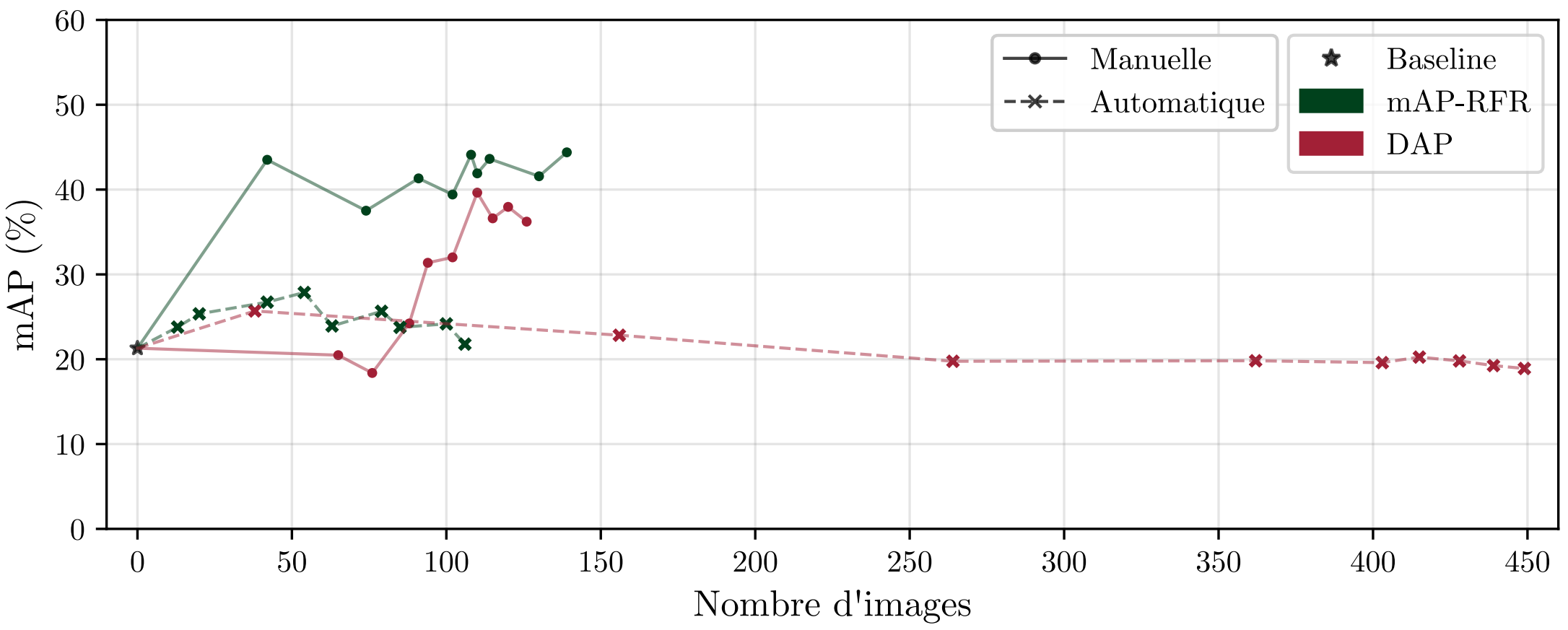

**Figure 6.7 –** Évolution des performances de détection de lignes de texte (mAP) sur l'ensemble de test du jeu de données Hugin-Munin pendant les itérations d'apprentissage actif pour différentes stratégies de sélection de données. Les courbes "Manuelle" correspondent à la sélection des exemples avec les confiances les plus faibles et une annotation manuelle de ces exemples. Les courbes "Automatique" correspondent à la sélection des exemples avec les confiances les plus élevées et l'utilisation des prédictions comme labels d'entraînement.

Le modèle obtenu avec l'estimateur DAP et les données annotées automatiquement est tout de même bien moins performant que celui obtenu avec les données annotées manuellement (-21 % de mAP). Pour l'estimateur mAP-RFR, l'écart est moins important puisque les performances sont dégradées de seulement 5,5 % de mAP en passant des données annotées manuellement aux labels automatiques. Ainsi, nous constatons que l'estimateur mAP-RFR est le plus robuste pour cette tâche puisqu'il permet de détecter de manière fiable les bonnes ainsi que les mauvaises prédictions. Cet estimateur permet d'obtenir l'amélioration de performance la plus intéressante sans annotation manuelle.

### 6.4.2 DÉTECTION DE LIGNES DE TEXTE

Les Figure 6.7 et Table 6.7 présentent les résultats obtenus pour les deux stratégies de sélection pour la tâche de détection de lignes de texte. Les courbes et valeurs correspondant à la sélection basée sur les faibles confiances sont les mêmes que celles présentées en section 6.3.3. D'après la Table 6.7, l'utilisation des prédictions comme labels d'entraînement mène à une légère amélioration des performances par rapport au modèle de référence (-6 % de CER pour mAP-RFR et -11 % de CER pour DAP). Ceci permet de valider l'utilité des estimateurs mAP-RFR et DAP dans la sélection des mauvais autant que des bons exemples.

Comme pour la détection de pages, les modèles obtenus avec les données annotées automatiquement sont moins performants que ceux obtenus avec les données annotées manuellement (+1,7 % de CER pour mAP-RFR et +8,7 % de CER pour DAP).

## 6.5 CONCLUSION

Dans ce chapitre, nous avons comparé quatre estimateurs de confiance pour les modèles de détection d'objets. Nous avons montré que, dans un contexte d'apprentissage actif, ces estimateurs peuvent être utilisés pour entraîner des modèles atteignant des performances élevées pour la détection d'objets en termes d'IoU et de mAP tout en ne nécessitant qu'un faible effort d'annotation manuelle. Lorsque les métriques optimisées sont étroitement liées à l'objectif, comme pour la mAP et la détection de pages, nous avons montré que l'estimateur mAP-RFR permet d'obtenir de meilleures performances de détection que celles basées sur le *dropout* de Monte Carlo, tout en ayant un coût de calcul réduit. Cependant, cet estimateur est supervisé et doit être entraîné, ce qui n'est pas le cas pour DAP, DOV et PCE. Dans le cas d'une adaptation à de nouvelles données, il est donc avantageux, dans un premier temps, d'utiliser l'estimateur DAP basé sur le *dropout*. Si les résultats n'atteignent pas les performances attendues, il semble alors plus intéressant d'utiliser un estimateur entraîné tel que mAP-RFR. D'autre part, lorsque les métriques sont moins étroitement liées à l'objectif, comme pour la détection des lignes de texte, les méthodes basées sur le *dropout* sont plus compétitives.

À l'avenir, nous envisageons d'adapter l'estimateur mAP-RFR afin qu'il estime la confiance au niveau de l'objet directement de façon à ne plus rejeter les images mais les objets. Cela permettrait de savoir exactement quels objets posent un problème et de les corriger. De plus, il serait intéressant de créer automatiquement des vecteurs de description d'objets à travers des représentations apprises. Enfin, nous avons montré que l'utilisation de métriques orientées vers la tâche finale permet d'évaluer l'impact des modèles de détection sur les résultats finaux. Il semblerait donc intéressant de sélectionner les images ou les objets en se basant sur les résultats de reconnaissance de texte. Dans cette optique, nous prévoyons de mettre en place un nouvel estimateur qui reflète les résultats de la reconnaissance de texte.

# 7

# DÉTECTION SÉQUENTIELLE D'OBJETS DANS DES IMAGES DE DOCUMENTS

Les systèmes à base de Transformers proposés récemment, et détaillés dans le Focus 2.11, obtiennent désormais les meilleures performances de l'état de l'art tant sur des tâches de traitement de la langue que des tâches de classification d'images. Un de leurs avantages réside dans leur capacité à modéliser et à générer des séquences et même des objets structurés. Il semble désormais possible de prédire automatiquement la structure complète d'une image de document, avec l'ensemble de ses éléments organisés de manière hiérarchique. De plus, ces systèmes sont capables de prédire séquentiellement les coordonnées des objets à détecter (Chen et al., 2022), sans avoir à passer par une prédiction pixel à pixel. Bien qu'il n'y ait pas, à notre connaissance, de travaux proposés dans la littérature afin de réaliser une telle tâche, une prédiction directe de coordonnées présente de nombreux avantages comparée à une prédiction standard niveau pixel. C'est pourquoi, dans ce chapitre, nous avons choisi d'explorer les modèles Transformers pour construire un nouveau modèle de détection séquentielle d'objets dans les images de documents.

Un premier point ayant motivé nos travaux dans ce sens est lié à la capacité d'un tel système à passer outre les problèmes liés aux boîtes englobantes qui se touchent et se superposent. En effet, le modèle n'est pas appris avec des images de labels mais directement avec les coordonnées des éléments à détecter, telles que les boîtes englobantes ou les lignes de base. Similairement aux approches par régression de boîtes englobantes, il devient possible de détecter plusieurs objets d'une même classe au même endroit sur l'image. Un autre avantage de cette approche tient au fait qu'elle permet d'apprendre un ordre de lecture implicitement représenté par la séquentialité du processus de détection des éléments. Le modèle est, en effet, entraîné à détecter les éléments dans l'ordre imposé par la séquence des objets représentés dans la vérité terrain. Cette séquentialité de la vérité terrain définit donc un ordre de détection, et donc de lecture des objets présents dans l'image. Resitué dans le contexte de la reconnaissance de documents, il devient possible d'apprendre à prédire les lignes de texte dans l'ordre de lecture du texte. Enfin, comme énoncé plus tôt, cette approche permet d'avoir une sortie structurée des résultats. Si nous imaginons un problème à deux classes telles que les paragraphes et les lignes de texte, il est possible d'apprendre un modèle qui détecte le début d'un paragraphe, toutes les lignes qu'il contient puis la fin de ce paragraphe avant de passer au suivant. Nous pouvons donc obtenir directement une détection hiérarchique des éléments sur une image.

La plupart des systèmes proposés traitant la détection d'objets prédisent un masque de probabilités à la résolution de l'image d'entrée. Bien que cette tâche de détection puisse être traitée à l'aide de Vision Transformers (voir le Focus 2.14), qui utilisent un encodeur Transformer suivi d'un décodeur CNN, elles ne bénéficient pas de l'avantage principal des Transformers qui réside en leur capacité à prédire des éléments de manière séquentielle, capacité induite par le décodeur. De même, dans le domaine de la vision, la plupart des travaux proposés dans la littérature ont exploré les architectures Transformers pour constituer de nouveaux extracteurs de caractéristiques, des encodeurs Transformers, et ainsi tenter d'améliorer les architectures convolutives. D'autres travaux prédisent un nombre fixe d'objets (Carion et al., 2020) ou utilisent la sortie du décodeur comme entrée d'un CNN afin d'avoir une prédiction dense (Zheng et al., 2020). Dans ce domaine, la séquentialité du processus de décision n'est pas non plus exploitée.

Les architectures Transformers, et plus particulièrement les décodeurs Transformers, fonctionnent selon un nouveau paradigme qui traite un élément en entrée séquentiellement, au rythme d'une séquence d'attention visuelle. Cela nécessite donc de repenser le type de sorties attendues qui doivent nécessairement être structurées sous forme d'une séquence d'objets. Il semble complexe de réaliser une prédiction pixel à pixel de manière séquentielle puisqu'elle induirait des temps et coût de traitement très élevés. Cependant, l'application de ce paradigme pour résoudre un problème d'analyse de document est relativement directe puisque, dans la plupart des applications, les sorties du modèle de détection et de reconnaissance ont besoin d'être organisées dans l'ordre naturel de lecture. La tâche de détection doit donc être reformulée afin de profiter pleinement de la capacité de prédiction séquentielle de ces nouvelles architectures. C'est pourquoi, nous présentons, dans la section 7.1, une étude et comparaison de différentes modélisations du problème de détection d'objets permettant une prédiction séquentielle.

Comme nous venons de l'évoquer, très peu de systèmes ont été proposés dans la littérature permettant de prédire séquentiellement les objets présents dans des images. Le seul modèle réalisant une telle tâche est Pix2Seq (Chen et al., 2022), détaillé dans le Focus 2.15, appliqué aux images de scènes naturelles. Ce système possédant un grand nombre de paramètres et nécessitant un pré-entraînement sur des milliers d'images, il n'est pas directement applicable à nos jeux de données réduits d'images de documents. Ainsi, inspiré par Pix2Seq, nous nous sommes intéressés à la mise en place d'un système permettant de prédire séquentiellement les objets tout en possédant un nombre réduit de paramètres afin d'être entraîné sur des jeux de données réduits. Ce système est détaillé dans la section 7.2.

## 7.1 MODÉLISATION DE LA TÂCHE DE DÉTECTION

Dans cette section, nous présentons et comparons différentes modélisations possibles de la tâche de détection d'objets. Dans un premier temps, nous comparons plusieurs modélisations de la position et de la forme des objets. Nous comparons ensuite plusieurs stratégies de prédiction des coordonnées. Enfin, nous présentons la modélisation des classes des objets que nous avons retenue.

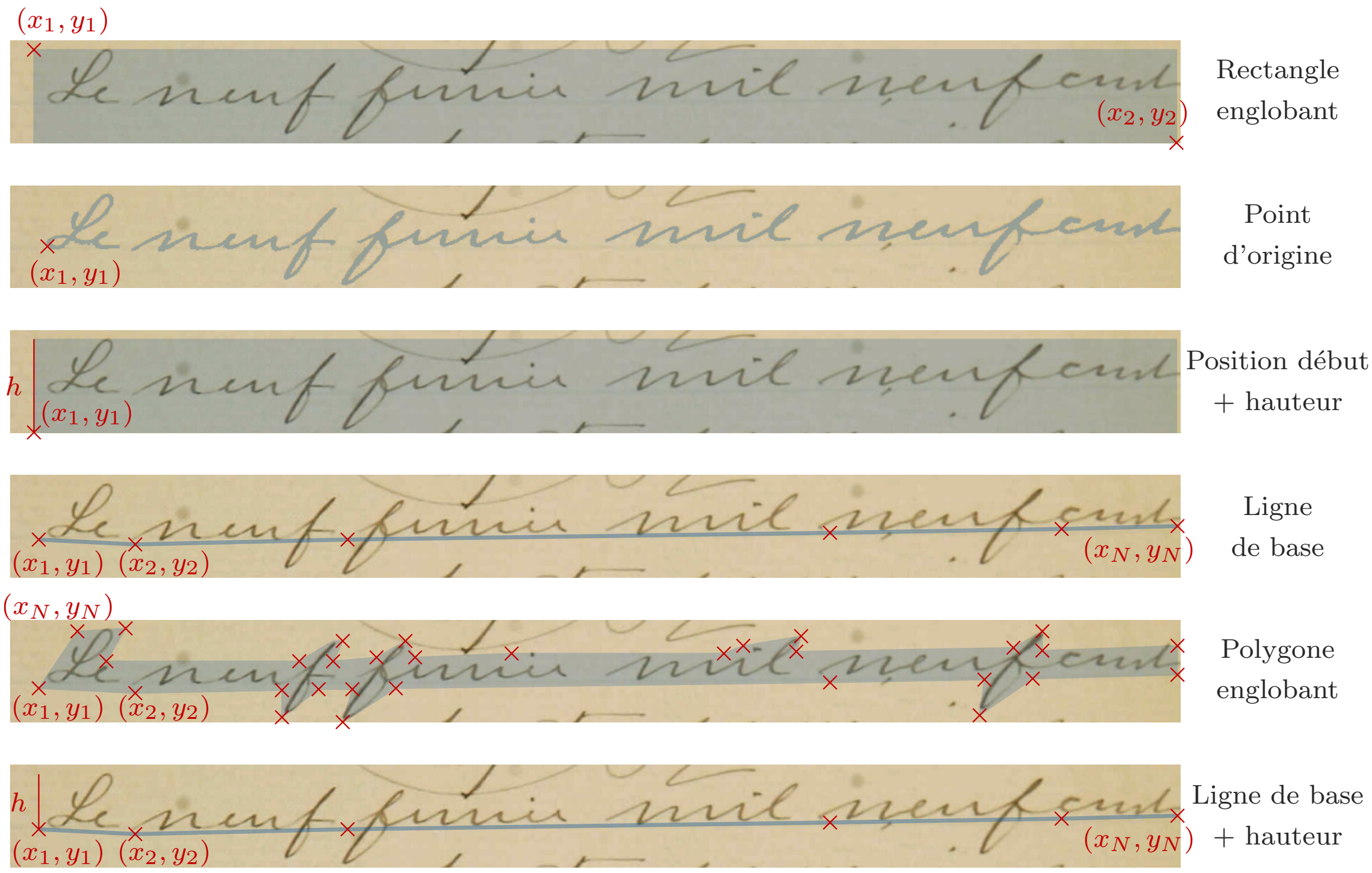

**Figure 7.1 –** Représentation de différentes modélisations de la position et de la forme des objets à détecter. Exemple pour la détection d'une ligne de texte.

### 7.1.1 MODÉLISATION DE LA POSITION ET DE LA FORME DES OBJETS

Afin de réaliser une prédiction séquentielle des éléments présents sur une image de document, il est nécessaire de passer d'une prédiction pixel à une prédiction de plus haut niveau, au niveau de l'objet. Pour cela, différentes formulations du problème de détection d'objets ont été proposées dans la littérature. Nous les présentons, dans le cadre d'une détection de lignes de texte, sur la Figure 7.1 et détaillons leurs caractéristiques dans la Table 7.1.

Dans le domaine de la vision, la grande majorité des systèmes tels que les modèles R-CNN (Girshick, 2015 ; Girshick et al., 2014 ; Ren et al., 2015) et YOLO (Redmon et al., 2016 ; 2017 ; 2018) définissent les objets à détecter par leur rectangle englobant. C'est également le cas de Pix2Seq (Chen et al., 2022) qui prédit la séquence suivante : ordonnée du point supérieur gauche, abscisse du point supérieur gauche, ordonnée du point inférieur droit, abscisse du point inférieur droit et classe de l'objet. Cette détection permet une extraction directe de l'objet mais n'est pas correctement applicable à des éléments non rectangulaires.

Lors de la compétition ANDAR-TL de détection de lignes de texte (Murdock et al., 2015), la tâche de détection correspond à l'identification des points d'origine des lignes de texte, à savoir la ligne de base du premier caractère d'une ligne. D'un autre côté, dans Moysset et al. (2017), les auteurs proposent une localisation des lignes de texte basée sur des régressions dans lesquelles seules les positions du début des lignes de texte et leurs hauteurs sont prédites. Le reconnaisseur de texte est alors chargé de reconnaître le texte de la ligne et de s'arrêter lorsqu'il n'y a plus de texte à reconnaître. Ces propositions permettent d'envisager une détection optimisée des éléments puisque, pour chaque objet, seules deux ou trois valeurs sont à prédire. Cependant, elles ne permettent pas une détection complète de l'objet puisque la largeur est inconnue. Il serait donc nécessaire d'avoir des traitements supplémentaires afin d'extraire les objets de l'image. Sans cela, il serait impossible d'appliquer un reconnaisseur de texte standard sur les lignes de texte par exemple.

**Table 7.1 –** Tableau récapitulatif de différentes modélisations de la position et forme des objets à détecter. La colonne PRÉDICTION indique les valeurs à prédire pour un objet. La colonne "Extraction directe" indique si l'objet peut directement être extrait en sortie du détecteur ou si des traitements supplémentaires sont nécessaires tels que l'estimation de la largeur et/ou de la hauteur de l'objet. La colonne "Optimisation mémoire" indique si la quantité de mémoire nécessaire pour prédire un objet est importante ou non, cette quantité étant directement corrélée au nombre de valeurs à prédire. La dernière colonne indique si la détection est applicable à des objets non rectangulaires ainsi qu'à des lignes inclinées ou incurvées.

| MODÉLISATION | PRÉDICTION | Extraction directe | Optimisation mémoire | Objets non-rectangulaires |
|---|---|---|---|---|
| Rectangle englobant<br>Pix2Seq (CHEN et al. (2022)) | $(x_1, y_1, x_2, y_2)$<br>$\rightarrow$ 4 valeurs | ✓ | ✓ | ✗ |
| Point d'origine<br>(MURDOCK et al., 2015) | $(x_1, y_1)$<br>$\rightarrow$ 2 valeurs | ✗ | ✓ | ✗ |
| Position du début + hauteur<br>(MOYSSET et al., 2017) | $(x_1, y_1, h)$<br>$\rightarrow$ 3 valeurs | ✗ | ✓ | ✗ |
| Ligne de base<br>(DIEM et al., 2019) | $(x_1, y_1, ...x_N, y_N)$<br>$\rightarrow$ N valeurs | ✗ | ✗ | ✓ |
| Polygone englobant | $(x_1, y_1, ...x_N, y_N)$<br>$\rightarrow$ N valeurs | ✓ | ✗ | ✓ |
| Ligne de base + hauteur | $(x_1, y_1, ...x_N, y_N, h)$<br>$\rightarrow$ N+1 valeurs | ✗ | ✗ | ✓ |

Enfin, la détection basée sur les lignes de base (DIEM et al., 2017 ; DIEM et al., 2019) ou les polygones englobants présente l'avantage de localiser précisément les contours d'objets rectangulaires ou non, tels que des lignes inclinées et incurvées. Cependant, ces propositions sont très coûteuses en mémoire puisque le système doit prédire un nombre de points inconnu à l'avance, et qui peut-être très variable d'un objet à l'autre, en fonction de la taille et de la forme des objets à localiser.

Cette représentation pose également un problème de paramétrage du Transformer. En effet, dans un Transformer, la taille maximale de la séquence pouvant être générée pour une image est fixée durant la phase d'entraînement afin de réduire la mémoire utilisée. Durant l'inférence, il est donc impossible de prédire plus de valeurs que cette limite. Bien que celle-ci puisse être fixée à plusieurs milliers de valeurs, il est toujours possible de rencontrer un document avec un très grand nombre d'objets, menant à une séquence plus longue. Dans le cas d'une détection de rectangles englobants, cette limite est facile à fixer puisque seules quatre valeurs sont à prédire pour chaque objet. Ainsi, le problème est réduit à quelques dizaines de valeurs à prédire par image. Le problème se complexifie lorsque nous souhaitons prédire des polygones plus précis puisque nous ignorons à l'avance combien de valeurs sont nécessaires pour prédire chaque objet. Il serait possible de simplifier les polygones englobants afin de fixer le nombre de coordonnées les définissant, cependant, cela mènerait à un traitement supplémentaire et à une perte de précision. De plus, de nombreuses questions en découlent telles que le nombre de points à utiliser pour décrire un polygone annoté, leurs espacements, l'évaluation des points obtenus pour le calcul de la fonction de perte.

**Table 7.2 –** Stratégies de prédiction séquentielle des rectangles englobants. Pour un rectangle donné $i$ de coordonnées $(x_0^i, y_0^i, x_1^i, y_1^i)$, à chaque pas $t$, une coordonnée unique, un point ou le rectangle complet peut être prédit.

| Séquence | $t = 0$ | $t = 1$ | $t = 2$ | $t = 3$ | $t = 4$ | $t = 5$ | $t = 6$ | ... |
|---|---|---|---|---|---|---|---|---|
| Coordonnée | $x_0^0$ | $y_0^0$ | $x_1^0$ | $y_1^0$ | $x_0^1$ | $y_0^1$ | $x_1^1$ | ... |
| Point | $(x_0^0, y_0^0)$ | $(x_1^0, y_1^0)$ | $(x_0^1, y_0^1)$ | $(x_1^1, y_1^1)$ | ... | | | |
| Rectangle | $(x_0^0, y_0^0, x_1^0, y_1^0)$ | $(x_0^1, y_0^1, x_1^1, y_1^1)$ | ... | | | | | |

Pour toutes ces raisons, nous pensons que la détection des rectangles englobants par la prédiction du point supérieur gauche et du point inférieur droit, semblable à Pix2Seq, représente un bon compromis entre performance et précision de la localisation. Cette formulation est assez simple et rapide à appliquer, et peut permettre à un système d'apprendre malgré la quantité relativement faible de données annotées. En effet, plus le système doit prédire de points, plus il sera en difficultés et nécessitera un grand nombre de données d'apprentissage.

### 7.1.2 Stratégie de prédiction des coordonnées : singleton vs n-uplet

La modélisation de la tâche de détection explicitement définie, il est maintenant nécessaire de choisir la stratégie de prédiction des rectangles englobants. En effet, le système devra être capable de prédire séquentiellement les coordonnées des rectangles englobants. Cependant, la Table 7.2 présente trois stratégies différentes afin de réaliser cette tâche. Ainsi, pour un rectangle donné $i$ de coordonnées $(x_0^i, y_0^i, x_1^i, y_1^i)$ avec $(x_0^i, y_0^i)$ les coordonnées du point supérieur gauche et $(x_1^i, y_1^i)$ les coordonnées du point inférieur droit, il est possible de prédire, à chaque pas de la séquence :

— Un singleton correspondant à une coordonnée d'un des deux points du rectangle ;
— Un couple de valeurs correspondant à un des deux points du rectangle ;
— Un quadruplet correspondant aux coordonnées du rectangle complet.

La première stratégie, qui consiste à prédire un singleton à chaque pas dans la séquence, permet d'avoir un modèle possédant quelques paramètres en moins par rapport à la prédiction de couples qui elle-même nécessite moins de paramètres que la prédiction de quadruplets. En effet, la dernière couche du modèle produisant les coordonnées finales sera différente d'une stratégie à l'autre. Cependant, la prédiction de singletons requiert davantage d'itérations puis-qu'elle nécessite deux fois plus d'itérations que la prédiction de couples, elle-même nécessitant deux fois plus d'itérations que la prédiction du quadruplet.

Dans Pix2Seq, les auteurs ont choisi de traiter cette tâche de telle sorte qu'à chaque pas, une seule valeur de coordonnée est prédite. Ainsi, quatre prédictions sont nécessaires afin de prédire un objet. Dans la suite de ce chapitre, nous décidons d'adopter la même stratégie.

### 7.1.3 STRATÉGIE DE PRÉDICTION DES COORDONNÉES : CLASSIFICATION VS RÉGRESSION

Une autre stratégie à étudier dans la conception du modèle concerne le type de prédiction souhaité. En effet, la prédiction de coordonnées peut être réalisée de deux manières. La première consiste à réaliser une régression où le but est de prédire une coordonnée de la boîte sur l'image. La seconde consiste à considérer chaque pixel de l'image d'entrée comme étant une classe distincte et à réaliser une classification. Le but est alors de maximiser les probabilités de la classe correspondant à la coordonnée de la boîte dans l'image.

L'avantage de la régression est qu'elle nécessite légèrement moins de paramètres que la classification puisqu'une seule valeur est produite par le modèle, directement considérée comme la coordonnée finale. Ainsi, la dernière couche du modèle ne produira qu'une seule valeur. Cependant, les valeurs prédites ne sont pas bornées, il est donc possible que le modèle prédise des valeurs en dehors de l'image. Pour pallier ce problème, les coordonnées peuvent être normalisées, ce qui permet également d'obtenir une cohérence des labels entre les images pouvant être de tailles variables.

La classification est quant à elle plus simple à mettre en œuvre. Dans Pix2Seq, les auteurs choisissent de traiter les images ainsi, en redimensionnant les images dans une taille fixe et en considérant une classe pour chaque valeur possible en abscisse et en ordonnée. Dans un premier temps, nous avons choisi d'utiliser cette même stratégie.

Dans Chen et al. (2022), les auteurs considèrent les classes permettant de représenter les positions des objets en abscisse et en ordonnée comme appartenant à un "vocabulaire". Cela leur permet de distinguer les positions des objets des "classes", utilisées pour représenter les classes des objets à détecter telles que, dans leur application, les classes "chaise" ou "voiture". Nous utilisons ces mêmes termes dans la suite de ce chapitre. Dans Pix2Seq, les auteurs utilisent un vocabulaire partagé pour les deux axes et pour toutes les classes de position, la taille du vocabulaire est donc égale au nombre maximal de pixels sur les deux axes. Pour une image de taille 600×600 pixels, le vocabulaire a donc une taille de 600. De la même manière, nous disposons, dans nos expériences, d'un vocabulaire $V$ de taille $T_V = max(H, W)$ avec $H$ et $W$ respectivement les hauteur et largeur de l'image d'entrée.

### 7.1.4 STRATÉGIE DE PRÉDICTION DE LA CLASSE DES OBJETS

En plus des coordonnées des objets présents dans les images, il est nécessaire de prédire leurs classes. En effet, dans les chapitres précédents, nous avons principalement abordé la tâche de détection de lignes de texte uniquement, ainsi une seule classe est à prédire. Cependant, certaines tâches plus complexes considèrent davantage de classes d'objets qui doivent être prédites par le modèle.

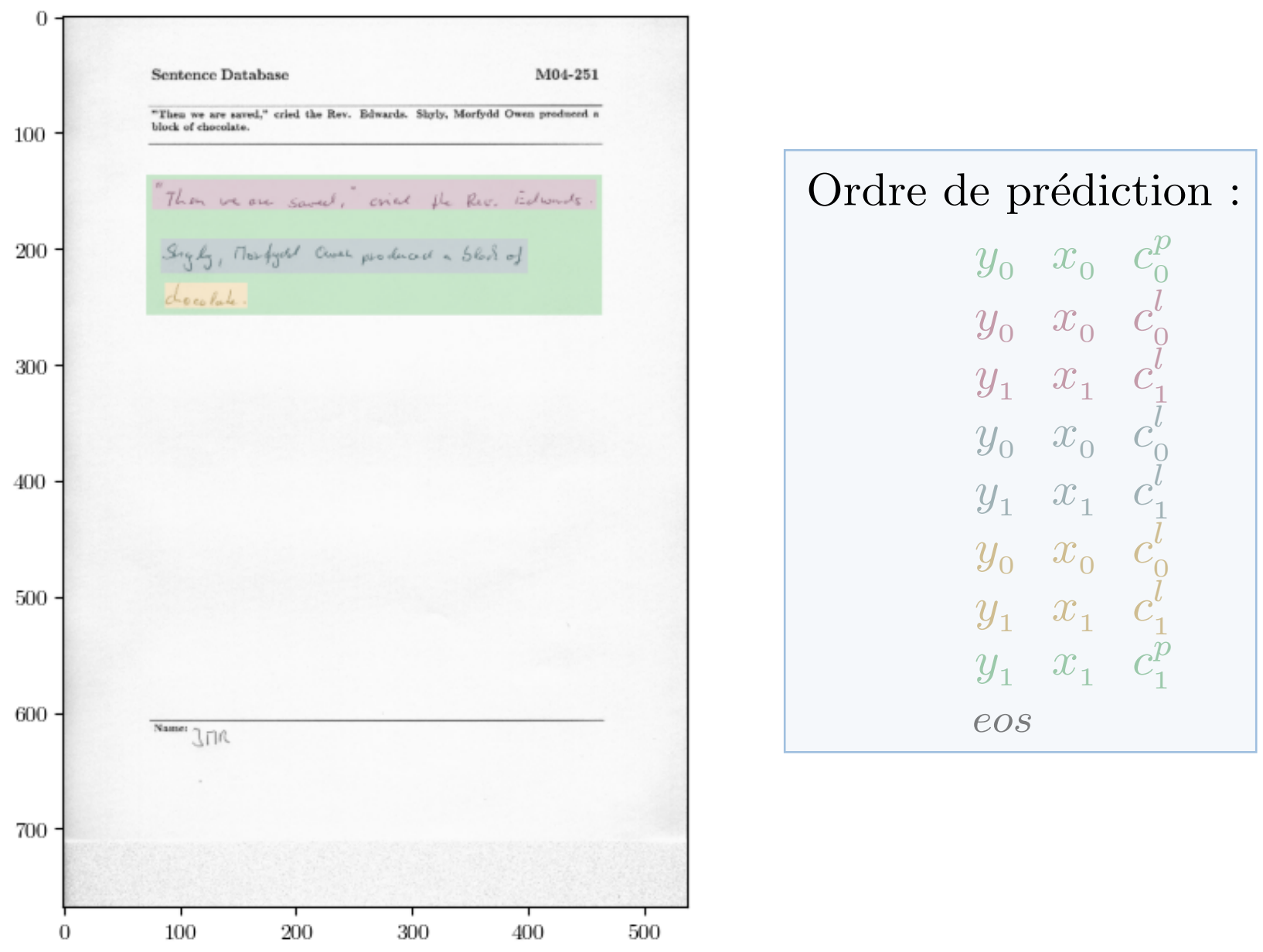

**Figure 7.2 –** Exemple de séquence à deux classes : paragraphe et ligne de texte. L'ordre de prédiction préserve la hiérarchie des objets : point supérieur gauche du paragraphe, point supérieur gauche de la première ligne de texte, point inférieur droit de la première ligne de texte, ..., point inférieur droit du paragraphe, fin de séquence ($eos$).

Ces classes peuvent être représentées de différentes manières. Ainsi, dans Pix2Seq, le modèle prédit la classe de l'objet après les quatre coordonnées du rectangle englobant. Chaque objet est donc défini par cinq prédictions successives. Cette représentation ne permet cependant pas de représenter la hiérarchie des objets présents sur une image.

Afin de représenter au mieux ces informations, nous proposons de représenter un objet par ses deux points supérieur gauche et inférieur droit avec, pour chacun de ces points, une classe indiquant s'il s'agit du premier ou du second point de l'objet. Cette représentation permet au modèle d'apprendre, en plus de l'ordre de lecture, la hiérarchie des objets. La Figure 7.2 présente un exemple de séquence construite pour deux classes d'objets : paragraphe et ligne de texte. Chaque ligne représente un point avec son ordonnée, son abscisse et la classe correspondante. Pour chaque classe d'objet, deux classes sont définies : une classe indiquant le début de l'objet et une classe indiquant la fin. Ainsi, un objet paragraphe est défini par deux points, le premier avec la classe $c_0^p$ et le second avec la classe $c_1^p$. De la même manière, les lignes de texte sont définies par les classes $c_0^l$ et $c_1^l$.

Cette représentation permet également de reconstruire les boîtes englobantes de manière plus fiable, même dans le cas d'une prédiction manquante ou supplémentaire. En effet, si le modèle prédit deux points de début de ligne à la suite, lors de la reconstruction des objets, un des deux points devra être mis de côté. Sans ces indicateurs de début et de fin d'objet, il serait impossible de détecter ce phénomène. Le processus étant séquentiel, l'ensemble des boîtes reconstruites après le point erroné seraient fausses.

En conclusion, nous avons choisi de représenter un objet par les deux séquences "$y_0$, $x_0$, $c_0$" et "$y_1$, $x_1$, $c_1$" avec :

- $y_0$ et $y_1$, respectivement les ordonnées des points supérieur gauche et inférieur droit du rectangle englobant de l'objet ;
- $x_0$ et $x_1$, respectivement les abscisses des points supérieur gauche et inférieur droit du rectangle englobant de l'objet ;
- $c_0$ et $c_1$, les jetons de début et de fin de la classe de l'objet.

## 7.2 ARCHITECTURE DU SYSTÈME PROPOSÉ : DOC2SEQ

Dans cette section, nous présentons le modèle que nous avons développé, appelé Doc2Seq. Nous détaillons l'architecture ainsi que les choix que nous avons faits lors de sa conception.

Très peu de modèles ont été proposés pour la détection séquentielle d'objets dans les images. Seul Pix2Seq (Chen et al., 2022) a été proposé, appliqué aux images de scènes naturelles. Il s'agit d'un modèle comportant un très grand nombre de paramètres (341 millions pour le meilleur modèle) qui montre des résultats satisfaisants lorsqu'il est pré-entraîné sur des milliers d'images. Or, nous ne disposons pas d'une telle quantité d'images annotées. C'est pourquoi, il est nécessaire que notre système comporte moins de paramètres afin de pouvoir être entraîné sur les jeux de données d'images de documents. C'est dans cet objectif que nous avons conçu Doc2Seq, dont l'architecture est présentée en Figure 7.3. Il s'agit d'un modèle hybride encodeur-décodeur où l'encodeur extrait les caractéristiques importantes de l'image d'entrée et le décodeur prédit séquentiellement les éléments à partir de l'image encodée et des prédictions précédentes. L'encodeur génère une matrice de caractéristiques 2D de l'image d'entrée. Un encodage positionnel 2D est additionné à cette matrice afin de conserver l'information spatiale, avant d'être aplani en une séquence 1D de caractéristiques. Comme pour un FCN, cette représentation est calculée une seule fois et sert d'entrée au décodeur. Le décodeur suit un processus récurrent : étant donné l'image encodée et les éléments prédits précédemment ($\hat{y}_0$, $\hat{y}_1$, ..., $\hat{y}_{t-1}$), il produit les caractéristiques de l'élément suivant. Enfin, la branche de classification produit des probabilités à partir de la sortie du décodeur et l'élément prédit $\hat{y}_t$ est celui qui a la plus forte probabilité. Chacun de ces composants est détaillé dans les paragraphes suivants.

### 7.2.1 ENCODEUR DOC-UFCN

L'encodeur de Doc2Seq est identique à l'encodeur de Doc-UFCN présenté dans le chapitre 4. Il est donc composé de quatre blocs dilatés comportant chacun cinq convolutions dilatées consécutives. Chaque bloc est suivi d'une couche de *max-pooling*, sauf le dernier.

Nous avons choisi d'utiliser l'encodeur de Doc-UFCN car, d'après les expériences présentées précédemment, il a démontré de bonnes capacités d'extraction de caractéristiques sur les images tout en possédant un nombre réduit de paramètres, ce qui nous permet d'entraîner le système complet sur des jeux de données restreints. De plus, nous avons opté pour un FCN comme encodeur car ces modèles peuvent traiter des entrées de tailles variables.

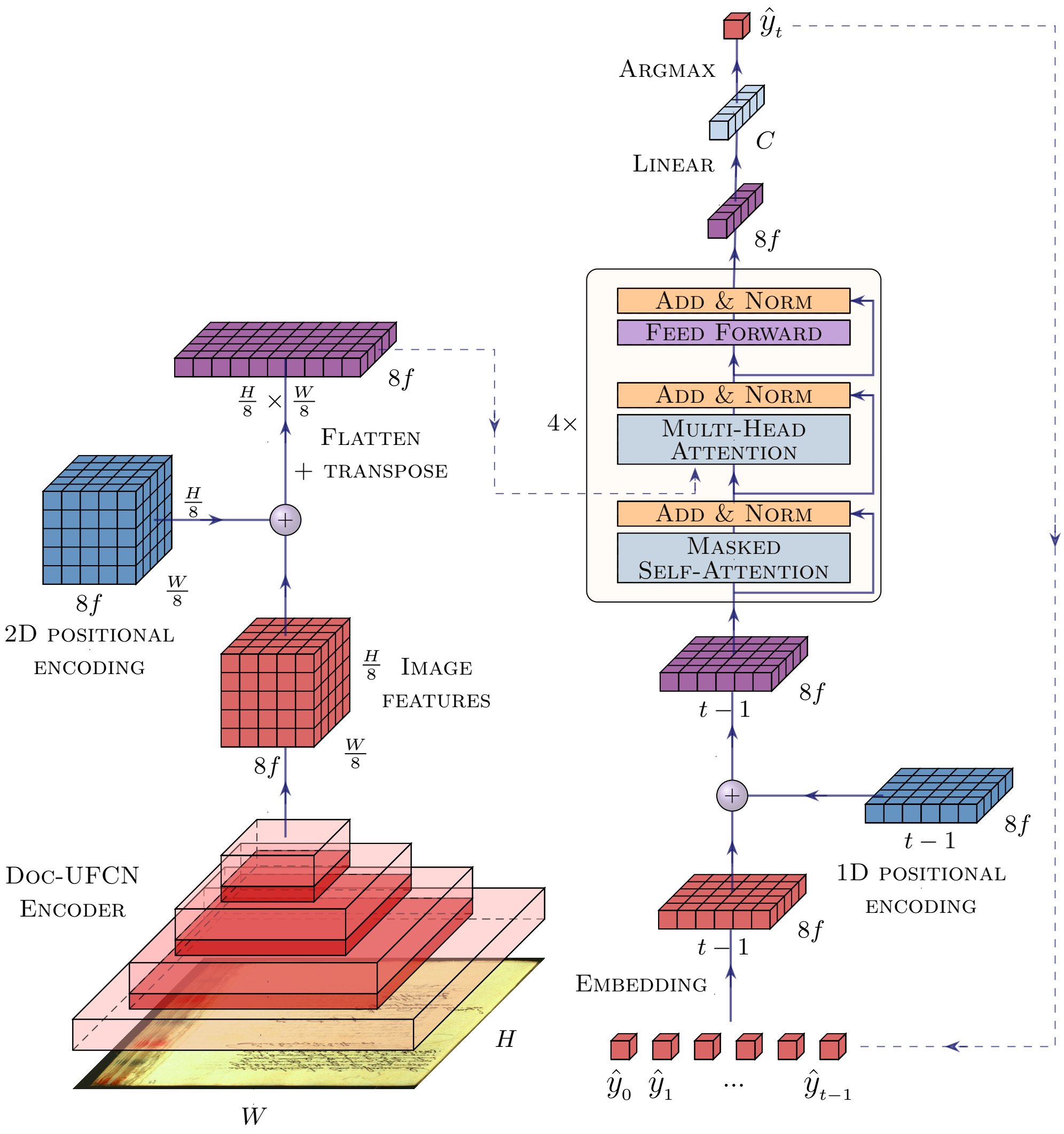

**Figure 7.3 –** Schéma de l'architecture du modèle Doc2Seq avec respectivement $H$ et $W$ les hauteur et largeur de l'image d'entrée, $f$ le nombre de cartes de caractéristiques et $\hat{y}_i$ les prédictions.

Certains systèmes tels que les *Vision Transformers* remplacent les encodeurs convolutifs par des encodeurs Transformer. Ces systèmes sont appliqués sur des patchs d'image à la résolution originale projetés dans la dimension $d_{model}$. Bien que cet encodeur ait montré de légèrement meilleures performances dans CHEN et al. (2022) par rapport à un encodeur CNN, il augmente significativement le nombre de paramètres et nécessite donc une plus grande quantité de données d'entraînement.

Dans nos expériences, l'encodeur prend en entrée une image de document de taille $(H \times W \times 3)$ avec $H$ la hauteur, $W$ la largeur et *3* le nombre de canaux de l'image. Il produit une matrice de caractéristiques de taille $(\frac{H}{8} \times \frac{W}{8} \times 8f)$ avec $f = 32$ comme pour Doc-UFCN.

### 7.2.2 ENCODAGE POSITIONNEL 2D

Une fois l'image encodée, sa matrice de caractéristiques est additionnée à une matrice de codage positionnel 2D afin de garder en mémoire à quelle partie de l'image chaque pixel correspond. Le Transformer original a été conçu pour traiter des séquences en entrée en 1D. Pour lui donner des représentations 2D en entrée, il suffit de transformer les cartes de caractéristiques en les sérialisant ligne par ligne. Cependant, il est important d'associer à ces représentations un encodage positionnel cohérent avec l'information originale, c'est-à-dire un encodage 2D. C'est ce que nous avons réalisé comme proposé par SINGH et al. (2021). Ainsi, il s'agit toujours d'un codage fixe basé sur les fonctions cosinus et sinus, mais, au lieu de coder une position 1D sur tous les canaux, la première moitié est dédiée à l'encodage positionnel vertical et la seconde à l'encodage positionnel horizontal (voir équations 7.1).

$$
\begin{aligned}
\mathrm{PE}(pos_x, pos_y, 2i) &= \sin(w_i \cdot pos_y) \quad \forall i \in \left[0, \frac{d_{model}}{4}\right] \\
\mathrm{PE}(pos_x, pos_y, 2i+1) &= \cos(w_i \cdot pos_y) \quad \forall i \in \left[0, \frac{d_{model}}{4}\right] \\
\mathrm{PE}(pos_x, pos_y, \frac{d_{model}}{2} + 2i) &= \sin(w_i \cdot pos_x) \quad \forall i \in \left[0, \frac{d_{model}}{4}\right] \\
\mathrm{PE}(pos_x, pos_y, \frac{d_{model}}{2} + 2i+1) &= \cos(w_i \cdot pos_x) \quad \forall i \in \left[0, \frac{d_{model}}{4}\right]
\end{aligned}
\tag{7.1}
$$

avec :

$$
w_i = \frac{1}{10000^{\frac{2i}{d_{model}}}}
$$

La position de l'élément dans la séquence 2D est donnée par $pos_x$ et $pos_y$. $d_{model}$ correspond à la dimension d'encodage de l'image d'entrée et des éléments au sein du Transformer. Dans notre cas, $d_{model} = 8f = 256$.

La matrice de caractéristiques ainsi enrichie de la position des éléments est ensuite aplanie afin de pouvoir être utilisée lors du décodage.

### 7.2.3 DÉCODEUR TRANSFORMER

Pour le décodeur, nous utilisons un Transformer standard puisqu'il permet la prédiction de séquences de longueurs variables. Celui-ci est constitué d'un empilement de quatre couches de décodeur Transformer. Chaque couche suit une architecture standard avec un mécanisme d'auto-attention et un mécanisme d'attention croisée. L'auto-attention modélise les dépendances entre les éléments de la séquence prédite, contrairement à l'attention croisée, utilisée pour extraire des informations visuelles de l'encodeur, sur la base des prédictions précédentes. En d'autres termes, étant donné les prédictions précédentes, elle indique où le modèle doit regarder pour prédire le prochain élément.

Le décodeur suit donc un processus récurrent où à chaque itération, il prend en entrée les caractéristiques visuelles aplanies et les éléments prédits précédemment ($\hat{y}_0$, $\hat{y}_1$, ..., $\hat{y}_{t-1}$) et produit un vecteur de caractéristiques pour la prédiction de l'élément au pas de temps $t$.

Chaque prédiction précédente est encodée dans un vecteur de taille $8f$ grâce à une couche d'*embedding* apprise, puis les vecteurs sont concaténés afin de former une matrice de taille $(t - 1 \times 8f)$. Ainsi, les caractéristiques visuelles et les vecteurs des prédictions précédentes ont la même dimension $d_{model} = 8f = 256$.

Comme pour un Transformer standard, la matrice des *embeddings* est additionnée à une matrice de codage positionnel 1D de même taille qui permet de savoir où se trouve cette prédiction dans la séquence. Cet encodage positionnel 1D est détaillé dans le Focus 2.13.

### 7.2.4 BRANCHE DE CLASSIFICATION

La branche de classification génère des probabilités à partir de la sortie du décodeur. Elle est composée d'une couche linéaire qui permet de modifier la dimension de la sortie de $d_{model} = 256$ à $T_C = T_V + 2 \times C + 2$, $C$ étant le nombre de classes considérées dans l'expérience. Deux sorties supplémentaires sont ajoutées afin de représenter les jetons de fin de séquence (EOS) et de *padding*. Le jeton de *padding* est utilisé afin de permettre un entraînement par batchs dans lesquels les séquences doivent être de même longueur. Cette couche linéaire est suivie d'une fonction *argmax* qui assigne à $\hat{y}_t$ la position ou la classe d'objet pour laquelle la probabilité est maximale.

## 7.3 DÉTAILS D'IMPLÉMENTATION ET STRATÉGIES D'ENTRAÎNEMENT

Dans cette section, nous donnons des détails techniques sur l'implémentation utilisée lors de nos expériences.

Doc2Seq est implémenté à l'aide du framework PyTorch. Il est entraîné avec un *learning rate* initial de $5e-5$, l'optimiseur Adam et la fonction de coût d'entropie croisée. Les poids sont initialisés grâce à l'initialisation Glorot. Nous utilisons le *teacher forcing*, qui est une stratégie d'apprentissage de modèle qui utilise la vérité terrain comme entrée au lieu de la sortie du modèle de l'itération précédente. Cela permet de paralléliser les calculs en prédisant en parallèle tous les éléments de la séquence de sortie, et donc de réduire le temps d'entraînement.

Au total, le modèle comporte 6,6 millions de paramètres répartis comme suit :

— 3,5 millions venant de l'encodeur Doc-UFCN ;
— 0,2 million pour la couche d'*embedding* ;
— 2,6 millions venant du décodeur Transformer ;
— 0,2 million pour la branche de classification.

Ces valeurs sont données dans un contexte dans lequel une seule classe d'objets est considérée ($C = 1$). Les nombres de paramètres venant de la couche d'*embedding* et de la branche de classification varient légèrement en fonction de ce nombre de classes $C$.

### 7.3.1 TAILLE DES IMAGES EN ENTRÉE

Puisque nous utilisons l'encodeur du modèle Doc-UFCN, nous avons décidé d'utiliser la taille d'image en entrée ayant obtenu les meilleurs résultats dans les expériences présentées dans les chapitres précédents. Ainsi, les images d'entrée sont redimensionnées en images plus petites telles que la plus grande dimension de l'image soit égale à 768 pixels tout en conservant le ratio de l'image originale. Les coordonnées des objets présents sur les images sont également mises à l'échelle. Ainsi, nous disposons d'un vocabulaire de taille $T_V = 768$.

### 7.3.2 AUGMENTATION DE DONNÉES

Durant l'entraînement des modèles, nous utilisons une stratégie d'augmentation des données. Tout d'abord, des augmentations sont appliquées à l'image d'entrée telles qu'un ajout de bruit et de flou gaussien, un changement de luminosité, une inversion des canaux de couleur ou une mise en niveaux de gris. De plus, des transformations linéaires, telles que des translations et des rotations, sont appliquées.

Enfin, nous augmentons les séquences d'entrée pendant l'apprentissage pour inclure des jetons bruités. Cela améliore la robustesse du modèle contre les prédictions bruitées et dupliquées. Les séquences sont augmentées de trois manières :

- Ajout de bruit sur les coordonnées des boîtes englobantes (translation et redimensionnement aléatoires avec une probabilité de 0,3) ;
- Suppression aléatoire de 20 % des boîtes ;
- Inversion des jetons de classes de début et de fin avec une probabilité de 0,1.

### 7.3.3 DÉCODEUR TRANSFORMER

Nous utilisons quatre couches de décodage avec la dimension $d_{model} = 8f = 256$. Chaque couche de décodage possède quatre têtes d'attention et utilise une activation ReLU.

Comme les différentes images comportent souvent un nombre différent d'objets, les séquences générées auront des longueurs différentes. Pour indiquer la fin d'une séquence, nous incorporons donc un jeton de fin de séquence (EOS). Ainsi, le processus de prédiction se termine lorsque le jeton EOS est prédit ou après un nombre prédéfini de valeurs prédites.

### 7.3.4 CHOIX DU MEILLEUR MODÈLE

Dans le cadre d'une application de reconnaissance de document, l'objectif principal est d'obtenir le texte contenu dans celui-ci ainsi que sa position sur l'image. Comme montré dans le chapitre 5, il est nécessaire d'évaluer les modèles de détection de lignes de texte grâce aux métriques de reconnaissance. Ainsi, nous pouvons évaluer l'impact des résultats de détection sur les résultats finaux.

Dans nos expériences, nous avons souhaité poursuivre dans cette direction en intégrant un reconnaisseur pré-entraîné dans les processus de sélection des meilleurs modèles et d'évalua-tion. Ainsi, chaque expérience impliquant une détection de lignes de texte intègre un modèle de reconnaissance de texte niveau ligne. Pour cela, un modèle de reconnaissance est tout d'abord entraîné sur les lignes transcrites provenant du même jeu de données que celui considéré dans l'expérience. Ensuite, à partir de la 500e époque et toutes les cinq époques, le modèle de reconnaissance est appliqué à l'ensemble des lignes prédites sur l'ensemble de validation et un CER@page est calculé (voir algorithme 5.1). Le modèle final est celui obtenant le CER le plus bas. Cette stratégie de sélection du meilleur modèle est comparé à une sélection standard basée sur la fonction de coût dans la section 7.4.3.

Table 7.3 – Statistiques du jeu de données IAM utilisé pour la détection de lignes de texte.

| Jeu de données | Images | | | Lignes | | |
|---|---|---|---|---|---|---|
| | train | valid | test | train | valid | test |
| IAM Marti et al. (2002) | 747 | 220 | 232 | 6 482 | 1 926 | 1 965 |

Enfin, ce même modèle de reconnaissance est appliqué durant la phase d'évaluation afin d'obtenir le CER@page sur l'ensemble de test.

## 7.4 EXPÉRIENCES ET RÉSULTATS

Nous décrivons, dans cette section, les résultats préliminaires obtenus avec Doc2Seq pour la détection de lignes de texte sur le jeu de données IAM (Marti et al., 2002).

### 7.4.1 JEU DE DONNÉES

La Table 7.3 présente les statistiques du jeu de données issu de la base IAM utilisé pour l'entraînement et l'évaluation du modèle Doc2Seq. Nous avons choisi ce jeu car il est annoté au niveau ligne et nous disposons des transcriptions pour chaque ligne de texte. De plus, il s'agit d'un jeu de données assez simple, annoté en rectangles englobants.

Durant l'entraînement, les lignes de texte sont ordonnées de haut en bas afin que le modèle apprenne cet ordre de lecture.

### 7.4.2 ENTRAÎNEMENT DES MODÈLES DE DÉTECTION

Le modèle est entraîné avec des mini-batchs de taille 12 pour réduire le temps d'apprentissage sur un maximum de 1500 époques. De plus, dans un processus d'entraînement standard, le meilleur modèle est choisi comme étant celui obtenant les meilleures performances sur l'ensemble de validation. Il est choisi selon les valeurs de la fonction de coût ou d'une métrique directement liée à la tâche. Dans nos expériences, nous choisissons le meilleur modèle comme étant celui obtenant le plus faible CER. Nous comparons l'impact de ce choix par rapport à une sélection standard basée sur la valeur de la fonction de coût dans les paragraphes suivants.

**Table 7.4** – Résultats de reconnaissance de textes manuscrits sur le jeu de données IAM. Nous présentons également les résultats du modèle de Moysset et al. (2019), modèle obtenant les résultats à l'état de l'art sans modèle de langue.

| Système | CER (%) | | | WER (%) | | |
|---|---|---|---|---|---|---|
| | train | valid | test | train | valid | test |
| PyLaia | 0,32 | 6,50 | **7,68** | 1,26 | 19,12 | **19,82** |
| Moysset et al. (2019) | – | **4,62** | 7,73 | – | **17,31** | 25,22 |

#### Modèle de reconnaissance PyLaia

Un modèle de reconnaissance PyLaia (Puigcerver, 2017) est entraîné au préalable sur les mêmes données et en suivant la même répartition dans les ensembles d'entraînement, de validation et de test. Il est ensuite intégré à l'entraînement de Doc2Seq et appliqué aux boîtes prédites sur l'ensemble de validation. Le modèle PyLaia a été choisi car il obtient des résultats satisfaisants sur les textes manuscrits tout en étant assez rapide ce qui permet de l'intégrer à l'entraînement. De plus, le système est facilement interfaçable avec le code PyTorch de Doc2Seq.

Les résultats du modèle de reconnaissance niveau ligne sont présentés dans la Table 7.4. Cette table montre des performances assez satisfaisantes que nous considérons suffisantes afin d'évaluer et de comparer les modèles de détection.

### 7.4.3 Résultats et discussion

Dans cette section, nous présentons, tout d'abord, les résultats quantitatifs du modèle Doc2Seq. Nous visualisons ensuite les prédictions et analysons les erreurs obtenues.

Durant l'inférence, les valeurs prédites sont regroupées par six (les quatre coordonnées et les deux classes de points) afin de créer les rectangles englobants. Le modèle a très bien appris sur le jeu de données IAM puisque qu'aucune valeur n'est prédite en plus et que les jetons de classes ont tous été correctement prédits.

Pour comparaison, nous avons entraîné un modèle Doc-UFCN sur les mêmes données IAM. Le modèle a été entraîné durant 150 époques dans les mêmes conditions que celles décrites dans les chapitres précédents :

- Images redimensionnées telles que la plus grande dimension de l'image soit égale à 768 pixels ;
- Taux d'apprentissage initial de $5e-3$, mini-batchs de taille 4, optimiseur Adam, fonction de coût Dice et arrêt anticipé (*early stopping*).

De plus, afin d'avoir des résultats comparables, les rectangles englobants des composantes connexes prédites par Doc-UFCN sont extraites et le même schéma d'évaluation que Doc2Seq est appliqué.

**Table 7.5** – Résultats des modèles de détection de lignes sur le jeu de données IAM, donnés en fonction du critère de sélection. Les colonnes "Manuel" présentent les résultats du reconnaisseur sur les lignes annotées manuellement contrairement aux colonnes "Prédit" qui présentent les résultats du reconnaisseur sur les lignes prédites automatiquement.

| Système | Critère de sélection | Set | AP@.5 | AP@.75 | mAP | CER (%) Manuel | CER (%) Prédit | WER (%) Manuel | WER (%) Prédit |
|---|---|---|---|---|---|---|---|---|---|
| Doc2Seq | CER | train | 0,98 | 0,83 | 0,68 | 0,31 | 1,21 | 1,43 | **3,82** |
| | | valid | **0,98** | 0,68 | 0,60 | 5,72 | **6,98** | 20,11 | 22,67 |
| | | test | **0,98** | 0,71 | 0,62 | 6,65 | **7,58** | 20,97 | **22,62** |
| Doc2Seq | Entropie croisée | train | **0,99** | 0,82 | 0,69 | 0,31 | **1,17** | 1,43 | 3,86 |
| | | valid | 0,97 | 0,70 | 0,61 | 5,72 | **6,98** | 20,11 | **22,49** |
| | | test | **0,98** | 0,70 | 0,62 | 6,65 | 7,66 | 20,97 | 22,68 |
| Doc-UFCN | CER | train | 0,98 | 0,61 | 0,57 | 0,31 | 3,37 | 1,43 | 9,13 |
| | | valid | 0,97 | 0,59 | 0,57 | 5,72 | 8,47 | 20,11 | 25,48 |
| | | test | 0,95 | 0,61 | 0,55 | 6,65 | 10,64 | 20,97 | 27,34 |
| Doc-UFCN | DICE | train | 0,92 | **0,89** | **0,71** | 0,31 | 6,83 | 1,43 | 9,69 |
| | | valid | 0,92 | **0,87** | **0,70** | 5,72 | 12,68 | 20,11 | 28,24 |
| | | test | 0,88 | **0,83** | **0,68** | 6,65 | 16,59 | 20,97 | 30,68 |

### RÉSULTATS QUANTITATIFS

La Table 7.5 présente les performances des modèles obtenus pour le jeu de données IAM. Dans cette Table, nous montrons les résultats de deux modèles issus du même entraînement mais sélectionnés selon un critère différent. Les modèles sont évalués par différentes métriques :

— Les métriques objet fournies par COCO [1], notamment la précision moyenne (AP) pour différentes valeurs de seuil : AP@.5, AP@.75 et AP@[.5,.95] (mAP) ;
— Les métriques de reconnaissance niveau page : CER et WER.

Comme dans les chapitres précédents, les colonnes "Manuel" présentent les résultats du reconnaisseur sur les lignes annotées manuellement. Elles correspondent donc au CER entre les transcriptions et les résultats d'HTR appliqué sur les mêmes lignes de texte. Ainsi, ces valeurs représentent les meilleures atteignables dans le cas d'un détecteur de lignes idéal.

Les deux modèles Doc2Seq présentent des résultats très satisfaisants. En effet, les valeurs d'AP sont relativement élevées pour de la détection de lignes de texte dans les images de documents. De plus, les valeurs de CER sur les ensembles de validation et de test sont très proches des valeurs obtenues sur les lignes annotées manuellement. Cela signifie que les lignes obtenues sont localisées avec précision sur les images. En effet, pour le modèle ayant obtenu le CER le plus faible, nous perdons moins d'un point de pourcentage de CER (de 6,65 % à 7,58 %) en passant des lignes manuelles aux lignes prédites sur l'ensemble de test.

De plus, les deux modèles Doc2Seq obtiennent des résultats similaires. En effet, pour les métriques AP, les deux modèles obtiennent, en moyenne, un point de pourcentage d'écart. Le modèle sélectionné sur la base du CER présente un faible gain de performances en CER

1. https://github.com/cocodataset/cocoapi

et WER sur l'ensemble de test, ce qui était attendu puisqu'il a été choisi afin de minimiser le CER. Cependant, le modèle sélectionné sur la base de l'entropie croisée présente des résultats très satisfaisants, ce qui valide l'utilisation de la classification comme type de prédiction couplé à la fonction de perte d'entropie croisée. Le calcul du CER durant l'entraînement, qui nécessite le texte des documents transcrits ainsi qu'un modèle de reconnaissance entraîné, ne semble donc pas nécessaire pour obtenir un modèle très performant. Bien que cela permette d'optimiser les résultats de reconnaissance, il est envisageable d'entraîner un modèle performant sans utilisation de reconnaisseur de texte.

Les résultats obtenus par le modèle Doc-UFCN sont légèrement meilleurs en termes de précision moyenne par rapport aux modèles Doc2Seq, notamment sur l'ensemble de validation. Cependant, les résultats finaux de reconnaissance niveau page sont bien moins bons que ceux des modèles Doc2Seq. En effet, pour le critère de sélection basé sur la fonction de coût, nous notons une augmentation de +5,70 points de pourcentage de CER sur l'ensemble de validation et de +5,75 points de WER. Ces écarts sont plus faibles lorsque nous comparons les modèles sélectionnés sur la base du CER, cependant, le modèle Doc-UFCN reste bien moins bon que le modèle Doc2Seq.

Ainsi, pour des résultats niveau objet équivalents, le modèle Doc2Seq obtient de bien meilleures performances en reconnaissance de texte. Cela s'explique, entre autres, par sa capacité à apprendre l'ordre de lecture, ordre non disponible dans les prédictions de Doc-UFCN pour lequel nous avons dû ordonner les boîtes de haut en bas. Ces résultats montrent une nouvelle fois l'intérêt des métriques orientées vers la tâche finale dans l'évaluation et la comparaison de modèles de détection.

Le modèle Doc2Seq présente un temps de prédiction moyen de 284 ms par image sur une carte graphique Tesla V100-SXM2-16GB pour le jeu de données IAM. Dans les mêmes conditions, Doc-UFCN est lui deux fois plus rapide, avec un temps moyen de 134 ms par image. Le temps d'inférence présenté par Doc2Seq est très raisonnable sachant que le modèle permet l'extraction directe des objets dans l'ordre de lecture demandé.

#### VISUALISATION DES PRÉDICTIONS

La Figure 7.4 montre les résultats visuels sur quatre images de l'ensemble de test du jeu de données IAM. Les boîtes annotées manuellement sont représentées en bleu et les boîtes prédites par le modèle sont en rouge. Visuellement, les prédictions sont très proches des boîtes annotées pour les trois images en haut et en bas à gauche. Nous remarquons très peu de problèmes liés à la largeur des boîtes et à leur position selon l'axe des abscisses. De plus, nous notons que les principales erreurs viennent de la hauteur des boîtes ainsi que leur position selon l'axe des ordonnées. En effet, sur les images en haut à droite et en bas à gauche, nous voyons que certaines boîtes sont trop hautes par rapport aux boîtes annotées correspondantes. L'image en bas à gauche montre des boîtes mal localisées, trop basses selon l'axe des ordonnées.

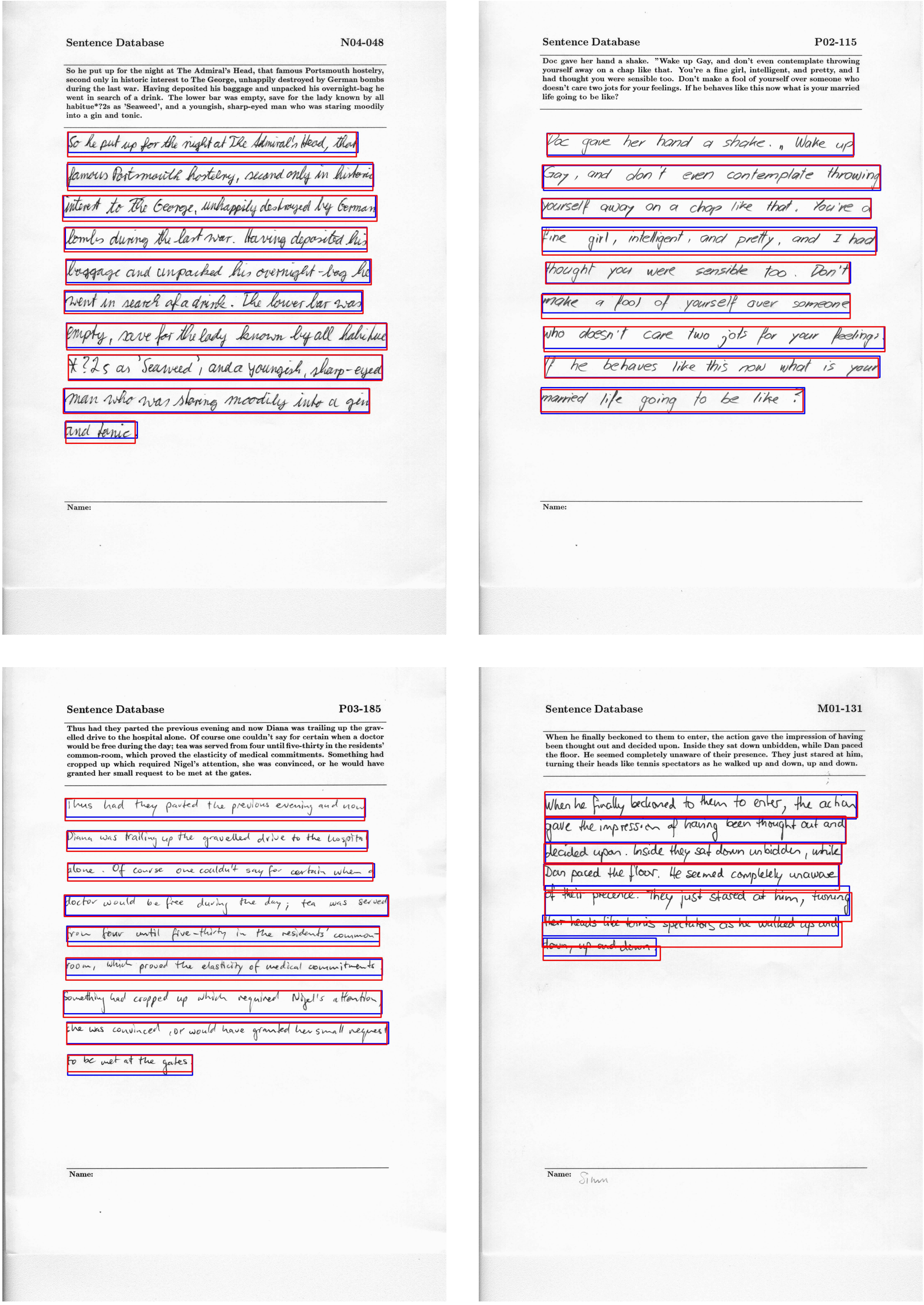

**Figure 7.4 –** Détections de lignes produites par le modèle Doc2Seq, sélectionné sur les valeurs du CER, sur quatre images de l'ensemble de test du jeu de données IAM. Les rectangles englobants annotés manuellement sont représentés en bleu et les boîtes prédites en rouge.

Le modèle prédit, tout d'abord, l'ordonnée puis l'abscisse de chaque point. Ainsi, notre hypothèse est que, pour prédire l'abscisse d'un point, le modèle a déjà connaissance de son ordonnée, il sait donc exactement où regarder sur l'image (sur quelle ligne d'ordonnée) pour se positionner. Les coordonnées prédites selon l'axe des abscisses sont donc plus précises. Au contraire, afin de prédire la première coordonnée, le modèle peut regarder partout sur l'image, ou du moins en dessous des lignes précédemment prédites, ce qui peut mener à des localisations moins précises. Cette hypothèse s'applique également au second point à prédire.

Afin de vérifier cette hypothèse, il serait nécessaire d'entraîner un second modèle à prédire d'abord l'abscisse puis l'ordonnée de chaque point. Ainsi, nous pourrions voir si les problèmes seraient désormais liés à la largeur des boîtes et leur localisation selon l'axe des abscisses. Ces expérimentations sont actuellement en cours.

## 7.5 CONCLUSION

Dans ce chapitre, nous avons présenté un nouveau système de détection d'objets dans les images de documents. Celui-ci se base sur les Transformers, algorithmes les plus performants actuellement dans de nombreux domaines. Nous avons montré que ce système pouvait être entraîné sur un jeu de données restreint d'images de documents. Il permet d'obtenir des premiers résultats prometteurs. En effet, il montre des temps d'inférence raisonnables tout en obtenant des performances à l'état de l'art et en permettant une prédiction séquentielle.

Nous avons également proposé une modélisation complète du problème de détection d'objets. Cette modélisation permet de représenter les objets de manière simple et une détection efficace de ceux-ci. Cette modélisation a le potentiel à se généraliser à d'autres tâches plus complexes, qui feront l'objet de nos futures recherches.

# 8 CONCLUSIONS ET PERSPECTIVES

## 8.1 CONCLUSIONS

Dans cette thèse, nous avons proposé deux systèmes à base de réseaux neuronaux afin de résoudre la tâche de détection d'objets dans les images de documents. Le premier modèle, Doc-UFCN, a présenté de grandes capacités de détection sur de nombreux jeux de données manuscrits hétérogènes. Il a également montré une grande robustesse en obtenant des résultats très compétitifs sur de nouveaux documents hors échantillon. Le second modèle, Doc2Seq, a également obtenu des premiers résultats encourageants qui permettent de traiter la détection d'objets à un plus haut niveau.

Le développement et l'application de ces modèles ont mené à des études plus globales concentrées sur les données et leurs annotations, les métriques d'évaluation et les scores de confiance.

Nous avons répondu aux principales problématiques liées à la détection d'objets dans les images de document dans un cadre industriel. La première problématique concerne le développement de modèles avec de grandes capacités de généralisation. En effet, dans ce cadre dans lequel de nouveaux projets sont régulièrement mis en place, il n'est pas envisageable d'annoter de nombreuses données pour chacun de ces projets, d'où la nécessité de développer des modèles plus génériques, montrant des performances élevées sur des documents très hétérogènes.

Dans cette optique, nous avons entraîné plusieurs systèmes sur un grand volume de données très différentes. Ces entraînements ont mené à des modèles plus robustes, obtenant de meilleures performances que des modèles spécifiques entraînés sur un jeu de données unique. Comme la plupart des modèles de type FCN, ces modèles ont cependant montré des difficultés à prédire des éléments qui se touchent ou se chevauchent. Pour cette raison, nous avons proposé une uniformisation des annotations, ainsi qu'une scission des boîtes afin de réduire ces chevauchements dans les annotations, utilisées durant la phase d'entraînement. Ces traitements permettent la prédiction de boîtes plus précises et non fusionnées.

Une seconde problématique induite par le cadre de production dans lequel se situe cette thèse concerne l'efficacité des modèles de détection : ceux-ci doivent fournir des détections de grande précision tout en montrant des temps de traitement réduits et en pouvant être entraînés sur des jeux de données restreints. Dans la littérature, de nombreux modèles ont été proposés pour la détection d'objets dans les images de documents, cependant, la plupart requièrent un grand nombre de données annotées. Pour pallier ce problème, des systèmes utilisant des poids pré-entraînés sur des images de scènes naturelles, tels que dhSegment, ont été proposés, mais ces systèmes montrent des temps d'inférence encore trop élevés pour une utilisation à l'échelle industrielle.

Pour répondre à ces problématiques, nous avons proposé le système Doc-UFCN. Ce système a montré des temps d'inférence réduits et obtenu des performances à l'état de l'art. Celui-ci peut être entraîné sur peu de données en comparaison avec les systèmes dédiés à la détection d'objets dans les images de scènes naturelles.

Dans de nombreux projets, il y a peu, voire aucune donnée annotée. Comme énoncé plus tôt, il est nécessaire d'avoir un détecteur assez générique afin de traiter ces documents plus facilement. Cependant, les modèles génériques peuvent parfois ne pas être suffisamment performants sur ces nouveaux documents, qui peuvent avoir une mise en page très différente de celles des documents que le modèle a rencontrés durant sa phase d'entraînement. Pour cela, l'apprentissage actif a été introduit, permettant d'entraîner itérativement des modèles en ajoutant, à chaque itération, de nouvelles données annotées sélectionnées dans le but d'améliorer les résultats du modèle de détection tout en réduisant le coût d'annotation manuelle. Dans ce cadre, il est nécessaire que le modèle fournisse les détections tout en estimant automatiquement leur qualité.

Nous avons proposé et évalué quatre estimateurs de confiance. Ceux-ci ont permis d'entraîner des modèles atteignant des performances élevées pour la détection d'objets tout en ne nécessitant qu'un faible nombre d'images annotées manuellement. Nous avons également démontré que deux de ces estimateurs permettent de sélectionner les détections les plus précises afin d'être utilisées dans un entraînement autosupervisé. Les modèles ainsi entraînés ont permis d'obtenir des gains significatifs de performances par rapport aux modèles génériques tout en ne nécessitant aucune donnée annotée.

La reconnaissance de documents consiste généralement en l'application successive de différents modèles. Dans ce cadre, les lignes de texte produites par un modèle de détection sont généralement fournies à un modèle de reconnaissance de texte. Ainsi, l'amélioration de la détection des lignes de texte doit permettre d'améliorer les résultats de reconnaissance, or les deux tâches ne sont pas étroitement liées. Il est donc nécessaire d'évaluer, à chaque étape du traitement, son impact sur les résultats de l'étape suivante ou finale.

Cette problématique a été très peu étudiée dans la littérature. Afin d'avoir une évaluation de la détection de lignes de texte davantage cohérente avec la tâche finale, nous avons donc proposé des métriques liées à la reconnaissance de texte. Ces métriques permettent de voir directement l'impact de la tâche de détection sur les résultats finaux. De plus, nous avons constaté que l'utilisation de ces métriques durant l'entraînement de modèles de détection permet d'optimiser les résultats du reconnaisseur de texte.

Enfin, les modèles à base de réseaux de neurones profonds de type Transformers ont récemment été proposés. Ceux-ci ont été introduits pour des tâches de traitement de la langue et notamment la tâche de traduction de texte. Ils ont été initialement établis afin de pallier le trop faible contexte disponible pour traiter les longues séquences de texte par les réseaux récurrents, systèmes largement utilisés jusqu'alors pour traiter ces tâches. Par la suite, certains travaux ont cherché à adapter ces modèles aux tâches de vision, motivés par leur capacité de modélisation des dépendances des éléments en entrée réalisée à l'aide du mécanisme d'attention. Bien que ces travaux aient montré des avancées significatives pour les tâches de classification d'images, très peu se sont intéressés à leur application aux tâches de détection d'objets.

Dans cette thèse, nous nous sommes intéressés à adapter ces modèles à base de Transformers à la tâche de détection d'objets dans les images de documents. Nous avons donc proposé Doc2Seq, un modèle hybride combinant un encodeur convolutif et un décodeur Transformer. Ce modèle a permis d'obtenir des premiers résultats encourageants, tout en respectant l'ensemble des contraintes évoquées précédemment. Il permet de modéliser correctement les dépendances entre les différentes parties de l'image d'entrée mais aussi celles entre l'image d'entrée et les coordonnées prédites. Ce système apporte également d'autres avantages tels que sa capacité à produire des résultats séquentiels et structurés.

## 8.2 PERSPECTIVES

Dans de nombreux domaines d'application des réseaux de neurones profonds, les modèles sont entraînés sur des milliers, voire des millions d'exemples. En effet, le premier Vision Transformer proposé, pour la tâche de classification d'images, a nécessité un pré-entraînement sur 303 millions d'images. De la même manière, Pix2Seq a été entraîné sur le jeu de données de référence MS-COCO 2017 comportant 118 000 images d'entraînement. De plus, leur meilleur modèle a été obtenu grâce à un pré-entraînement sur les données Objects365, qui représentent 600 000 images d'entraînement. Ces modèles ont montré des performances très élevées, montrant l'intérêt d'utiliser de telles quantités de données.

Dans cette optique, une perspective de cette thèse est de collecter encore davantage de jeux de données, toujours plus divers, et d'étudier la capacité du modèle à apprendre à partir de ces données. La plupart de ceux utilisés jusqu'ici étaient principalement historiques, il serait également envisageable de collecter des documents modernes afin d'obtenir un détecteur de lignes de texte très générique.

Nous souhaiterions également comparer cette approche à une stratégie de collecte dans laquelle un nombre plus restreint d'exemples serait utilisé, mais qui chercherait à maximiser la diversité des mises en page et des contenus. Durant cette étude, il serait également intéressant d'évaluer l'impact du balancement des différentes données dans l'ensemble d'entraînement.

De plus, nous avons proposé, durant cette thèse, quatre estimateurs de confiance permettant de sélectionner les images à annoter. Pour le moment, ceux-ci sont utilisés dans le cadre d'une adaptation d'un modèle générique à un nouveau domaine, afin de sélectionner les images à annoter pour mettre en place un nouveau système. Une perspective serait d'utiliser les confiances estimées afin de suivre la qualité des résultats en production. De plus, dans un cadre non supervisé, les confiances estimées par le modèle de détection peuvent permettre de vérifier que celui-ci s'améliore, durant les différentes itérations, grâce à sa confiance moyenne.

De nombreux modèles de détection d'objets traitent les images de documents à partir de sous-résolutions. C'est également le cas des deux modèles que nous avons proposés, Doc-UFCN et Doc2Seq. Bien que, dans la plupart des cas, cette sous-résolution soit suffisante pour obtenir des résultats satisfaisants, dans le cas où les objets sont très petits et très proches, la détection est impossible puisque de nombreuses fusions sont produites. Il serait intéressant de comparer différentes sous-résolutions mais aussi une approche par patchs, bien que celle-ci soit beaucoup plus coûteuse en ressources et en temps d'inférence. De même, nous pourrions estimer la sous-résolution optimale pour un jeu de données ou pour une image de manière automatique.

Nous souhaitons également évaluer le modèle Doc2Seq sur d'autres jeux de données et d'autres tâches, et notamment des problèmes plus complexes avec des objets imbriqués tels que des paragraphes et lignes de texte. Il serait intéressant de le tester sur d'autres tâches telles que l'analyse de mise en page de tableaux avec la détection des éléments structurels tels que les lignes et colonnes de titre et les pieds de tableaux [1].

De plus, l'ensemble des modèles de détection traitent les images de manière isolée. Ils ne possèdent aucune mémoire quant aux prédictions réalisées sur les images précédentes. Or, dans le cadre du traitement de séries (ouvrages ou collections), il pourrait être bénéfique de considérer, lors du traitement d'une nouvelle image, des propriétés établies sur d'autres images ou au niveau de la série. Le système Doc2seq permet d'envisager un tel traitement. En effet, l'utilisation des prédictions précédentes dans le décodeur Transformer permet d'imaginer des tâches de plus haut niveau. Ainsi, lors du traitement d'un livre ou d'une collection, la prédiction d'une image pourrait être initialisée par les éléments prédits sur l'image précédente ou par une moyenne des positions prédites sur l'ensemble des pages précédentes. Cela permettrait de transférer des informations d'une image à l'autre et d'avoir des résultats homogènes. Nous souhaitons étudier cette possibilité dans de futurs travaux.

1. http://www.socface.org/

# BIBLIOGRAPHIE